School of Chemical Sciences and Engineering
The University of New South Wales

# Adaptation and Self-Organization in Evolutionary Algorithms

by

James M. Whitacre

**submitted in partial fulfillment of**

**the requirements for the degree of**

**Doctor of Philosophy**

**July 2007**

# ACKNOWLEDGEMENTS

There are a number of very important people whom I would like to acknowledge. First, I would like to sincerely thank my supervisor, Associate Professor Q. Tuan Pham for his guidance, support, and encouragement. His critical eye in the earlier drafts of this thesis helped to greatly strengthen the final version.

I would like to thank Dr Ruhul Sarker for the support and advice he has provided over the last two years. I also want to thank him for taking an interest in my initial research ideas and for his willingness to take on a co-supervisory role in this research.

I want to acknowledge a number of people who have provided fruitful discussions and advice over the last few years including Hussein Abbass, Alan Blair, David Green, and Eric Whitacre. I also want to thank Hartmut Pohlheim for providing a number of the fitness landscape graphics used in this thesis.

Special thanks to my special gal Alice, especially for putting up with my sometimes long work hours. Finally, I would like to thank my wonderful family for their support, encouragement and love (especially my mom and sister).

# ABSTRACT

The objective of Evolutionary Computation is to solve practical problems (e.g. optimization, data mining) by simulating the mechanisms of natural evolution. This thesis addresses several topics related to adaptation and self-organization in evolving systems with the overall aims of improving the performance of Evolutionary Algorithms (EA), understanding its relation to natural evolution, and incorporating new mechanisms for mimicking complex biological systems.

Part I of this thesis presents a new mechanism for allowing an EA to adapt its behavior in response to changes in the environment. Using the new approach, adaptation of EA behavior (i.e. control of EA design parameters) is driven by an analysis of population dynamics, as opposed to the more traditional use of fitness measurements. Comparisons with a number of adaptive control methods from the literature indicate substantial improvements in algorithm performance for a range of artificial and engineering design problems.

Part II of this thesis involves a more thorough analysis of EA behavior based on the methods derived in Part I. In particular, several properties of EA population dynamics are measured and compared with observations of evolutionary dynamics in nature. The results demonstrate that some large scale spatial and temporal features of EA dynamics are remarkably similar to their natural counterpart. Compatibility of EA with the Theory of Self-Organized Criticality is also discussed.

Part III proposes fundamentally new directions in EA research which are inspired by the conclusions drawn in Part II. These changes involve new mechanisms which allow self-organization of the EA to occur in ways which extend beyond its common convergence in parameter space. In particular, network models for EA populations are developed where the network structure is dynamically coupled to EA population dynamics. Results indicate strong improvements in algorithm performance compared to cellular Genetic Algorithms and non-distributed EA designs. Furthermore, topological analysis indicates that the population network can spontaneously evolve to display similar characteristics to the interaction networks of complex biological systems.

# TABLE OF CONTENTS

# LIST OF FIGURES

# LIST OF TABLES

# ABBREVIATIONS

| | |
|---|---|
| BA | Barabasi-Albert Network Model (Section 5.1.3.1) |
| cGA | cellular Genetic Algorithm (Section 2.3.7.2) |
| DC | Deterministic Crowding (Section 2.3.7.1) |
| DD | Duplication and Divergence Network Model (Section 5.1.3.2) |
| DE | Differential Evolution (Section 2.3.4) |
| dGA | distributed Genetic Algorithm (Section 2.3.7.2) |
| EA | Evolutionary Algorithm (Section 2.3) |
| EC | Evolutionary Computation (Section 2.3) |
| ECC | Error Correcting Code Problem (Appendix A) |
| ETV | Event Takeover Value (Section 3.2.2) |
| GA | Genetic Algorithm (Section 2.3) |
| MMDP | Massively Multimodal Deceptive Problem (Appendix A) |
| MTTP | Minimum Tardy Task Problem (Appendix A) |
| RCGA | Real-Coded Genetic Algorithm (Section 2.3.6) |
| SOC | Self-Organized Criticality (Section 4.3) |
| SOTEA | Self-Organized Topology Evolutionary Algorithm (Chapter 5) |

# Chapter 1 Introduction

Evolutionary Algorithms (EA) are a class of stochastic optimization methods which loosely follow principles of natural selection in order to solve challenging problems. Over the years, a strong track record (e.g. see [1], [2]) has brought them popularity in academia and has also started to bring acceptance from industry [3], [4]. Today a number of companies which specialize in providing optimized business solutions are now using EA techniques [5], [6], [7], [8], [9], [10], [11].

Although Evolutionary Algorithms have achieved impressive performance in many application domains, these achievements are partly the result of careful algorithm design which often involves substantial efforts in defining a problem's representation and/or the careful design of an EA's genetic operators. These significant design efforts are a reflection of the fact that an EA is presently not able to robustly adapt its search behavior to fit a particular optimization problem. One promising avenue for addressing this problem is learn how open-ended adaptability and robustness occurs in natural evolutionary processes and to incorporate these mechanisms into an EA.

To achieve such a goal, it is expected that a number of key features of natural evolution will need to be integrated into an EA, some of which are not yet fully understood.[1] Although we still lack a complete understanding of evolution, the post-genomic era has provided a number of important insights into complex biological systems as well as a better understanding of the evolutionary processes that created these systems. With these recent developments in mind, it was agreed upon at a recent workshop that a concerted effort should now be made to integrate the latest understanding of evolutionary processes into EA design [22].

If such efforts bear fruit over the coming years, it is anticipated that EA will become a more flexible, autonomous, and robust algorithm for solving today's learning, control, design, and scheduling tasks. As the gap between natural evolution and EA behavior is narrowed even further, it is anticipated that EA research could also become of strategic importance for a number of frontier technologies where optimization methods are required but are not presently capable of providing viable solutions.

[1] Examples of important features of natural evolution that should be of particular interest to EC research are discussed in [12], [13], [14], [15], [16], [17] and studied in [18], [19], [20], [21].

## 1.1 Aims and Outline

The overarching aim of this research is to make progress in narrowing the gap between EA and natural evolution. Hence, the research questions raised in this thesis are aimed at incorporating (or understanding) some of the non-trivial aspects of natural evolution that are missing in Evolutionary Algorithms. The key aims and research questions raised in this thesis are described below.

**Understanding and Designing an Adaptive System:** The effectiveness of an adaptive system can be measured by its ability to maintain competitiveness in a changing environment. Natural adaptive systems have ingrained within them an ability to advantageously change internal components when exposed to changing external forces. However, it is not completely understood what features are required to make an adaptive process effective in Evolutionary Algorithms. An example of such an adaptive process is observed in the adaptive methods used for the automated control of EA design parameters. Within this context, this thesis aims to answer how the interactions between such adaptive systems and their environment can be translated into useful information for driving internal changes to these adaptive systems.

To date, research into methods for adapting EA design parameters has focused on the use of fitness measurements for controlling adaptive behavior. Chapter 3 presents an alternative measurement, called ETV (Event Takeover Value), which is derived from empirical evidence of an individual's impact on population dynamics. The ETV is able to measure an individual's impact on EA population dynamics through an analysis of EA genealogical graphs.

During a preliminary analysis of population dynamics (using ETV), an unexpected behavior is uncovered in Evolutionary Algorithms. It is found that there is a surprising scarcity of individuals in an EA population that cause even moderate changes to population dynamics. Instead, most individuals actually have a negligible impact on the system. After thorough testing (mostly presented in Chapter 4), it is concluded that this is a fundamental property of EA dynamics and that most interactions between the EA population and its environment are effectively neutral and non-informative.

Based on this conclusion and through the use of statistical arguments, the new adaptive system is modified to filter out data from individuals that have a small impact on population dynamics. As a result, only important interactions between the adaptive system and its environment are used to drive adaptive changes in the system. Experiments conducted on a number of artificial test functions and engineering design problems indicate that the new method for adapting EA design parameters is superior to a number of adaptive methods selected from the literature.

**Understanding EA Population Dynamics:** The aim of Chapter 4 is to gain a better understanding of the population dynamics of Evolutionary Algorithms using the ETV measurement derived in Chapter 3. One important observation from this chapter is that the probability distribution of an individual's impact on population dynamics fits a power law with most individuals having a negligible impact. This chapter investigates this feature of EA dynamics more closely with the goal of determining what experimental conditions lead to power laws and what conditions lead to deviations from a power law.

By knowing what aspects of EA design can impact this characteristic of EA population dynamics, it is expected that this information can be used to improve EA robustness, in part, by driving EA behavior towards a more accurate reflection of natural evolution. After comparisons are made between EA results and somewhat related observations from natural evolution, it is concluded that some aspects of the two systems share similar patterns of behavior but only when certain conditions are met. In particular, the population topology is found to be a significant factor in the ETV results and it is found that EA populations that are fully connected (i.e. Panmictic) are unable to mimic the spatial properties of natural evolutionary dynamics.

The experimental results from this chapter also provide evidence that the spatial properties of EA dynamics and its genealogy are self-organized and possible explanations for this behavior are given based on the Theory of Self-Organized Criticality.

**Mimicking the Structural Organization of Complex Systems:** Although the majority of experimental factors tested in Chapter 4 do not significantly influence EA population dynamics, one factor which did alter its behavior was the introduction of spatial restrictions within the population. Interaction restrictions also occur in biological systems, however it is well known that network approximations of these systems have very different topological properties compared to current spatially distributed EA populations.

Structure is an emergent property of complex biological systems and plays a fundamental role in the robustness and general behavior of these systems. Chapter 5 reviews contemporary understanding of how structure emerges in nature with the goal of determining how similar structural organization can be integrated with EA design in order to improve its robustness and behavior.

Hence, one of the primary aims of this chapter is to determine how an EA population structure can self-organize to exhibit topological characteristics similar to complex biological systems. This aim is achieved by modifying the population topology using localized rules that are coupled to EA population dynamics. Two different models of topological organization are studied and each is found to display interesting behaviors. In particular, the first model is found to generate non-random selection pressure patterns within the population topology and also is able to sustain very high levels of genetic diversity. The second model is intentionally designed to evolve population structures with high levels of modularity. Results from testing this algorithm on a suite of test problems indicate that the new EA design strongly outperforms a number of other algorithms including cellular Genetic Algorithms and several non-distributed EA designs.

The following chapter provides general background material for this thesis. The chapter starts with a brief introduction to optimization including a discussion of what conditions make optimization challenging and why nature-inspired optimization methods are useful within certain contexts. A justification is also provided for the specific focus of this thesis on Evolutionary Algorithms as opposed to other nature-inspired methods.

# Chapter 2 General Background of Evolutionary Algorithms

This chapter reviews (briefly) the background concepts and ideas that underlie the work conducted in this thesis. Each subsequent chapter also introduces and critically reviews material that is relevant to the research questions being addressed within that chapter. The intention is to allow each chapter to be largely self-contained in order to improve clarity of the material and to keep terminology and concepts fresh in the reader's memory as they are introduced and subsequently explored.

## 2.1 Optimization Framework

In general, optimization problems involve setting a vector $x$ of free parameters of a system in order to optimize (maximize or minimize) some objective function $F(x)$. A solution to a problem can also be subject to the satisfaction of inequality constraints $g(x)$, equality constraints $h(x)$, as well as upper and lower bounds on the range of allowable parameter values. Given a minimization problem consisting of $n$ parameters, $q$ inequality constraints, and $r$ equality constraints, the problem can be defined as shown below.

$$Min\ F(x), \quad x = (x_1, \ldots, x_n) \qquad \textbf{(2-1)}$$

*Subject to:*

$$g_k(x) \le 0, \quad k \in \{1,...,q\} \qquad \textbf{(2-2)}$$

$$h_j(x) = 0, \quad j \in \{1,...,r\} \qquad \textbf{(2-3)}$$

$$x_i^L \le x_i \le x_i^U, \quad i \in \{1,...,n\} \qquad \textbf{(2-4)}$$

This is the basic structure of the single objective optimization problems considered in this thesis. There are no specific conditions attached to the variable type and the function characteristics although many of the problems tested have multimodal fitness landscapes and significant levels of parameter epistasis. Other conditions which are commonly addressed in optimization research but will not be specifically addressed here include dynamic objective functions and multiple conflicting objectives.

Optimization problems are typically broken down into classes based on fitness landscape characteristics. For some combinations of characteristics, search algorithms can be designed with a search bias that can effectively exploit these landscape features and allow the problem to be solved to optimality (or near optimality) with relatively little computational effort. Examples of such simplifying characteristics include linear separability, convex feasible spaces, and smooth unimodal landscapes.

However, many real world optimization problems have characteristics which are not as susceptible to simplifying assumptions. A list of arguably the most important of these characteristics is given below. It is important to note that many of the characteristics do not necessarily pose a significant challenge when they occur in isolation, however the presence of several of these conditions can make a problem very difficult to solve.

**Characteristics which make optimization problems challenging**

Dimensionality

Multimodality

Complex Constraints on Feasibility

Epistasis

Deception

Uncertainty

Computational Costs

Dynamic Objectives

Multiple Objectives

**Dimensionality**: The more parameters that must be varied in order to optimize a problem, the larger the dimensionality of the solution space. This can result in the problem size increasing by orders of magnitude. However, the importance of dimensionality greatly depends on the existence of parameter epistasis. If the additional parameters can be solved separately from the other parameters in the problem then the increase in complexity will only be additive and for the most part negligible.

**Multimodality**: Multimodality refers to fitness landscapes that contain multiple fitness peaks (i.e. locally optimal solutions). These peaks play a critical role in the performance of almost all optimization methods. Their prevalence through many problems of interest has been the primary impetus for research into alternatives to deterministic directed search and gradient-based search algorithms.

**Complex Constraints on Feasibility**: Constraints determine which solutions are considered feasible within the solution space and they can play a significant role in the difficulty of an optimization problem. When constraints are nonlinear, they often result in a patchwork of feasible solutions where isolated islands are surrounded by infeasible solution space. The location of the optimal solution within this patchwork can be an important factor in dictating how difficult the problem is to solve for a given algorithm.

Nonlinear constraints add to problem difficulty in a way that is somewhat similar to the inclusion of multiple objective functions however they also introduce unique difficulties in a problem. Unlike the objective function where "almost optimal" is generally good enough, almost feasible is rarely accepted.

**Epistasis:** Epistasis is a term used to indicate the degree of interaction between parameters in an objective function (or in constraint functions). Problems lacking epistatic interactions are completely separable (i.e. decomposable) meaning that each of the parameters can be solved in isolation. Many real-world problems have at least some degree of epistasis. Epistasis is also a common contributor to multimodality and general problem difficulty. The impact that epistasis has on problem difficulty and on EA behavior is a significant topic of investigation in this thesis which is dealt with in more detail in Chapter 5.

**Deception**: Deception is traditionally a term used to describe a feature of fitness landscapes that make them difficult to solve using Evolutionary Algorithms. This difficulty is due to the challenge of maintaining building blocks of genetic material that are needed later in the search process in order to find the optimal solution. However, the concept of deceptiveness can be generalized to apply to any algorithm where the search bias ingrained in the algorithm makes the optimal solution more difficult to find as the search progresses. In other words, deception is the result of an algorithm's search bias being fundamentally inappropriate for searching the given fitness landscape. A common form of deceptiveness is when hill climbing in a fitness landscape consistently drives the search away from the optimal solution.

**Uncertainty**: Uncertainty refers to a lack of confidence that the fitness landscape generated by an objective function accurately reflects the true landscape of the problem being solved. Uncertainty in a problem can come in many different forms. For instance, many real world problems are representations of a physical system where a model is developed using a number of assumptions and simplifications. In this case, uncertainty can come from the accuracy of the governing equations that are used to model the physical problem. Another source of uncertainty is discretisation (i.e. granularity) of the parameter space which is necessary for numerical optimization using computers. Discretisation can significantly reduce the size of the parameter space however this can also eliminate any chance of sampling the best parameter combinations. Depending on parameter sensitivity, granularity can also play a role in defining prominent characteristics of a fitness landscape which in turn could alter the dynamics and performance of a search process.

Uncertainty can also come from noise in the objective function evaluation which can be inherent in the system, caused by measurement errors, or caused by numerical errors. A review of additional types of uncertainty that are experienced in optimization research and how they are addressed in Evolutionary Algorithms is provided in [23].

**Computational Costs**: Many real world optimization problems have large computational costs associated with the objective function and/or constraint evaluation. These costs are generally due to simulation of a real system. Metamodels such as Kriging models (e.g., see [24]) can be used to approximate the fitness landscape so that the more computationally expensive simulations are needed less frequently. However, increasing computational efficiency in this way will also add uncertainty to the evaluation of the objective function. This tradeoff means there are limits to the amount of increased computational efficiency that can be afforded by metamodels. Another possibility is to increase the granularity of the search space however there are tradeoffs with this approach as well which were previously discussed.

**Dynamic Objectives:** Dynamic objective functions involve fitness landscapes that can change over time. When presented with a dynamic objective function, it is generally assumed that it is not possible to know how the problem definition will change or how this will impact the fitness landscape being searched. In this context, it is no longer sufficient to design a solver that can effectively search a given landscape. Instead, a solver must also maintain some degree of search robustness and flexibility which can allow the search process to quickly account for changes in the fitness landscape. In some respects, the desirable search features for a dynamic optimization problem are more inline with the features observed in natural evolution and which are demanded by the natural environment and less similar to the conditions treated in traditional optimization (e.g. mathematical programming).

This is not to say that traditional optimization techniques are not used in this domain, however they commonly assume that dynamical uncertainty can be represented using probabilistic models (e.g. stochastic programming). In most cases however, such models are unable to account for the emergent phenomena that is present in dynamic optimization problems involving complex systems (e.g. social systems, climate change, warfare, and organizational dynamics).

**Multiple Objectives:** Most real world problems are not defined as having a single objective. Instead there are often multiple conflicting objectives which can not be combined into a single metric. Common examples of such objectives include various measures of cost, performance, efficiency, risk, and heuristic objectives based on human experience. This complicates a search process because we are generally no longer presented with a problem containing a single optimal solution or a single selection pressure (i.e. driving force) for searching through the solution space. Such conditions can introduce unique challenges but also unique opportunities, especially for population based search processes. For more information on multi-objective problem characteristics and Evolutionary Algorithms designed for this problem domain, we refer the reader to [25], [26].

### 2.1.1 Reconciling Optimization Research in a World of "No Free Lunches"

Presented with the challenges listed above, it is important to ask whether an algorithm can be designed to effectively deal with all of these characteristics simultaneously and in all of their varied forms. In other words, is it possible to create an effective general purpose optimization algorithm?

An important development along this line of questioning was the No Free Lunch Theorems for Optimization (NFL) [27]. Given some basic assumptions (e.g. see [28]), NFL states that no optimization algorithm is better than any other when its performance is averaged over all possible problems. If one assumes that this equality holds true for the subset of real-world problems, then NFL would place severe limitations on the amount of progress that is possible in optimization research.

However, experience over the years suggests that NFL has only a partial bearing on the real world. On the one hand, experience has shown that real-world problems cover a broad range of problem types and that even for problems which appear to be similar, the best approach to solving them can often be very different. In short, empirical evidence supports the notion that no best approach to optimization exists.

On the other hand, most real-world problems do display some basic similarities in fitness landscape features such as the presence of correlated landscapes[2] (also see [30]). Experience also has shown that not all optimization algorithms are equal and in fact some appear to be quite good at solving a reasonable range of problems (also see [30]). In summary, NFL should act as a guide when conducting optimization research however the goal of developing more effective optimization algorithms can be a reasonable aim if sufficient justification is provided.

## 2.2 Justification of EA Research

Following from the previous discussion, it appears that a strong argument should be given to justify research that focuses on advancing a particular class of optimization algorithms. A common and certainly valid justification would be one that is based on empirical evidence of strong algorithm performance. Indeed many nature-inspired algorithms and particularly Evolutionary Algorithms (EA), have been found to be effective in a number of important niche applications. As a result, the use of these methods in solving real world problems has steadily grown over the years.

[2] One common feature of almost all real-world problems is the existence of correlated landscapes which is to say that one can expect (on average) that similarities between solutions in parameter space will produce similarities in objective function value. For correlation metrics and a review, see [29].

Today, a number of nature-inspired optimization methods exist. Examples include Ant Colonies, Immune Systems, Particle Swarms, and Simulated Annealing. Each are interesting as topics of investigation in their own right, and deserve further study. However, the decision to use EA as the algorithmic framework for this research was not a decision that was taken lightly nor was it a decision based solely on current empirical evidence. The decision to study Evolutionary Algorithms was instead largely based on the desirable qualities of its natural counterpart.

**Evolution and Optimization:** Many biological systems in nature are viewed as having powerful problem solving abilities. Nature-inspired optimization methods attempt to mimic these behaviors, however few of the systems being mimicked have a clear relation to optimization. On the other hand, natural evolution has a number of important similarities to optimization that are now well recognized.

The first link between optimization and evolution was made in relating the natural environment to a fitness landscape which was suggested by Sewall Wright back in 1932 [31]. He postulated that the adaptation of species was similar to climbing up a fitness landscape which occurs due to genetic mutations and is driven by natural selection. It is quite simple (although not strictly accurate) to also think of the genome as a parameterization of life and to think of the thriving of a species as being due to its success in accomplishing some set of objectives.

Looking at the diversity of life forms and the diversity of environments where life has flourished suggests that, although individual species are great specialists, the forces driving evolution are a powerful generalist. This ability to continually adapt and evolve new specialized behaviors is not possible in today's optimization algorithms although it is a highly desirable feature. In short, EA was selected as the topic of investigation because natural evolution has a capacity to robustly "solve" a range of problems in the natural environment which are well outside the capacity of today's algorithms.

Other natural systems, such as the behavior of ants or the immune system, are viewed as very capable but highly specialized systems that have come about as a result of evolutionary processes. Hence, it is doubtful whether the overall potential of these other algorithmic frameworks is comparable with the potential from mimicking natural evolution.

Based on the premise that natural evolution has unique and advanced problem solving capabilities, this thesis focuses on ways to mimic natural evolution in an artificial environment for purposes of optimization. This thesis tackles this topic in a multifaceted approach looking at issues such as i) building an effective feedback adaptive process ii) comparing the dynamical behavior between EA and natural evolution and iii) creating models for the emergence of nature-inspired EA population structures. It is hoped that this work will help others to look at EA from a different perspective and will help to generate more effective algorithms for exploiting the power of natural evolutionary processes. Having now provided the motivation and justification for this thesis, the remainder of this chapter provides a basic review of Evolutionary Algorithms for optimization.

## 2.3 Evolutionary Algorithms

### 2.3.1 A Brief History

The term Evolutionary Algorithms is used to describe a range of stochastic optimization methods which employ principles of natural selection and reproduction in biology to evolve solutions to problems. Research in the field of Evolutionary Computation (EC) started as early as the late 1950s [32], [33], [34], although much of the fundamental work, which is generally recognized as the origins of EC research, took place several years later.

Three of the algorithmic frameworks developed in the early days of EC research are still in active use today and include Genetic Algorithms (GA) [35], [36], Evolutionary Programming (EP) [37], [38] and Evolution Strategies (ES) [39], [40]. Although there are differences between each of the algorithms, their similarities are much more striking and most research using one algorithm class is generally applicable to the others.

Instead of reviewing each of these algorithmic classes, the following review of EA reflects the scope of the research presented in this thesis which deals primarily with the topic of parameter optimization using population-based search heuristics. For a more thorough review of EC research we refer the reader to [41].

### 2.3.2 General Description

Taking terminology from genetics, an Evolutionary Algorithm initially starts with a population of individuals, with each individual representing a solution to the problem being solved. Each individual has a chromosome made up of genes or parameters, and the set of all possible combinations of these genes makes up the genotypic space (i.e. solution space or parameter space). The individuals within a population are selected to reproduce and participate in the next generation in a process similar to the Darwinian principle of survival of the fittest. New individuals (referred to as *offspring*) are generated from selected *parents* using what has become a library of reproduction operators (i.e. search or variation operators); some of which are similar to genetic mutation and recombination.

The selection of individuals is based on their fitness or phenotype which is typically defined by the objective function value and is calculated using the genes of the individual. This fitness then impacts an individual's chances of survival and/or procreation. By creating a bias toward selecting the best solutions for populating the next generation, the algorithm is often able to exploit information contained in these more fit solutions in order to reach an optimal or near optimal solution.

A more concrete understanding of Evolutionary Algorithms is possible using the pseudocode in Figure 2-1 which loosely follows the framework outlined in [42]. For this pseudocode, the parent population of size $\mu$ at generation $t$ is represented by $P(t)$. For each new generation, an offspring population $P`(t)$ of size $\lambda$ is created using reproduction operators and evaluated to determine the objective function values for each offspring. The parent population for the next generation is then selected from $P`(t)$ and $Q$, where $Q$ is a subset of $P(t)$. $Q$ is derived from $P(t)$ by selecting those in the parent population with an age less than $\kappa$.

```
t=0
Initialize P(t)
Evaluate P(t)
Do
        P`(t) = Variation(P(t))
        Evaluate P`(t)
        P(t+1) = Select(P`(t) U Q)
        t=t+1
Loop until termination criteria
```

**Figure 2-1 Pseudocode for a basic Evolutionary Algorithm design**

Although some EA designs do not fit the framework listed above, many common designs do. For instance, a *non-elitist generational* EA design refers to conditions where $\kappa$=1 and $\lambda > \mu$, a *steady state* EA design refers to conditions where $\kappa=\infty$ and $\lambda$ =1, a *generation gap* EA has $1< \kappa < \infty$, and a *pseudo steady state* EA design typically involves $\kappa=\infty$ and $\lambda = \mu$. Also, when an EA design is used with elitism, this simply means that the best individual in a population is given its own value for $\kappa$ which is set to $\kappa=\infty$.

Before reviewing variation and selection schemes that are commonly used in EA design, it is important to first provide a clearer understanding of the parameters $\lambda$, $\mu$, and $\kappa$. To do this, it is helpful to temporarily neglect the mechanisms used for selecting parents and for creating new offspring. By doing this, search with an EA can be understood through its relation to a simple branching process.

Extending this analogy, active nodes describe points in the branching process from which new branches can potentially be grown and for an EA, the number of active nodes is controlled by $\mu$. The parameter $\mu$ can also be thought of as providing an upper bound on the memory or the amount of genetic material present in the system. In a single time step or generation, the total number of new branches is controlled by $\lambda$. Only active nodes have the capacity to influence where new branches occur and the composition of the new nodes (thereby making the branching process a Markov Chain). Furthermore, active nodes are forced to become inactive after $\kappa$ time steps. This limits the amount of time that a node can directly influence the creation of new nodes.

In short, these three parameters impact the algorithm by constraining the dynamics to meet certain conditions of this branching process. For instance, a constant value for $\mu$ means that the size (or memory) of the system can not change while setting $\kappa=\infty$ means an individual has the potential to directly influence future dynamics for arbitrarily long periods of time.

The actual influence of these parameters on other qualities of the system ultimately depends on how selection and variation procedures are executed in the algorithm although some general comments can be made. For instance, decreasing $\kappa$ generally causes the search process to become more influenced by basins of attraction in the fitness landscape while increasing $\kappa$ causes the search to be more influenced by point attractors (i.e. local optima). Increasing $\mu$ can increase the amount of parallel search behavior that can potentially take place in an EA however the actual amount of parallelism depends greatly on other aspects of EA design. Increasing $\lambda$ can increase the amount of innovation or changes to the makeup of the population that can potentially occur, but again the actual amount of innovation depends on other genetic operators.

The remaining sections of this chapter introduce each aspect of EA design in more detail. The next section discusses different selection methods that have been devised for selecting *P*. Section 2.3.4 discusses variation methods which are also referred to as search operators or reproduction operators. Constraint handling is discussed in Section 2.3.5 (an important topic in fitness evaluation) while options for parameter encoding are discussed in Section 2.3.6. Some advanced features in EA design are presented in Section 2.3.7 with a focus on interaction constraints between EA population members. Section 2.3.8 presents ways in which performance can be measured in Evolutionary Algorithms and Section 2.3.9 discusses the uses and applications of Evolutionary Algorithms.

### 2.3.3 Selection Methods

Selection methods have the task of deciding how much each individual in the population will act to bias future search steps that are taken by the algorithm. There are a number of selection methods that have been considered in EA research and some of the more common schemes are briefly described in this section. In general, selection simply involves the creation of one population *P`* by selecting individuals from another population *P*. Selection that is done with replacement means that individuals in *P* can be selected multiple times so that multiple copies of an individual can exist in *P`*. On the other hand, selection without replacement means that, at most, only one copy of an individual from $P$ can exist in $P$\`. Several common methods for selecting individuals are described below.

#### 2.3.3.1 Proportional Selection

Proportional selection selects an individual from a population with a probability proportional to its fitness. Given a population size of $N$, the probability $p_i$ that individual $i$ with fitness $f_i$ is selected in a single selection event is defined by (2-5). To use proportional selection with minimization problems, it is necessary to define a scaling function. Scaling functions are also generally needed when proportional selection is used in order to address this method's selection pressure sensitivity to a population's distribution of phenotypes. Due to the necessity of problem-specific scaling functions, proportional selection is difficult to implement in practice.

$$p_i = \frac{f_i}{\sum_{j=1}^{N} f_j} \qquad \textbf{(2-5)}$$

#### 2.3.3.2 Linear Ranking Selection

Linear ranking selection is an alternative selection procedure which does not have the scaling problems present in proportional selection. In linear ranking selection, solutions are ranked from most fit (Rank=1) to worst fit (Rank=$N$) and are selected with a probability that is linearly proportional to its ranking. In this way, an individual's probability of selection is based on how its fitness ranks among others in the population instead of being based on the magnitude of the fitness value. Given a population size of $N$ and parameters $\eta^+$ and $\eta^-$ which control the overall selection pressure, the probability that individual $i$ is selected in a single selection event is given by (2-6). In cases where linear ranking is used in this thesis and no values are specified for $\eta^+$ and $\eta^-$, it is assumed that $\eta^+$ =1 and $\eta^-$ =0. Other formulations for defining linear ranking are possible such as the original definition which is given in [43], [44].

$$p_i = \frac{\eta^- + (\eta^+ - \eta^-)(N - Rank_i)}{\sum_{j=1}^{N} \eta^- + (\eta^+ - \eta^-)(N - Rank_j)} \tag{2-6}$$

#### 2.3.3.3 Exponential Ranking Selection

Exponential ranking is sometimes used to introduce a stronger selection pressure than is possible with linear ranking. As the name suggests, the probability of selection follows an exponential function of rank so that there is a much greater chance of being selected if a member has a high ranking in the population. Selection pressure is controlled by $c$ so that as $c \rightarrow 1$, the difference in selection probability between the best and worst solutions is lost and as $c \rightarrow 0$, selection probability differences become increasingly larger and follow an exponential curve along the ranked solutions.

$$p_i = \frac{c^{Rank_i}}{\sum_{j=1}^{N} c^{Rank_j}} \tag{2-7}$$

#### 2.3.3.4 Tournament Selection

Tournament selection works by randomly sampling a subset of the population with sample size $q$ and then selecting the best individual from that sample. The size of $q$ will impact the selection pressure from this method. A commonly used form of tournament selection is binary tournament selection where $q$=2.

Tournament selection and tournament-based variants have a number of desirable properties that make them a good choice when designing an Evolutionary Algorithm. Tournament selection is simple to use and many advanced features in an EA can be implemented using tournaments such as the use of crowding and age restrictions. Furthermore, tournaments do not require global information in order to make selection decisions thereby making these designs more efficient to execute when run in a physically parallel environment. Finally, the selection pressure of this method can be easily tuned by changing the tournament size $q$.

#### 2.3.3.5 Truncation Selection

Truncation selection works by selecting with equal probability from among a fraction *T*, *T* ∈ [0,1] of the best individuals in the population. With a population of size *N* sorted based on rank, the selection probability is given by (2-8). If selection is conducted without replacement (as is done throughout this thesis), then each of the *T*N* best individuals are selected one time only thereby causing the selection method to be deterministic.

$$p_i = \begin{cases} \frac{1}{TN}, & if\ 1 \le i \le TN \\ 0, & else \end{cases} \qquad \textbf{(2-8)}$$

#### 2.3.3.6 Modified Tournament Selection

The experimental work in this thesis uses a modified form of tournament selection that is defined by the pseudocode below.

**Modified Tournament Pseudocode**

- starting with $\lambda + \mu$ individuals, conduct $\lambda$ tournaments
- for each tournament, select the worst individual in the tournament and remove it from the $\lambda + \mu$ population
- after $\lambda$ tournaments, we are left with a new parent population of size $\mu$
- randomly select from the parent population to generate $\lambda$ new offspring

This procedure is essentially equivalent to a canonical GA using elitism and tournament selection without replacement. The major difference with the canonical GA is in the application of elitism: the surviving elite are chosen statistically (by tournament) rather than deterministically. Furthermore, conventional elitism could be seen to over-favor the fitter members, which have a larger share of offspring per generation and survive more generations. In the modified tournament selection this favoritism happens naturally: a fitter member has more offspring simply by surviving longer - a phenomenon observed in many species.

#### 2.3.3.7 Characterization and Comments

Selection schemes introduce bias into a search process primarily by selecting points in solution space that are of relatively high fitness. However, selection schemes can also bias a search in other ways and so it is important to be able to quantify different aspects of the search bias present in a selection scheme.

One aspect of a search bias is the loss of genetic diversity which can be approximated by measuring the proportion of individuals that are not selected during a selection phase [45]. Selection schemes are also commonly characterized by their selection pressure which indicates the extent to which the scheme is biased towards preferring more fit individuals. For example, the selection intensity (as defined in [46]) measures the increase in mean fitness resulting from a selection phase. Another possibility is to measure the takeover time which is the time required for the best individual to take over a population (restricted experimental conditions apply, e.g. see [47]). Additional ways to characterize selection methods have also been proposed, some of which can be found in [48], [45], and [49].

A number of commonly used selection schemes were briefly described in this section and include proportional, linear ranking, exponential ranking, tournament, and truncation selection. Although there are differences in these selection schemes, their similarities seem to be much more striking. Most have a parameter which (to a rough approximation) tunes the selection pressure in a manner similar to the others. Most are also global selection methods where selection is based only on an individual's fitness.

More advanced selection methods do exist which take into account other factors such as age, genealogy, spatial locality and genotype in order to encourage different forms of diversity or parallel search behavior. Some of these advanced methods are reviewed in Section 2.3.7.

### 2.3.4 Search Operators and Variation

Search operators work by taking information from one or more individuals in the population as a basis for sampling new points in solution space. Early studies of Evolutionary Algorithms involved the use of crossover and/or mutation however many of the algorithms in use today employ a diverse range of search operators. Although an exceedingly large number of search operators have been introduced in the literature over the years, only a few have been able to find traction with the broader EC community. Some of the more popular search operators are found in Differential Evolution [50], Covariance Matrix Adaptation [51], and Estimation of Distribution Algorithms [52]. These operators generally employ the use of multiple parents and are highly successful when assumptions about the fitness landscape are met.

#### 2.3.4.1 Characterization

A set of ten search operators are used in experiments throughout this thesis and are described in Appendix B. As an alternative to reviewing each of these search operators in detail, as well as others that have been introduced in the literature, it is possibly more illuminating to discuss the search operators in more general terms based on their behavior and intended usage. A few directions along these lines are provided below.

**Intent and Search Bias**: Probably the most important questions to ask about an operator are; what sort of search bias is created by the operator and what sort of landscape is the operator expecting to search. A number of standard search operators have been developed over the years which can be better understood by attempting to answer these questions.

As an example, gene swapping operators like single point crossover and uniform crossover (defined in Appendix B) are expecting that a partial decomposition of the problem exists (which does occur to some extent in many problems). However, these operators also expect that the problem can be separated along the specified dimensions of the parameter space (e.g. without the need for linear transformation) which is less often the case. These operators also expect to have access to a population that adequately samples each of the sub-problems or so-called building blocks.

Another good example are search operators with hill-climbing characteristics. These operators expect that the fitness landscape will be somewhat smooth in the region in which the population is distributed in parameter space. One example is Wright's heuristic crossover [53] which generates solutions by linear interpolation between two parents or extended line crossover [46], [54] which generates solutions by linear extrapolation (both defined in Appendix B).

**Exploitation and Exploration**: Another important search operator property is the extent to which a search operator creates offspring that are biased to reflect the genetic material of the parents. In discrete spaces this can be approximated by counting how many of the genes in the final offspring are identical to the genes in the parents. In continuous spaces, such an assessment can be more difficult to make. Search operators which create offspring that largely reflect the parents are often labeled as exploitive. This label is also used for operators that are biased to predominately reflect the features of the more fit parent. In either case, the operators are "exploiting" a landscape feature that is common in many optimization problems, namely that similarities in genotype tend to correspond with similarities in phenotype (i.e. correlated landscapes).

The opposite of this is exploration which generally is used to describe offspring that are different from their parents. However, it is worth pointing out that this definition of exploration does not mean that an algorithm is necessarily capable of exploring new regions of the solution space. From a population perspective, what really defines exploration is the ability to create genetic material that is not only different from the parents, but is also different from other individuals in the population. Moving to a global perspective, an accurate definition of exploration should actually change based on the actions and history of the search process. However, because EA is a memory-less search process, this latter definition of exploration can not be measured or enforced meaning that search by an EA can become trapped in regions of parameter space for extended periods of time.

**Stochasticity**: The execution of most search operators involves a random variable whose value is drawn from some predefined distribution. This allows the operator to display a range of behaviors and greatly reduces the chances that the same inputs (i.e. parents) will generate the same output (i.e. offspring). This can potentially help to improve the robustness of a search process compared to deterministic operators which always give the same output when given identical inputs.

#### 2.3.4.2 Multiple Operators

Another option for improving the robustness of a search process is to introduce multiple search operators, each containing a unique search bias. Recent studies have indicated that the presence of multiple operators can help improve general performance of Evolutionary Algorithms [55], [56].

On the other hand, the addition of multiple search biases (or the addition of stochastic components to a search operator) are only advised if it is not possible to determine what the most effective search bias is for a given problem. If such a bias can be obtained or learned then the algorithm will perform substantially better on that problem. As a simple example, the search bias obtained from using a gradient based search operator should be greatly preferred over a more robust operator like random search when searching a smooth unimodal fitness landscape.

However, for many complex problems, it is expected that one particular search bias may be insufficient for effectively searching throughout the entire fitness landscape. In these conditions, multiple search operators may be more effective. Since EA is often applied to complex problems with poorly understood fitness landscapes, it is expected that Evolutionary Algorithms should generally be designed using multiple search operators.

#### 2.3.4.3 Search Operator Probabilities

To develop an effective search bias in an EA design, it is necessary to select a set of search operators for traversing the fitness landscape as well as select the usage probabilities for executing those operators.[3] Since probability parameters can take on values equal to (or close to) zero, the task of selecting appropriate search operators and tuning the probability parameters can be thought of as similar tasks.

Setting these parameters is often done by trial and error or by using an efficient design of experiments. However, instead of running the algorithm many times in order to establish an effective EA design, it is worth considering whether an appropriate search bias can be learned for a problem while the algorithm is being run. One promising option is to develop mechanisms that allow the probability parameter settings to adapt to the environment so that the overall search bias reflects what has so far been useful for traversing that particular fitness landscape.

[3] This can be more or less important depending on the amount of effort given to parameter encoding, which is a complementary aspect of EA design that alters the fitness landscape.

Notice that in this case, the adapted parameter settings used in the EA design are no longer generally robust but instead become specialized for the particular problem being solved. Since adapting and tuning search operator probabilities is a major topic of investigation in this thesis, a more detailed review will be presented in Chapter 3.

#### 2.3.4.4 Local Search and Expert Search Hybrids

Search operators can also involve more than one objective function evaluation. Such search operators are generally classified as local search or hill climbing methods and are designed to exploit local features of the fitness landscape. These methods often resemble classic directed search methods, gradient-based search methods or use expert knowledge in order to intelligently select new solutions to evaluate.

Local search operators are implemented in a lot of different ways depending on their purpose. In some cases they are used only on the best solution at the end of an EA run as a means to fine tune the final solution more quickly than is otherwise possible using standard EA search operators. Other times local search is used on all individuals throughout an EA run with the intention of modifying the fitness landscape from the perspective of the rest of the algorithm. In this case, the local search operators are sometimes used to only modify the phenotype (i.e. Baldwinian Evolution) instead of altering both the genotype and phenotype (i.e. Lamarckian Evolution). EA designs that are hybridized with local search operators are often referred to as Memetic Algorithms and are reviewed in [57].

### 2.3.5 Constraint Handling

Many real-world optimization problems require a set of constraints to be placed on the parameters being optimized. Constraints can be simple bounds on the values a parameter is allowed to take but they can also be complex nonlinear relationships between multiple parameters which can fragment the feasible solution space. Finding feasible solutions can be a challenging problem in itself and so the manner in which feasibility is treated is crucial to the effectiveness and efficiency of an optimization algorithm. In this section, several constraint handling techniques are presented and discussed.

#### 2.3.5.1 Rejection of infeasible Solutions

A rather naïve approach to constraint handling in non-convex feasible spaces is to treat constraints as they are typically treated in linear, convex problems; that is, to strictly enforce the constraints by requiring feasibility for every solution generated. For cases where it is not possible to directly solve the system of nonlinear constraints, this approach can result in a significant computational burden.

#### 2.3.5.2 Constraint Handling by Repair

Constraint handling by repair involves the development of a procedure for turning infeasible solutions into feasible solutions and can sometimes be as difficult to solve as the original problem. In some instances where expert knowledge is available, repairing infeasible solutions can be relatively straightforward and prove quite useful.

#### 2.3.5.3 Penalty Functions

The most common and popular method for handling constraints is to incorporate penalty functions into the objective function. Solutions that violate one or more constraints will be penalized by altering their objective function value so that it represents a less fit solution.

A few alternatives to static penalty functions are also available such as creating a dynamic penalty function that responds to changes in the population [58] or one that changes by a fixed schedule [59]. It is also worth noting that constraints can also be treated as additional objectives [60] or as pseudo-objectives to an optimization problem [61]. A review of constraint handling techniques that have been used with Evolutionary Algorithms is provided in [62] and [63].

#### 2.3.5.4 Stochastic Ranking

Almost all constraint handling methods work by applying a selection pressure that aims to drive the population toward regions of the solution space that are both of high fitness and feasible. For most EA selection methods (other than proportional selection) the selection pressure is based on fitness rankings and not the relative difference between fitness values. Hence, the majority of constraint handling techniques (e.g. penalty functions) can be understood as altering selection pressure by altering the rankings of individuals.

However, as discussed in [64], the use of penalty functions is prone to over-dominance or under-dominance. In the case of over-dominance, the penalty is too strong and all feasible solutions are preferred over infeasible solutions. In the case of under-dominance, the penalty for infeasibility is not strong enough to impact the rankings of individuals such that the objective function value is the only driver of population dynamics. Selecting the appropriate penalty weights is not only hard but the optimal penalty is also likely to be dynamic due to the non-stationary distribution of fitness values and the non-stationary distribution of constraint violations within the population.

An effective alternative is provided by the Stochastic Ranking method of Runarsson and Yao [64] and is used throughout this thesis. Stochastic Ranking works by ranking population members using a stochastic sorting procedure that considers both the objective function and constraint violations. A pseudocode for Stochastic Ranking is provided in Figure 2-2. The decision to swap adjacent individuals (when at least one is infeasible) occurs based on the objective function with a probability $P_f$ and otherwise is based on the extent of constraint violation. As $P_f \rightarrow 0$, the ranking of population members becomes dominated by the goal of attaining feasibility and as $P_f \rightarrow 1$, ranking becomes completely based on the objective function. The parameter $P_f$ allows for direct control over the extent of ranking changes that occur within a population due to infeasibility. Hence, this approach to constraint handling eliminates any problems with fitness scaling that plague penalty function methods (e.g. over-dominance and under-dominance).

```
For j = 1 To N
   For k = 1 To λ - 1
      sample u ∈ U(0,1)
      If (φk = φk+1 = 0) or (u < Pf) Then
         If (Fk > Fk+1 ) Then
            swap(k, k+1)
         End If
      Else
         If (φk > φk+1) Then
            swap(k, k+1)
         End If
      End If
   Next k
   If (no swap) Then Exit For
Next j
```

**Figure 2-2 Pseudocode for Stochastic Ranking procedure where U(0,1) is a uniform random number generator, $\varphi_k$ and $F_k$ are the total constraint violation and objective function value (resp.) for individual *k*, *N* is the number of sweeps through the population, $\lambda$ is the population size, and $P_f$ is the probability that infeasible solutions will be evaluated based on objective function values. The original description provided in [64] recommends $N \geq \lambda$ and $P_f$ = 0.45.**

It is worth mentioning that approaches have also been proposed in [65] that can deal with the issue of over-dominance and under-dominance when using penalty functions. In this case, mock competitions take place between members in an EA population in order to determine how large the penalty function must be in order to balance the competing forces from the objective function and the penalty function.

### 2.3.6 Parameter Encoding

Genetic encoding (also referred to as parameter encoding) is a term that is used to describe how solutions are represented in the chromosome. Genetic encoding methods are generally classified as either direct or indirect encodings. With direct encoding, each gene in the chromosome is a parameter that is directly used without alteration for objective function evaluation. In other words, binary parameters of the optimization problem are represented by binary valued genes, integer parameters are presented by integer valued genes, and continuous parameters are represented by floating point numbers, also called real-coded genes. Since many design problems involve continuous variables, real number representation is very common in EA research. A review and analysis of real-coded GAs (RCGA) is provided in [66].

Indirect encoding, on the other hand, involves a mapping process where genes must be transformed or processed in some fashion before fitness evaluation can occur. A simple indirect encoding scheme is provided in the canonical GA. Here the chromosome is represented by a binary string which is broken up into *n* equal length segments corresponding to *n* genes. Each binary segment is decoded to an integer value which is then rescaled to fit within the boundary constraints of the corresponding parameter.

Although arguments based on the Schema Theorem [36] have been given for preferring the binary encoding scheme of the canonical GA, direct encoding is more commonly used in EA designs today. However, some specialized indirect encoding schemes have been successfully applied such as that seen in Genetic Programming [67], [68] and in the evolution of artificial neural networks [69], [70], [71]. Similar success has yet to be seen for parameter optimization problems, however some interesting studies have occurred in recent years [72], [73].

#### 2.3.6.1 Gene Expression Research

Indirect gene encoding can be seen as an analogue to the genotype-phenotype mapping process in living systems and is an important open topic in EA research. Indirect gene encodings are of great importance because they can influence features of the fitness landscape potentially making a problem more or less difficult for a particular algorithm to search.

**Observations from Nature:** Several details of the genotype-phenotype mapping process in nature have been uncovered over the years and this should help to guide future directions of artificial gene expression research. Some features of the mapping process are discussed below however this is not meant to be an exhaustive review of the topic.

One important feature of the mapping process is the dominating presence of canalizing functions. This ubiquitous feature acts to dampen perturbations (e.g. from the environment) to the genotype-phenotype mapping process allowing it to quickly return to its intended dynamical trajectory. Canalization can also be understood as an important source of dynamical robustness. The term dynamical robustness is used in reference to the stability of phenotypic expression in the face of environmental perturbations. Evidence of dynamical robustness is seen for example in the yeast cell-cycle [74]. Artificial models have also displayed this property to some degree as seen for instance in [75].

A feature that is similar to dynamical robustness is mutational robustness which is also known as genetic neutrality. Mutational robustness can be observed in the fitness landscape of natural evolving systems and is created (in part) by a many to one mapping from genotype to phenotype. There is evidence that high levels of neutrality are present in natural evolution and this feature is believed to have a dramatic impact on evolutionary dynamics as was first theorized by Kimura [76]. Most studies of neutrality have so far focused on RNA folding as seen for instance in [77], [78].

It has also been proposed in [79] that with sufficient neutrality, neutral networks of single point mutations can percolate throughout genotype space meaning that a large portion of the space can be reached without requiring changes in fitness. Such landscape features are less likely to force population convergence to a static region of genotype space and could play an important role in maintaining population diversity as well as improving evolvability. In fact, they suggest in [80] that the presence of neutral networks can cause entropic barriers to replace fitness barriers meaning that adaptive improvements become less a question of "if" and more a question of "when".

It is worth mentioning that key features of the mapping process such as canalizing functions were shown to be easily created in Boolean networks [81] however similar mechanisms have not yet been considered in any indirect encoding schemes for EA. Also in [82], it was found that genetic robustness or neutrality is prevalent in some classes of distributed dynamical systems.

### 2.3.7 Interaction Constraints in EA Populations

Traditionally, EA population dynamics occur without restrictions or constraints on which individuals in the population can interact (e.g. through selection and reproduction). As a consequence, the population is tightly coupled and can become stuck or stalled for many fitness landscapes of interest. In order to provide for a more robust and parallel search process, restrictions in competition and/or mating are sometimes used.

This section reviews a number of advanced aspects of EA design which can be grouped under heading of interaction constraints within EA populations. The history and intent of the methods discussed below are often very different but they share a commonality that makes them an important development in EA research.

Each of the topics reviewed below introduces a new form of diversity or heterogeneity into an Evolutionary Algorithm. In many cases, these changes to the algorithm have resulted in substantial improvements in the robustness of EA performance. In addition to issues of diversity in genotypes, these methods also offer other types of diversity such as diversity in phenotypes, ages, genetic operators, and selection pressures. In some cases, the age-old exploration-exploitation tradeoff no longer applies due to the fact that a range of exploitive and explorative behaviors can now be displayed within a single system.

#### 2.3.7.1 Crowding and Niche Preserving Methods

One set of methods for restricting interactions in an EA are crowding methods which are generally thought of as a subclass of niching methods. Crowding methods work by forcing individuals to compete for survival and/or reproduce with others in the population that are similar. Similarity is generally defined based on the genotype or phenotype however historical (i.e. genealogical) similarity is also sometimes used. An example of the later would be restricting an offspring to only compete with its parents. A number of crowding methods have been developed including the original “standard crowding” proposed in [83], Deterministic Crowding [84], restricted tournament selection [85] and probabilistic crowding [86].

Deterministic Crowding (DC) is an interesting extension to standard crowding in that the method guarantees that each individual in the population, or a direct descendent of the individual, will survive to the next generation. From a genealogical perspective, this is an interesting change to the algorithm because stochastic effects are eliminated which otherwise would cause a continual loss of lineages from the population.[4] The result is a substantially parallel search process. Pseudo code for DC is presented in Figure 2-3.

```
For each generation
        Do N/2 times
                Select 2 parents, p1 and p2 (randomly, no replacement)
                Create 2 offspring, c1 and c2
                If  (d(p1, c1) + d(p2, c2)) < (d(p1, c2) + d(p2, c1))
                        If  f(c1) > f(p1)  replace p1 with c1
                        If  f(c2) > f(p2)  replace p2 with c2
                Else
                        If  f(c2) > f(p1)  replace p1 with c2
                        If  f(c1) > f(p2)  replace p2 with c1
                End If
        Loop
Next
```

**Figure 2-3 Pseudo code for Deterministic Crowding. Distance (either genotype or phenotype) is given as *d*(), fitness is given as *f*(), and *N* is the population size.**

The more general classification of niching is used to described strategies which allow an EA population to converge on multiple optima but do not necessarily involve a clear restriction of interactions within an EA population. For instance, Sharing Methods [87] are a form of globally-controlled niching where individuals are forced to share their fitness with other individuals based on distances in genotype (or phenotype) space. Because an intimate knowledge of the fitness landscape is needed to appropriately establish the correct sharing strategy and parameter settings, sharing methods are not easily used. Other niching methods also exist such as Clearing which was proposed in [88] and is similar to Fitness Sharing.

#### 2.3.7.2 Spatially Distributed Populations

Another approach to restricting interactions in an EA is to define the population on a graph so that operators such as selection and reproduction are only able act within localized regions defined by the graph topology. EA designs where the population is spatially distributed in some manner are referred to as distributed Evolutionary Algorithms (dEA).

---

[4] This is not to say that population convergence can no longer occur. Convergence is also be driven by search operators.

There are two general approaches to introducing spatial restrictions in an EA population which are referred to in this thesis as island models and network models.[5] Reviews on distributed EA designs as well as physically parallel implementations of Evolutionary Algorithms are provided in [89] and [90].

**Island Models:** Island models work by breaking the EA population into subpopulations or islands, with the dynamics of each subpopulation loosely coupled to the others. With Island Models, genetic operators can only be used between individuals within the same island. Occasionally, individuals are selected to move to a new island thereby allowing one island to influence the dynamics of another. A directed graph topology is usually established for the islands so that only specified islands can pass individuals to other specified islands. Some of the earliest work on island model EA populations was conducted in [91].

**Network Models**: With network models, the population is defined on a graph where each node represents an individual in the population. Network models are almost exclusively defined on a cellular grid and are often referred to as cellular Genetic Algorithms (cGA). These distributed EA designs work by having genetic operators such as selection and reproduction restricted to occur within local neighborhoods on the network. Some of the earliest work on network models for EA populations was conducted in [92] and [93].

The network topology is known to significantly impact the behavior and performance of a cGA as was demonstrated in [94]. Since population topology and its impact on EA behavior is a major topic of investigation in this thesis, a more detailed review will be presented in Chapter 5.

Population updating strategies also can influence the overall dynamics of the system as well as the selection pressure as seen in [95] and [96]. It is also worth noting that other distributed dynamical systems have displayed sensitivity to the population updating strategy as seen for instance in the related field of complex systems research [97], [98].

---

[5] In practice, combinations of the two classes are common and are referred to as Hierarchical Evolutionary Algorithms.

An example of a cGA with synchronous updating is provided in the pseudocode below. First, a population *P*() of size *N* is defined on a graph (usually a 1-D or 2-D grid). For a single generation, each cell is subjected to standard genetic operators. For a given cell *i*, parents are selected from within its neighborhood, search operators are selected, and an offspring is created and evaluated. The better fit between the offspring and *P*(*i*) is stored in *Temp*(*i*) until all *N* cells are calculated. The grid is then updated (synchronously) for the next generation by replacing *P*() with *Temp*(). This process repeats until some stopping criteria is reached.

```
Initialize P()
Evaluate P()
Initialize Population Topology
Do
        For i=1 to N
                Select Parents from Neighborhood(P(i))
                Select Search Operators
                Create and Evaluate offspring
                Temp(i) = Best_of(offspring, P(i))
        Next i
        Replace P() with Temp()
Loop until termination criteria
```

**Figure 2-4 Pseudocode of a synchronous cGA**

#### 2.3.7.3 Other Restrictions

**Age Restrictions:** Other mechanisms for restricting interactions have also been considered such as age-based restrictions which constrain interactions to only occur between individuals of similar age. The age in this case refers to the total age of a search path such that offspring inherit the age of their parents + 1. An example of this approach is seen in the Age-Layered Population Structure (ALPS) presented by Hornby in [99]. Hornby has found that age restrictions can allow for an effective utilization of new genetic material when it is introduced to an EA population.

**Environmental Restrictions:** As individuals in an EA population continue to internalize more and more algorithmic features that originally were defined globally, more interaction restrictions become possible. One option is to create heterogeneous island models where individuals are grouped based on similarity of parameter space granularity, search operator type, or selection pressure [100]. Classification of an island's exploitive or explorative character can then be used to restrict the flow of information between islands.

### 2.3.8 Performance Metrics and Analysis

In optimization research, there have been intermittent efforts aimed at developing standards for performance evaluation (e.g. see [101] and references therein). This section reviews several ways that performance can be measured in Evolutionary Algorithms. The discussion is restricted to only address issues that arise within the particular experiments conducted in this thesis and so some context and background is necessary. A review of other methods for analyzing the behavior and performance of an EA is available in [102].

The discussion of performance metrics changes depending on whether *a priori* knowledge is available about the problem such as knowledge of optimal genotype(s) or phenotype(s). Here it is assumed, as is often the case for many real world optimization problems, that no *a priori* knowledge is provided. As a consequence, it is assumed that performance can not be measured by how close an algorithm gets to reaching the optimal solution, how fast it approaches the optimal solution, or how often it finds the optimal solution. Furthermore, it is assumed that only one best solution exists for a problem and this discussion neglects any additional considerations of tradeoff surfaces which occur when dealing with multiple objectives.

#### 2.3.8.1 Time Dependency

Developing a useful performance metric requires careful treatment of an important tradeoff between short-term and long-term performance. It is of general interest in optimization research to understand how an algorithm performs over different time scales. Along with classification of the problem being searched, knowing the performance at different time scales also provides clues as to the other types of problems or conditions where the algorithm may prove useful.

As a consequence, algorithm performance is often represented as a function of time or computational effort. Time is typically measured as the number of function evaluations (or some multiple, e.g. generations) with the implied and generally valid assumption that computational costs not associated with objective function evaluation are negligible.

A common mistake when presenting results is to only present performance at a single point in time (namely the time when the algorithm stops running). This neglects performance at different timescales and introduces a significant bias into any conclusions drawn from these results. Although a single performance measure is sometimes more satisfying to the reader, it does not provide a good sense of the actual usefulness of an algorithm.

#### 2.3.8.2 Defining Performance

For a given instance in time, it is necessary to define a metric that is able to capture the salient details of algorithm performance. If an Evolutionary Algorithm performed exactly the same way every time it was run then its performance could be defined by the best solution found as a function of computational effort. However, Evolutionary Algorithms are a stochastic search method and their performance is sensitive to initial conditions. To obtain a measure of expected algorithm performance requires a sampling of experimental replicates to be taken with different initial conditions. Hence for a given instance in time, it is necessary to develop a performance metric which measures some group property of a sample of solution quality values taken from a set of experimental replicates.

Most experimental results are presented by simply comparing the mean or median solution quality between different optimization algorithms. This is particularly useful for comparing results that are stated in different publications however this is not the best way to determine which algorithm within a set of experiments has superior performance.

Making comparisons using a group statistic like the median does not take into consideration the other properties of the sample distribution (i.e. moments). This becomes particularly relevant when the distributions deviate from normality or the sample variance is large (both common occurrences in performance distributions of EA solution quality results). This bias is rarely (if ever) acknowledged in the presentation of EA experimental results.

For making statements about the superiority of one algorithm over another, a suitable alternative is to compare algorithms based on the ranking of solution quality values that are taken from a set of experimental runs. Taking the median ranking of an algorithm's solution quality (with each run of an algorithm ranked against the other algorithm's tested) provides a robust measure of how strong an algorithm performed compared to the others. For even better comparisons, one can use rank-based statistical tests (e.g. the Mann-Whitney U Test) to determine the confidence level for stating that one algorithm is superior to another based on ranking. This approach was used for example in [103] and [104].

The experimental results in this thesis have been presented using several of the approaches described above so that the reader can obtain a well-rounded picture of algorithm performance. Some performance metrics focus on final algorithm performance while others present performance as a function of time (e.g. with performance profiles). Statistical tests are also used in order to gain a sense of which algorithms are superior to others and the confidence with which such statements can be made.

### 2.3.9 Uses and Applications of EA

Although Evolutionary Algorithms provide a useful tool for studying natural evolution, the purpose of this research, and most EC research, is towards its application in solving optimization problems.

#### 2.3.9.1 When EA is used

There is no indisputable set of conditions that dictate when an Evolutionary Algorithm should be applied to an optimization problem however some fairly clear guidelines can be established.

**Benefits:** Evolutionary Algorithms are often effective in poorly defined problems where little is known about the fitness landscape. They are also effective for problems with substantial levels of noise or other sources of uncertainty [105] and for problems containing a significant level of parameter epistasis. Evolutionary Algorithms do not require gradient information which also makes them useful for non-differentiable problems.

Evolutionary Algorithms are effective on many problems because they allow for a global search through parameter space while exploiting basic landscape features (e.g. partial decomposition) that are present in many problems. Their stochastic nature also provides an inherent robustness to the search process by introducing a search bias that is expected to be true on average but does not have to be strictly true for the algorithm to perform well. Another related benefit is their capacity to allow for some degree of parallel search to take place. A population of solutions allows the algorithm to search in multiple directions at a time and reduces its chances of becoming trapped or stuck due to the presence of local optima. A population of solutions can also help the algorithm deal with fitness functions that have multiple objectives [106] or dynamic objectives [105]. They are also less sensitive to numerical errors compared to gradient-based and direct search methods [107].

**Drawbacks:** Although there are many benefits to using an EA, some cautionary notes are also warranted. One drawback to using an EA is that it must be tailored to fit a specific problem. Although most design issues do not require expert knowledge of the fitness landscape, some expert knowledge is required in order to design search operators that match well with the problem being solved. Furthermore, this aspect of EA design is crucial to the behavior and performance of the final algorithm. The remaining aspects of EA design largely consist of establishing the correct parameter settings. Some have claimed that the non-intuitive nature of setting these parameters can make Evolutionary Algorithms hard to implement in practice (by non-experts).

Another drawback of EA is that they generally do not scale well and so are rarely used on problems containing a large number of parameters (e.g. 1000+). The stochastic aspects which make EA robust also make it perform poorly if some regularity within the fitness landscape can be exploited (but is neglected during EA design/hybridization efforts). Even the slightest improvement in search bias can make huge differences in performance as the scale of a problem increases. A common source of regularity is the presence of smoothness at different scales in a fitness landscape. When such approximately smooth features are present, a more directed search process can sometimes be much more effective. On the other hand, it is quite common to integrate local search mechanisms or expert knowledge into an EA design which can help to address this drawback of Evolutionary Algorithms.

Similarly, EA has generally not been used on problems where computational resources only allow for a small number of function evaluations. Since EA is primarily designed to be a global search heuristic, it tends to have relatively poor performance over short time scales compared to deterministic methods. However, it is worth noting that computational costs are steadily becoming less of an issue when considering whether to use an EA. This is due to the increased availability of parallel computing resources as well as the development of more efficient surrogate models, both of which are conditions that EA is particularly suited to exploit.

In summary, Evolutionary Algorithms are not a panacea for solving all complex problems in the world however they do provide search characteristics that are important for solving many challenging problems. In general, EA should be treated as an adaptable framework for solving difficult problems instead of viewing them as a collection of ready-to-use algorithms [108].

#### 2.3.9.2 Where EA is used

Popular application domains for Evolutionary Algorithms include data mining, classification, scheduling, planning, and design. Although these are active areas of applied EC research, it is also worth pointing out that there is a diverse range of problems being solved by Evolutionary Algorithms in both academia and industry with new applications continually surfacing.

To get a sense of more specific applications where Evolutionary Algorithms are used, one can simply look at workshops that have taken place at international EC conferences over the years. For instance, taking a look at the Genetic and Evolutionary Computation Conference (GECCO), one will find workshops have dealt with applications related to the petroleum industry, medical applications, mechatronic design, fault tolerance, robotic vision, evolvable hardware machines, circuit design, sensor evolution, damage recovery in robots, modeling financial markets, structural design, and software testing.

The popularity of EA in a particular application domain can also indicated by the presence of application-specific EA surveys and reviews. A non-exhaustive list of such surveys includes EA applied to computer-aided molecular design [109], job-shop scheduling [110], project scheduling [1], aerospace problems [111], data mining and knowledge discovery [112], control systems engineering [113], [114], chemistry applications [115], macroeconomic models [116], and the modeling and control of combustion processes [117].

Ample evidence of its versatility can also be found in the hundreds of unique industry-driven applications where EA has been successfully applied. A sampling of applications include breast cancer detection [118], design of the world's fastest race car [119], optimizing schedules for bringing new products to market [3], designing pharmaceutical drugs [120], processing MRI brain images [121], and the optimization of Intensity-Modulated Radiation Therapy IMRT [122].

Evolutionary Algorithms can also be an important tool in academic research and have been used to build improved models of atomic force fields [123], improve interpretation of mass spectrometry data [124], and for the design of better molecular scale catalysts [125]. Many other examples can be found in the *Applications of Evolutionary Computation* book series [126].

# Chapter 3 Adaptation of EA design

Designing an Evolutionary Algorithm involves a number of activities, from the development of an appropriate genetic encoding and/or an appropriate set of search operators, to establishing the correct selection pressure. For some aspects of EA design, this involves the setting of a number of parameters which control aspects of EA behavior. Parameter tuning is an important task in EA design because the optimal parameter settings will vary from one problem to the next and the use of poor parameter settings can significantly impact algorithm performance.

This chapter focuses on ways in which robust EA search behavior can be attained by allowing traditionally static EA design parameters to adapt to their environment. Section 3.1 starts by reviewing adaptive control of EA design parameters. The review focuses particularly on a general framework for the adaptation of search operator usage probabilities.

Section 3.2 presents a new adaptive control procedure that is driven by empirical measures of an individual's impact on population dynamics. The method follows a principle of empirical search bias which is presented in this thesis as an alternative to the fitness driven search bias present in most optimization algorithms.

Through a preliminary analysis of population dynamics, an unexpected behavior is also uncovered in Evolutionary Algorithms. It is found that there is a surprising scarcity of individuals in an EA population that cause even moderate changes to population dynamics. Instead, most individuals actually have a negligible impact on the system. From tests using a number of experimental conditions, it is concluded that this is a fundamental property of EA dynamics such that most interactions between the EA and its environment are effectively neutral and non-informative.

Based on this conclusion and with the use of statistical arguments, the adaptive system is modified to filter out data from individuals with little impact on population dynamics (i.e.

neutral interactions). Experiments in Section 3.3, which are conducted on a number of artificial test functions and engineering design problems, indicate that the new adaptive control method is superior to several other adaptive methods in the literature.

## 3.1 Approaches to Adaptation: Literature Review

This section reviews past research on the tuning and adaptive control of EA design parameters. The section starts off with a justification for such research efforts and then presents a number of design parameters that can be adjusted in order to tune EA behavior. Different classes of parameter control techniques are also introduced, and finally, a framework is presented for the adaptive control of search operator usage probabilities.

### 3.1.1 Impetus for EA design adaptation research

Designing an Evolutionary Algorithm involves a number of activities, from the development of an appropriate genetic encoding and/or an appropriate set of search operators, to establishing the correct selection pressure. For some aspects of EA design, this involves setting a number of parameters which control different aspects of EA behavior. The number of parameters is potentially very large and each can influence the optimal setting of the others. Furthermore, there is no reason to assume that static EA parameter settings are optimal, especially considering the non-stationary environment caused by the search process.

Early attempts at resolving this problem were focused on determining the best static parameter settings for an EA [83], [127]. Over time, it has become well recognized that the best parameter settings depend on the problem being solved. Other attempts have focused on so-called *competent* EA designs where general guidelines are used to quickly determine good parameter settings. These approaches however generally rely on a number of simplifications and only apply to very specific algorithm designs such as the canonical form of the Genetic Algorithm (e.g. see [128], [129]).

Ironically, the problem of parameter setting for EA design actually defines an optimization problem. However the "optimization" of EA design is particularly difficult because we generally do not have the computational resources to test a large number of algorithm designs. With this in mind, one alternative is to carry out a systematic investigation using an efficient Design Of Experiments (DOE) for finding the best parameter settings. In cases where algorithm design features are not as easily parameterized, the problem becomes one of selecting among a set of optimization algorithms. A number of methods have been proposed for algorithm design automation or the hybridization of multiple search algorithms. These approaches often involve some iterated learning process as seen in [130], [131], [132], [133].

For many optimization problems, very little if any computational resources can be devoted to parameter tuning, or more generally, to algorithm design automation, since optimizing the search method would take much more time than optimizing the problem at hand. Under these circumstances, one promising option is to consider ways to adapt EA design at the same time it is being used to solve a problem. It is not expected that an optimal EA design can be created by this approach however its potential to improve EA performance and reduce human design efforts makes such research of practical interest for EA practitioners. Evolutionary Algorithms which automatically tune one or more parameters are referred to in this thesis as Adaptive Evolutionary Algorithms.

### 3.1.2 Adjustable Parameters

Parameters can be established for many aspects of EA design which are broken down in this review into two general categories that are present in most optimization algorithms. The first is that of Selection which determines how search paths are added or lost in a search process. EA parameters associated with selection that have been adapted in the past include population size [134], population structure [135], [136 2007)], selection pressure [137], [138], [139], and penalty weights for constraint handling [59], [58].

The second general category is that of search operations and simply deals with how the algorithm moves from one point in parameter space to another. EA parameters associated with search operations include adapting crossover points [140], [141], [142], [143] adapting mutation step sizes [144], [145], and adapting the probability of using different search operators [146], [147], [148], [149], [150], [103], [104]. Reviews of adaptation and parameter control in Evolutionary Algorithms are available in [151], [152], [153], [154].

**Adapting Search Operator Probabilities**: Adapting search operator probabilities is an aspect of EA design which has been extensively studied due to the tedious nature of tuning these parameters, particularly when considering more than two search operators (ten in this chapter). This parameter tuning problem is also of real practical interest since most EA designs used in industry incorporate search operators that are custom made for a particular problem. Deciding which of these operators to use and how often to use them presents a difficult design challenge that is often dealt with by trial and error. Due to its significance, this chapter will focus on applying adaptive methods for controlling search operator probabilities.

### 3.1.3 EA Parameter Control Techniques

Methods for adapting EA design parameters can be broken down into three general classes which are known as Deterministic, Self-Adaptive, and Supervisory (or Feedback) adaptive control methods. Each of these methods are briefly discussed below.

#### 3.1.3.1 Deterministic Methods

With deterministic parameter control methods, parameters are adjusted by an external fixed schedule or by an heuristic based on EA properties during runtime. Although deterministic methods are included as a class of adaptive methods, there is no actual algorithm response to its environment and so its classification as an adaptive method is rather hard to justify. The success of deterministic methods is likely to be highly problem-specific and even run-specific and the issue of defining the best deterministic method becomes a challenging problem possibly rivaling that of the original optimization problem. A well-known example of a deterministic adaptive method is the cooling schedule used in the Simulated Annealing algorithm [155].

#### 3.1.3.2 Self-Adaptive Methods

With self-adaptive methods, information is encoded in the individual population members thereby allowing adaptation to occur while the EA is running. Research into self-adaptive EA originated with the self-adaptive mutation rates in Evolution Strategies [42].

Self-adaptive methods have some characteristics which are highly favorable. Most importantly, this approach can potentially allow for a diverse range of algorithm behaviors to be present within a single population. Such diversity has the potential to provide an overall robustness to search behavior and appears to have important similarities to what takes place in natural evolution.

One of the main challenges with self-adaptive methods is that current EA designs tend to have difficulty sustaining diversity in the population. Instead, dominant individuals tend to spread their genetic material throughout the population which drives population convergence. This process happens quite quickly in many cases and search behaviors which exploit local landscape characteristics can exacerbate the problem. Since diversity is a precondition for adaptation, population convergence can limit the adaptive capacity of self-adaptive methods.

Although it not typically classified as a self-adaptive method, a similar approach that should be mentioned is that of competitive evolution which was first proposed in [147]. In this approach, a set of subpopulations are created and computational resources are distributed based on fitness and improvement rates within the subpopulations. The framework naturally allows for different EA design parameters in the subpopulations or even different optimization algorithms altogether. The difference between this and self-adaptive methods is that with competitive evolution, selection occurs on a larger scale so that multiple individuals (in a subpopulation) will have identical parameter settings. Somewhat similar ideas have also been used in the Hierarchical distributed Genetic Algorithm [156] where subpopulations are given different EA design parameters and communication between populations is restricted in an intelligent manner.

#### 3.1.3.3 Supervisory Methods

Supervisory adaptive methods (also referred to as feedback adaptation) use measurements taken of EA performance during runtime in order to adapt or control parameter settings. Unlike self-adaptation where the adaptive mechanism is coupled to the EA population, supervisory methods work usually at a higher level than individuals and are an external mechanism that is uncoupled from the search space of the optimization problem.

Supervisory adaptation is used in this thesis for EA parameter control, however no claims are made regarding its superiority to self-adaptive methods. On the one hand, self-adaptive methods are admittedly more "nature-inspired", however they can also be more sensitive to convergence problems and the associated loss of variability in EA populations. In the words of Darwin "without variability, nothing can be effected" which is to say that adaptation is simply not possible without variation [157].

### 3.1.4 Supervisory Adaptation of search operator probabilities

This section presents a detailed framework for supervisory adaptation which is presented within the context of adapting search operator probabilities. This framework is similar to others such as that presented in [158].

#### 3.1.4.1 Operator Quality

Given a set of $N_{ops}$ search operators with probabilities $P_i$, $i = (1,\ldots,N_{ops})$, an adaptive method has the task of setting $P$ in order to optimally control the usage rates of the operators. When an operator $i$ is used, a reward $R$ is returned. Since the environment is non-stationary during evolution, an estimate of the expected reward for each operator is only reliable over a short span of time. This is addressed by introducing the operator quality $Q$, which is defined such that past rewards influence operator quality by an extent that decays exponentially with time $t$ as defined in (3-1). The $\alpha$ term in this equation controls the memory of the adaptive method where $\alpha \rightarrow 1$ results in no memory but maximum adaptation and $\alpha \rightarrow 0$ results in maximum memory but no adaptation. The initial value for $Q$ is $Q(t\!=\!0)\!=\!0$. These quality values are then directly used to define search operator probabilities.

$$Q_i(t+1) = Q_i(t) + \alpha\left[R_i(t) - Q_i(t)\right] \tag{3-1}$$

### 3.1.4.2 Operator Probability Setting

When adapting $P$, it is necessary to place a lower bound on the probability value in order to prevent it from reaching zero. Allowing the probability to reach zero prevents future assessment of the operator which is not advised considering that the environment is non-stationary. To address this, a lower bound $P_{Min}$ is used to define the minimum probability of operator usage. Since only one operator is used at a time, the sum of all probabilities must equal one meaning an upper bound $P_{Max}$ can also be defined as $P_{Max} = 1\text{-}N_{Ops}*P_{Min}$. A standard approach for setting search operator probabilities is the *Probability Matching Strategy* defined in (3-2) which sets probability values to be proportional to operator quality. For all experimental conditions where search operator probability values are adapted in this thesis, it is assumed that all probabilities are initialized as $P_i(t\!=\!0) = 1/N_{ops}$.

**Probability Matching**

$$P_i(t+1) = P_{Min} + \left(1 - N_{Ops} \cdot P_{Min}\right)\frac{Q_i(t)}{\sum_{j=1}^{N_{Ops}} Q_j(t)} \tag{3-2}$$

An alternative to the probability matching strategy is the *Adaptive Pursuit Strategy* proposed in [158]. This method was developed based on a perceived weakness in the probability matching strategy due to its sensitivity to $Q$ scaling. As a simple example of this, Thierens considers a case of two search operators where each operator has a stationary reward value. He demonstrates that, as the difference between operator rewards becomes small, the difference between search operator probabilities also becomes small. In his argument, Thierens suggests that if one of the operators is superior, it should be strongly favored, even if the extent of its superiority is small. The adaptive pursuit strategy defined in (3-3) provides a straightforward method for allowing the distinction between search operators to be maximized. The extent that the best operator dominates the search process is controlled in equation (3-3) through the parameter $P_{Max}$ while the rate at which probability values change is controlled by *β*. In short, the adaptive pursuit strategy allows the best operator to reach a set maximum value regardless of its relative performance compared to other operators.

**Adaptive Pursuit**

$$P_i(t+1) = P_i(t) + \beta\left(P_{Set} - P_i(t)\right) \qquad \textbf{(3-3)}$$

$$P_{Set} = \begin{cases} P_{Max} & if \quad Q_i(t) = Max\left(Q_1(t),\ldots,Q_{N_{Ops}}(t)\right) \\ P_{Min} & else \end{cases}$$

Implicit in the rationale for the adaptive pursuit strategy is an assumption that no significant interaction exists between search operators and their impact on algorithm performance (i.e. no epistatic interaction). However, recent studies in [55] and [56] have suggested that significant beneficial interactions do take place between search operators meaning that some operator combinations are superior to any single operator used in isolation. As a result, this puts into doubt whether our goal should be to overwhelmingly favor a single best search operator as is intended with the adaptive pursuit strategy.

### 3.1.4.3 Defining Operator Rewards

In the description of supervisory adaptation presented thus far, the reward *R* is simply the result of an operator's interaction with the environment. However the adaptive framework presented in this thesis draws a distinction between interactions with the environment and the interpretation of those interactions as is shown graphically in Figure 3-1.

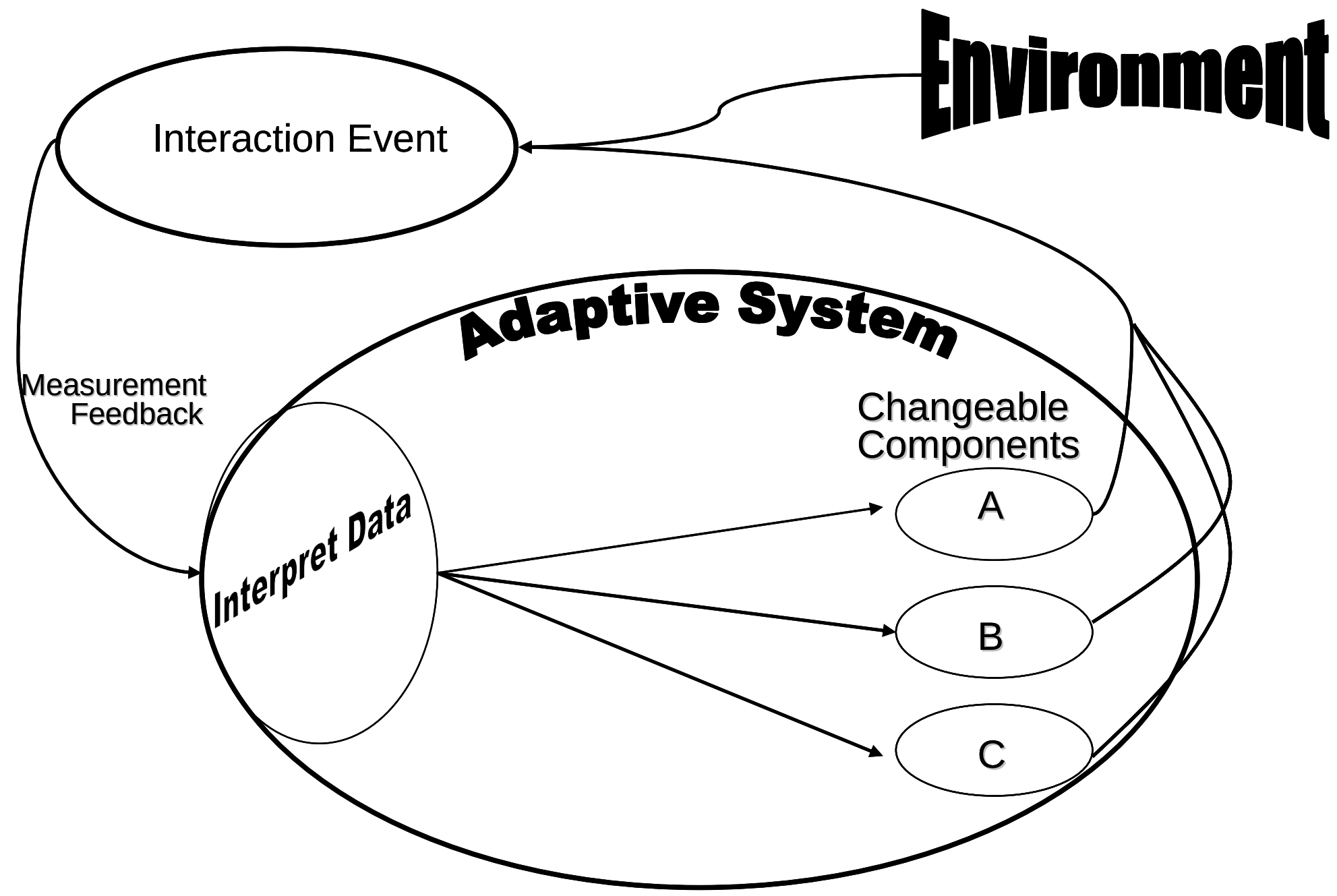

**Figure 3-1: Framework for a supervisory adaptive system. Here, an adaptive system receives measurement data as a result of its interactions with the environment. These measurements are then interpreted or assessed for relevance. Once interpreted, the data is then allowed to drive internal changes to the system. The mechanics for internal change are not shown in the figure but would consist of mechanisms such as the Quality function and the Probability Matching Strategy.**

Interaction events between an operator and its environment (i.e. offspring creation) are referred to simply as *events* and measurements of an event are labeled as $F$. To give meaning to an event, it must be interpreted within a particular context. Using this distinction between event measurements $F$ and their interpretation $I$, the reward for an operator $i$ at time $t$ can be defined in (3-4) as the average interpretation of a set of $M_i$ interactions with the environment. The $I_{archive}$ term in the equation simply stores the $I$ calculations from each operator.

$$R_i(t) = \frac{1}{M_i} \sum_{j=1}^{M_i} I_{archive}(i, j) \tag{3-4}$$

It is important to note that the reward at a particular time $t$ has now been redefined so that it represents multiple interactions between the adaptive system and its environment. Using this particular setup, each increment of time $t$ is now referred to as an adaptation cycle. The number of interactions that take place during an adaptation cycle is controlled by the adaptation cycle length $\tau$ which is the number of generations before operator probabilities are recalculated. The meaning of each of these terms within the overall adaptive framework is demonstrated in the pseudocode below.

```
Do
    'add standard genetic operators here
    Gen= Gen+1
    For each offspring
        i = offspring's search operator
        Calculate F (defined in Section 3.1.4.4)
        Calculate I (defined in Section 3.1.4.5)
        M_i = M_i + 1
        I_archive(i, M_i)= I
    Next offspring
    If Gen mod τ = 0 THEN
        t=t+1
        For each operator i
            Calculate R (defined in Section 3.1.4.3)
            Calculate Q (defined in Section 3.1.4.1)
            Calculate P (defined in Section 3.1.4.2)
            M_i = 0
        Next i
    End If
Loop until termination Criteria
```

The only calculation steps in the pseudocode above that have not yet been presented are for *F* and *I*. The next section briefly discusses possibilities for event measurements *F* while Section 3.1.4.5 describes interpretation methods *I* that have been used in the literature.

#### 3.1.4.4 Event Measurement

Measurements of search operator events typically consist of the fitness of the offspring that was created in an event. Fitness is generally based on the objective function value although fitness measurement can also take into account the feasibility of the offspring. Somewhat uncommon fitness measures based on the objective function are provided in [104].

Since practically all previous studies have used offspring fitness as the measurement of choice, event measurement *F* is assumed to be equivalent to the offspring fitness throughout this thesis. Section 3.2 is devoted to a new type of event measurement which is not based on standard measures of fitness.

### 3.1.4.5 Interpretation

The interpretation of search operator events involves defining an event measurement $F$ within a given context. As previously mentioned, this is then used to calculate the operator reward $R$ as defined in (3-4). It should also be mentioned that the interpretations methods, as described below, assume the problem being solved is a Maximization problem.

#### *3.1.4.5.1 Parent Context*

One common approach to interpreting a fitness measurement is to do so within the context of its parents. For instance, in $I_1$ the interpretation of the offspring measurement simply indicates (with a binary variable) whether the offspring was superior to one of its parents. Here it is assumed that the best parent is always chosen for comparison. Another option is to measure the magnitude of improvement between an offspring and its parent as seen in $I_2$.

$$I_1 = \begin{cases} 1 & if \quad F_{Offspring} > F_{Parent} \\ 0 & else \end{cases} \tag{3-5}$$

$$I_2 = F_{Offspring} - F_{Parent} \tag{3-6}$$

$$I_3 = Max\left(0, F_{Offspring} - F_{Parent}\right) \tag{3-7}$$

The interpretation $I_3$ equals $I_2$ if $I_2 > 0$ but otherwise is set to 0. The interpretation $I_3$ was first defined in [159] and is described in [160], as a combination of the probability of improvement and the expectation of improvement.

#### *3.1.4.5.2 Population Context*

It is also common to consider a measurement within the context of an individual's population. This is seen for example in $I_4$, which is an interpretation similar to $I_1$ except the context being considered is the median population fitness $F_{Median}$ instead of the parent fitness.

$$I_4 = \begin{cases} 1 & if \quad F_{Offspring} > F_{Median} \\ 0 & else \end{cases} \tag{3-8}$$

Also similar to $I_2$, interpretation $I_5$ considers the size of measurement improvement compared to the best individual in the population $F_{Best}$. However $I_5$ also scales the interpretation to take into account the distribution of measurements. This measurement was first introduced in [161].

$$I_5 = \frac{\left(F_{Offspring} - F_{Best}\right)}{\left(F_{Best} - F_{Median}\right)} \qquad \textbf{(3-9)}$$

Interpretations similar to $I_3$ have also been considered within a population context. For example, in [162], they created $I_6$ which is the same as $I_3$ except that $F_{Parent}$ is replaced with $F_{Best}$. Also, in [163] they created $I_7$ which is the same as $I_3$ except that $F_{Parent}$ is replaced with the $F$ measurement in the population that represents the 90$^{th}$ percentile $F_{90th}$ (i.e. $F_{90th}$ is the $F$ value that is greater than 90% of other $F$ values in the EA population).

$$I_6 = Max\left(0, F_{Offspring} - F_{Best}\right) \qquad \textbf{(3-10)}$$

$$I_7 = Max\left(0, F_{Offspring} - F_{90th}\right) \qquad \textbf{(3-11)}$$

Finally, another common interpretation is to simply rank an offspring's fitness $F$ within the EA population of size $N$ as seen in $I_8$.

$$I_8 = \sum_{i=1}^{N} \phi\left(Offspring, i\right) \qquad \textbf{(3-12)}$$

$$\phi\left(Offspring, i\right) = \begin{cases} 1 & if \quad F_{Offspring} > F_i \\ 0 & else \end{cases}$$

A number of past studies on the adaptation of search operator probabilities can be defined using the adaptive framework that has been laid out in this chapter. Several of these are presented in Table 3-1.

**Table 3-1 Partial list of methods that have been used to adapt search operator probabilities. Parameters that are not specified in the method are listed as (-), parameters that are not applicable are listed as (*) , and parameters that were varied in experiments are listed as the range of values tested. Parameter $\alpha$ in column two is the memory parameter given in (3-1), $\beta$ in column three is a parameter specific to the adaptive pursuit strategy and is defined by (3-3), $\tau$ in column four is the adaptation cycle length which defines the number of generations between the updating of search operator probabilities, $P_{Min}$ in column five is the lower bound on the allowed range of probability values, the "Event Measurement" in column six is described in Section 3.1.4.4, the "Interpretation" in column seven is the interpretation of event measurements as described in Section 3.1.4.5, and $N_{ops}$ in column eight is the number of search operators being adapted. The ETV event measurement and Outlier interpretation in the bottom two rows of the table are new event measurement and interpretation methods (resp.) proposed in this thesis and are described in Sections 3.2.2 and 3.2.4. "His. Credit" refers to Historical Credit Assignment which is described below in Section 3.1.4.6.**

| Reference | α | $\beta$ | τ | $P_{Min}$ | Event Measurement | Interpretation | $N_{ops}$ |
|---|---|---|---|---|---|---|---|
| [158] | 0.8 | 0.8 | 1 | 0.1 | - | - | - |
| [164] | * | * | 1 | | | His. Credit | |
| [160], [149] | 0.3 | * | 4 | 0.1 | $F$ | $I_3$ | 5 |
| [162] | 0.001-0.5 | * | 1 | 0 | $F$ | $I_6$ | 2 |
| [165] | * | * | 100/N | 0.01 | His. Credit | $I_4$ | 3 |
| [163] | 0.001-0.1 | * | 1 | - | $F$, His. Credit | $I_6$, $I_7$ | 5 |
| [104] | 0.5 | * | 20 | 0.02 | $F$ | $I_8$, Outlier | 10 |
| [103] | 0.5 | * | 20 | 0.02 | $F$, ETV | $I_8$, Outlier | 10 |

#### 3.1.4.6 Other Approaches

**Historical Credit Assignment:** An interesting alternative for defining operator rewards based on a principle of historical credit assignment is presented in [146] and also used in [164], [165]. In this adaptive method, each individual stores a search operator tree as shown in Figure 3-2. The operator tree records which search operators were used to create the ancestors of each individual. When a search operator event occurs, credit for the event is assigned backward to all search operators in the operator tree with the initial credit defined by $I_4$. To account for the diminished importance of past events, the actual credit a search operator receives is adjusted to be $\gamma^L * I_4$ where $L$ is the path length between the current event and the search operator receiving credit in the operator tree. The parameter $\gamma$ controls how quickly credit decays with distance in the operator tree.

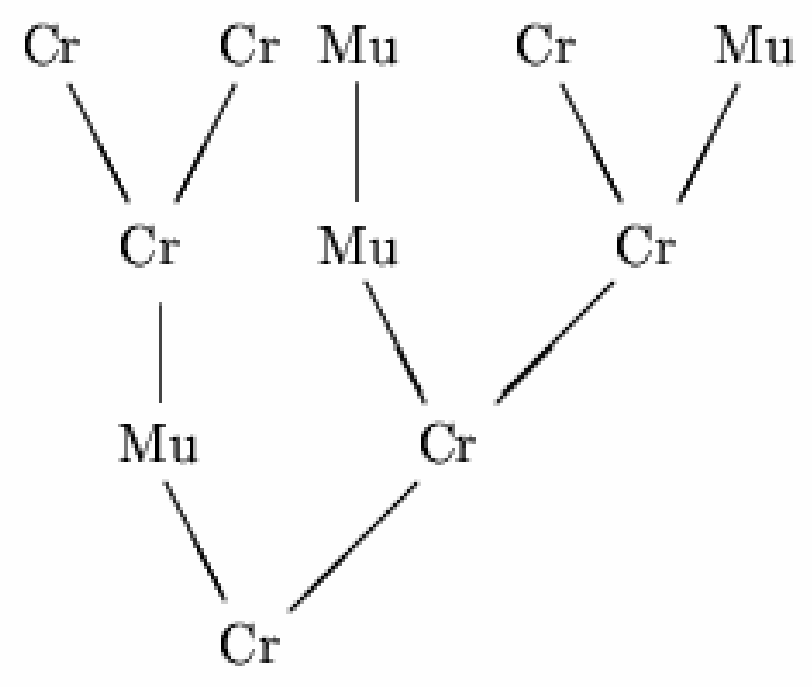

**Figure 3-2 Operator tree for an individual where only crossover (Cr) and mutation (Mu) events occur. The root node in the tree, Cr, is the search operator that was used to generate the individual that is storing the operator tree.**

Although this approach does not fit precisely within the "event measurement/ interpretation" adaptive framework used in this thesis, it represents the only method where events are measured based on the success of future offspring making this method quite unique. Interestingly enough, this method has a number of similarities to the new adaptive methods presented in the next section. It is important to emphasize however that the adaptive methods derived in this chapter were developed independently from the approach presented in [146]. Also, despite the similarities, there are a number of important differences in the approaches as will be seen as the new methods are introduced.

## 3.2 Measuring Population Dynamics for Adaptive Control

The following section presents a new type of event measurement that is notably distinct from the standard objective function value. Section 3.2.1 begins by clarifying why objective function values are a common form of event measurement used for driving parameter adaptation (and more generally used for guiding an optimization search process). Section 3.2.1 also provides some alternatives to fitness-based search including a newly proposed concept called *Empirical Search Bias*. The new event measurement is presented as an example of Empirical Search Bias in Section 3.2.2. Based on an analysis of the new measurement in Section 3.2.3, a new interpretation method is also developed which is presented in Section 3.2.4. The concepts of event measurement and measurement interpretation follow from the review in the last section and the framework outlined in Section 3.1.4.

### 3.2.1 Why is Objective Function a Standard measure for fitness?

As previously mentioned, adaptive methods typically use a measure of fitness based on an individual's objective function value in order to assess the merits of different search behaviors. To understand why objective function values are universally used as indicators of fitness requires an understanding of the assumptions implicit in any search process. The most important of these assumptions is referred to in this thesis as the *Hill-Climbing Assumption*.

#### 3.2.1.1 The Hill-Climbing Assumption

In order to search for an optimal solution to a problem, it is necessary to make assumptions about the fitness landscape of the problem being solved. One of the most common and successfully applied assumptions is that a solution's objective function value (Fitness) can approximate a solution's usefulness in searching for more fit solutions. Following this to its logical conclusion, this implies highly fit solutions will ultimately be useful in finding the optimal solution.[6] This also implies that a solution's reproductive worth (i.e. usefulness as a point to search from) and the solution objective function value are roughly equivalent measures. For unimodal landscapes, this assumption is often sufficient for guaranteeing an optimal solution will be found consistently and in a reasonable amount of time. However, for multimodal landscapes, the Hill-Climbing Assumption can fail to produce reliable or acceptable results.[7]

#### 3.2.1.2 Search Bias Assumption

An alternative approach is to look at solving the inverse problem which is that of optimizing search bias. For clarity, this will be called optimization based on the *Search Bias Assumption*. Here the goal is to find solutions and search mechanisms that are most likely to reach the optimal solution reliably and in a small number of steps. Instead of assigning credit to a highly fit solution, we look to assign credit to solutions that participate in finding high fitness solutions. The underlying assumption made here is that a solution that was helpful in finding good solutions has a chance of being helpful in finding even better solutions. Furthermore, we treat this assumption as if it can be successfully applied throughout the fitness landscape all the way up to the globally optimal solution. One obvious result of this approach is that a distinction is drawn between the reproductive value and the objective function value of a solution which makes this markedly distinct from optimization under the Hill-Climbing Assumption. The Search Bias Assumption is difficult to implement in practice although some options for doing so are proposed in [103].

---

[6] Similar arguments can also be applied to the 1st and 2nd derivatives of the objective function.

[7] In EA, most selection schemes involve relaxation of the Hill-Climbing Assumption. This involves treating the Hill-Climbing Assumption as being true in the average sense but not strictly true (i.e. a probability of it being true).

#### 3.2.1.3 Empirical Bias

The Hill-Climbing Assumption and the Search Bias Assumption represent opposite ends of a spectrum of possible bias for driving an optimization search process. On the one hand, the Hill-Climbing Assumption relies solely on the current states of the system in order to guide future search behaviors while the Search Bias Assumption places emphasis entirely on the initial conditions. In between is a realm where an intermediate reliance on history or search experience occurs (i.e. where history/experience partially guides the search process). This third option is referred to in this thesis as *Empirical Bias* and a proposal for implementing Empirical Bias is provided in the following sections.

For the implementation considered, the core of the EA design remains unchanged so that the overall search process is still driven by the Hill-Climbing Assumption. However, search operator usage rates are driven by empirical evidence of offspring importance as opposed to being driven by the fitness of these offspring.

For a population-based optimization algorithm like EA, an empirical measure of the importance of an individual can be obtained by measuring the individual's impact on population dynamics. Looking at population dynamics on a small timescale such as a single generation, an individual will only impact the population through competition for survival and/or competition to reproduce. However, if longer timescales are considered, we will find an individual's impact is largely a result of the survival and spread of its offspring.

The next section describes a procedure for measuring an individual's impact on population dynamics. This new event measurement, combined with a new interpretation method will be used to adapt search operator probabilities in the experimental work in this chapter.

### 3.2.2 Measuring Impact on Population Dynamics: The Event Takeover Value (ETV)

This section describes the *Event Takeover Value* (ETV) which is used for measuring an individual's impact on population dynamics. Throughout the discussion, the term *event* is used as before to describe the creation of a new individual. To help understand ETV, Figure 3-3 shows a directed graph which represents the family tree of an individual's lineage. Here different generations are indicated by positioning on the horizontal axis, nodes represent individuals created in a particular generation and the parents and offspring of an individual are indicated by connections to the left and right (resp.). Starting at the root node on the far left of Figure 3-3, one can observe how this individual's genetic material is able to spread through the population. At each generation, it is possible to count the number of individuals in the population that are historically linked to the root node. This can be thought of as an instantaneous measure of the individual's impact on population dynamics and is referred to as $ETV_{gen}$. A more detailed description of $ETV_{gen}$ is provided in the caption of Figure 3-3.

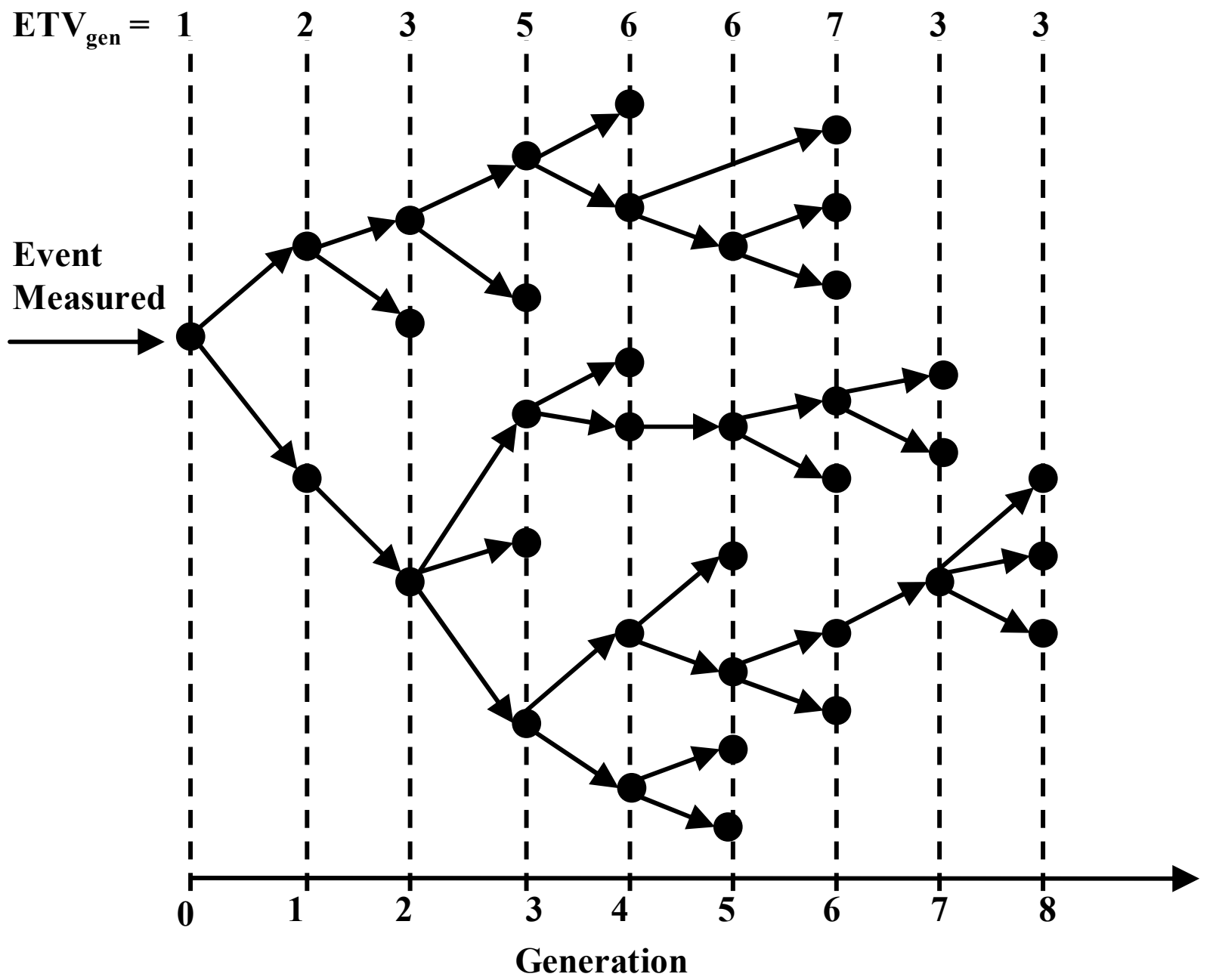

**Figure 3-3: Visualizing an individual's impact on population dynamics using genealogical graphs. An individual's impact for a given generation (horizontal axis) is defined as the number of paths leading from the measured node to the current generation. This is referred to as $ETV_{gen}$ and can be calculated for the "Event Measured" in the graph above by counting the number of nodes on the dotted vertical line for a given generation. As the population moves from one generation to the next, one can see that the number of individuals in the population that are descendants of the "Event Measured" will change with each new generation. In other words, the $ETV_{gen}$ value is dynamic. To clarify this point, $ETV_{gen}$ values are calculated for the "Event Measured" and are shown at the top of the graph. The maximum impact an event has on the population is the maximum $ETV_{gen}$ value that is observed. This graphical illustration assumes a generational population updating strategy such that an individual exists in a single generation only. This is done to simplify the illustration however other updating strategies could be used in which case some nodes would be stretched across multiple generations in the graph.**

Observing Figure 3-3, it appears that a reasonable calculation of an individual's impact on population dynamics would be to count the total number of descendants for a given individual. This is equivalent to summing up $ETV_{gen}$ for all generations where the individual's lineage remains alive. The problem with this measurement is that an individual's lineage occasionally is able to spread throughout the entire population so that the cumulative $ETV_{gen}$ value increases indefinitely. A useful alternative which is used in this thesis is to define ETV as the largest $ETV_{gen}$ value observed. This value naturally has an upper bound equal to the population size of the system. For the example given in Figure 3-3, the ETV value for the "Event Measured" would be ETV=7, which occurs in the sixth generation.

#### 3.2.2.1 Multiple Parents and Genetic Dominance

Figure 3-3 shows how an individual can impact population dynamics through the spread of its genetic material, however it does not consider the fact that offspring are often created from multiple parents. Also, when using multi-parent search operations, offspring tend to be genetically biased to be more similar to one parent than the other(s). An accurate measure of ETV should therefore account for the possibility of multiple parents as well as account for the possibility of dominance by one of the parents.

As an alternative to assigning a weighted importance to each of the parents, a dominant parent is chosen instead so that (for ETV calculation purposes) the offspring is seen as having only a single (dominant) parent. By using dominance, it is no longer necessary to address the issue of distributing credit among multiple parents meaning that Figure 3-3 is still a valid representation of the ETV measurement process. This also helps to simplify implementation of the ETV calculation steps as seen later.

Several ways for selecting the dominant parent have been tested including random selection, phenotypic similarity, and genotypic similarity between parents and offspring. In preliminary studies (results not shown), random selection resulted in mediocre EA performance as well as poor differentiation between search operator probabilities. Selecting the parent that was most genetically similar (by Normalized Euclidean Distance) to the offspring worked well while selecting the parent that was least genetically similar performed even more poorly than random selection. No significant difference in performance was observed between using genetic similarity and phenotypic similarity. In order to maintain consistency with the ETV measurement definition, genetic dominance is used in ETV calculations.

#### 3.2.2.2 Hitchhiking

Thus far, the ETV measurement implicitly assumes that an individual's impact on future dynamics does not degrade with the passage of time. However the stochastic nature of an EA makes this time dependency true and unavoidable. Addressing time dependence in credit assignment has previously been done using exponential decay functions in [146], [164], [165], and [103]. Another possible approach is to set a time window beyond which an individual's impact on population dynamics can no longer be measured.

Through careful study of the genealogical branching process, it has been found that certain branching structures can indicate exactly when confidence in the $ETV_{gen}$ measurement is lost. An example of these conditions is shown in Figure 3-4. Looking at the ancestors (i.e. nodes to the left) of the white node, it is noticed that all ancestors have the same $ETV_{gen}$ value and that this value is obtained solely due to their historical linkage to an important future event. Obtaining credit in this fashion is referred to in this thesis as *Genetic Hitchhiking*.

This phenomenon actually happens quite often. If an important event occurs, it will likely spread quickly throughout much of the population. However, all events prior to the important event also spread because they are historically linked. Care must be taken then to make sure an event has spread due to its own importance and not the importance of some later event. To account for this, $ETV_{gen}$ measurements of hitchhikers are disregarded.

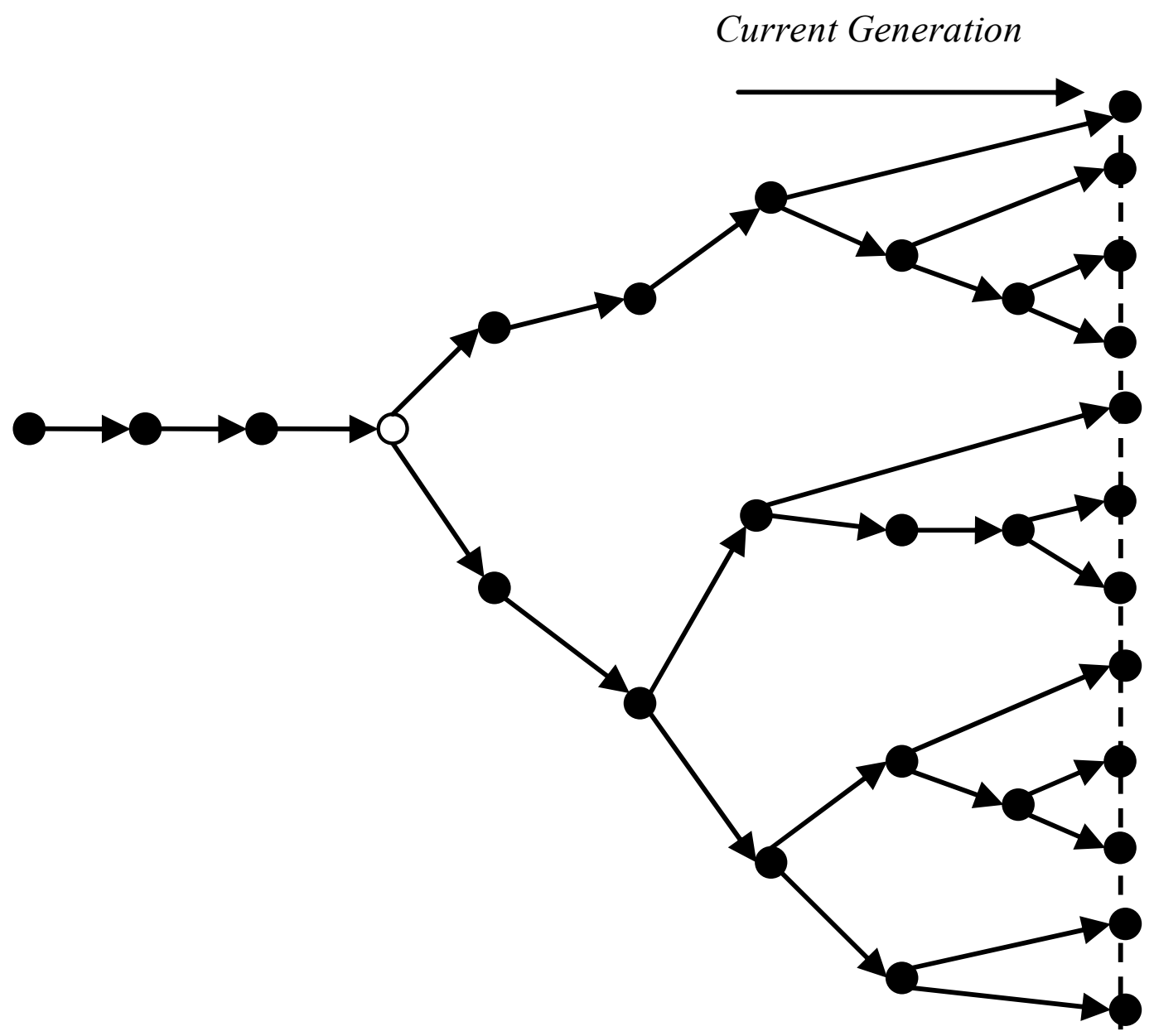

**Figure 3-4: Genetic Hitchhiking in EA population dynamics. Considering $ETV_{gen}$ measurements based on the current generation, one can easily see that all nodes to the left of the white node will have the same $ETV_{gen}$ value (i.e. they all have the same number of paths leading to the current population). However, these nodes are assigned their $ETV_{gen}$ values only because of a single important descendant (the white node). These linear structures in the genealogical branching process are a sign of genetic hitchhiking and can be seen in several different places in the graph above (seven genetic hitchhiking occurrences in total).**

### 3.2.2.3 ETV Calculation Procedure

To calculate ETV, a procedure is needed for recording genealogical information. The first step is to assign an *ID* to each event that uniquely identifies the offspring and indicates which search operator created it. Historical information in the form of these *ID* values is stored in each individual as an ordered list which represents the direct line of ancestry for that individual. An example of these ordered lists and their meaning within a genealogical tree is provided in Figure 3-5. When a new offspring is created, it inherits the historical records of the genetically dominant parent, and a new *ID* (representing the offspring) is added to the offspring's historical record.

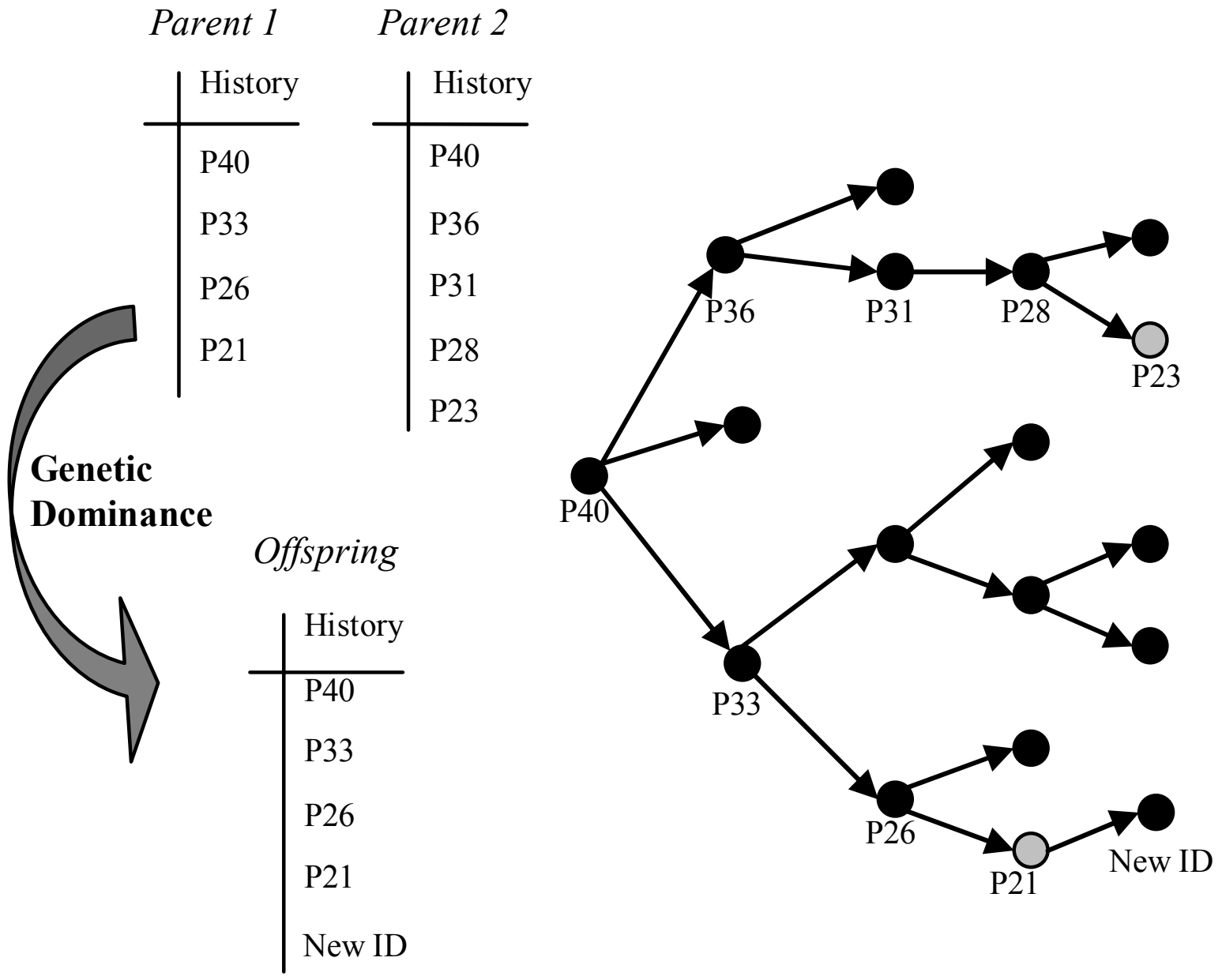

**Figure 3-5: Transfer of Historical Data. Each individual holds historical information in addition to genetic information. The historical information represents the direct line of ancestry for an individual. Examples of historical data lists are shown above for Parent 1 (*ID*=P21) and Parent 2 (*ID*=P23) and their meaning is demonstrated by the genealogical graph on the right. A new offspring only takes historical information from the parent that is genetically most similar (i.e. genetically dominant). In this example, Parent 1 is assumed to be the genetically dominant parent. In addition, the offspring creates a new ID to indicate its placement in the genealogical tree.**

By going over the historical records that are stored in the individuals in the current population and counting the number of times that the *ID* of an event is observed, the $ETV_{gen}$ for that event (and that generation) can be calculated. Given a maximum size $T_{obs}$ for the historical records list, an EA population size $N$, and individual population members $M$, the $ETV_{gen}$ measurement for event "*ID*" can be calculated using (3-13).

$$ETV_{gen}(ID) = \sum_{i=1}^{N} \sum_{j=1}^{T_{Obs}} \phi_{i,j} \qquad \textbf{(3-13)}$$

$$\phi_{i,j} = \begin{cases} 1 & if \quad ID = M_i(ID_j) \\ 0 & else \end{cases}$$

To check for genetic hitchhiking, the $ETV_{gen}$ value of an event must be compared with one of its offspring. If they are equal then the parent's $ETV_{gen}$ value is set to zero. Given two events $ID_1$ and $ID_2$, genetic hitchhiking can be defined by (3-14).

$$\begin{aligned} &IF\ (ID_1 = M_i(ID_j))\ AND\ (ID_2 = M_i(ID_{j-1})) \\ &AND\ (ETV_{gen}(ID_1) = ETV_{gen}(ID_2))\ THEN\ ETV_{gen}(ID_1) = 0 \end{aligned} \qquad \textbf{(3-14)}$$

The final step in the ETV calculation is to compare each $ETV_{gen}$ value with the archived ETV value. If $ETV_{gen}$ is larger than the archived ETV, then the ETV value is updated, otherwise the old value is retained. The ETV calculation for an event is completed when an event's $ETV_{gen}$ is found to be zero (i.e. a hitchhiker event).

#### 3.2.2.4 Computational Costs of ETV Calculation

The computational costs of the ETV calculations are reasonably small if properly implemented. These costs come primarily from i) the size of the "historical list" in each individual and ii) from the number of events that are being calculated at each generation (i.e. the size of the ETV archive).

**Historical List Size:** The size of the historical list $T_{obs}$ establishes the maximum number of ancestor events that can be stored in each individual in the population. If $T_{obs}$ is too small, it effectively reduces the amount of time that an ETV can be measured however if $T_{obs}$ is too large, it will negatively impact the computational costs of the procedure. The first step to improve computational efficiency was to determine how large $T_{obs}$ must be in order to calculate ETV. This was accomplished by running experiments with EA designs varying by population size $N$ ($N$=30 to $N$=400), selection pressures (binary tournament selection and random selection), and population updating (generational and steady state) on a random sampling of test functions taken from Table 3-2. $T_{obs}$ was set to 100, which is large enough to ensure ETV calculations are almost always finalized. For each event, the smallest $T_{obs}$ was recorded that would have allowed for the ETV to be calculated. Looking at the cumulative distribution of these values, it was found that 99.9% ($\pm$0.04) of all ETV calculations complete within $T_{obs}$=20. The completion time did not appear to be sensitive to any of the conditions varied in these experiments. For the 0.1% of ETV calculations that do not complete with $T_{obs}$=20, the measured ETV is expected to be an undervaluation for these events. However, from these tests it is clear that the required size of the historical lists is quite small ($T_{obs}$=20) and has little sensitivity to test conditions (including the population size).

**Computational Costs from ETV Archive:** At each generation, an event's ETV must be checked for a larger value and checked for evidence that the ETV calculation has completed. As a result, the computational costs will depend on the number of ETV that are actively being calculated at a single point in time.

In order to determine the computational cost from ETV, it was necessary to look at the average size of the ETV archive[8] as a function of time and as a function of population size. First looking at the time dependency, it was found that the ETV archive size always converges to a stable value. Focusing on these stable values, the ETV archive size was found to equal 11.4*N* with *N* being the population size.[9]

Also, to help understand ETV computational costs, experiments were run to determine the average number of generations needed to complete an ETV calculation. This was found to be 4.0 with no sensitivity to population size. From these tests, it was concluded that ETV computational costs scale linearly with population size.

#### 3.2.2.5 Related Research

Recently, genealogical graphs have also been used to help understand the dynamics in Artificial Life systems [166]. However, to the author's knowledge, no previous work on genealogical graphs has addressed the issue of genetic dominance in multi-parent reproduction nor has anyone previously used the concept of genetic hitchhiking when assessing the impact of an event in population dynamics.

[8] The archive holds the ETV value and *ID* number for all events held in the historical lists of the population

[9] Relationship between ETV archive size and *N* determined by linear regression ($R^2$ = 0.999) with five tests conducted over the range $20 \leq N \leq 400$.

### 3.2.3 ETV Analysis

#### 3.2.3.1 Fitness as a predictor of ETV

Considering that ETV measures an individual's impact on population dynamics, it would be useful to assess whether an individual's objective function is an accurate predictor of ETV.

**Experimental Setup:** To test this, EA runs were conducted where fitness-based ranking and ETV values are calculated and stored for each individual. To ensure the results were not sensitive to experimental conditions, it was necessary to test a variety of EA designs on a variety of test functions. A large number of ad hoc experiments have been conducted with EA designs varying by selection pressure, population size, and the number of search operators. Few noticeable distinctions were observed in these preliminary tests and so the results are only presented for the EA design described in Figure 3-6. Several test functions have also been considered in preliminary tests with results suggesting the relationship between ETV and fitness is sensitive to the fitness landscape, however this sensitively is rather low.

**Relationship between ETV and Rank:** As expected, the results shown in Figure 3-6 indicate that almost all individuals with a large impact on population dynamics (i.e. large ETV) are caused by individuals of high rank. However, while a trend exists between larger ranks and larger ETV, this does not mean that low ranking individuals never have a strong impact on population dynamics. Evidence for this is provided in the box plots in Figure 3-6 where one can see that ETV measurements rarely ever reach ETV > 5 and yet a number of low ranking individuals were able to obtain much larger ETV values.

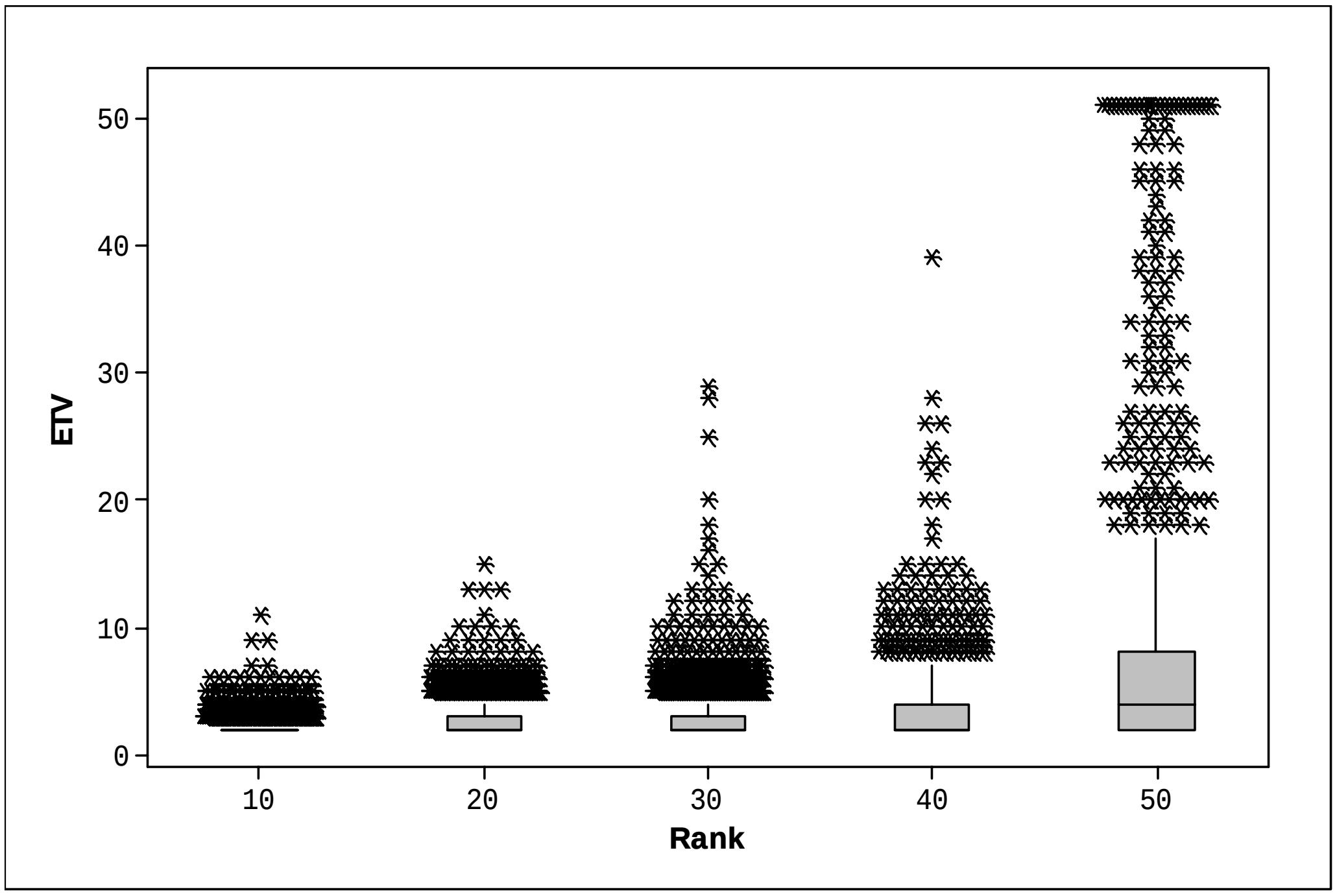

**Figure 3-6: Box plots of the size of an individual's impact on population dynamics (ETV) as a function of the individual's rank within the population where a rank of 1 represents the worst individual and a rank of *N* represents the best individual (based on objective function value). The data set was generated from a series of experiments involving a number of test functions listed in Appendix A. The EA used to generate the results was a real-coded, pseudo steady state EA design using binary tournament selection (without replacement) and a population size of *N* = 50. Results shown are a random sample of 5000 data points taken from a data set of 300,000. The box plots have the standard meaning with the bottom line in the box representing the first quartile, the middle line representing the median, and the upper line representing the third quartile. The symbol ✳ is used to represent outlier data points.**

**Sensitivity to fitness landscape:** To demonstrate how little the ETV-Rank relationship was sensitive to the fitness landscape, Figure 3-7 shows data from a simple unimodal test function and a highly deceptive test function. Here it can be seen that the relationship between ETV and rank is very similar for these two starkly different fitness landscapes. The only important difference comes in the location of extreme ETV outliers (e.g. ETV>40). For the deceptive problem (MMDP), the extreme ETV outliers are occasionally found in lower ranking individuals while for the simple unimodal test function (Quadratic Function), the extreme ETV outliers are more tightly associated with the highest ranked individuals.

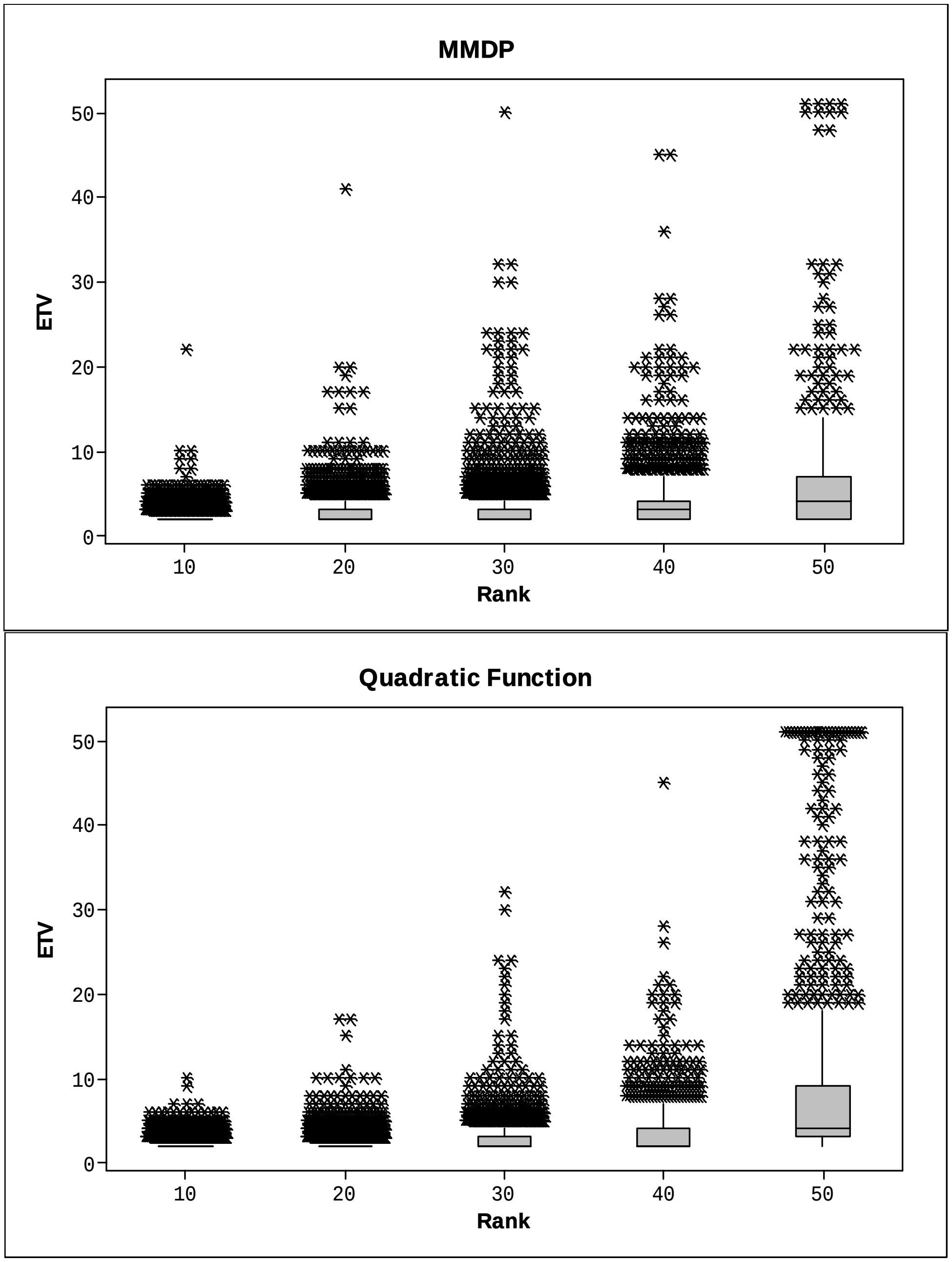

**Figure 3-7 Box plots of the size of an individual's impact on population dynamics (ETV) as a function of the individual's ranking within the population. Top: Results from running an EA on the Massively Multimodal Deceptive Problem (MMDP). Bottom: Results from running an EA on the Quadratic Test Function. Both test functions are defined in Appendix A. The EA used to generate the results was a real-coded, pseudo steady state EA design using binary tournament selection (without replacement) and a population size of *N* = 50. Results shown for each graph are a random sample of 5000 data points taken from a data set of approximately 15,000. The box plots have the standard meaning with the bottom line in the box representing the first quartile, the middle line representing the median, and the upper line representing the third quartile. The symbol ✳ is used to represent outlier data points.**

In summary, fitness-based ranking does provide some small indication of an individual's chances for impacting future dynamics, however its overall ability to predict future behavior is marginal. In general, fitness rankings have a strong tendency to overvalue the actual importance of individuals in future population dynamics.

#### 3.2.3.2 ETV Distribution

The results from Figure 3-6 indicate that regardless of rank, very few individuals have a large impact on population dynamics. Notice that even for the 20% highest ranked individuals (i.e. Box plot with the label Rank=50), the median ETV value is approximately five which is only 10% of the maximum ETV value. In order to better understand EA population dynamics, the distribution of ETV measurements is provided in Figure 3-8. The linearity of the ETV distribution on the log-log plot indicates the distribution fits a power law.

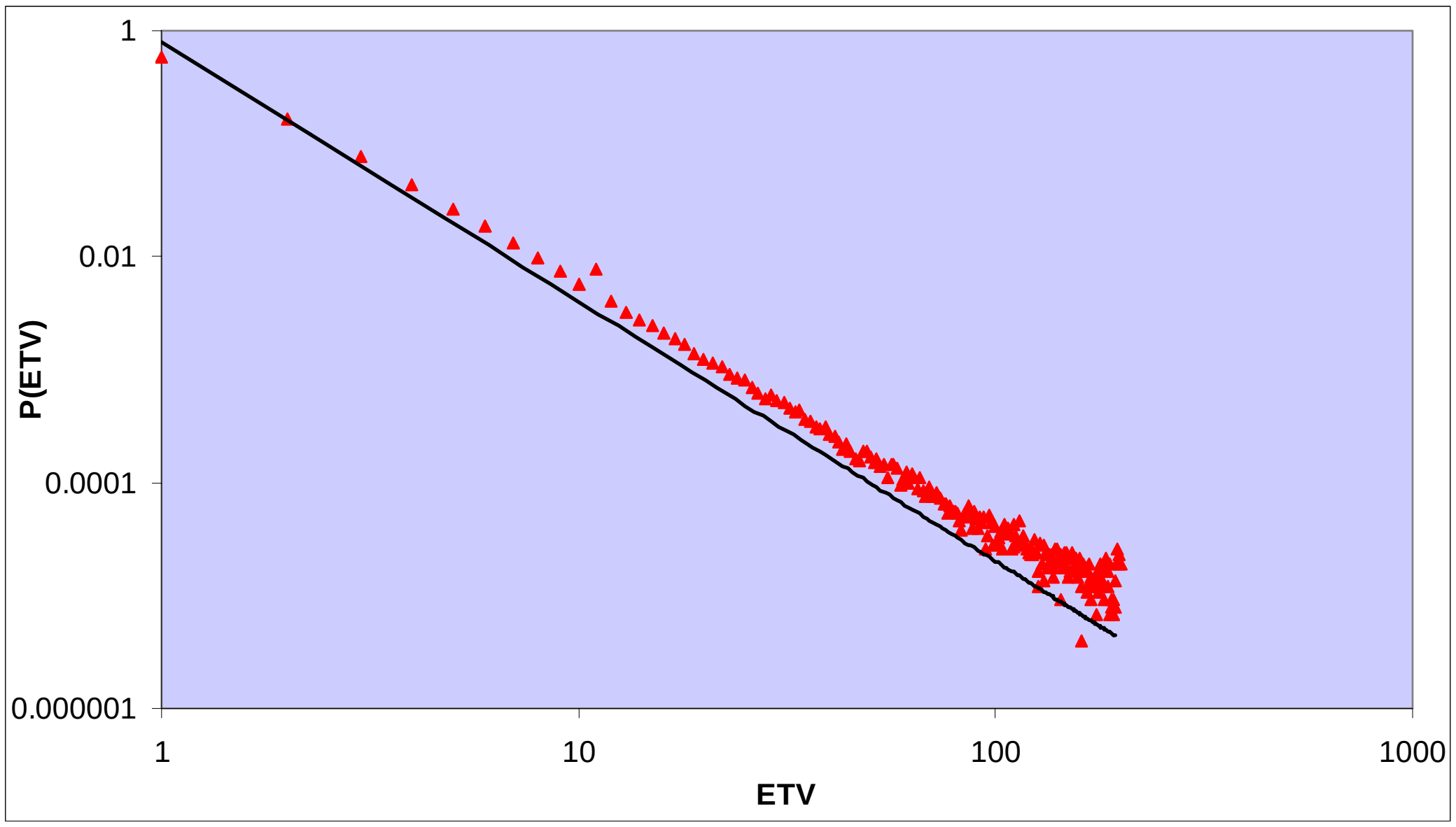

**Figure 3-8 ETV probability distribution from running an EA for 20,000 generations on the 30-D Hyper Ellipsoid test function. The EA design has a population size *N*=200, steady state population updating, and uses truncation selection. The solid line represents a power law with exponent 2.2.**

**Importance of ETV Distribution:** The existence of a power law ETV distribution (with exponent ~ 2) indicates that the large majority of individuals play a negligible role in influencing population dynamics whereas a vanishingly small number of individuals dominate population dynamics. This also indicates that most of the search is characterized by actions of questionable importance but is punctuated by the infrequent occurrence of important new discoveries. The power law ETV distribution has been confirmed for a broad range of EA designs which is presented and discussed in Chapter 4.

This result also has significant implications for how to best interpret the interactions between an adaptive system and its environment. For instance, consider the adaptive system that is used in this chapter, namely the adaptation of search operator probabilities. It is now known that no matter how good a search operator is, most of the times it is used, it will have little impact on population dynamics. This is not viewed as an indicator of poor operator performance but instead is understood as a reflection of the fundamental dynamics of the system. In other words, the majority of small impact interactions between a search operator and its environment are viewed as being not informative (effectively neutral interactions) whereas the small number of interactions which do have a large impact on population dynamics are viewed as being very informative and should be treated as valuable indicators of performance. The next section proposes a way to take into account these findings.

### 3.2.4 Interpreting ETV measurements

Based on the conclusions from the last section, the goal of this section is to interpret ETV data so that only informative, high impact events are able to influence search operator probabilities. This is accomplished by treating ETV measurements as being dominated by neutral measurements that fit some assumed distribution. Statistical arguments are then used to gauge whether an ETV measurement is important based on the extent that the ETV is an outlier of the neutral distribution. [10]

The first step is to determine the properties of the neutral distribution based on the ETV measurements gathered during an adaptation cycle. This is accomplished by taking all of the ETV measurements (gathered from the previous adaptation cycle) and calculating the mean and variance of the sample based on the assumption that neutral measurements dominate the data and fit a lognormal distribution. Other distribution assumptions have also been tested (e.g. Normal) with fairly similar results.

---

[10] For more information on statistical tests, see [167]

An ETV is an outlier when it is NOT expected statistically that the sample population contains at least one such measurement or any larger measurement. In other words, each ETV can be assigned a probability $p_\alpha$ that the ETV is an outlier of the sample distribution. A quantitative formulation of this definition will now be derived from statistics. Also note that for the calculations below, the ETV measurements are first transformed to log ETV so that neutral events can be assumed to fit a normal distribution.

#### 3.2.4.1 Outlier Calculation

Assuming a normal distribution for the data, individual measurements $ETV_j$ are tested against the sample mean $\mu$ and one-sided $p$ values, defined as $p_z$ in (3-16), are calculated using a $z$ statistic as defined in (3-15). The $s$ term in equation (3-15) represents the sample standard deviation. This calculated $p_z$ value indicates the probability of observing a measurement of size $ETV_j$ or greater. Hence, this simple statistical test can be used to determine the extent that an ETV value is an outlier. However, one must also account for the fact that the number of outliers observed for a given search operator will also depend on the number of times that the search operator was used (i.e. the search operator sample size). This is relevant because different operators will generally have different operator usage probability values and therefore will have different ETV sample sizes.

If a search operator $i$ has an ETV sample size $M_i$, the number of measurements $\alpha$ that are of size $ETV_j$ or greater follows a binomial distribution that is given by (3-17). The probability $p_\alpha$ of NOT observing a measurement greater than or equal to $ETV_j$ after $M_i$ observations is therefore the probability that $\alpha < 1$ given by (3-18, which can be calculated by the binomial cumulative distribution function.

$$z_j = \frac{ETV_j - \mu}{s} \quad \textbf{(3-15)}$$

$$p_z = P(z > z_j) \quad \textbf{(3-16)}$$

$$\alpha = Bin(M_i, p_z) \quad \textbf{(3-17)}$$

$$p_\alpha = P(\alpha < 1) \quad \textbf{(3-18)}$$

The final result $p_\alpha$ indicates the extent to which an ETV is an outlier that can not be easily accounted for by the stated distribution and the number of points sampled. Summing these $p_\alpha$ values over all events produced by a search operator indicates the extent that the operator can create exceptional offspring that have an unexpectedly large impact on population dynamics.

Measurement interpretation by the Outlier method is defined by (3-19). The reason that the $p_\alpha$ value is multiplied by $M_i$ in this equation is to allow this interpretation method to fit within the adaptive framework presented in Section 3.1.4. To be clear, this means that an operator's reward $R$ is equal to the sum of its $p_\alpha$ values (and not the average) when using this interpretation method. An average is not used because any sensitivity to the operator sample size $M_i$ has already been accounted for in the statistical arguments above.

$$I_{Outlier} = M_i p_\alpha\left(ETV_j\right), \quad j \in M_i \qquad \textbf{(3-19)}$$

**Impact of operator sample sizes:** Taking a hypothetical sample of ETV data that has been normalized using (3-15), Figure 3-9 illustrates how the calculation of $p_\alpha$ will interpret an ETV measurement for different ETV values and different operator sample sizes. For each of the sample sizes in Figure 3-9, the $p_\alpha$ calculation places almost no value on any measurements found below the sample mean ($z$ = 0). Also notice that for very high measurements ($z$ > 3), the $p_\alpha$ calculation approaches a value of 1 meaning it has high confidence that the measurement is an outlier (i.e. that the event had a large impact on population dynamics). Finally, for ETV values that are large but their classification as outliers is less certain, the sample size from which the measurement is taken will strongly influence the interpretation of the measurement.

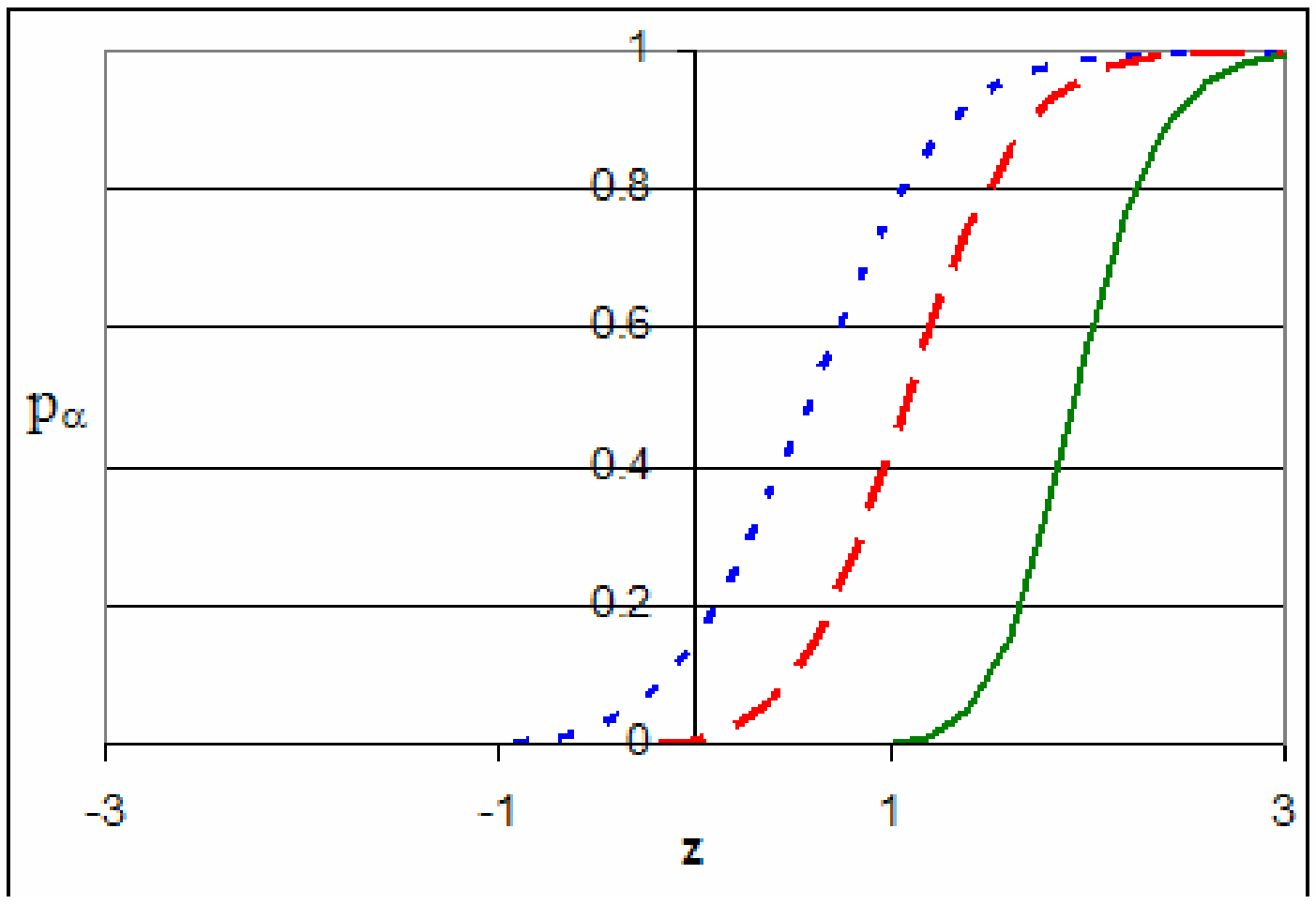

**Figure 3-9: $p_\alpha$ calculation curves for sample sizes $M_i$=5 (- - -), $M_i$ =10 (— —), and $M_i$ =20 (——).**

Although the Outlier interpretation method described here may seem overly complicated for those unfamiliar with statistical tests, it is really only a procedure for selectively using measurement outliers for adaptation. Many of the interpretation methods described in Section 3.1.4.5 are simple heuristics for placing emphasis on higher valued measurements and so in this way they are similar to the Outlier interpretation method. The difference with the present approach is that it actually quantifies the degree to which each event exceeds the average (in terms of probability of occurrence), and gives much more weight to "true" outliers.

## 3.3 Experiments

### 3.3.1 Experimental Setup

This chapter has thus far reviewed a framework for adaptation of search operator probabilities in an Evolutionary Algorithm and has presented an approach to adaptation based on the ETV measurement and the Outlier interpretation method. This section assesses the performance of the new supervisory adaptive method by testing it on a suite of artificial test functions and engineering design problems (listed in Table 3-2) and comparing these results with a number of adaptive and non-adaptive EA designs. Details of the EA designs used in these experiments are described next.

**Table 3-2 List of test functions used in experiments. Problem definitions, parameter settings, fitness landscape characteristics, and problem descriptions (for design problems) are provided in Appendix A.**

| Artificial Test Functions | Engineering Design Problems |
|---|---|
| Bohachevsky's | Turbine Power Plant |
| Quadratic | Welded Beam Design |
| Rosenbrock's Valley | Tension Compression Spring |
| Rastrigin | Gear Train Design |
| Schwefel | |
| Griewangk | **Application-Inspired Problems** |
| Massively Multimodal Deceptive Problem (MMDP) | Minimum Tardy Task Problem (MTTP) |
| Watson's | Error Correcting Code Problem (ECC) |
| Colville's | Frequency Modulation |
| System of linear equations | |
| Ackley's Path Function | |
| Neumaier's Function #2 | |
| 30-D Hyper Ellipsoid | |

#### 3.3.1.1 Core EA Design

The EA designs used in these experiments are described below with the core of the EA design given by the pseudocode in Figure 3-10. For each generation *Gen*, *N* new individuals are generated to form the offspring population. Offspring are generated by selecting one of the ten search operators given in Table 3-3. The operators are selected probabilistically in proportion to the operator probability value. Parents are then selected at random from the parent population, with the number of parents depending on the search operator used. The offspring is finally created and its objective function value is evaluated.

```
Initialize population
Evaluate population
Do
    'Reproduction
    For i=1 to N
        select a single search operator (based on probabilities)
        select parents (at random)
        Create offspring
        Evaluate offspring
    Next i
    Gen=Gen+1
    IF (constrained problem) THEN define fitness by Stochastic Ranking
    If Adapt THEN Adapt Operator Probabilities
    'Selection
    For i=1 to N
        select two individuals (at random from parents and offspring)
        keep more fit individual
    Next i
Loop until stopping criteria
```

**Figure 3-10 Pseudocode of EA design**

**Table 3-3: List of search operators used in EA designs. Full descriptions of each search operator are provided in Appendix B.**

| **Search Operators** |
|---|
| Wright's Heuristic Crossover |
| Simple Crossover |
| Extended Line Crossover |
| Uniform Crossover |
| BLX- α |
| Differential Evolution Operator (DE) |
| Swap |
| Raise |
| Creep |
| Single Point Random Mutation |

Selection takes place by combining the parent and offspring populations and then repeatedly selecting two individuals at random and removing the worse individual until the total population size is reduced to *N*. The selection procedure is very similar to binary tournament selection without replacement.

Populations were randomly initialized (with EA experimental replicates using random number seeds for blocking[11]), the population size was set to $N$=30, and the stopping criteria was set as a maximum of 3000 generations. Thus the final solution from each experiment is obtained after 90,000 objective function evaluations. Genes consisted of direct representations of the parameters being optimized (i.e. real coding).

For optimization problems with nonlinear constraints, fitness is determined using the stochastic ranking method presented in Chapter 2 and also described in [64]. Previous experience with this method has indicated that Stochastic Ranking works well for many problems using the parameter settings specified in [64].

#### 3.3.1.2 Search Operator Control

The algorithms tested in these experiments differ only in the settings of the search operator probability values. A number of methods for adapting the probability values were tested which are described in the pseudocode in Figure 3-11 and in Figure 3-12.

The standard adaptive procedure is described by the pseudocode in Figure 3-11 and works by taking the fitness measurement for each new offspring, interpreting the fitness using one of the interpretation formula provided in Section 3.1.4.5, and storing this interpretation with others from the same search operator. Every $\tau$ generations, the stored data is averaged to calculate the Reward $R$, as defined in (3-4), which in turn is used to calculate the Quality $Q$ for each operator as defined in (3-1). Finally the operator probability value $P$ is calculated using either the probability matching strategy defined in (3-2) or the adaptive pursuit strategy defined in (3-3).

---

[11] Blocking is a method for designing experiments in order to reduce the variability of results arising from some unimportant factor. For these experiments, the unimportant factor is the sensitivity of EA performance to the initial conditions of the population. In this case, blocking occurs by using the same set of random number seeds for tests on each EA design. More information on blocking can be found in chapter five of [167].

```
For each offspring
        i = offspring's search operator
        Calculate F (defined in Section 3.1.4.4)
        Calculate I (defined in Section 3.1.4.5)
        M_i = M_i +1
        I_archive (i, M_i)= I
Next offspring
IF Gen mod τ =0 THEN
        t=t+1
        For each operator i
                Calculate R (defined in Section 3.1.4.3)
                Calculate Q (defined in Section 3.1.4.1)
                Calculate P (defined in Section 3.1.4.2)
                M_i = 0
        Next i
END IF
```

**Figure 3-11 Pseudocode for standard search operator probability adaptation**

The order of calculation steps changes slightly when using ETV in place of the fitness measurement and so a separate pseudocode is provided in Figure 3-12. The change to the pseudocode is due to the fact that the ETV measurement of an offspring can take several generations to calculate. More information on the ETV calculation steps is provided in Section 3.2.2.3. When the ETV measurement is used, it is interpreted using either the Outlier interpretation method described in Section 3.2.4.1 or no interpretation is used, in which case the interpretation *I* is set equal to ETV.

```
Calculate ETV
If Gen mod τ =0 Then
        t=t+1
        For each completed ETV
                i = ETV's search operator
                F = ETV
                Calculate I     '(Options:  I_Outlier or I=ETV)
                M_i = M_i +1
                I_archive (i, M_i)= I
        Next ETV
        For each operator i
                Calculate R (defined in Section 3.1.4.3)
                Calculate Q (defined in Section 3.1.4.1)
                Calculate P (defined in Section 3.1.4.2)
                M_i = 0
        Next i
End If
```

**Figure 3-12 Pseudocode for search operator probability adaptation using ETV. ETV is defined in Section 3.2.2 and $I_{Outlier}$ is defined in Section 3.2.4.**

The parameter settings and other design details of the adaptive methods tested in these experiments are provided in Table 3-4. The adaptive methods chosen were done so in an attempt to sample a number of the design options described in the background material (Section 3.1.4) including measurement interpretations which use the parent context, the population context, and a standard ranking interpretation (see Section 3.1.4.5). Also included are several adaptive methods that use the adaptive pursuit strategy and the probability matching strategy for adjusting search operator probabilities (see Section 3.1.4.2).

**Table 3-4 Details of the adaptive methods used for adapting search operator probabilities are listed. Column one provides the label used to refer to each adaptive method. The second column indicates whether the adaptive method uses the adaptive pursuit strategy (Y) or the probability matching strategy (*N*). The measurement of an event is given in column three as either the fitness (*F*) or the Event Takeover Value (ETV). The interpretation of event measurements is either one of those listed in Section 3.1.4.5 or the Outlier method of Section 3.2.4.1. For the "ETV" adaptive method, the interpretation is equivalent to the ETV value. Each adaptive method has the task of setting the operator probabilities for the 10 search operators listed in Table 3-3. Each adaptive method uses parameter settings $\alpha$ = 0.8, $P_{Min}$=0.02 and $\tau$ =10. The adaptive pursuit strategy also has $\beta$ = 0.8. No attempt was made to tune these parameters and the values were chosen largely to maintain consistency with previous research in this topic. Preliminary testing indicated that the results are not strongly sensitive to the setting of $\alpha$ and $\tau$.**

| Adaptive EA design name | Adaptive Pursuit (Y/N) | Event Measurement | Measurement Interpretation |
|---|---|---|---|
| I(median)-Pursuit | Y | *F* | $I_4$ |
| I(parent)-Pursuit | Y | *F* | $I_1$ |
| I(rank)-Pursuit | Y | *F* | $I_8$ |
| I(median) | N | *F* | $I_4$ |
| I(parent) | N | *F* | $I_1$ |
| I(rank) | N | *F* | $I_8$ |
| ETV-Outlier | N | ETV | Outlier |
| ETV | N | ETV | ETV |

Two EA designs which do not adapt search operator probabilities are also considered. The first, referred to as Static-Ops2, only uses uniform crossover with probability 0.98 and single point mutation with probability 0.02. All other search operators have probability values of zero. The second design, Static-Ops10, uses all ten search operators listed in Table 3-3 with equal probability ($P$ = 0.1).

### 3.3.2 Results and Discussion

#### 3.3.2.1 General Performance Statistics

This section attempts to draw general conclusions about the performance of the adaptive EA designs tested in these experiments. The first statistic in column two of Table 3-5 states the percentage of problems that an EA was found to be the best algorithm out of those tested. The second statistic in the third column states the percentage of problems where an EA found the best solution in at least one its runs. These two statistics evaluate final algorithm performance, however it is also useful to make statements about performance at other timescales. To address this, Figure 3-13 presents general algorithm performance as a function of time.

**Table 3-5 Overall performance statistics for each of the adaptive and non-adaptive EA designs. Column two measures the percentage of problems where an EA design was the best EA design (comparisons based on median objective function value). Column three measures the percentage of problems where an EA design was able to find the best solution at least one time. The best solution is defined as the best found in these experiments and is not necessarily the global optimal solution.**

| EA Design | % of problems where EA | |
|---|---|---|
| | was best design | found best |
| I(median)-Pursuit | 10.4% | 40% |
| I(parent)-Pursuit | 2.9% | 35% |
| I(rank)-Pursuit | 7.9% | 40% |
| I(median) | 4.5% | 55% |
| I(parent) | 12.9% | 45% |
| I(rank) | 9.5% | 45% |
| ETV-Outlier | **27.0%** | **90%** |
| ETV | 15.4% | 35% |
| Static-Ops2 | 3.0% | 15% |
| Static-Ops10 | 6.6% | 45% |

Based on these general performance statistics, some important conclusions can be drawn. First, it is clear that the two operator EA design with static search operator probabilities (Static-Ops2) performs very poorly on almost every test function. This is a significant conclusion since the two operator non-adaptive EA is by far the most commonly used EA design. By simply including more operators without even tuning (or adapting) the probability parameters, it is found that substantial performance improvements occur.

Adding adaptive mechanisms for tuning the probability parameters provides significant performance improvements although the best adaptive method is problem specific. It is important to notice that while as a class, adaptive methods were more than twice as likely to be the best design for a given problem (compared to the non-adaptive EA designs), no single adaptive method was strongly favored over all others. However, it is still somewhat impressive that the ETV-Outlier adaptive method is found to be the best design on 27% of the test functions while the average for all other adaptive methods was 9.1%. It is also worth pointing out that the next best adaptive method, ETV, was the best design only 15.4% of the time.

If one is more interested in an algorithm's ability to find good solutions over multiple runs, then much stronger conclusions can be made from these results. From column three of Table 3-5, one can see that the ETV-Outlier method is able to find a best solution in 18 of the 20 problems tested while the second best algorithm, I(median), only finds a best solution 55% of the time.

Finally, if one is concerned with performance at different time scales, it is also clear that the ETV-Outlier adaptive method exhibits strong performance throughout the 3000 generations tested and that the non-adaptive method, Static-Ops2, exhibits poor performance throughout the 3000 generations tested. It is also interesting to note in Figure 3-13 that two of the three adaptive methods employing the adaptive pursuit strategy are no better (on average) than the non-adaptive EA design, Static-Ops10.

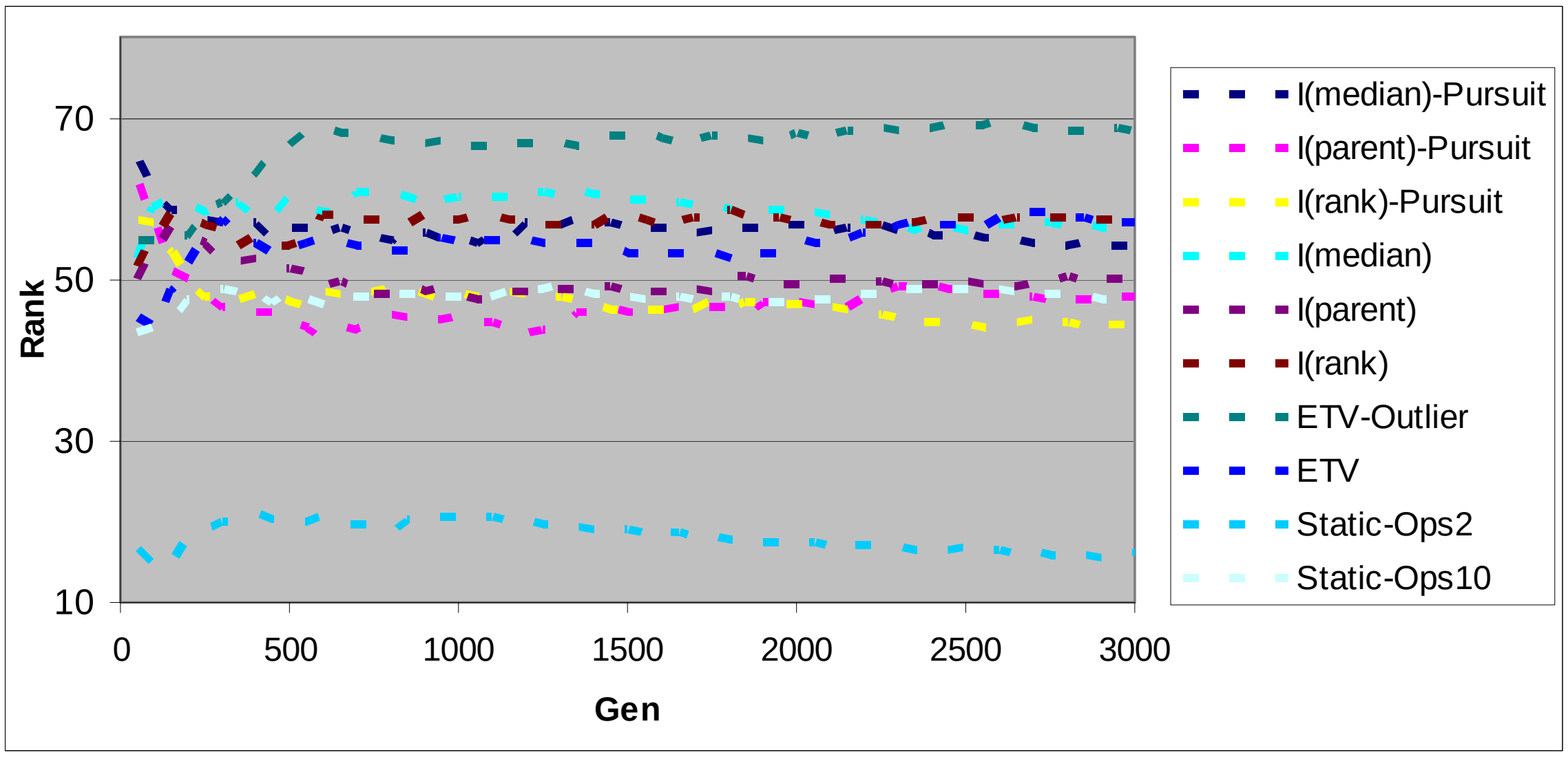

**Figure 3-13 General algorithm performance for both adaptive and non-adaptive EA designs shown as a function of the number of generations (Gen) of evolution. In order to aggregate performance data from different test functions, it was necessary to deal with differences in fitness scaling. This was addressed by using the following ranking procedure. Each algorithm is run 20 times on each test function listed in Table 3-2. At a given generation, every EA run is ranked among all runs conducted on that test function (with a higher ranking being better). The median rank of each EA design is then calculated for each test function. Finally, these median ranks are averaged over all test functions and plotted against the number of generations (Gen).**

More detailed performance results for individual test functions are broken down into three parts. In Section 3.3.2.2, the results from a selected set of artificial test functions are analyzed in detail with the goal of understanding the relationship between search operator probability profiles and algorithm performance. In Section 3.3.2.3, performance on the remaining artificial test functions is presented and briefly described. Finally, Section 3.3.2.4 looks at algorithm performance on a selected set of engineering design problems.

### 3.3.2.2 Operator Probability Profile Analysis

The optimal features of a search operator probability profile are generally not known, however it is possible to venture a few educated guesses as to what such a profile might look like. First, the adaptive method should recognize those search operators which are not at all effective as a means of searching the fitness landscape. For those operators, it is desirable to keep their probability values very close to $P_{min}$ (i.e. the smallest allowed value). Although it is unknown which operators are suited for a particular problem, it is believed that the range of search operators used in these experiments (listed in Table 3-3) is broad enough so that at least one or more operators are not suited for each problem.

Also, as evolution proceeds, one can expect that the ruggedness and other fitness landscape characteristics will change as the population occupies different regions of parameter space. Hence one would expect that, on some problems, an adaptive method should modify which search operators it prefers as it adapts to changes in the environment. For some artificial test functions, the fitness landscape is fairly well understood and this knowledge is used in the following section to assess each of the adaptive methods.

The Rosenbrock, Schwefel, Griewangk, and Ackley test functions have been selected for this analysis. Search operator probability profiles are taken from the same experiments used to generate the performance results. The probability values that are shown for each search operator represent the median value from 20 experimental replicates.

#### *3.3.2.2.1 Rosenbrock Test Function*

The Rosenbrock test function is a smooth unimodal test function with the global optimum residing inside a long and narrow parabolic shaped valley as seen in Figure 3-14. Since the landscape is smooth and unimodal, finding the valley is trivial, however the curvature of the valley makes convergence to the global optimum slow and difficult. In most runs the population tends to gather along the bottom of the valley then follow it towards the bottom. Search operators which exhibit hill climbing behaviors are expected to be more effective in this fitness landscape since the problem is unimodal. The extended-line operator, differential evolution, and Wright's heuristic crossover all have hill climbing characteristics (similar to directed search) so one would suspect these to be favored over the other operators. Gene swapping operators like single point crossover, uniform crossover, and swap are not expected to perform well.

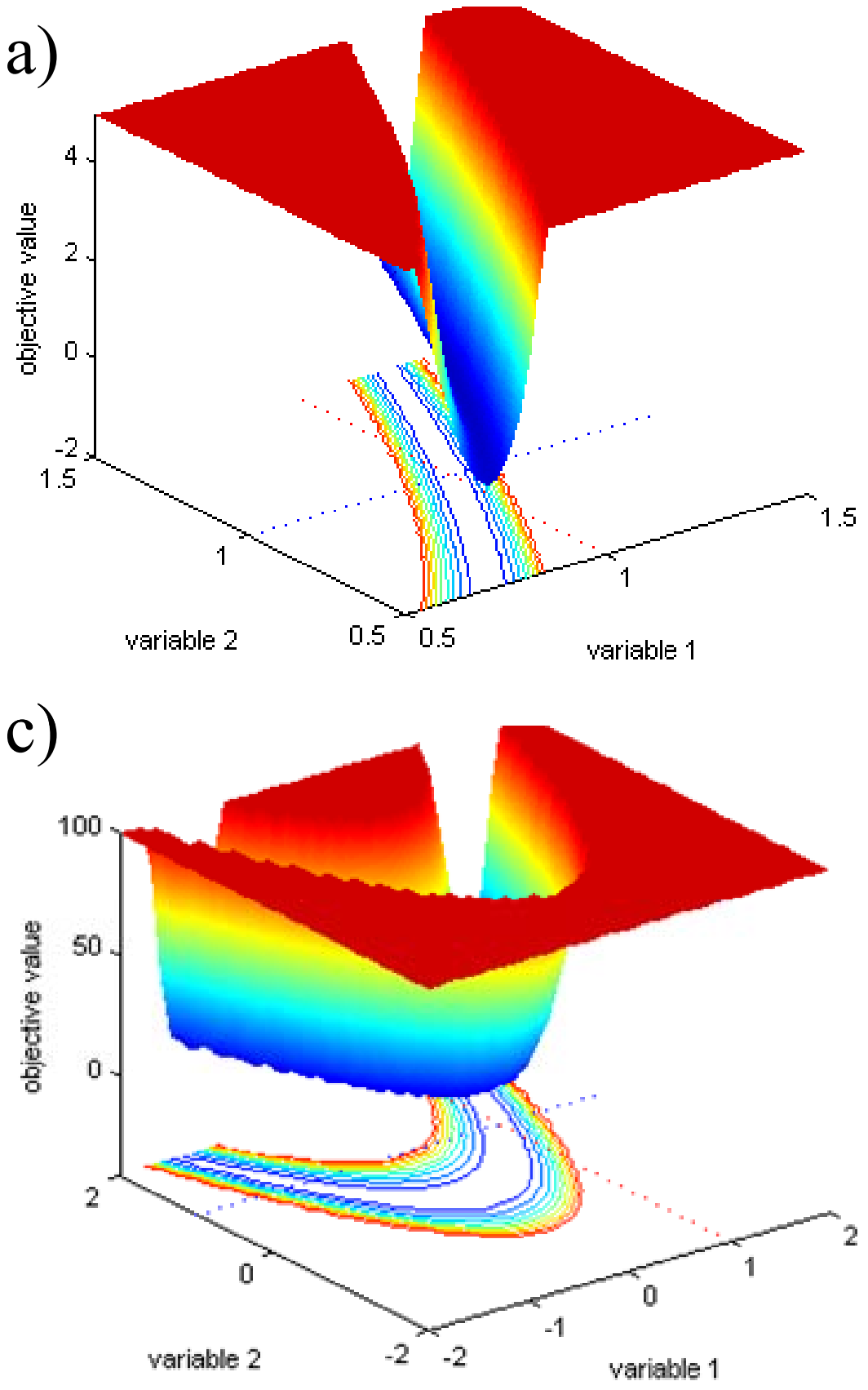

**Figure 3-14 Rosenbrock fitness landscape shown in a two dimensional parameter space. The two bottom graphs are shown for variable 1 and variable 2 varying over the entire parameter range [-2,2]. Graphs on the top focus on the parameter region containing the global optimum. The two graphs on the left show a restricted range of objective function values (vertical axis) to help in visualizing the fitness landscape. Images were kindly provided by Hartmut Pohlheim and were generated using the GEATbx toolbox in Matlab® [168]. Low resolution images can also be found at http://www.geatbx.com/docu/fcnindex-01.html#P85_2637.**

For many of the adaptive methods presented in the figures below, Wright's heuristic crossover and extended line crossover are strongly favored over other search operators. These operators act as interpolation and extrapolation search actions (resp.) which are biased towards the more fit parent making them quite effective on smooth landscapes like Rosenbrock. It is also noticed that the differential evolution operator and BLX are also favored, although to a lesser extent.

For the adaptive methods I(rank) and ETV, there is very little difference between search operator probabilities. These are also the two worst adaptive methods for this problem as seen in Figure 3-15. On the other hand, I(median) and ETV-Outlier find the greatest differences between search operator probabilities. These are also two of the three best adaptive methods for this problem. These results taken together (particularly methods ETV and ETV-Outlier) indicate that poor measurement interpretations can prevent adaptive methods from making important distinctions between operators.

It is also worth noting that there exists a form of symmetry in the Rosenbrock fitness landscape which causes the environment to be approximately stationary during most of evolution. As a consequence, one would expect the same search operators to be consistently preferred throughout the run. It is noticed however that for two of the three adaptive methods employing the adaptive pursuit strategy, very little consistency is present in the selection of search operators and in fact the operator profiles appear quite chaotic. This behavior is a general characteristic of the adaptive pursuit strategy which is repeatedly seen in the other test functions analyzed in this section. These two methods also seem to suffer in performance compared with their less chaotic cousin, I(median)-pursuit, as seen in Figure 3-15.

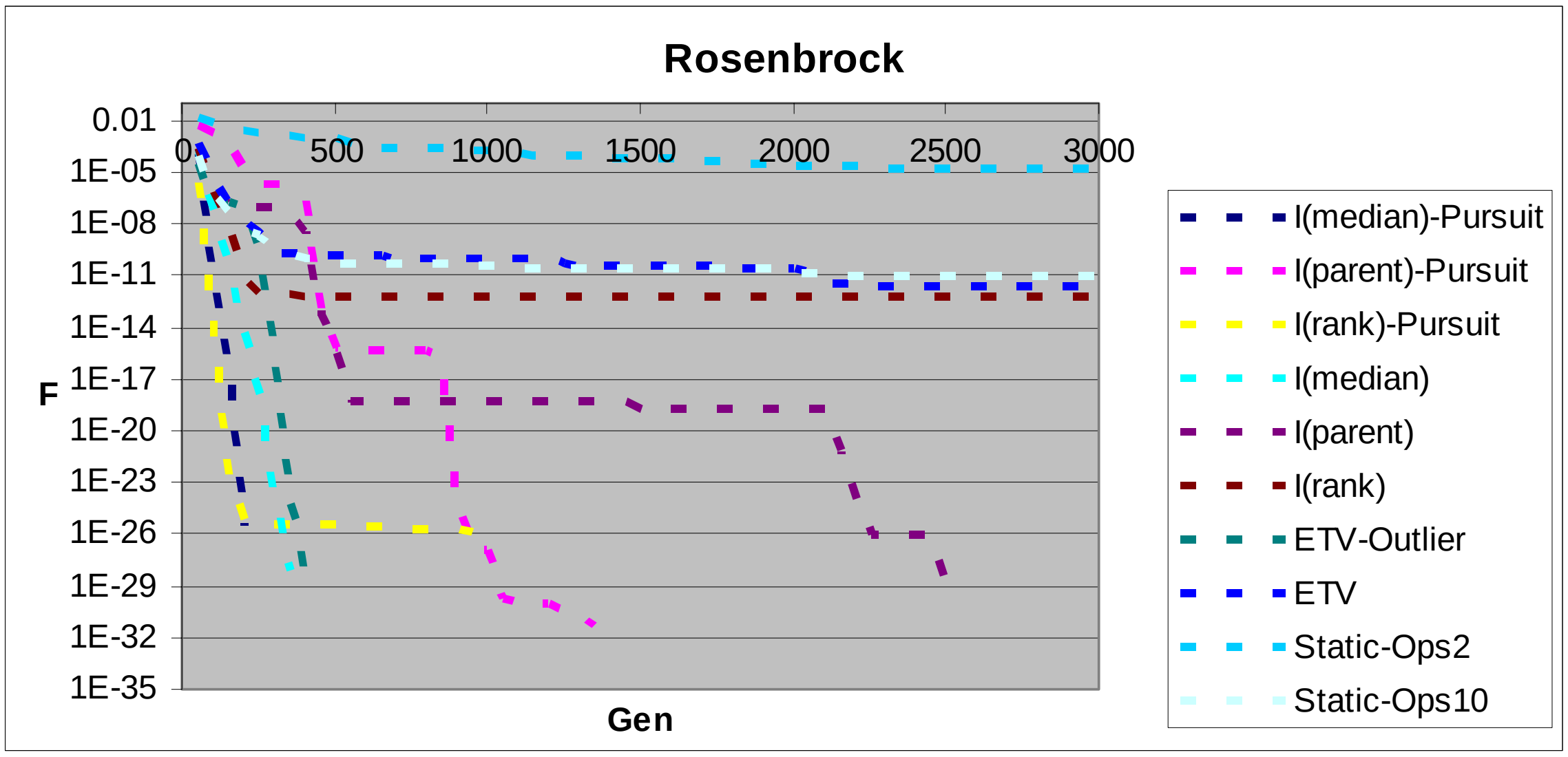

**Figure 3-15 Performance of adaptive and non-adaptive EA designs on the Rosenbrock test function. The global optimal solution is *F*=0. The optimal *F* value can not be shown due to log scaling on the *F* axis so performance profiles are seen to terminate when the global optima is reached.**

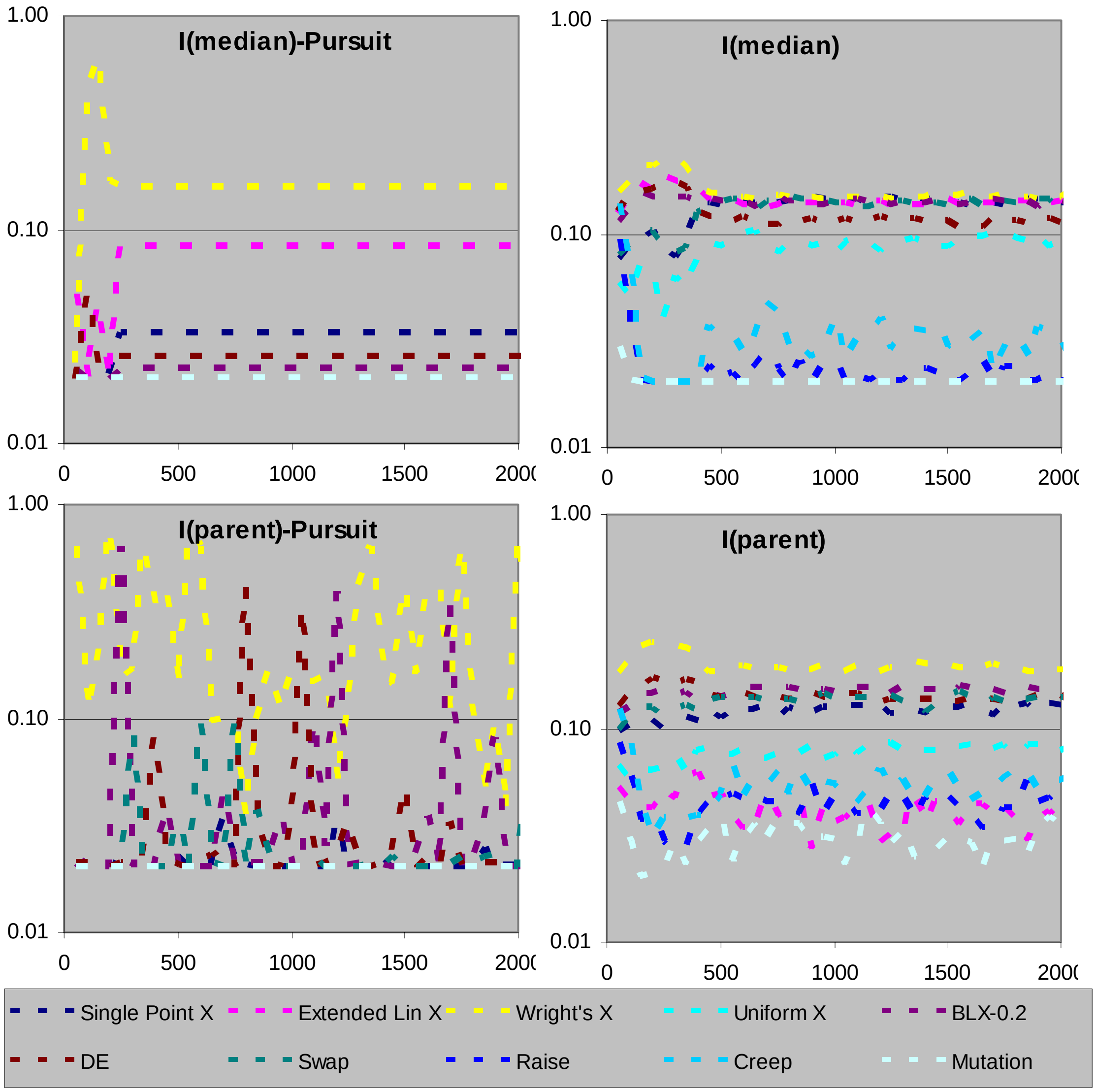

**Figure 3-16 Search operator probability profiles for adaptive methods I(median)-Pursuit, I(median), I(parent)-Pursuit, and I(parent) on the Rosenbrock test function. Probability values are shown on a logarithmic scale over the first 2000 generations of evolution.**

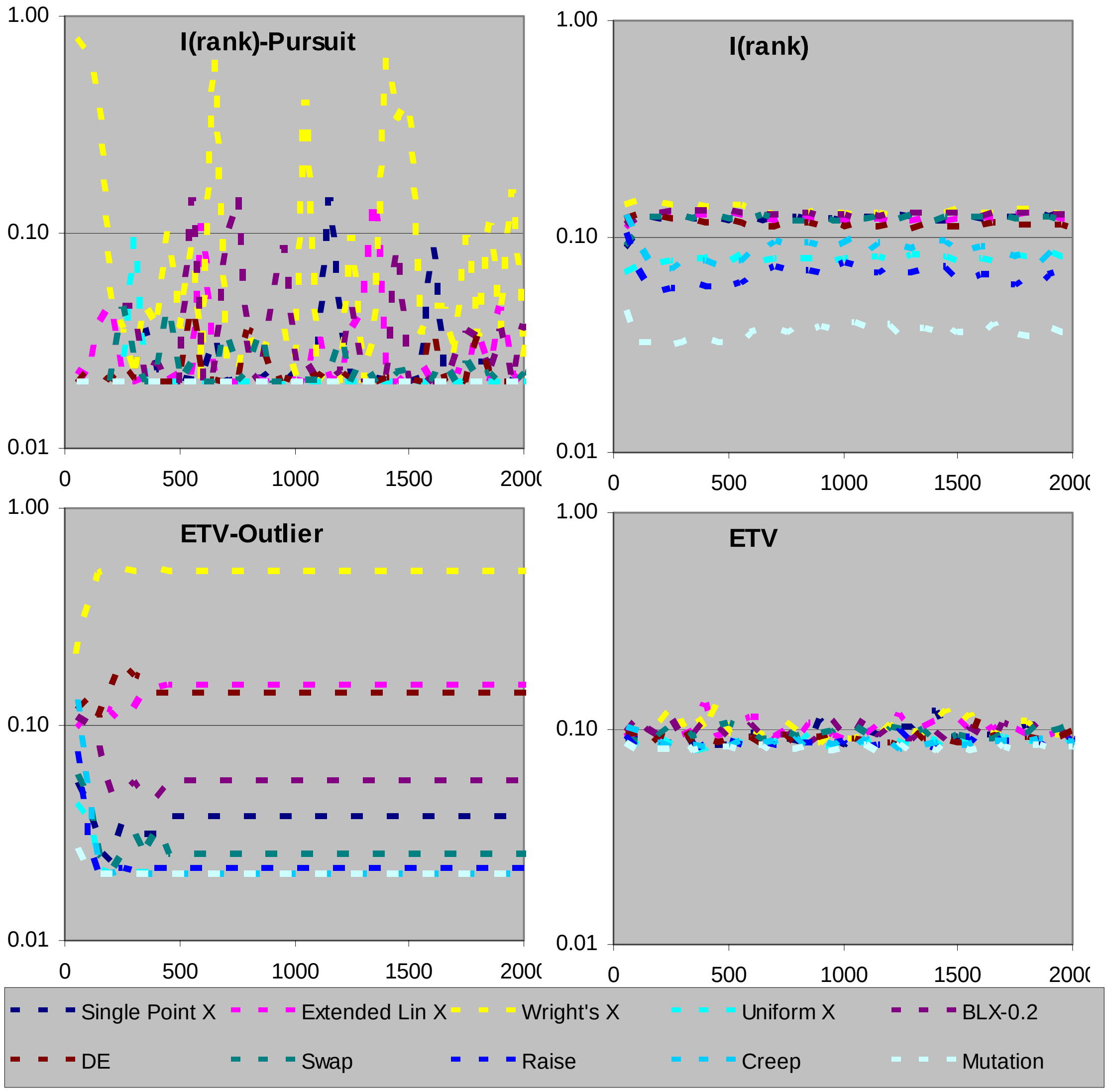

**Figure 3-17 Search operator probability profiles for adaptive methods I(rank)-Pursuit, I(rank), ETV-Outlier, and ETV on the Rosenbrock test function. Probability values are shown on a logarithmic scale over the first 2000 generations of evolution.**

### *3.3.2.2.2 Schwefel Test Function*

The Schwefel test function has a multimodal fitness landscape as seen in Figure 3-18. Local optima are distributed throughout parameter space with many containing fitness values that are similar to the global optima making the problem challenging.

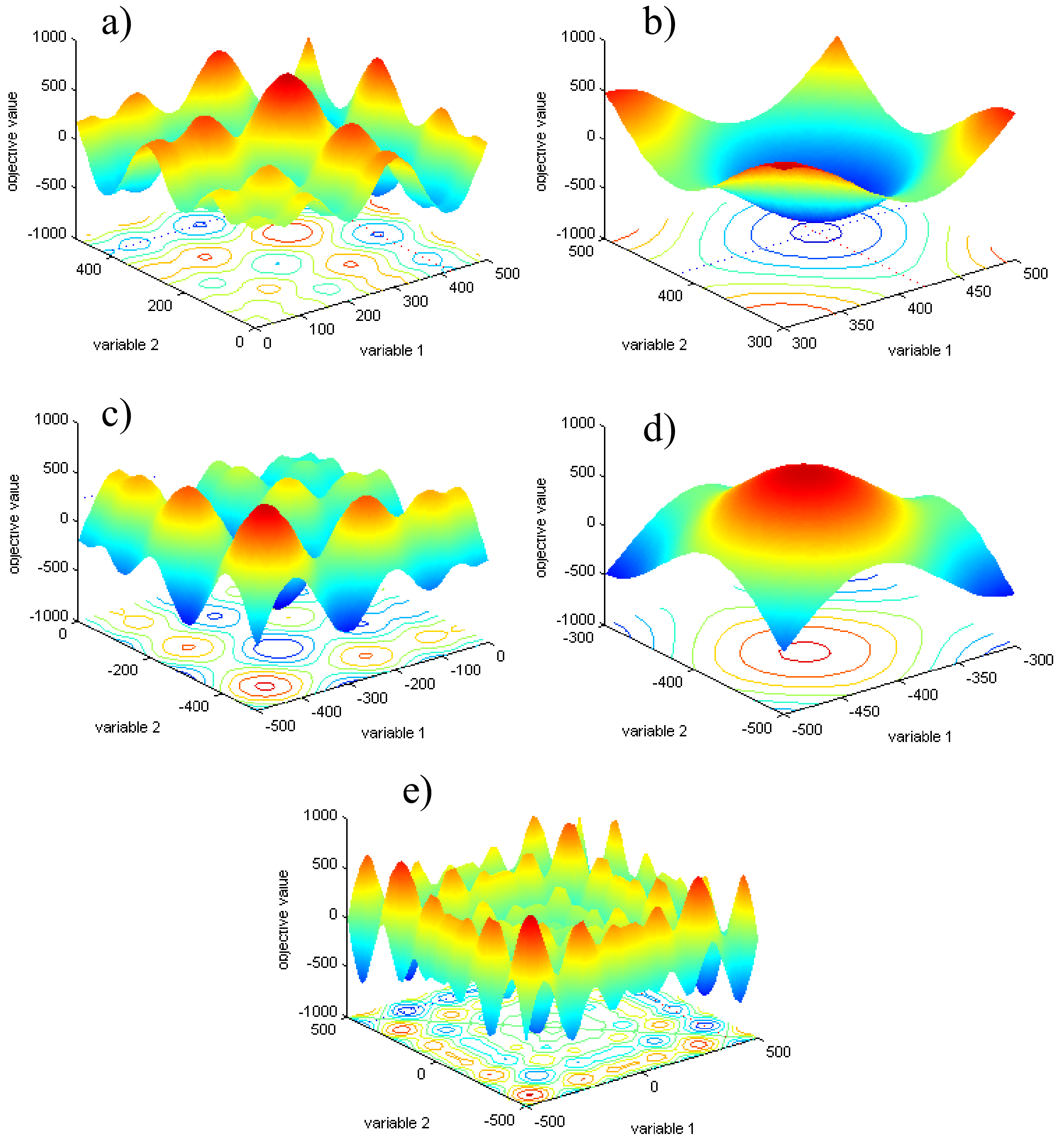

**Figure 3-18 Schwefel fitness landscape shown in two dimensions of parameter space. The landscape is shown for variable 1 and variable 2 varying over different parameter ranges. The entire range is shown in the bottom graph with each parameter varying over [-500,500]. The vertical axis shows the objective function value (minimization) with the global optimal solution located at the origin of parameter space. Images were kindly provided by Hartmut Pohlheim and were generated using the GEATbx toolbox in Matlab® [168]. Low resolution images can also be found at http://www.geatbx.com/docu/fcnindex-01.html#P85_2637.**

For the adaptive methods that can make clear distinctions between search operators, it appears that two general trends occur. In the first trend, which is observed in I(median), I(parent) and I(rank), it is found that the creep operator is consistently preferred throughout the span of evolution. These methods also demonstrate poor performance on this test function.

The second general trend is seen in I(rank)-pursuit, ETV-Outlier and to a lesser extent in I(median)-pursuit. For these methods, the creep operator is strongly favored initially but in the later stages of evolution, operators with hill climbing characteristics (Wright's heuristic crossover, extended line crossover, differential evolution) are found to be strongly favored. This behavior is most clearly visible in the search operator probability profile for the ETV-Outlier adaptive method. The three adaptive methods that exhibit this second trend in behavior also have the best performance as seen in Figure 3-26.

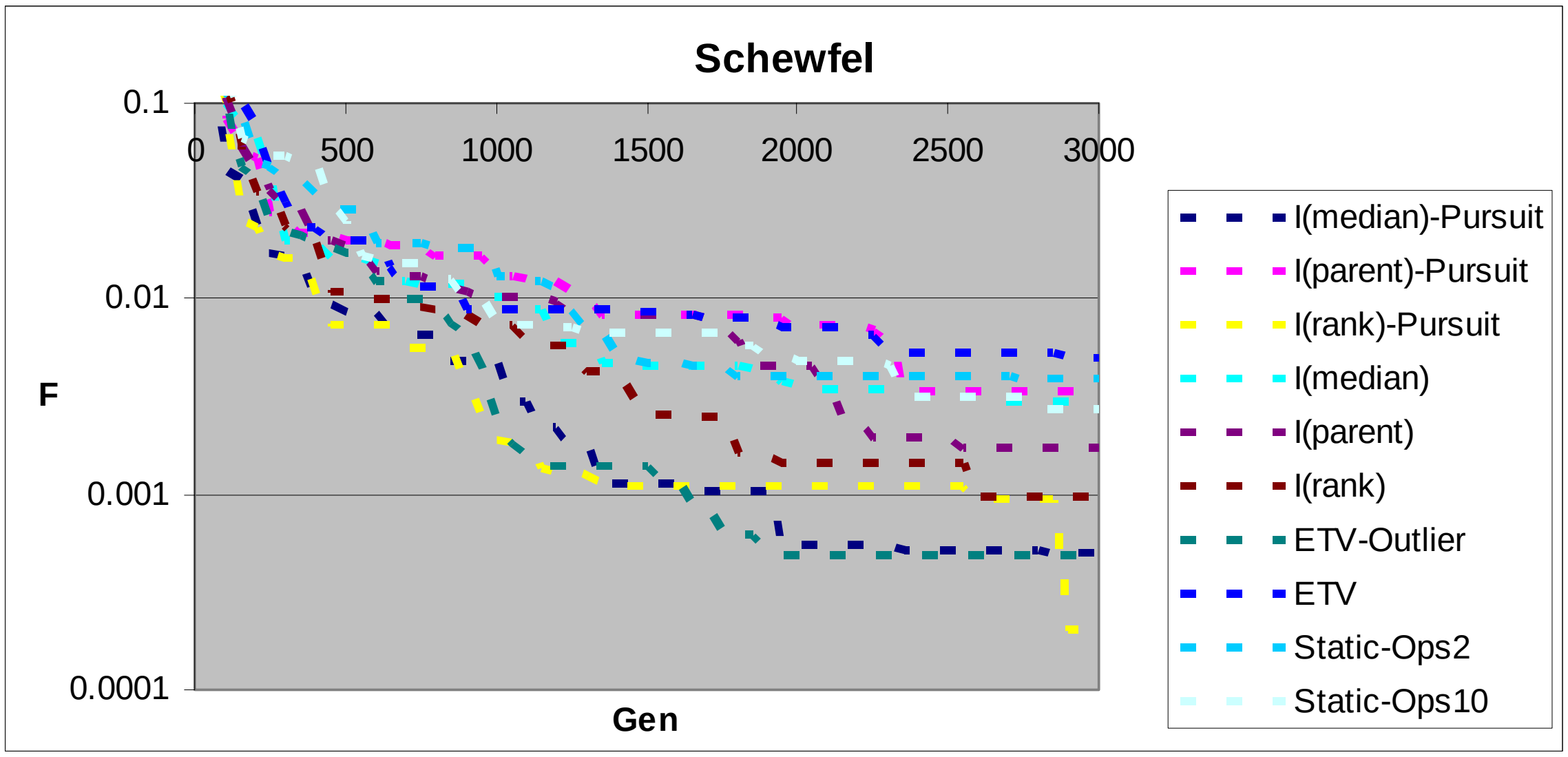

**Figure 3-19 Performance of adaptive and non-adaptive EA designs on the Schwefel test function. The global optimal solution is at *F*=0.**

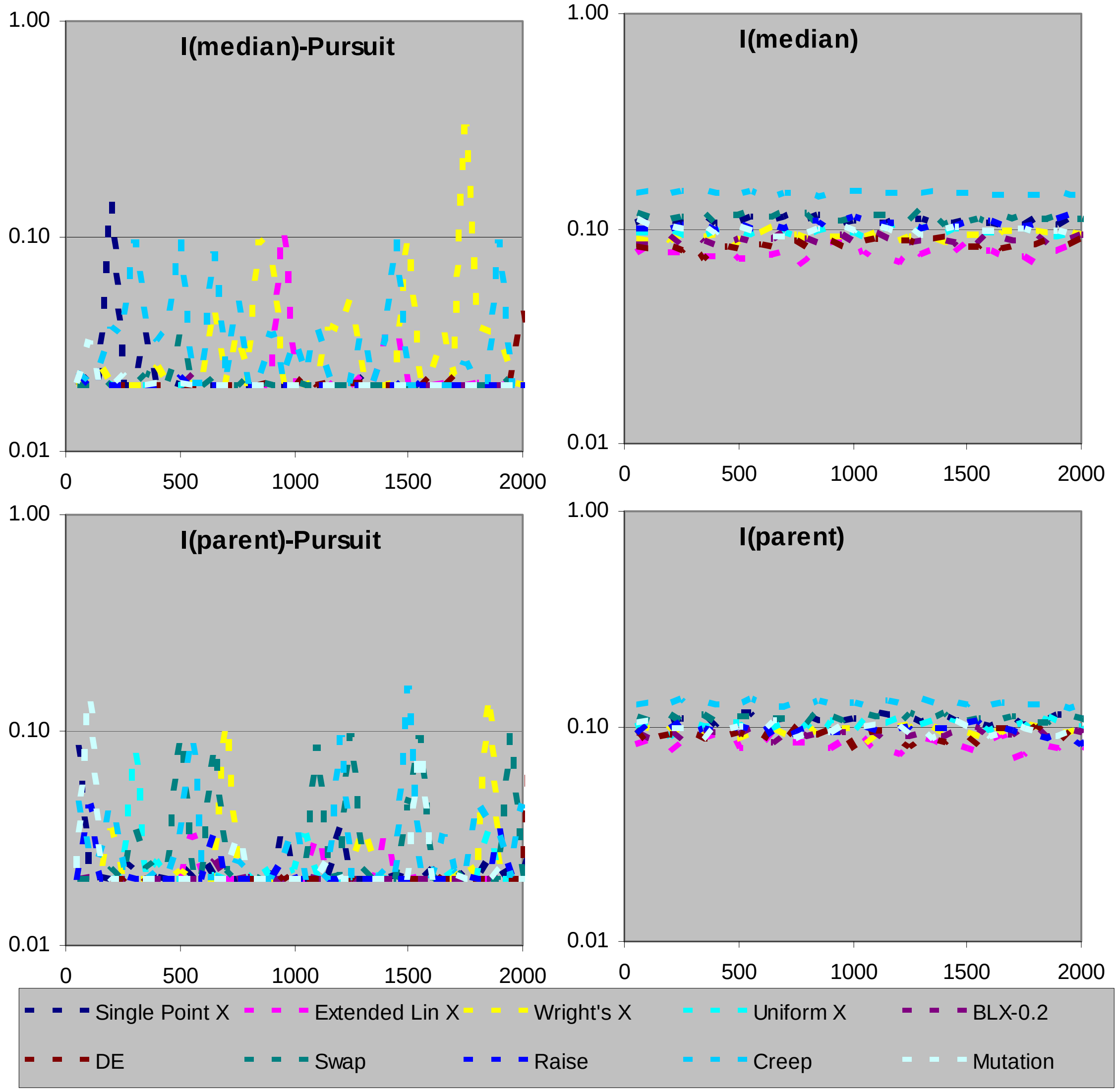

**Figure 3-20 Search operator probability profiles for adaptive methods I(median)-Pursuit, I(median), I(parent)-Pursuit, and I(parent) on the Schwefel test function. Probability values are shown on a logarithmic scale over the first 2000 generations of evolution.**

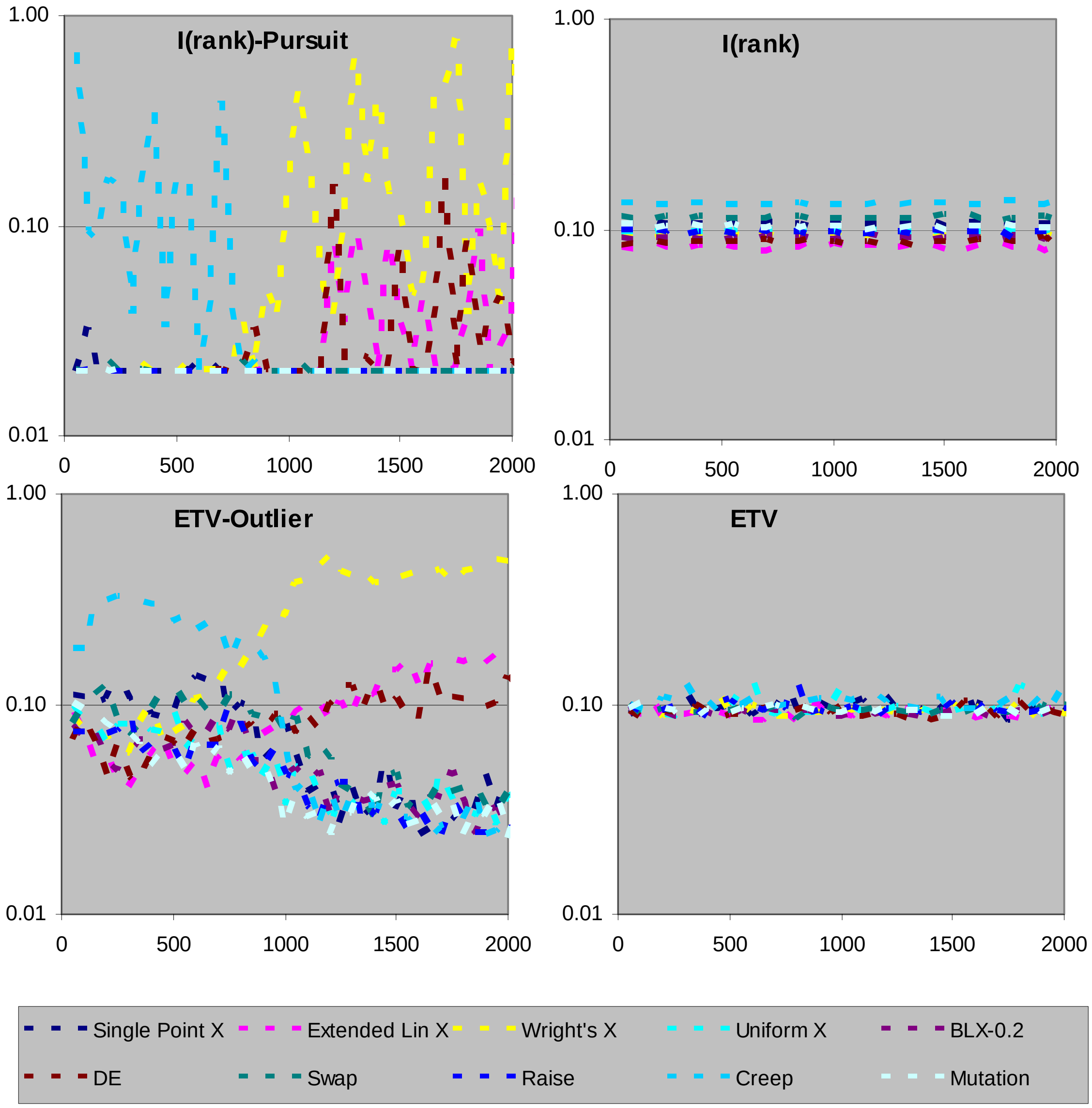

**Figure 3-21 Search operator probability profiles for adaptive methods I(rank)-Pursuit, I(rank), ETV-Outlier, and ETV on the Schwefel test function. Probability values are shown on a logarithmic scale over the first 2000 generations of evolution.**

### *3.3.2.2.3 Griewangk Test Function*

From Figure 3-22a, one can see that the fitness landscape for the Griewangk test function is smooth at large parameter scales. With the initial EA population randomly distributed throughout parameter space, it is expected that interpolative actions would be particularly useful in the early stages of evolution. However, as the population converges to a more localized region of parameter space, the landscape becomes very rugged as evidenced by the peaks in Figure 3-22b. Under these conditions, less exploitive operators are expected to be useful such as uniform crossover and BLX. Finally, when the population eventually converges to a single peak like one of those shown in Figure 3-22c, the landscape again becomes smooth so that highly exploitive operators are expected to again be preferred.

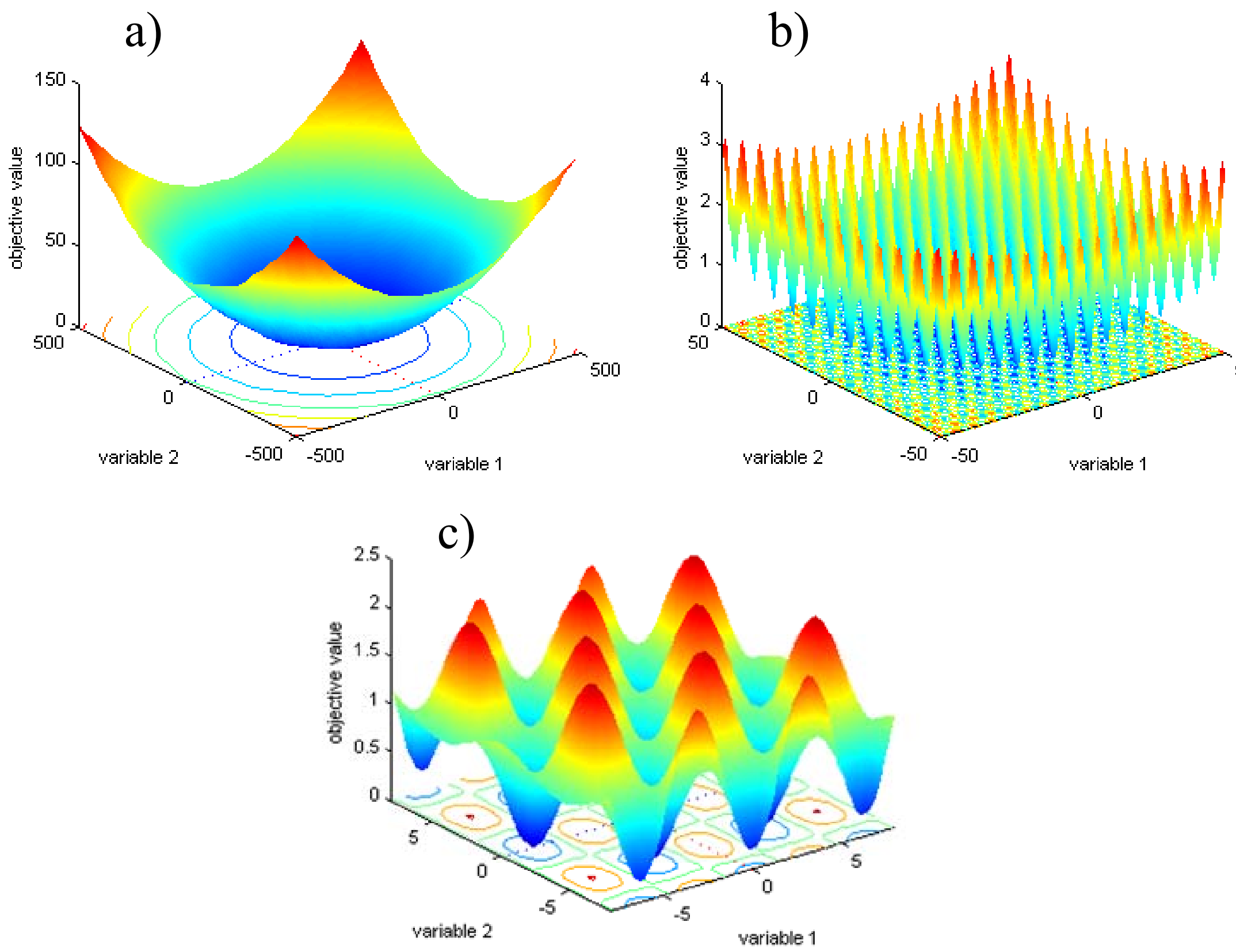

**Figure 3-22 Griewangk fitness landscape shown in two dimensions of parameter space. a) The landscape is shown for variable 1 and variable 2 varying over their complete range [-500,500]. b) The landscape is shown for variable 1 and variable 2 varying over the range [-50,50]. c) The landscape is shown for variable 1 and variable 2 varying over the range [-8,8]. The vertical axis shows the objective function value (minimization) with the global optimal solution located at the origin of parameter space. Images were kindly provided by Hartmut Pohlheim and were generated using the GEATbx toolbox in Matlab® [168]. Low resolution images can also be found at http://www.geatbx.com/docu/fcnindex-01.html#P85_2637.**

As anticipated, many of the adaptive methods initially prefer Wright's heuristic crossover which indicates an ability to exploit global landscape features (This is most notably seen in the ETV-Outlier method). However, most adaptive methods are not able to distinguish between search operators throughout the rest of the run. But notice in Figure 3-23 that most adaptive methods stop improving within the first 500 generations meaning that operator adaptation was only significant to performance during this initial phase of evolution.

Although speculative, it is possible that the ETV-Outlier adaptive method was best able to initially exploit global landscape characteristics but that this also helped to expedite population convergence and ultimately was detrimental to final performance of the algorithm. Considering that adaptation can only take into account performance data over short time scales, one might suspect that for some problems, an effective adaptive mechanism can actually impair algorithm performance. For the Griewangk test function, some of the least adaptive methods (e.g. Static-Ops10, I(rank), ETV) also have the best performance which seems to support this claim.

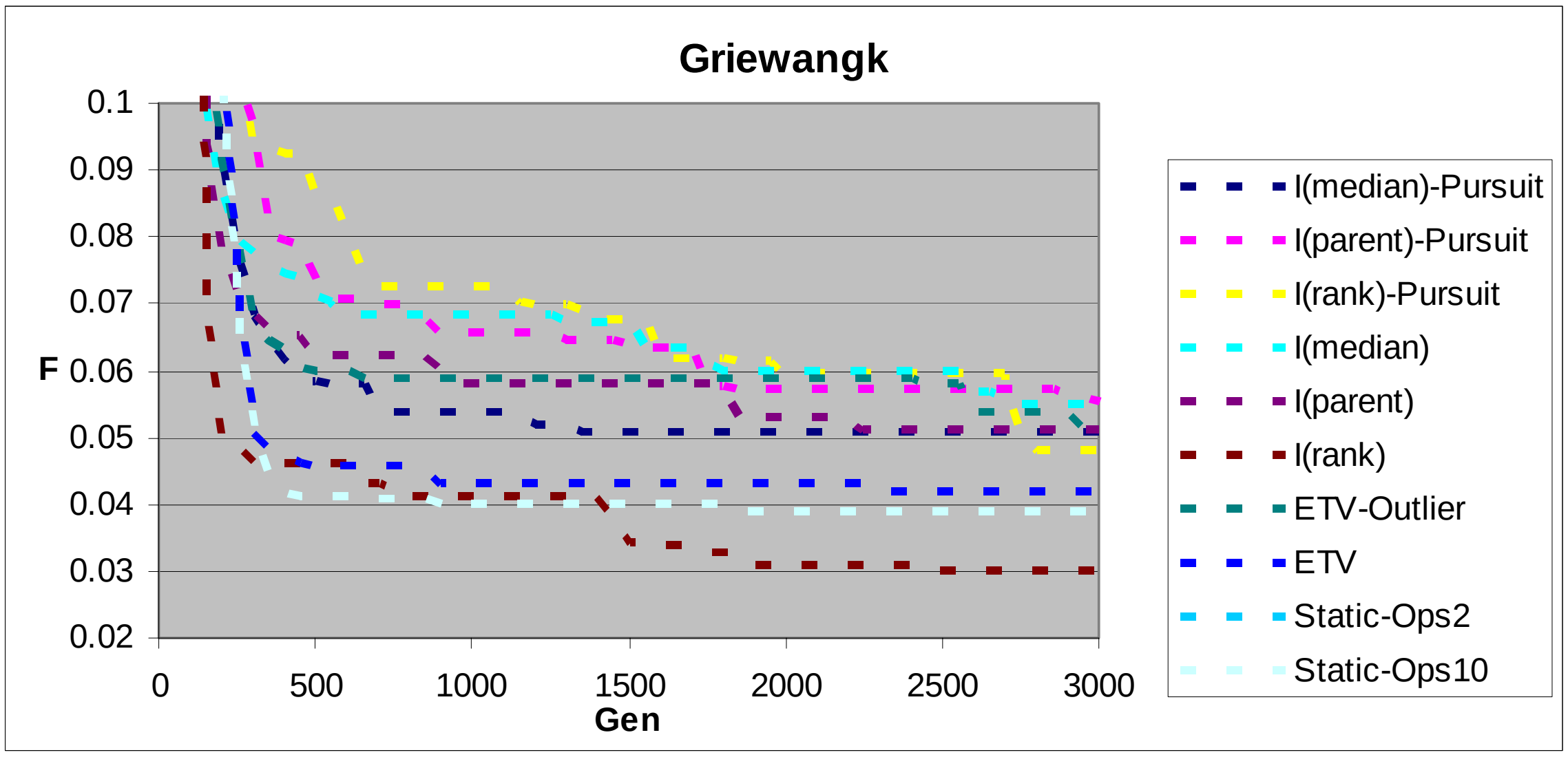

**Figure 3-23 Performance of adaptive and non-adaptive EA designs on the Griewangk test function. The global optimal solution is at *F*=0.**

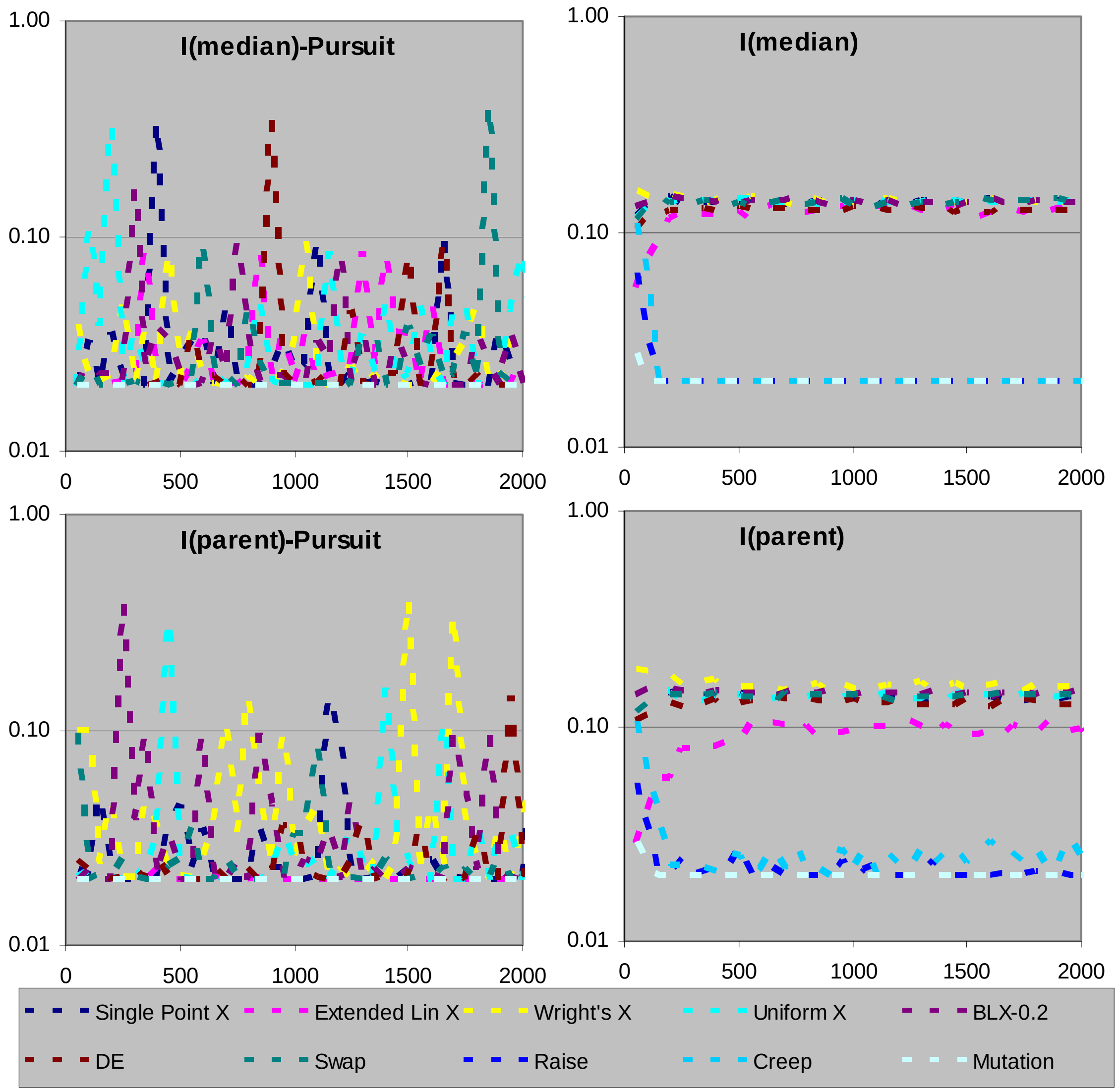

**Figure 3-24 Search operator probability profiles for adaptive methods I(median)-Pursuit, I(median), I(parent)-Pursuit, and I(parent) on the Griewangk test function. Probability values are shown on a logarithmic scale over the first 2000 generations of evolution.**

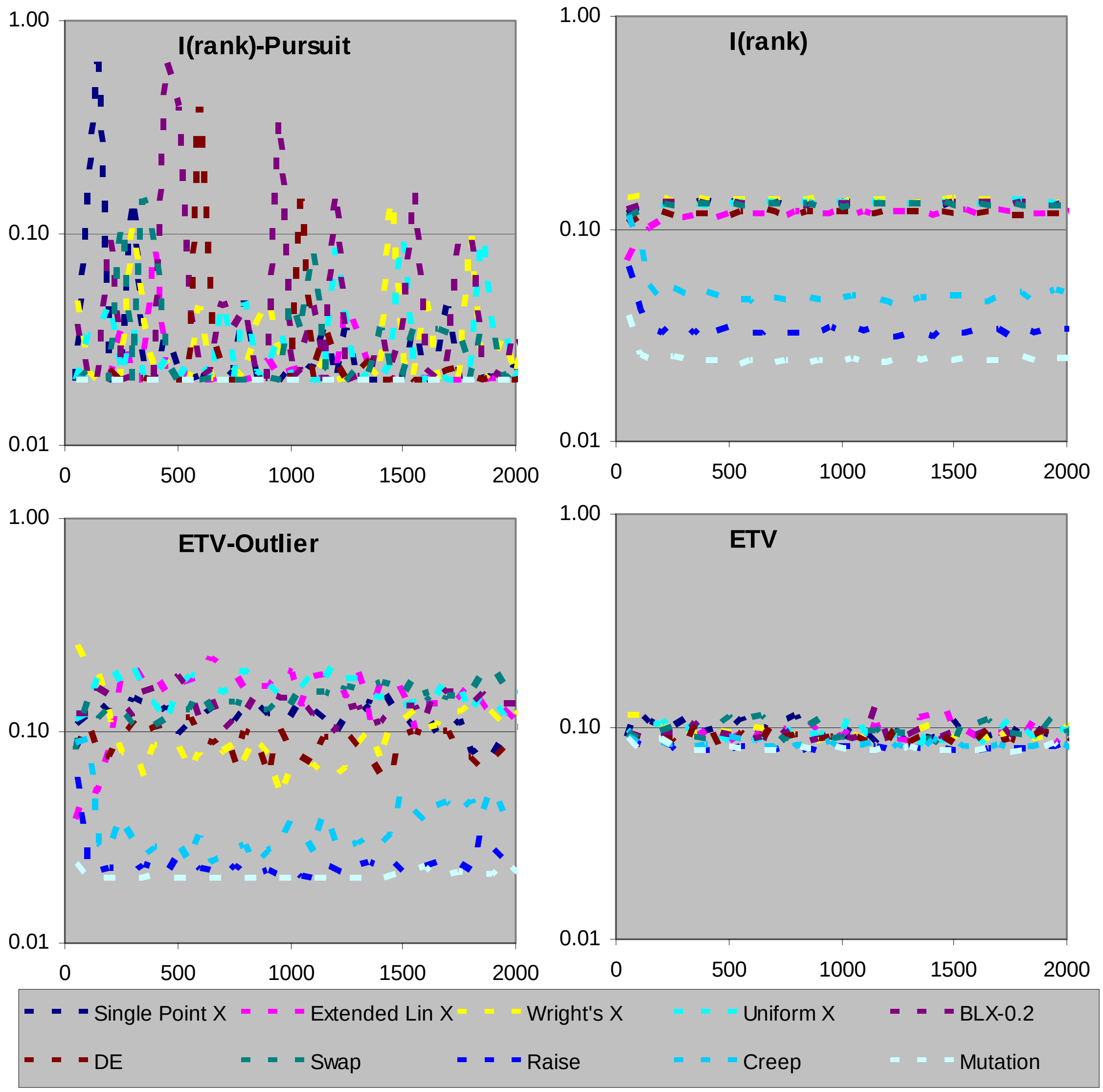

**Figure 3-25 Search operator probability profiles for adaptive methods I(rank)-Pursuit, I(rank), ETV-Outlier, and ETV on the Griewangk test function. Probability values are shown on a logarithmic scale over the first 2000 generations of evolution.**

### *3.3.2.2.4 Ackley's Path Function*

With Ackley's path function, the fitness landscapes is smooth on a global scale as seen in Figure 3-26a, however once within the region containing the global optima, the landscape becomes increasingly rugged as seen in Figure 3-26b. As a result, one might expect some exploitive search operators would be initially beneficial, however as the basin of attraction for the global optima is discovered, more explorative search operators would then become more useful. Also, since a large attractor is located in the center of parameter space, one would expect that an interpolation operator like Wright's heuristic crossover will be initially very effective.

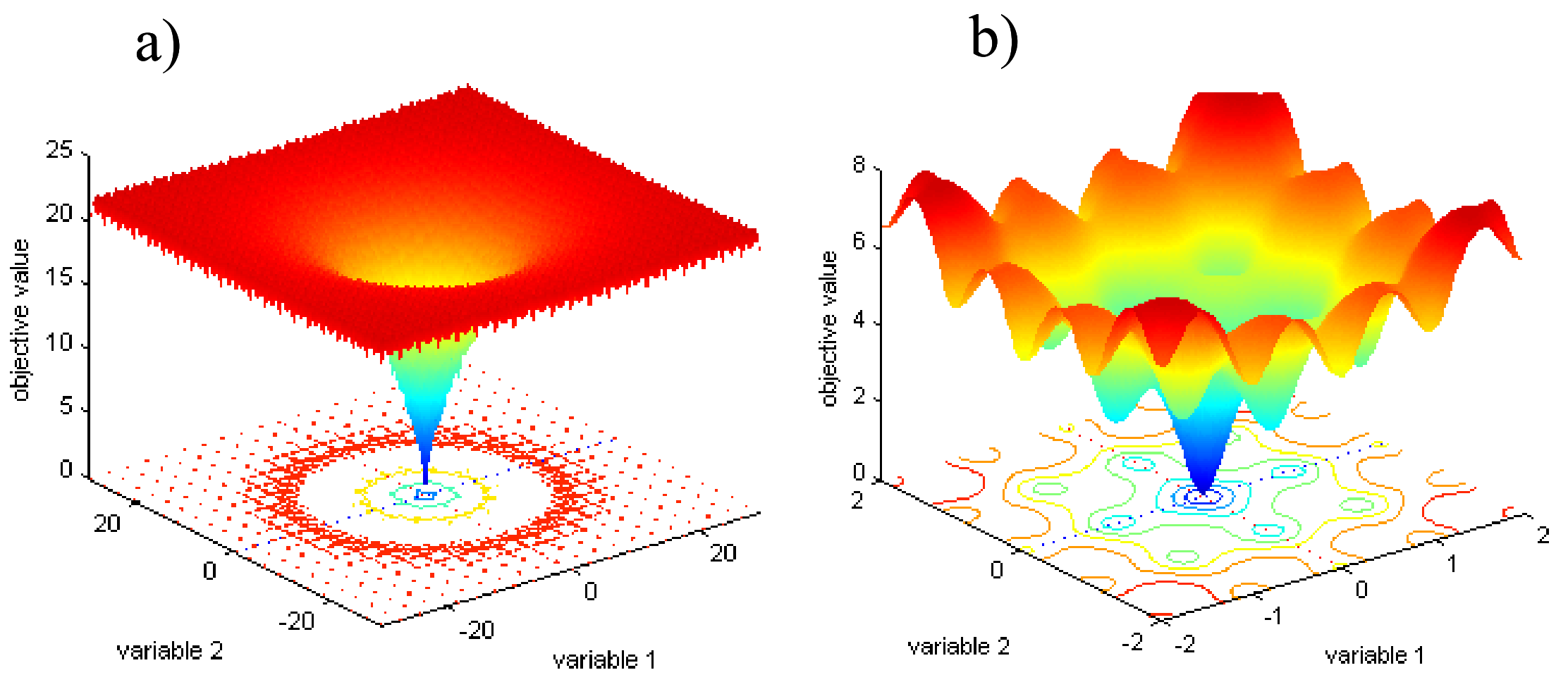

**Figure 3-26 Ackley's fitness landscape shown in two dimensions of parameter space. a) The landscape is shown for variable 1 and variable 2 varying over their complete range [-30,30]. b) The landscape is shown for variable 1 and variable 2 varying over the range [-2,2]. The vertical axis shows the objective function value (minimization) with the global optimal solution located at the origin of parameter space. Images were kindly provided by Hartmut Pohlheim and were generated using the GEATbx toolbox in Matlab® [168]. Low resolution images can also be found at http://www.geatbx.com/docu/fcnindex-01.html#P85_2637.**

For the three best adaptive methods shown in Figure 3-27, I(median)-pursuit, I(median), ETV-Outlier, and to a lesser extend with I(rank)-pursuit, there appears to be a very brief initial phase (20 to 100 generations) where an exploitive operator is preferred (either differential evolution or Wright's heuristic crossover). The fact that this period is extremely brief is not surprising since all methods reach objective function values of $F < 10$ in the first 50 generations which indicates that the population has already converged to the parameter region shown in Figure 3-26b. Next, a slightly longer phase (200 to 400 generations) is observed where more explorative gene swapping operators are preferred. Finally, for the last 500 to 700 generations before reaching the global optimal solution, the best adaptive methods again prefer highly exploitive search operators (Wright's heuristic crossover, extended line crossover, differential evolution). These changes in search operator preferences during evolution are believed to accurately reflect changes in the environment and are most clearly seen in the adaptive methods I(median) and ETV-Outlier.

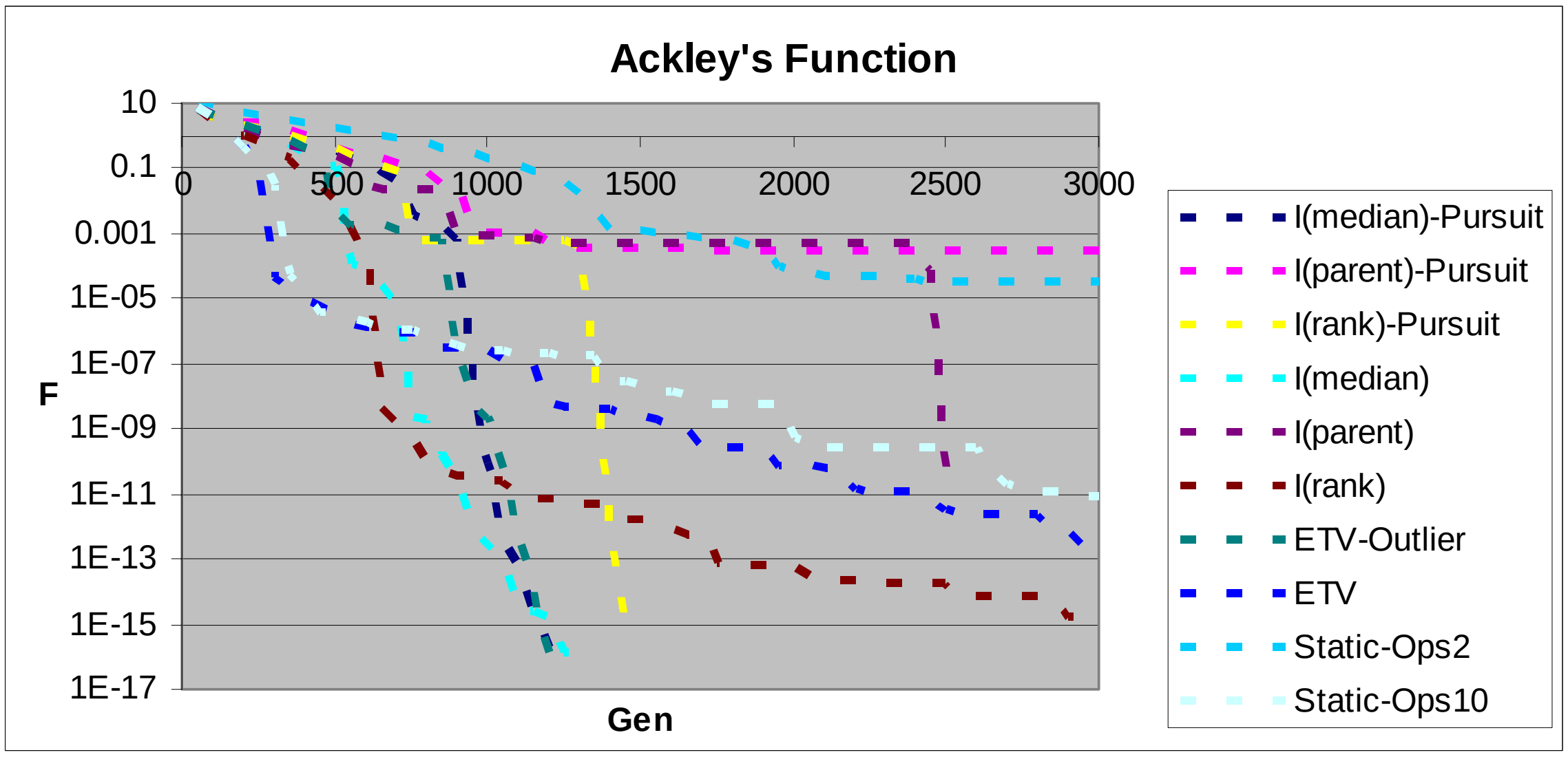

**Figure 3-27 Performance of adaptive and non-adaptive EA designs on Ackley's Path Function. The global optimal solution is at *F*=0. The optimal *F* value can not be shown due to log scaling on the *F* axis so performance profiles are seen to terminate when the global optima is reached.**

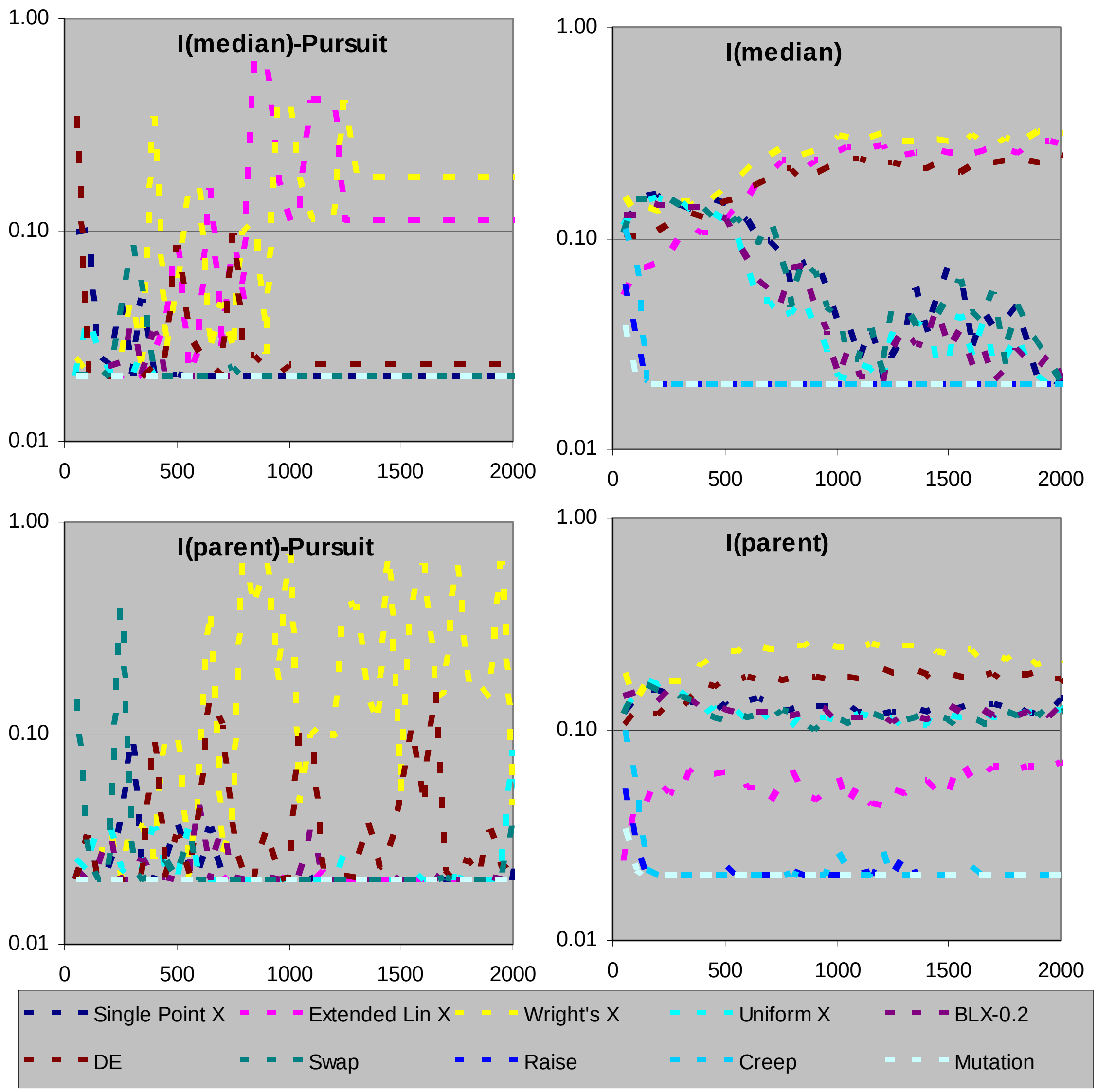

**Figure 3-28 Search operator probability profiles for adaptive methods I(median)-Pursuit, I(median), I(parent)-Pursuit, and I(parent) on Ackley's test function. Probability values are shown on a logarithmic scale over the first 2000 generations of evolution.**

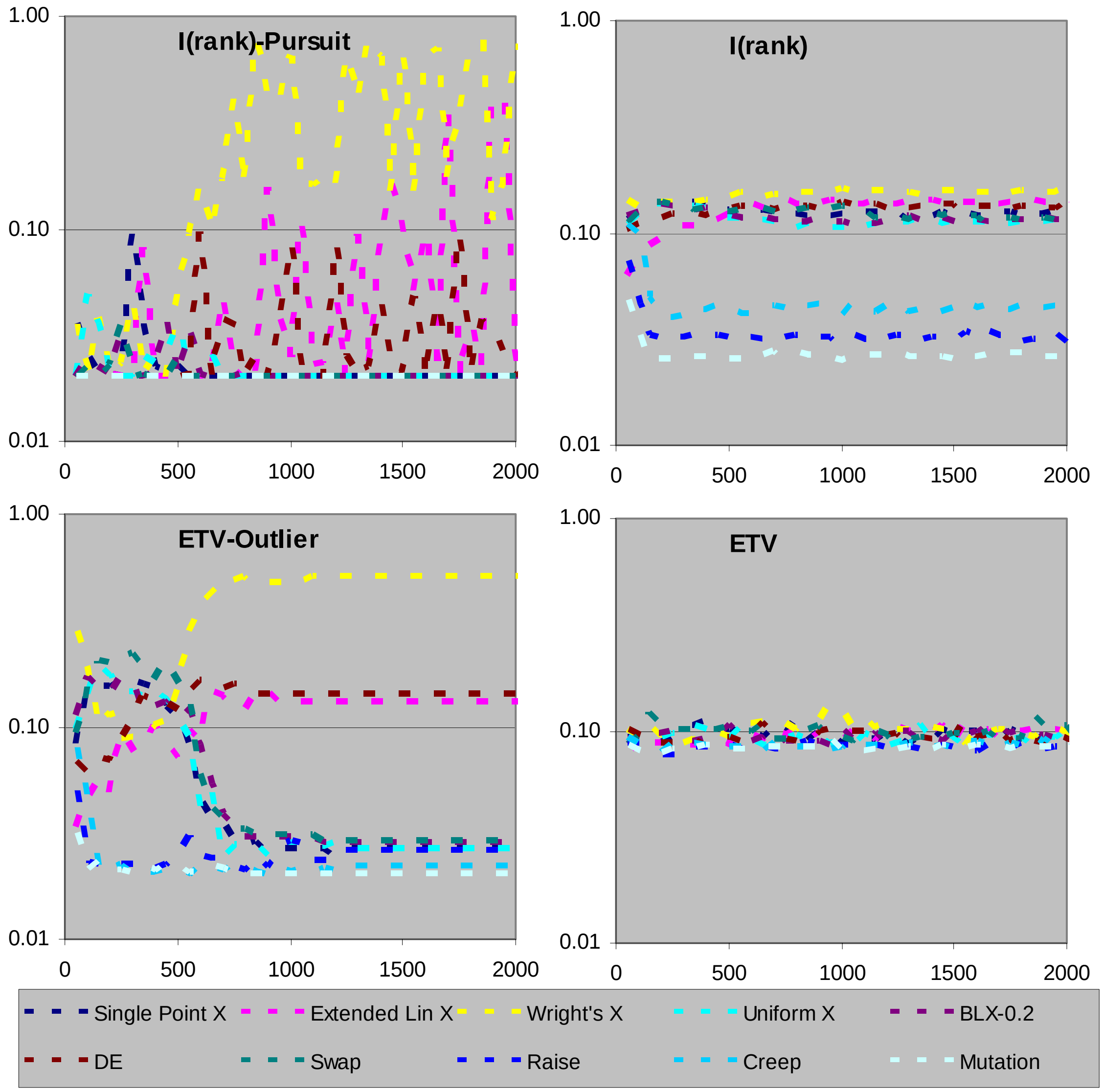

**Figure 3-29 Search operator probability profiles for adaptive methods I(rank)-Pursuit, I(rank), ETV-Outlier, and ETV on Ackley's test function. Probability values are shown on a logarithmic scale over the first 2000 generations of evolution.**

### 3.3.2.3 Performance Results on Artificial Test Functions

For many of the artificial test functions presented in this section, the ETV-Outlier adaptive method is found to perform strongly throughout the 3000 generations considered. Performance statistics provided in Table 3-6 also demonstrate superior final performance from the ETV-Outlier method for this set of test functions. Even for problems where the method does not clearly dominate, it generally was able to perform at least as well as the other EA designs. It is interesting to note that ETV-Outlier shows its worst performance on the most deceptive problem, MMDP. Performance graphs are presented roughly in the order of best to worst performance for the ETV-Outlier method.

The non-adaptive EA with two search operators, Static-Ops2, is often found to be significantly worse than all other algorithms. Also, it is interesting to note that the adaptive method ETV has nearly identical performance to Static-Ops10 for every problem. This result should not be surprising considering that ETV (without Outlier interpretation) has little ability to distinguish between search operators, as was indicated in the previous section.

**Table 3-6 Overall performance statistics for each of the adaptive and non-adaptive EA designs run on the artificial test functions. Column two measures the percentage of problems where an EA design was the best EA design (comparisons based on median objective function value). Column three measures the percentage of problems where an EA design was able to find the best solution at least one time. The best solution is defined as the best found in these experiments and is not necessarily the global optimal solution. Results for the non-adaptive EA designs are shown in the bottom two rows while the rows labeled as ETV and ETV-Outlier show results for the new adaptive methods developed in this thesis.**

| EA Design | % of problems where EA | |
|---|---|---|
| | was best design | found best |
| I(median)-Pursuit | 16.7% | 46.2% |
| I(parent)-Pursuit | 2.6% | 38.5% |
| I(rank)-Pursuit | 12.8% | 38.5% |
| I(median) | 15.4% | 61.5% |
| I(parent) | 0.0% | 38.5% |
| I(rank) | 7.7% | 46.2% |
| ETV-Outlier | **29.5%** | **92.3%** |
| ETV | 7.7% | 30.8% |
| Static-Ops2 | 0.0% | 7.7% |
| Static-Ops10 | 7.7% | 38.5% |

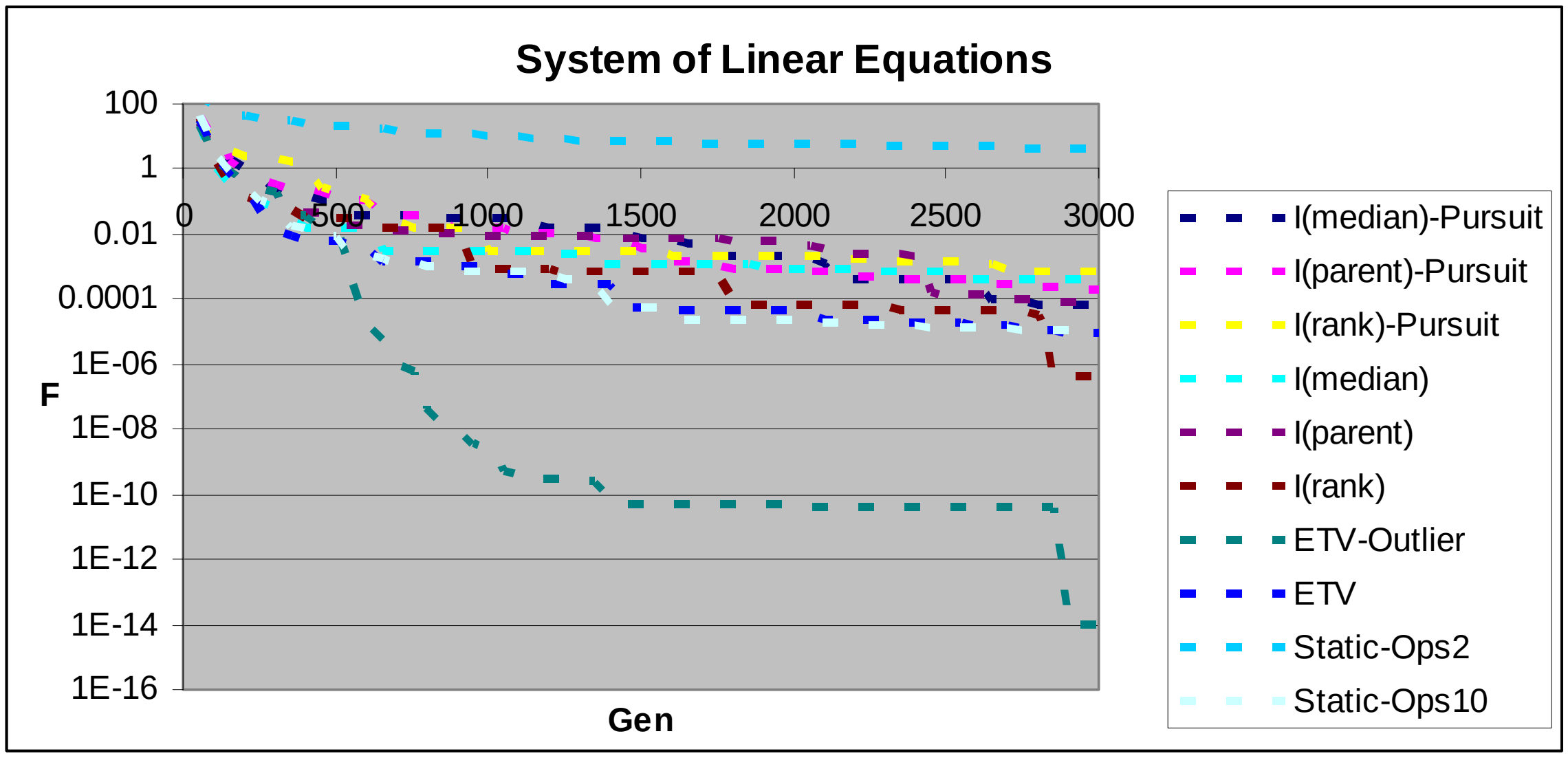

**Figure 3-30 Performance of adaptive and non-adaptive EA designs on the System of Linear Equations test function. The global optimal solution is at *F*=0.**

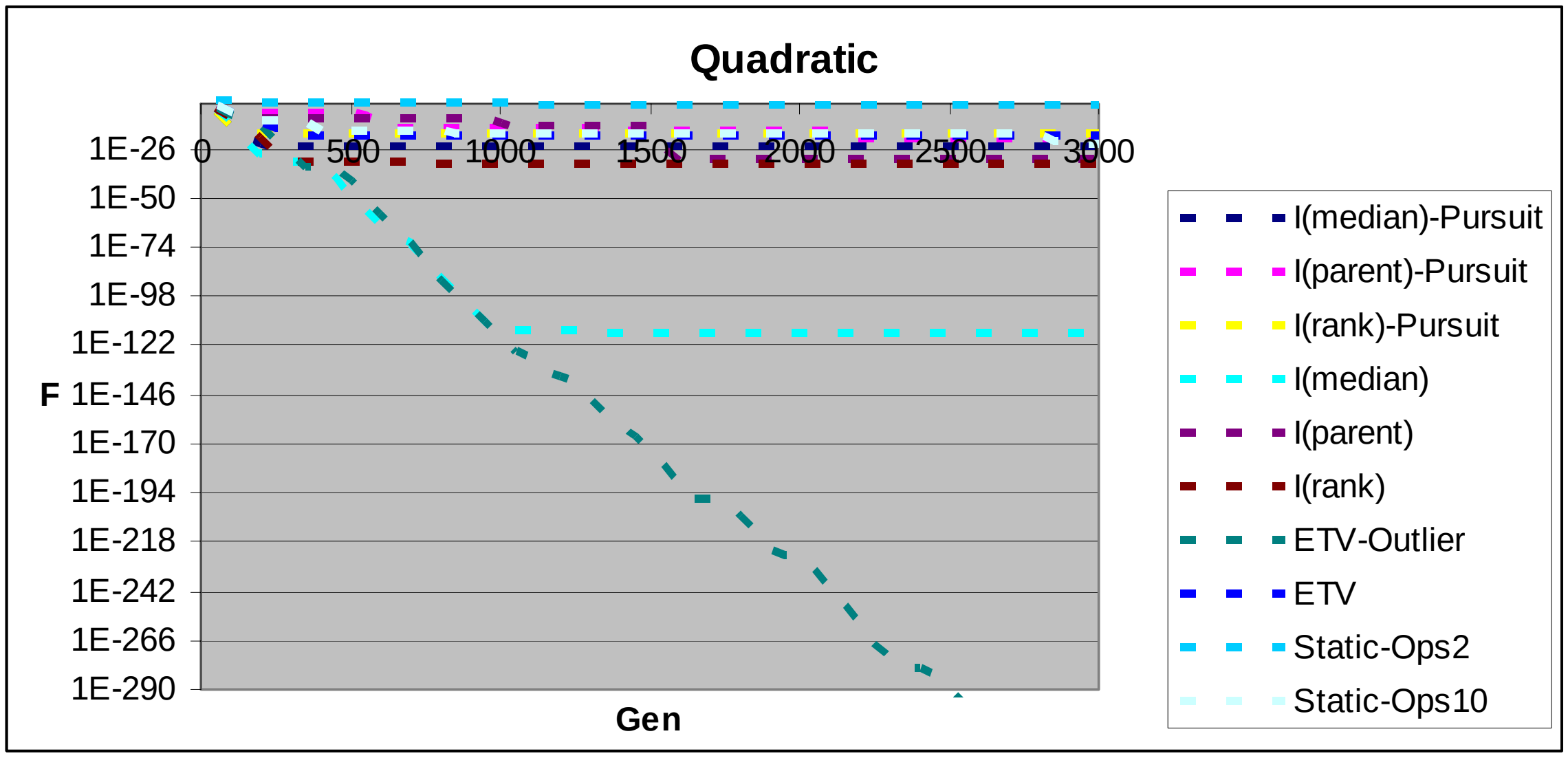

**Figure 3-31 Performance of adaptive and non-adaptive EA designs on the Quadratic test function. The global optimal solution is at *F*=0.**

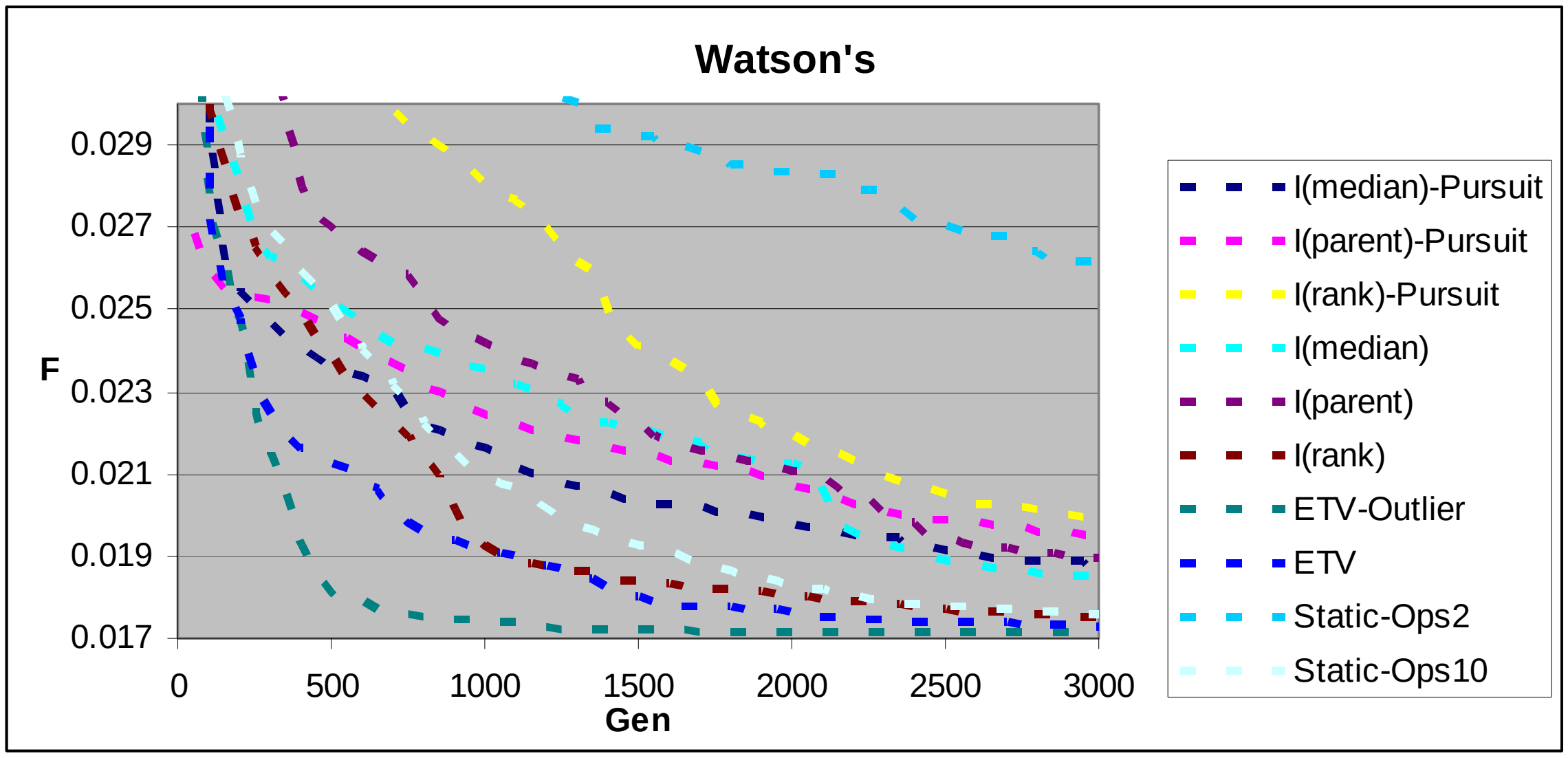

**Figure 3-32 Performance of adaptive and non-adaptive EA designs on Watson's test function. The global optimal solution is at *F*=2.288E-3.**

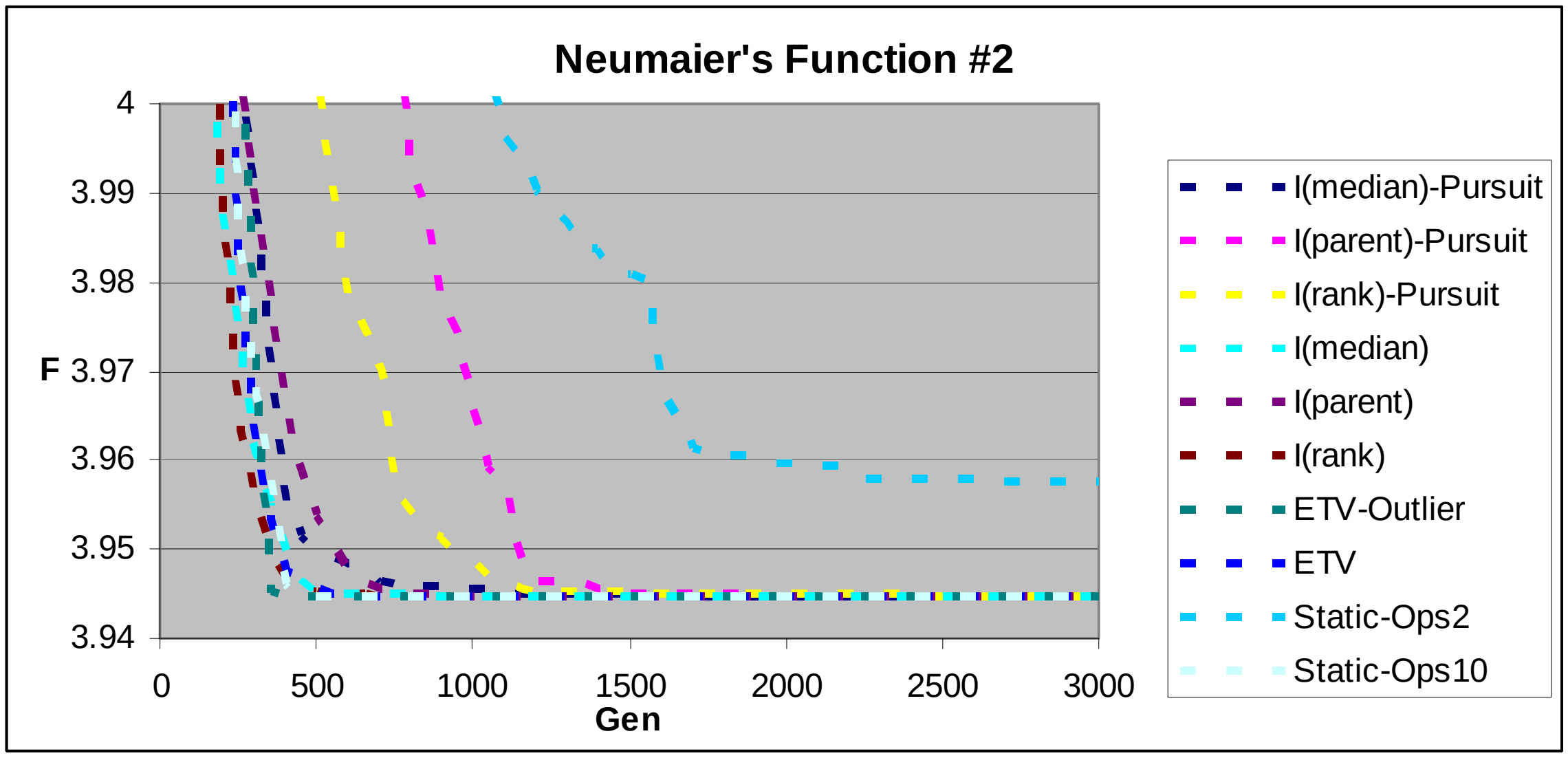

**Figure 3-33 Performance of adaptive and non-adaptive EA designs on Neumaier's function #2. The global optimal solution is unknown (see Appendix A).**

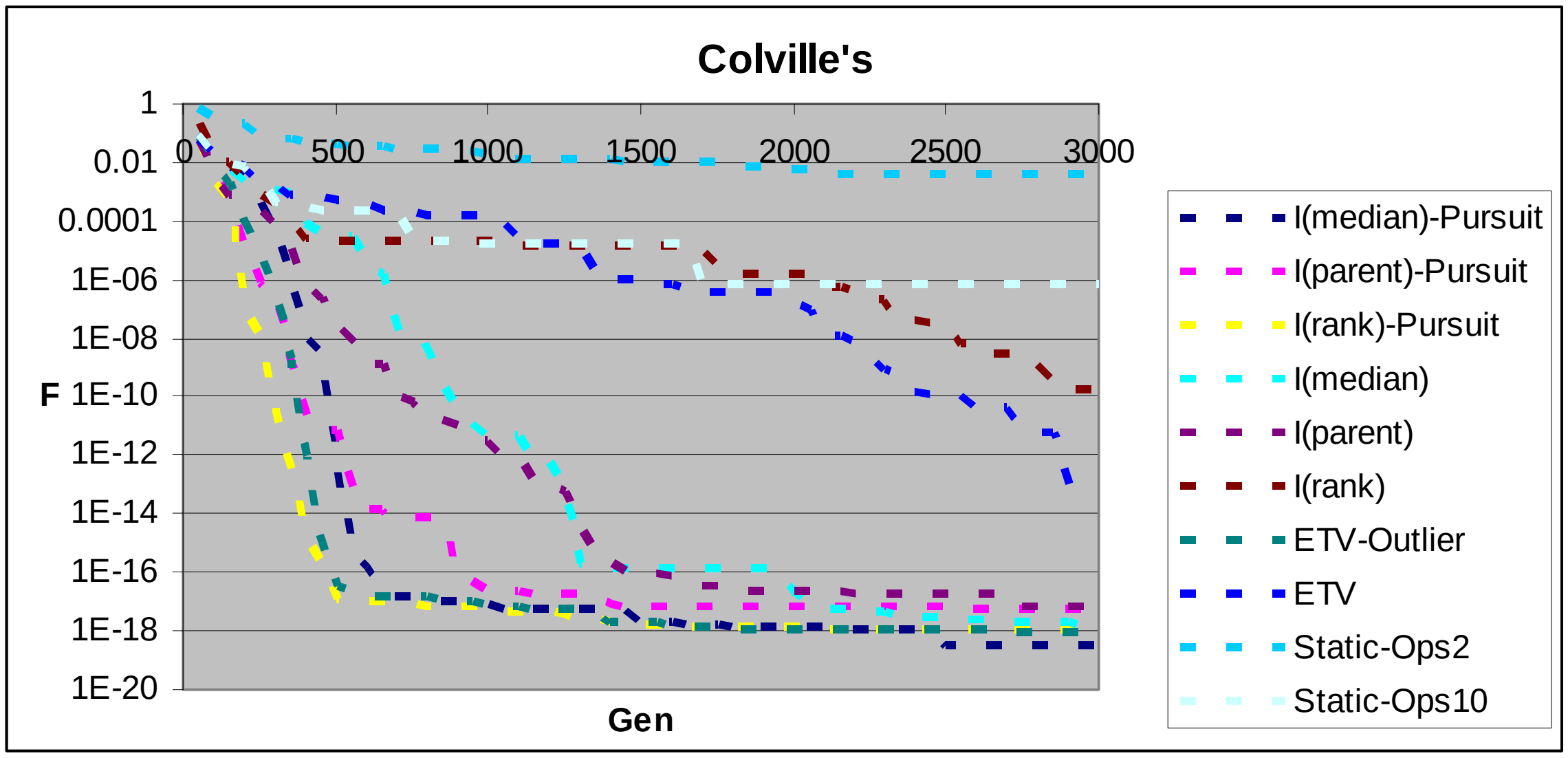

**Figure 3-34 Performance of adaptive and non-adaptive EA designs on Colville's test function. The global optimal solution is at *F*=0.**

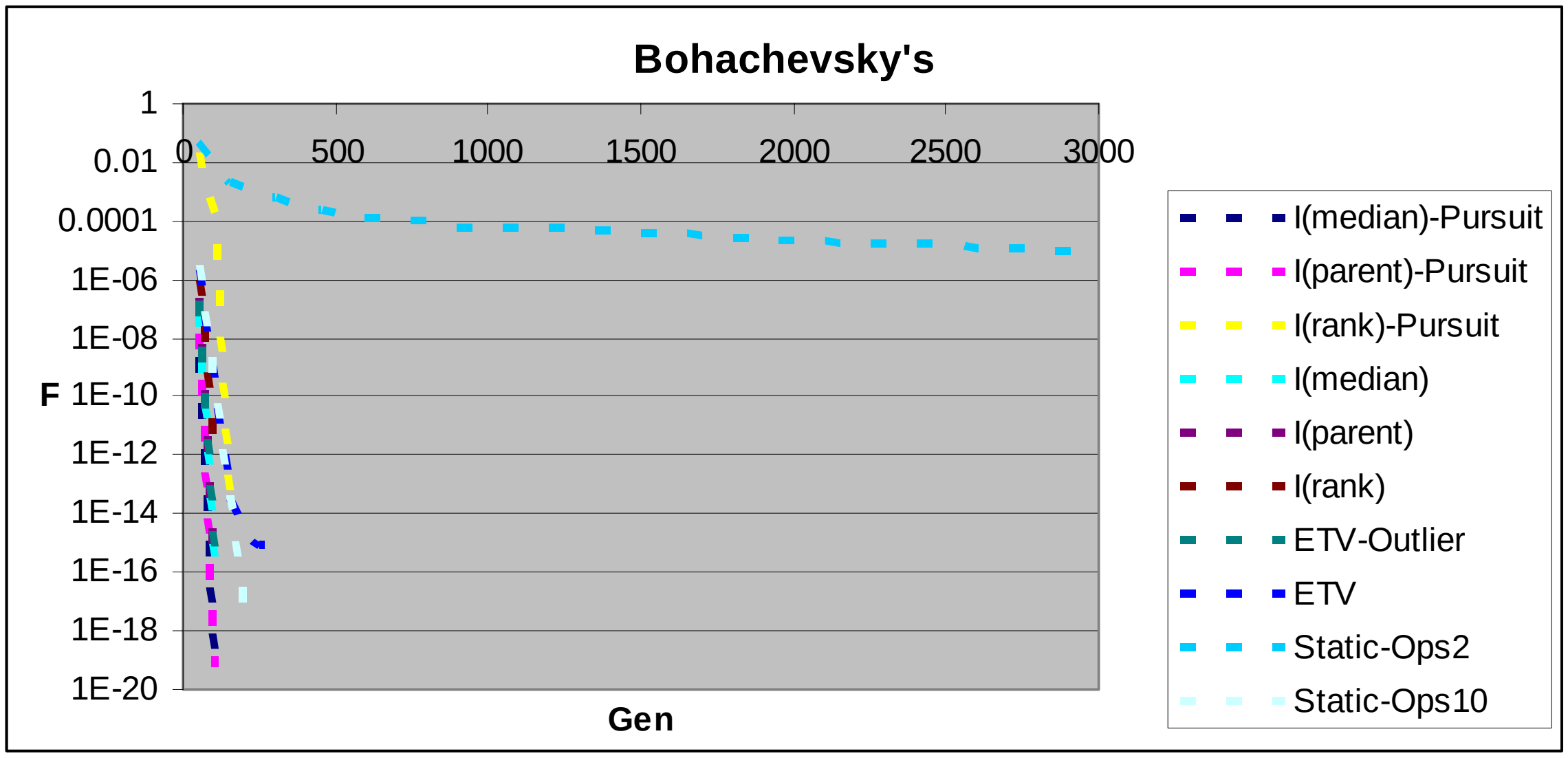

**Figure 3-35 Performance of adaptive and non-adaptive EA designs on Bohachevsky's test function. The global optimal solution is at *F*=0.**

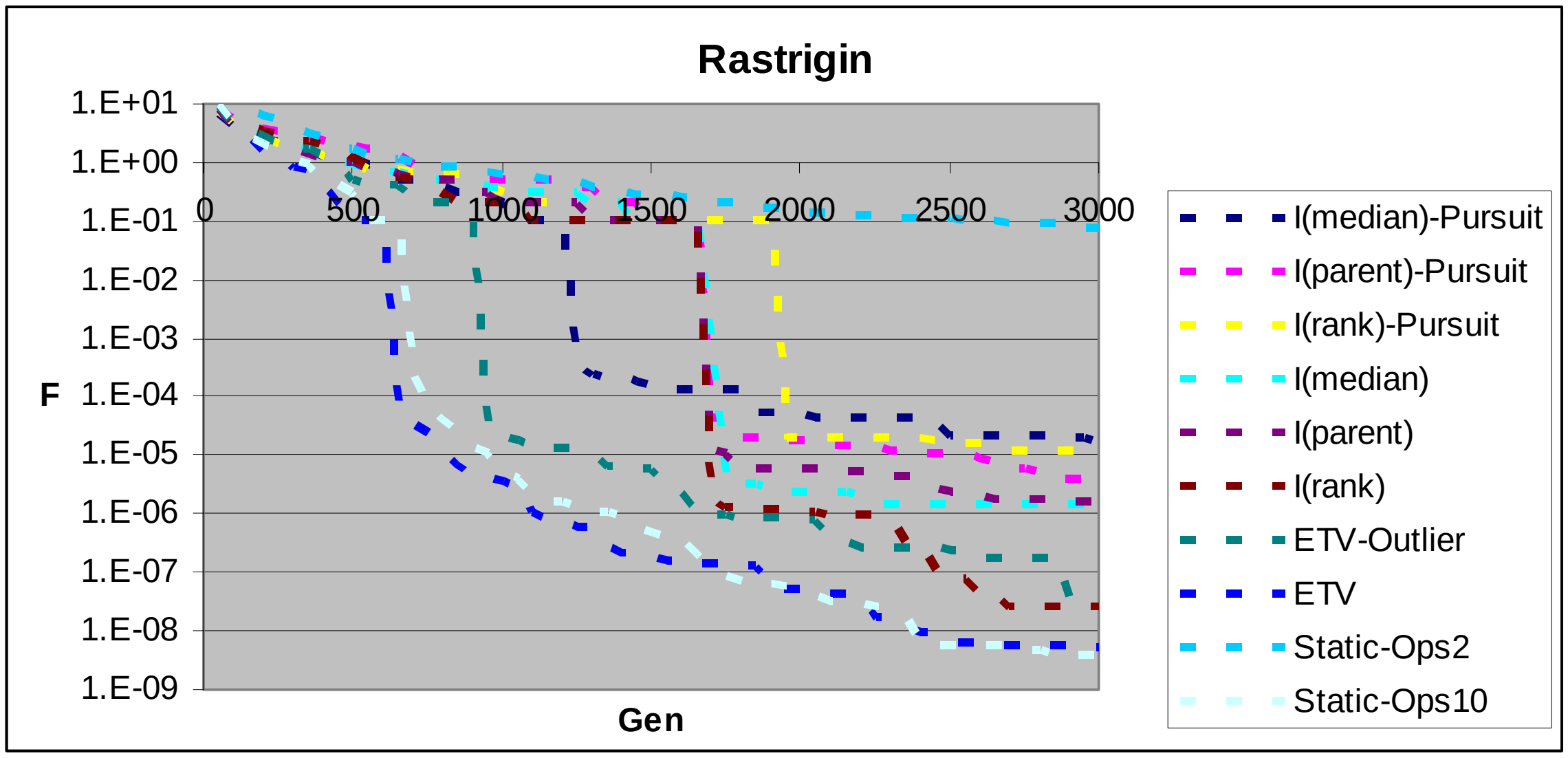

**Figure 3-36 Performance of adaptive and non-adaptive EA designs on the Rastrigin test function. The global optimal solution is at *F*=0.**

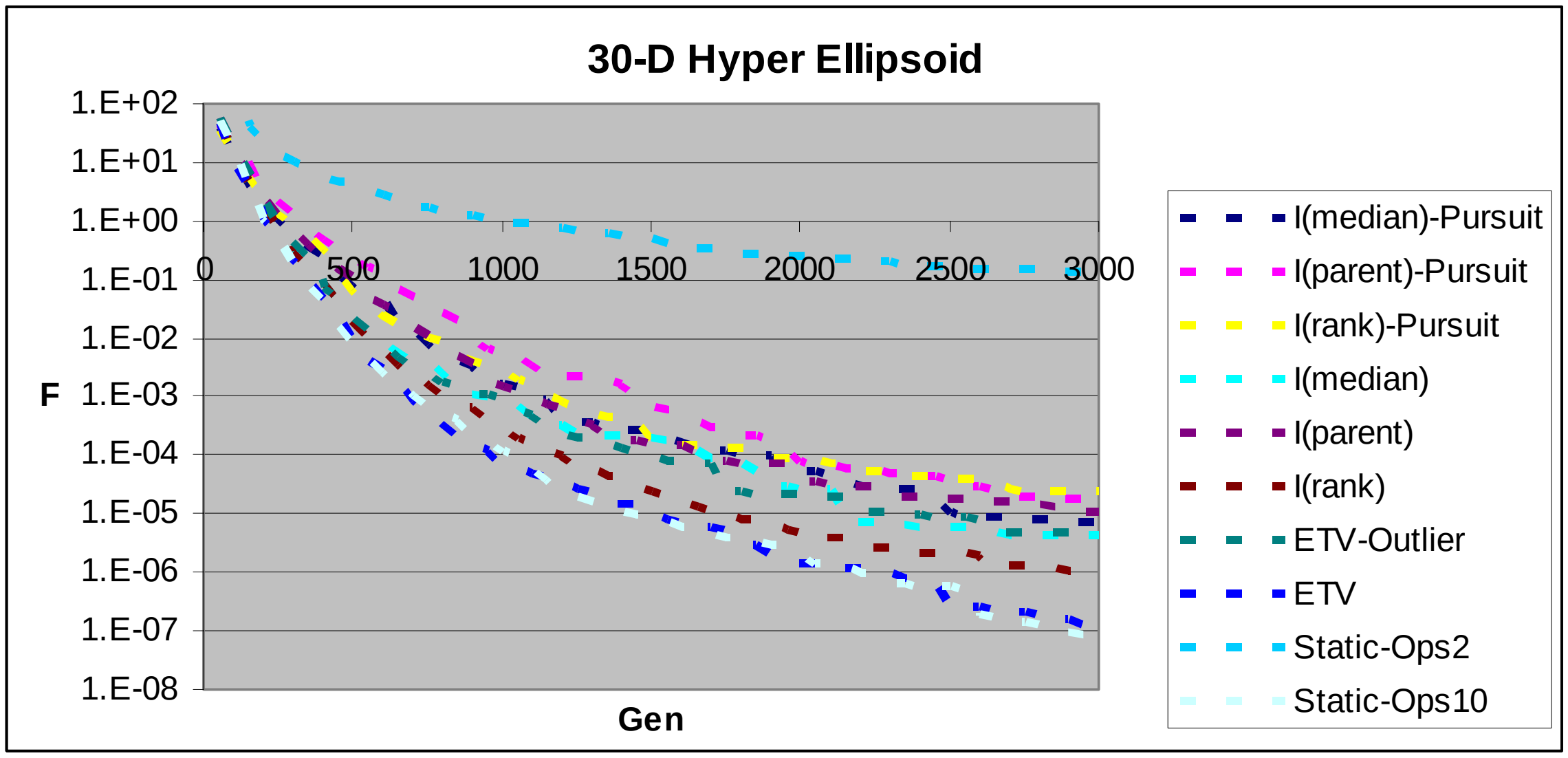

**Figure 3-37 Performance of adaptive and non-adaptive EA designs on the 30-D Hyper Ellipsoid test function. The global optimal solution is at *F*=0.**

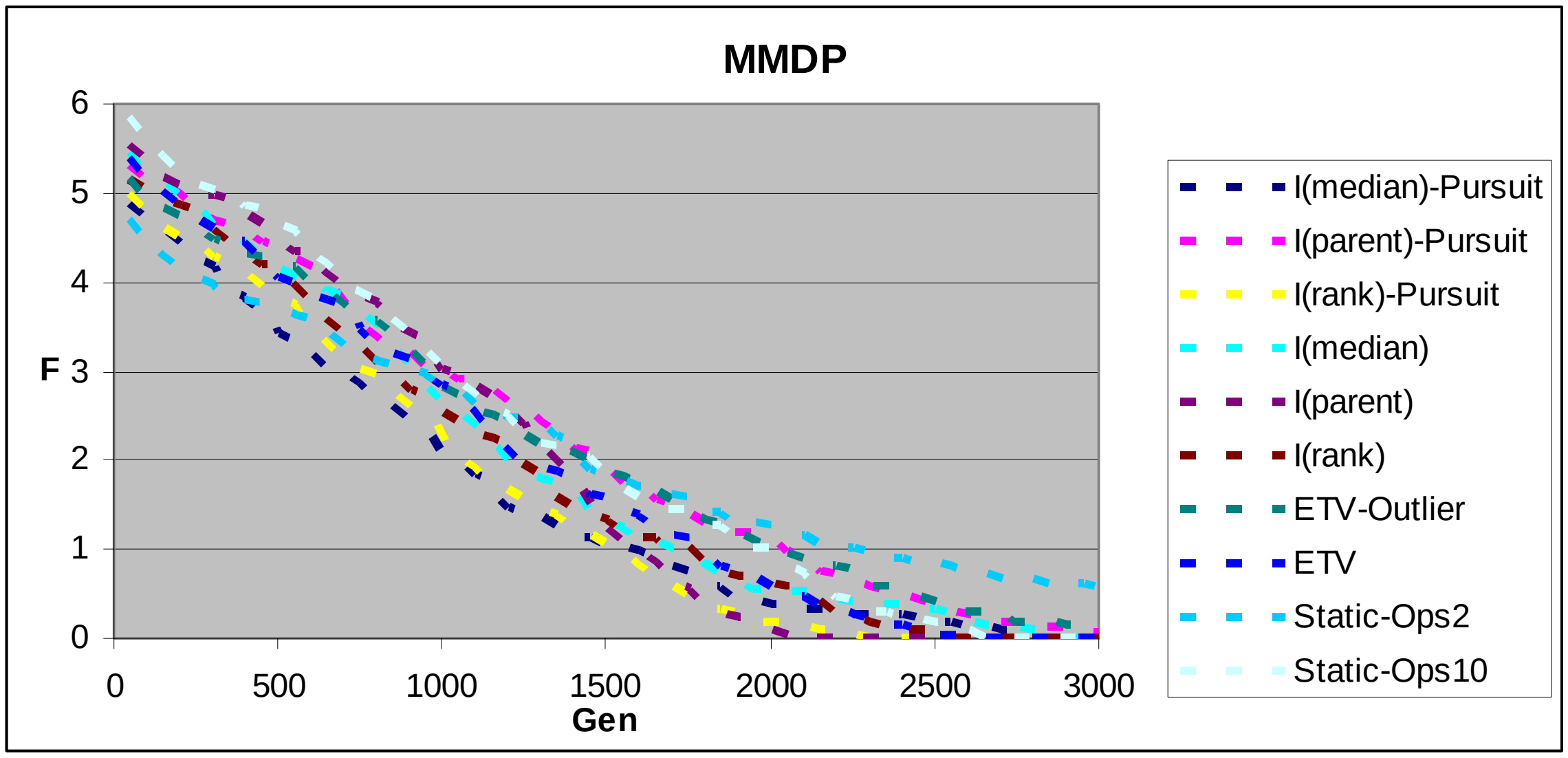

**Figure 3-38 Performance of adaptive and non-adaptive EA designs on the Massively Multimodal Deceptive Problem (MMDP). The global optimal solution is at *F*=0.**

### 3.3.2.4 Performance Results on Engineering Design Problems

Experiments are also conducted on a number of engineering design problems (and other application-inspired test functions) in order to gauge the effectiveness of the adaptive methods on practical optimization problems. Each of these problems are described in detail in Appendix A.

**Table 3-7 Overall performance statistics for each of the adaptive and non-adaptive EA designs run on the engineering design problems. Column two measures the percentage of problems where an EA design was the best EA design (comparisons based on median objective function value). Column three measures the percentage of problems where an EA design was able to find the best solution at least one time. The best solution is defined as the best found in these experiments and is not necessarily the global optimal solution.**

| EA Design | % of problems where EA | |
|---|---|---|
| | was best design | found best |
| I(median)-Pursuit | 0.0% | 28.6% |
| I(parent)-Pursuit | 4.8% | 28.6% |
| I(rank)-Pursuit | 0.0% | 42.9% |
| I(median) | 0.0% | 42.9% |
| I(parent) | **33.3%** | 57.1% |
| I(rank) | 0.0% | 42.9% |
| ETV-Outlier | **33.3%** | **85.7%** |
| ETV | 21.4% | 42.9% |
| Static-Ops2 | 7.1% | 28.6% |
| Static-Ops10 | 0.0% | 57.1% |

The results from these experiments are noticeably different from results on the artificial test functions. Comparing the "best design" column of Table 3-7 and Table 3-6, it is interesting to note that three of the four algorithms that perform best on the artificial test functions are now the worst algorithms for the engineering problems. On the other hand I(parent), which is tied as the worst method in the artificial test functions, is now tied with ETV-Outlier as the best method for the engineering problems. In the face of these strong reversals in algorithm performance, it is worth noting that ETV-Outlier maintains its status as the best algorithm in both sets of test problems. Similar behavior is observed with the performance metric in the third column of the same tables, however the performance reversals are not as pronounced in this case.

It is also worth mentioning that the non-adaptive methods faired better on the engineering problems, especially in their ability to find the best solution to a problem at least one time (i.e. the "found best" metric). In fact, Static-Ops10 is tied for being the second best algorithm for this performance metric.

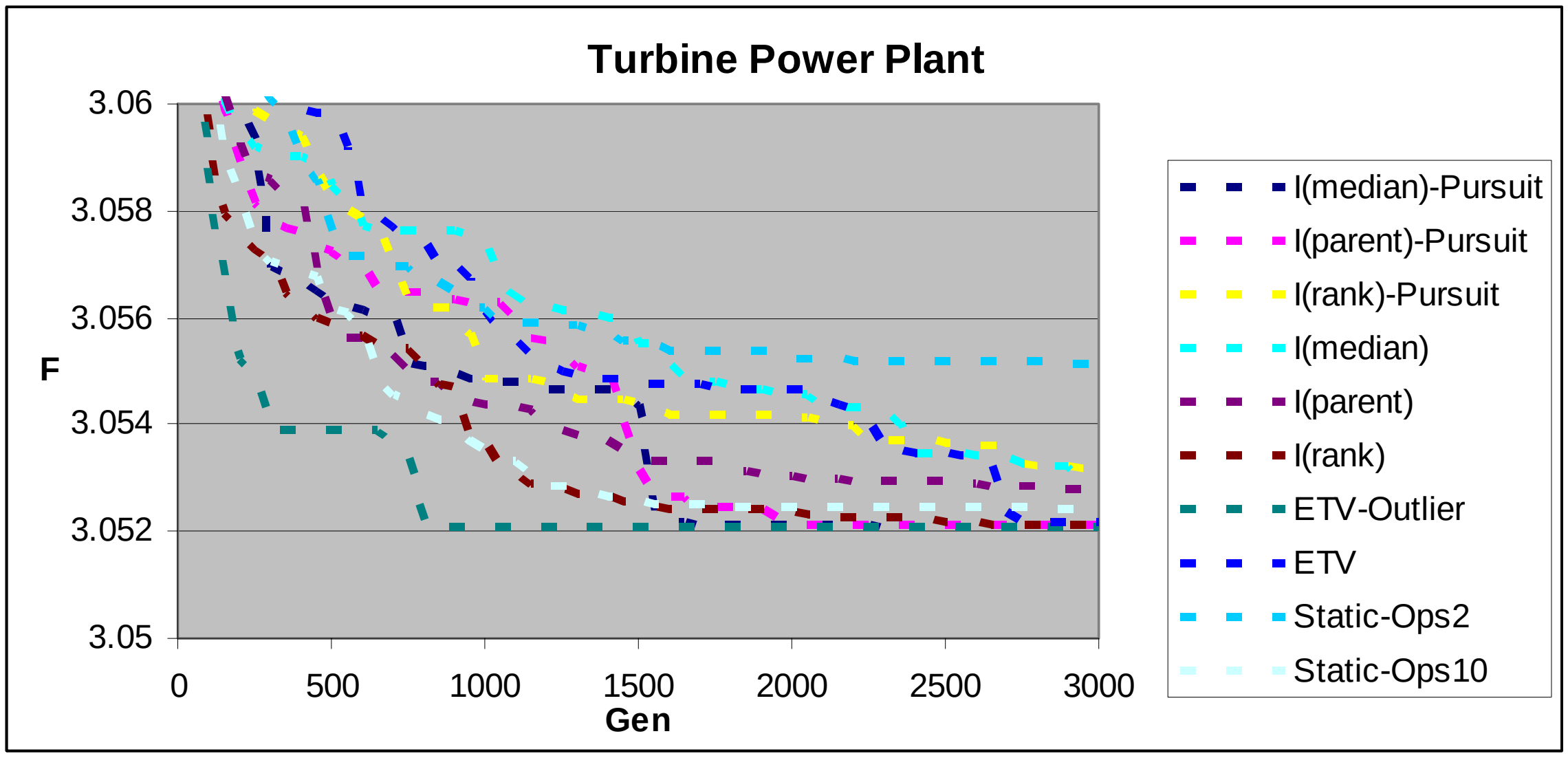

**Figure 3-39 Performance of adaptive and non-adaptive EA designs on the Turbine Power Plant Problem. The global optimal solution is at *F*=3.05.**

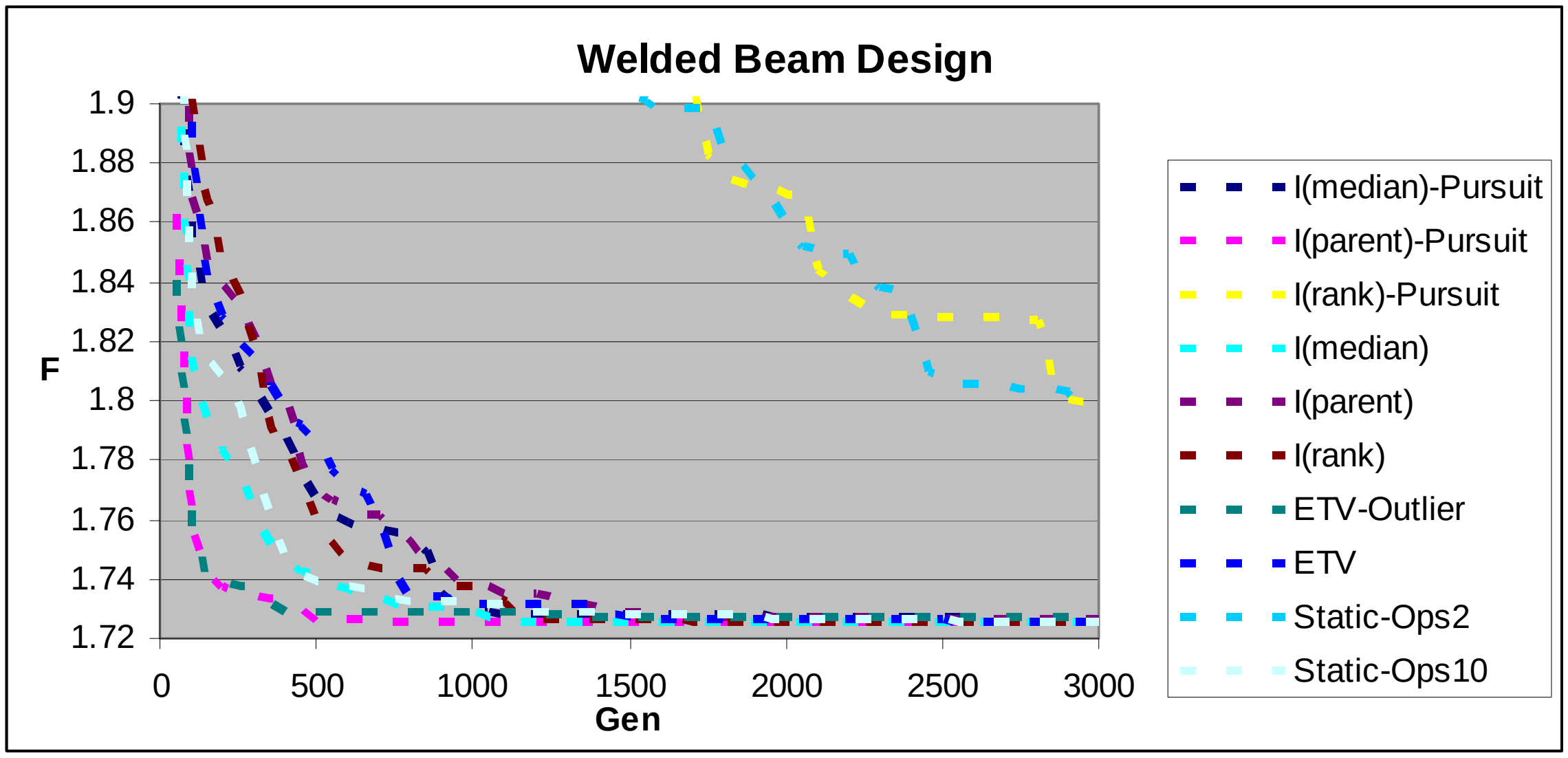

**Figure 3-40 Performance of adaptive and non-adaptive EA designs on the Welded Beam Design problem. The global optimal solution is unknown.**

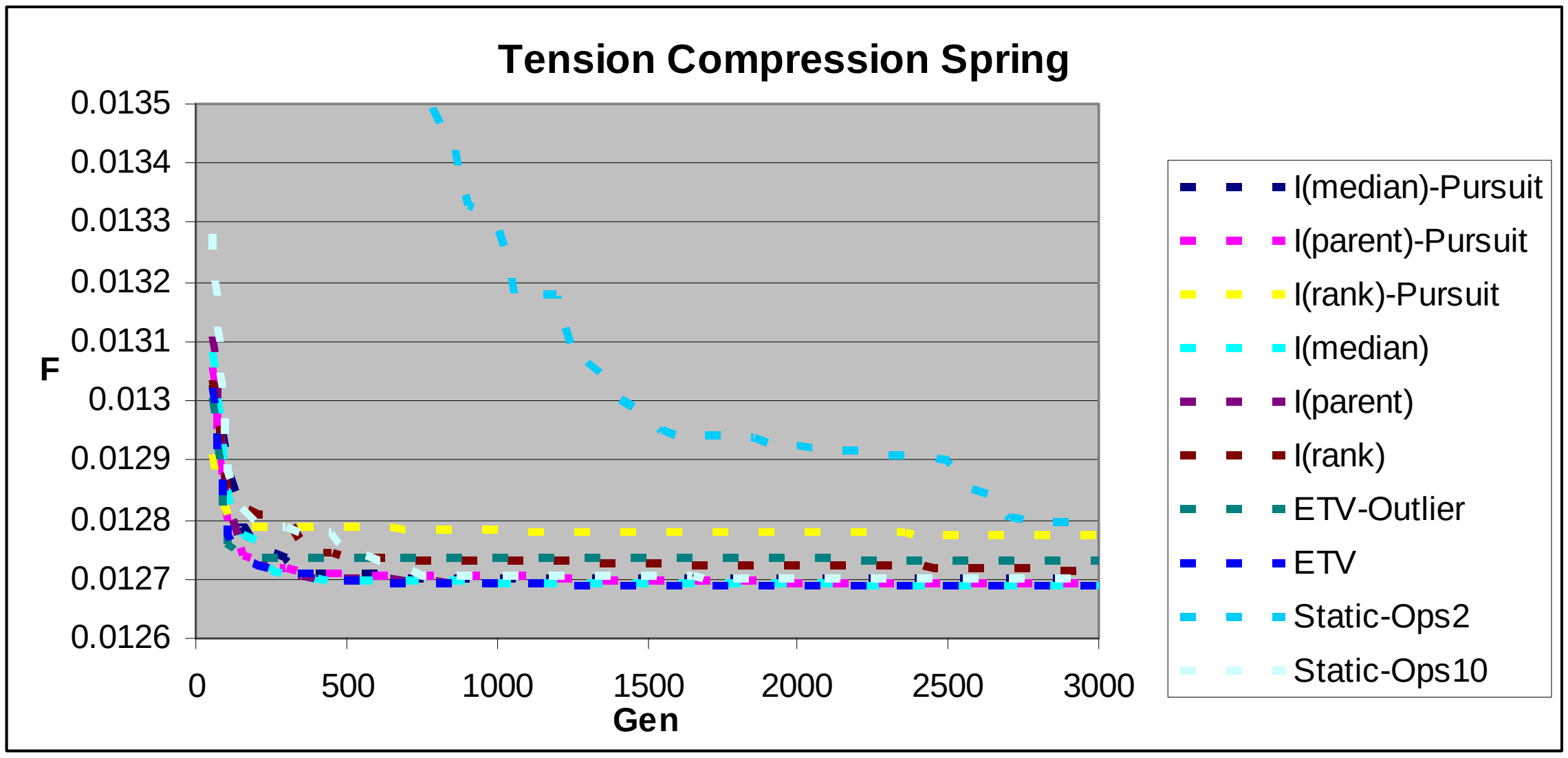

**Figure 3-41 Performance of adaptive and non-adaptive EA designs on the Tension Compression Spring problem. The global optimal solution is unknown.**

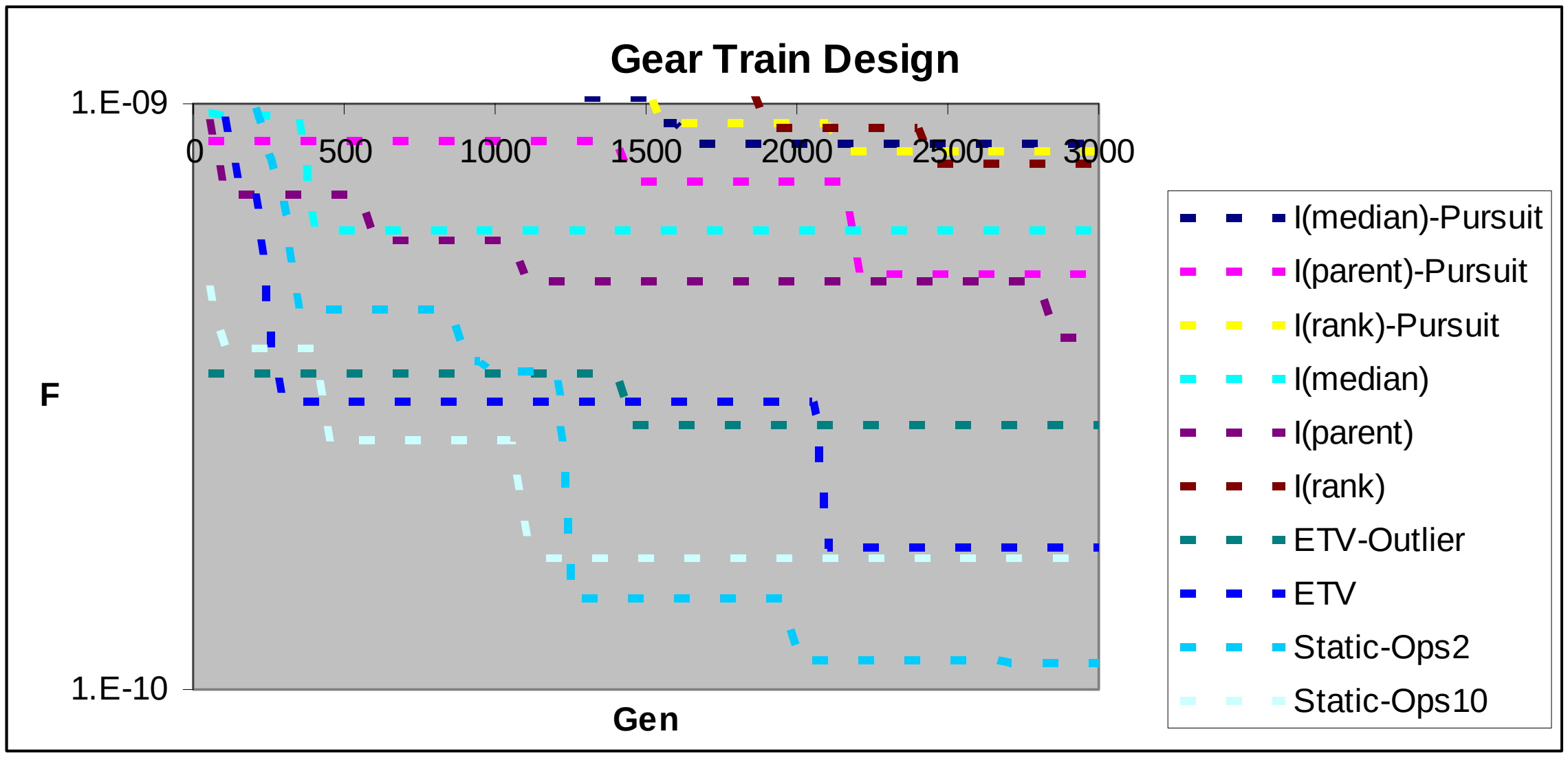

**Figure 3-42 Performance of adaptive and non-adaptive EA designs on the Gear Train Design problem. The global optimal solution is $F$=2.70 x10$^{-12}$.**

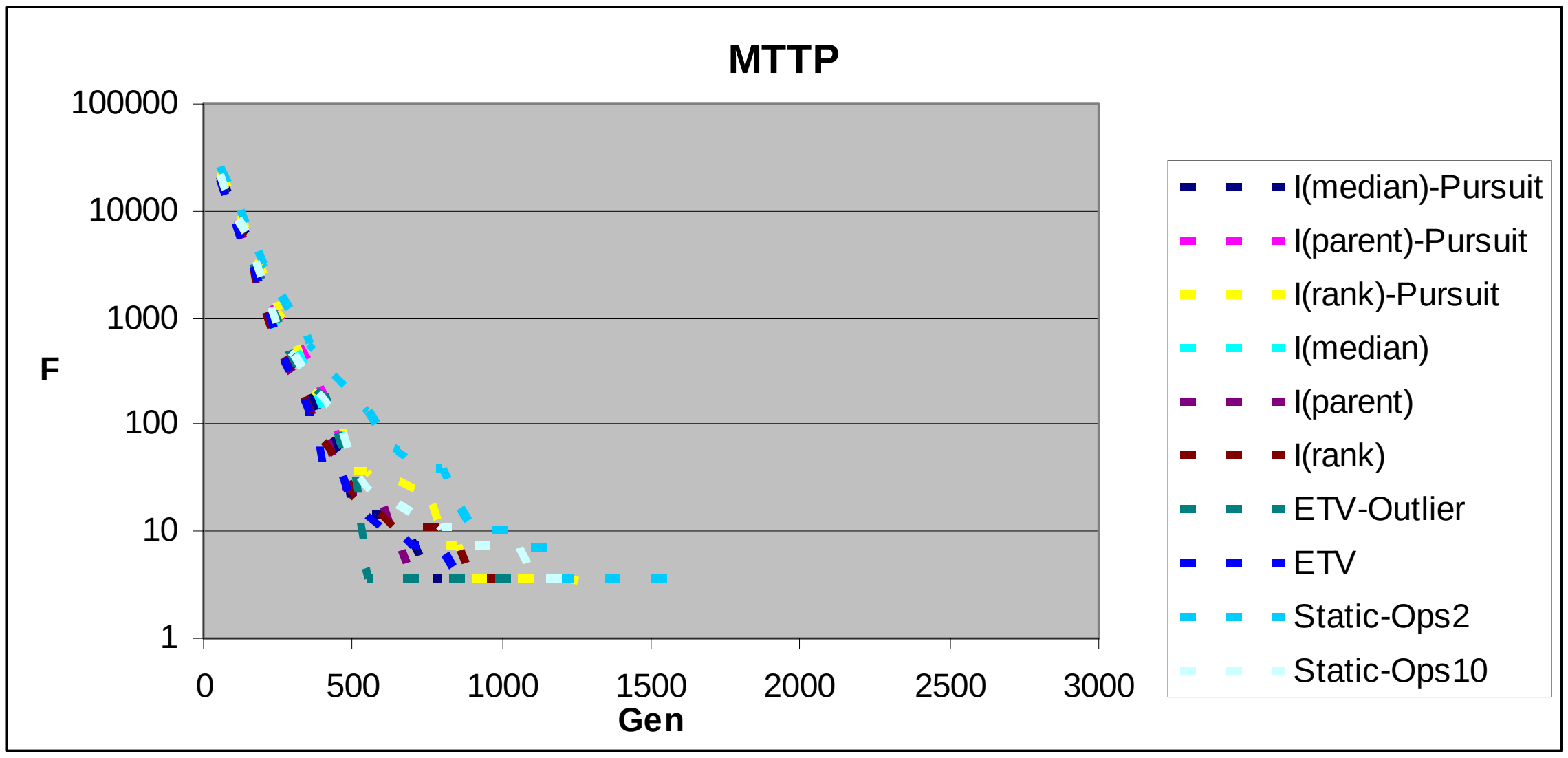

**Figure 3-43 Performance of adaptive and non-adaptive EA designs on the Minimum Tardy Task Problem (MTTP). The global optimal solution is *F*=0.**

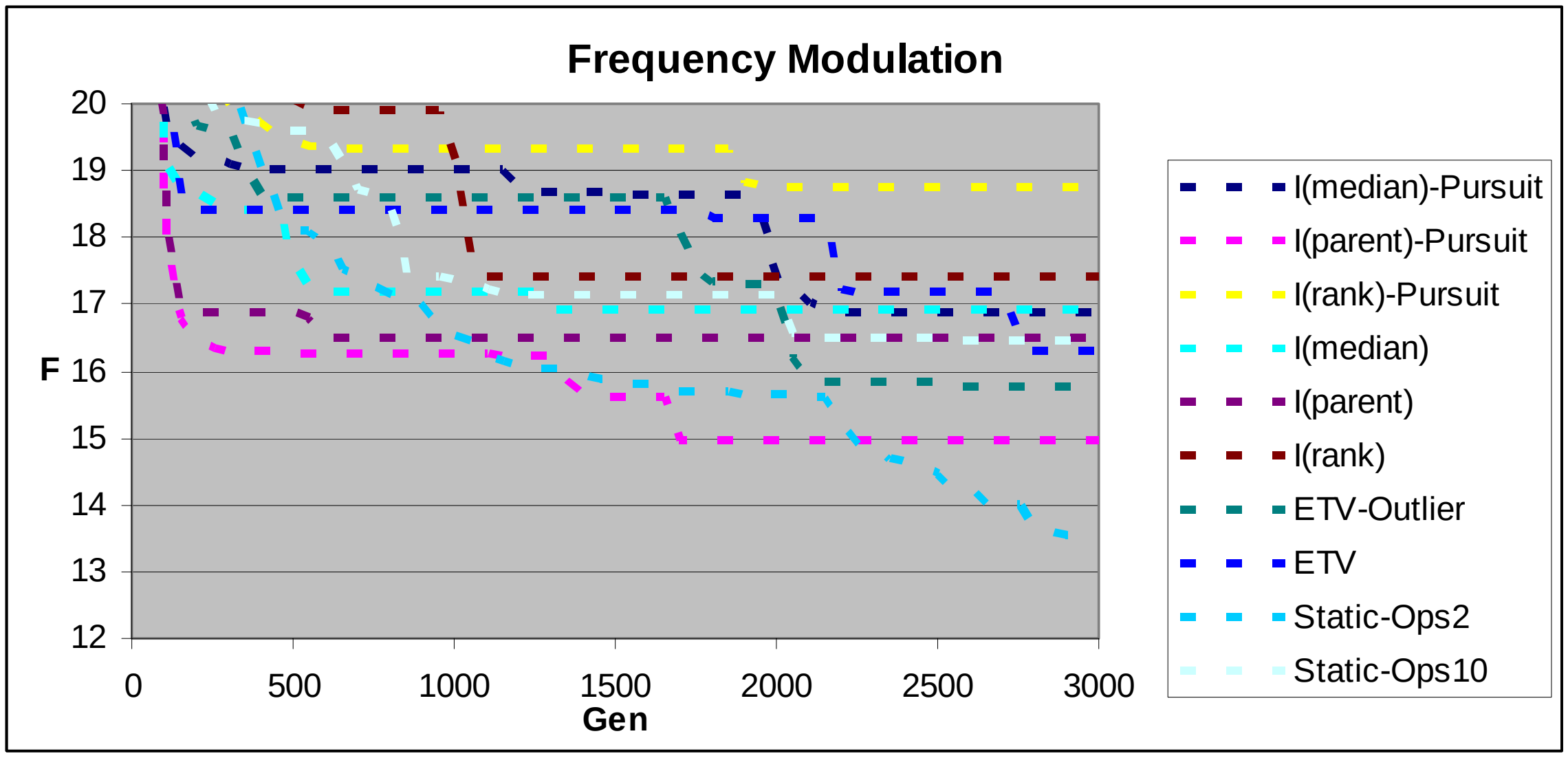

**Figure 3-44 Performance of adaptive and non-adaptive EA designs on the Frequency Modulation problem. The global optimal solution is *F*=0.**

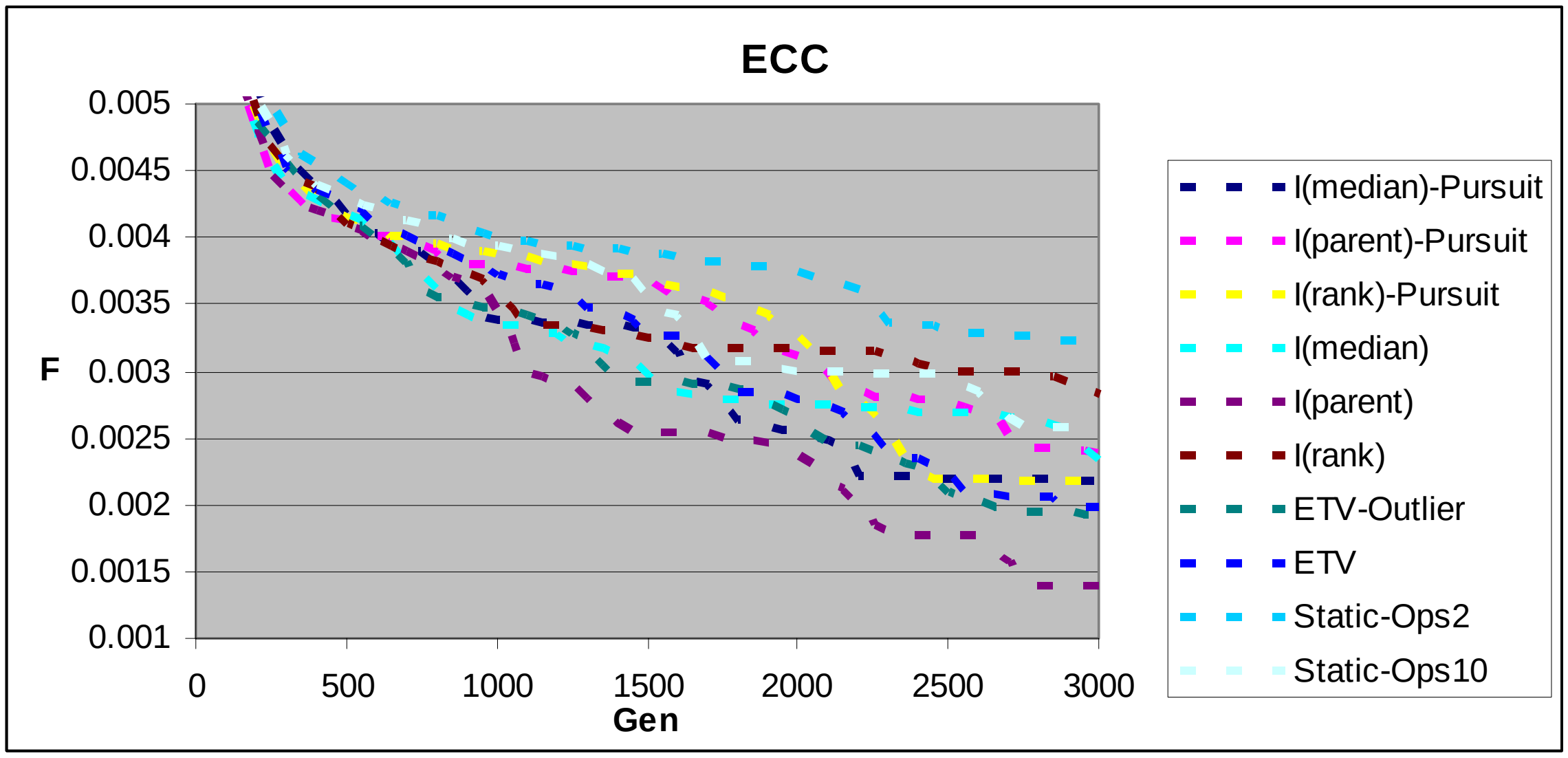

**Figure 3-45 Performance of adaptive and non-adaptive EA designs on the Error Correcting Code (ECC) problem. The global optimal solution is *F*=0.**

### 3.3.3 Discussion and Conclusions

A number of important conclusions can be drawn from the results in this chapter which are highlighted below and discussed within the context of EC research.

**Keys to Effective Adaptation of Search Operators:** Some useful conclusions can be drawn about the characteristics of an effective adaptive procedure based on the probability profile analysis from Section 3.3.2.2. From this analysis, it is concluded that an effective adaptive method is one that 1) is able to select appropriate operators for exploiting the features of a given fitness landscape, 2) changes only in response to environmental changes and 3) is able to resolve the "right amount" of difference in the search operator usage rates. Such a description of an effective adaptive method is not surprising however what is surprising is that very few of the adaptive methods were able to exhibit these behaviors.

For example, adaptive methods using the adaptive pursuit strategy almost always created erratic probability profiles with changes in operator preference that did not reflect changes in the environment. Although these methods generally failed at condition 2, they were still able (in most cases) to select appropriate operators for traversing the landscape (thereby satisfying condition 1).

When the ETV adaptive method was used without Outlier interpretation, it was very poor at resolving differences between search operators (i.e. it does not satisfy condition 3). This is not surprising given the ETV distribution results (e.g. see Figure 3-8) which have indicated that the ETV measurement is, on average, very similar for each event. Nor was it surprising that for a number of problems, its performance was very similar to that observed with Static-Ops10 (where all search operators are used with equal probability). However, when the ETV-Outlier method was used, large differences in probability values were resolved between the operators (satisfying condition 1). This adaptive method exhibited strong responses to changes in the environment (satisfying condition 2) yet these transitions were largely non-existent in many of the other adaptive methods. Finally, the superior performance of the ETV-Outlier method provides evidence that it also satisfies condition 3. It is argued that the ETV-Outlier method exhibited this set of behaviors because of its ability to focus solely on important events when making decisions of which operators to prefer. This requires both an accurate measure of an event's importance (provided by ETV) and the ability to filter out events that represent neutral interactions between the adaptive system and its environment (provided by Outlier interpretation).

**Performance Sensitivity to Test Function and NFL Implications:** Algorithm performance depended strongly on the set of test functions being considered. Many adaptive methods performed well on the first set of problems but were very bad on the second set (and vice versa). This did not occur nearly as much with ETV-Outlier which was unexpected. This adaptive method performed significantly better than the other adaptive and non-adaptive methods, although there were rare instances where this was not the case. It has been argued in this chapter that the ETV-Outlier method is an effective adaptive procedure which implies that it is able to exploit landscape features for short-term performance gains.

However, it is worth pointing out that short-term performance gains do not guarantee long-term performance gains for certain fitness landscapes, particularly those with deceptive features. Indeed, it is speculated that deceptive landscape features can prevent an otherwise effective adaptive method from providing long term benefits to the performance of an optimization algorithm. Some evidence of this may actually be seen in the results on the Massively Multimodal Deceptive Problem (MMDP) where the ETV-Outlier method's performance is particularly poor.

In a sense, this suggested tradeoff between short and long-term performance could be demonstrating some form of NFL-related limitations that are expected to occur with any solver. However, it is also postulated here that the subset of real-world problems, similar to the artificial ones selected for these experiments, are actually a biased sampling from the space of all possible fitness landscapes and as such will display certain landscape features more often than others. For example, the subset of real-world problems is expected to be dominated by correlated versus uncorrelated and deceptive landscape characteristics [30]. Under these assumptions, exploitive adaptive mechanisms such as the ETV-Outlier could be expected to work fairly well in many optimization problems as is evidenced to some degree in this thesis. On the other hand, a real-world problem's fitness landscape is not necessarily defined as being either deceptive or not deceptive. Instead, it is expected to have varying degrees of both features, which suggests that's a robust search process should be able to maintain high levels of both explorative and exploitive behaviors within a single search process. Such behaviors are not explicitly accounted for within the ETV-Outlier adaptive method nor has this issue been adequately addressed elsewhere in the adaptive literature.

As previously implied, it is possible that an adaptive method could actually be detrimental to a search algorithm within certain contexts. To adequately assess potential shortcomings, it is not only necessary to test the adaptive mechanism on a diverse set of fitness landscapes but also important to test a diverse range of search operators. By including highly exploitive search operators for instance, it is possible to see if exploitive behavior will work against the adaptive method (e.g. by encouraging premature convergence within multimodal fitness landscapes). This was accounted for in this thesis by creating the highly exploitive swap and creep operators which were included in the list of search operators used by the adaptive methods within this chapter. A possibly better test of an adaptive mechanism's limitations might be to include local search operators which involve a greedy and more exploitive multi-step search within a single operator.

Although the conditions tested in this thesis did not appear to expose particular weaknesses in the ETV-Outlier method, this does not mean that this adaptive method can fully address the tradeoff between short and long-term performance in all problems. Ultimately, this tradeoff can only be dealt with by using either an iterative search which learns about the fitness landscape or possibly by extending the timescale of fitness measurements which are used by the adaptive method (thereby making the timescale for short-term performance gains not as short). Considering that the ETV measurement itself can take multiple generations to calculate (compared to a single function evaluation for other adaptive methods) it is possible that this longer measurement time scale is a contributor to ETV's exceptional performance.

**Comparing ETV and fitness measurements:** The Event Takeover Value or ETV has been put forth as a method for measuring an individual's impact on population dynamics. Comparisons between ETV and fitness-based ranking measurements have shown that a correlation does exist however the scaling and distribution of the measurements is dramatically different. Most importantly, it was found that very few individuals have any significant impact on population dynamics. This was interpreted to mean that most interactions between the adaptive system and its environment are effectively neutral. This conclusion fits well with observations of other adaptive systems in nature and it is possible that power law scaling (see Figure 3-8) is a general feature of interactions between many adaptive systems and their environment. This phenomena may even be the motivating force for the repeated emergence of threshold phenomena in biological systems (e.g. in gene regulation, neural activation).

As an alternative to defining an arbitrary threshold, statistical arguments were used in this chapter to quantify the importance of interactions between an adaptive system and its environment. This eliminated any need for threshold tuning and provided strong performance gains in the ETV-Outlier adaptive method.

**ETV adaptation as a generic tool for optimization:** There are a few important issues that still need to be addressed before the ETV-Outlier adaptive method can be readily implemented as a generic add-on tool for multi-search operator metaheuristics. First, although the results in this chapter were promising, it is necessary to test the adaptive method on additional problems, particularly application-inspired problems. There are many EA application domains where a large number of specialized search operators have been proposed in the literature (e.g. scheduling problems) and where it is not clear which search operators should be used for which problem instances. These problems would provide an ideal environment for testing the performance of an adaptive method in its ability to advantageously select and control operator usage.

There are also some potential drawbacks with the ETV-Outlier method that should be pointed out. One possible concern with the ETV-Outlier method is its computational efficiency. Although Section 3.2.2.4 indicates that memory costs are actually small and that memory and computational costs scale linearly with population size, the method still could be deemed to be somewhat computationally costly when used on simple test functions (which is where most EC research takes place). It is also worth pointing out that the adaptive method is quite complicated in comparison to the elegance of the original GA, requiring a substantial degree of record keeping and statistical tests. This does not make it difficult to implement per se but could make the algorithm difficult to understand and act as a potential deterrent to its use.

**Final Remarks:** It is generally understood that individuals in an EA population have a usefulness in the overall search process which extends beyond their individual genotype and phenotype. However, few if any previous attempts have been made in measuring how individuals impact the search process or have considered ways in which this information might be used to improve algorithm performance. This chapter attempted to make some inroads into this topic using metrics derived from genealogical graphs. It was also pointed out that this new ETV measurement involves a new type of search bias assumption that was labeled as Empirical Bias and is notably distinct from the standard Hill Climbing Assumption. The experimental results provided evidence that the adaptive method ETV-Outlier has many of the characteristics that are desired in an adaptive procedure and are arguably missing in previous methods for adapting EA design parameters.

# Chapter 4 Large Scale Features of EA Population Dynamics

The previous chapter presented the Event Takeover Value (ETV) as a way to measure an individual's impact on EA population dynamics. The ETV is able to approximate an individual's impact on population dynamics through an analysis of EA genealogical graphs. From preliminary tests in Chapter 3, it was found that the ETV probability distribution fits a power law with an exponent of approximately 2. This distribution indicates that a large proportion of individuals do not significantly impact EA population dynamics while a small minority of individuals dominate population dynamics.

The aim of this chapter is to gain a better understanding of the population dynamics of Evolutionary Algorithms using the ETV measurement derived in Chapter 3. In particular, this chapter investigates what experimental conditions can significantly impact the ETV distribution. After a broad range of conditions are tested in Section 4.1, it is concluded that only i) the population topology and ii) the introduction of completely new (i.e. randomly created) individuals can result in significant changes to the ETV distribution. If the EA population topology is a fully connected graph or if no new individuals are inserted into the population then the ETV distribution is found to be well approximated by a power law. However, when these conditions are not met, the ETV displays power law deviations for large ETV sizes. From these power law deviations, it is concluded that these EA designs are not capable of being dominated by a small number of individuals and hence are able to exhibit a higher degree of parallel search behavior.

Section 4.2 reviews and discusses several studies on the spatial and temporal properties of natural evolutionary dynamics which are found to exhibit similarities to the results presented in this chapter. Although the actual form of the measurements used to study natural evolution is not identical to the ETV measurements used here, these results do suggest that power law behavior and scale-invariant properties are prevalent in evolution.

Section 4.3 describes the Theory of Self-Organized Criticality and presents the theory as a possible explanation for the spatial and temporal patterns observed in EA and natural evolution. The chapter is concluded with Section 4.4 which discusses the relevance of these results to EA research and provides some motivations for the final chapter of this thesis.

## 4.1 Analysis of EA dynamics using ETV

This section studies EA population dynamics using the ETV measurement. Section 4.1.1 first describes the experimental conditions that are used throughout this chapter. Section 0 then investigates the experimental conditions that affect the distribution of ETV sizes in EA population dynamics. Section 4.1.3 follows with an investigation of the conditions affecting the distribution of ETV ages where the age is the total amount of time that an individual is able to influence EA population dynamics.

### 4.1.1 Experimental Setup

The experiments presented in this chapter were conducted using a number of artificial test problems. Definitions and problem descriptions are provided in Appendix A. A number of Evolutionary Algorithm designs have also been used in these experiments as elaborated on below.

#### 4.1.1.1 Panmictic EA designs

The Panmictic EA design refers to the standard EA design where spatial restrictions are not imposed on the population. A high level pseudocode is given below with the parent population of size $\mu$ at generation $t$ defined by $P(t)$. For each new generation, an offspring population $P`(t)$ of size $\lambda$ is created through variation of the parent population. The parent population for the next generation is then selected from $P`(t)$ and $Q$, where $Q$ is subset of $P(t)$. Q is derived from $P(t)$ by selecting those in the parent population with an age less than or equal to $\kappa$.

**Pseudocode for Panmictic EA designs**

```
t=0
Initialize P(t)
Evaluate P(t)
Do
    P`(t) = Variation(P(t))
    Evaluate (P`(t))
    P(t+1) = Select(P`(t) U Q)
    t=t+1
Loop until termination criteria
```

**Population updating**: The generational (Gen) EA designs that were tested in these experiments used elitism for retaining the best parent and parameter settings $\mu=N/2$, $\lambda=N$, $\kappa=1$ ($\kappa=\infty$ for best individual). The steady state (SS) EA design that was used in these experiments actually involves a pseudo steady state population updating strategy with parameter settings $\mu=\lambda=N$, $\kappa=\infty$.

**Selection**: Selection occurs by either binary tournament selection without replacement (Tour), truncation selection (Trun), or random selection (Rand). Random selection is implemented in the same fashion as binary tournament selection except the winner of a tournament is chosen at random (without regard for fitness of the individuals).

**Search Operators:** For each EA design, an offspring is created by using a single search operator that is selected at random from the list in Table 4-1. Search operator descriptions are provided in Appendix B.

**Crowding**: Crowding in Panmictic populations was implemented using Deterministic Crowding (DC) which is described in Chapter 2.

#### 4.1.1.2 Spatially Distributed Populations

All distributed EA designs that are tested in this chapter involve a cellular Genetic Algorithm (cGA) which is described in the pseudocode below. The algorithm starts by defining the initial population *P* on a ring topology with each node connected to exactly two others. For a given generation *t*, each node in the population is subject to standard genetic operators. Each node *N1* is selected as a parent and a second parent *N2* is selected among all neighbors within a radius *R* using linear ranking selection. An offspring is created using the two parents plus a single search operator selected at random from the list in Table 4-1. The better fit between the offspring and *N1* is then stored in a temporary list Temp(*N1*) while genetic operators are used on each of the remaining nodes in the population. To begin the next generation, the population is updated with the temporary list. This process repeats until some stopping criteria is met.

**Pseudocode for cGA**

```
t=0
Initialize P(t) (at random)
Initialize population topology (ring structure)
Evaluate P(t)
Do
        For each N1 in P(t)
                Select N1 as first parent
                Select N2 from Neighborhood(N1,R)
                Select Search Operator (at random)
                Create and evaluate offspring
                Temp(N1) = Best_of(offspring, N1)
        Next N1
        t=t+1
        P(t) = Temp()
Loop until stopping criteria
```

**Crowding**: Distributed EA designs that include crowding procedures are modified so that the offspring competes with the parent (*N1* or *N2*) that is most similar in phenotype.

**Table 4-1 Names of the seven search operators used in the cellular GA and Panmictic EA designs are listed below. More information on each of the search operators can be found in Appendix B.**

| Search Operator Names |
|---|
| Wright's Heuristic Crossover |
| Simple Crossover |
| Extended Line Crossover |
| Uniform Crossover |
| BLX- α |
| Differential Operator |
| Single Point Random Mutation |

### 4.1.2 ETV Size Results

This section is concerned with determining what experimental conditions can influence the ETV distribution. There are many aspects of an EA design that have been modified or extended over the years meaning that any attempt at making broad statements about EA population dynamics requires a broad range of experimental conditions to be tested. Because a large number of experiments were necessary, only selected results are presented based on their capacity to illuminate system behavior. Section 4.1.2.1 looks at the impact that EA design features have on the ETV distribution while Section 4.1.2.2 investigates the impact of the fitness landscape. Section 4.1.2.3 follows up with an investigation of whether the ETV distribution is sensitive to the amount of time that evolution is observed.

#### 4.1.2.1 Impact of EA design

The first EA design factors tested consisted of selection methods and population updating strategies for Panmictic EA designs, with results shown in Figure 4-1a and Figure 4-1b. Selection pressures varied from very weak (e.g. random selection) to very strong (e.g. truncation selection) and the population updating strategy varied from infinite maximum life spans (steady state) to single generation life spans (generational). The most remarkable conclusion from these results is that the ETV distribution has very little sensitivity to these design factors and consistently takes on a power law distribution. Particularly surprising was the results using random selection, which has no sensitivity to the fitness landscape of the test problem being used. When random selection is used, the ETV distribution appears to take on a slightly smaller distribution tail although a power law is still clearly observed.

Experiments were also conducted to determine the impact of the population size. As seen in Figure 4-2a, EA designs which differ only in the value of *N* have nearly identical ETV distributions. The insensitivity to *N* was also observed for the other EA designs tested in Figure 4-1a and Figure 4-1b with *N* varying from 50 to 400 (results not shown).

The results in Figure 4-2 present what was found to be the most important factor impacting the ETV distribution. These experiments, which were run using the cellular Genetic Algorithm, found that spatial restrictions result in power law deviations for large ETV sizes. Furthermore, the extent of the deviation was clearly dependent upon the degree of spatial restrictions in the system. As seen in Figure 4-2b, the use of random selection changes the exponent of the power law (that best approximates the data) however power law deviations are still present.

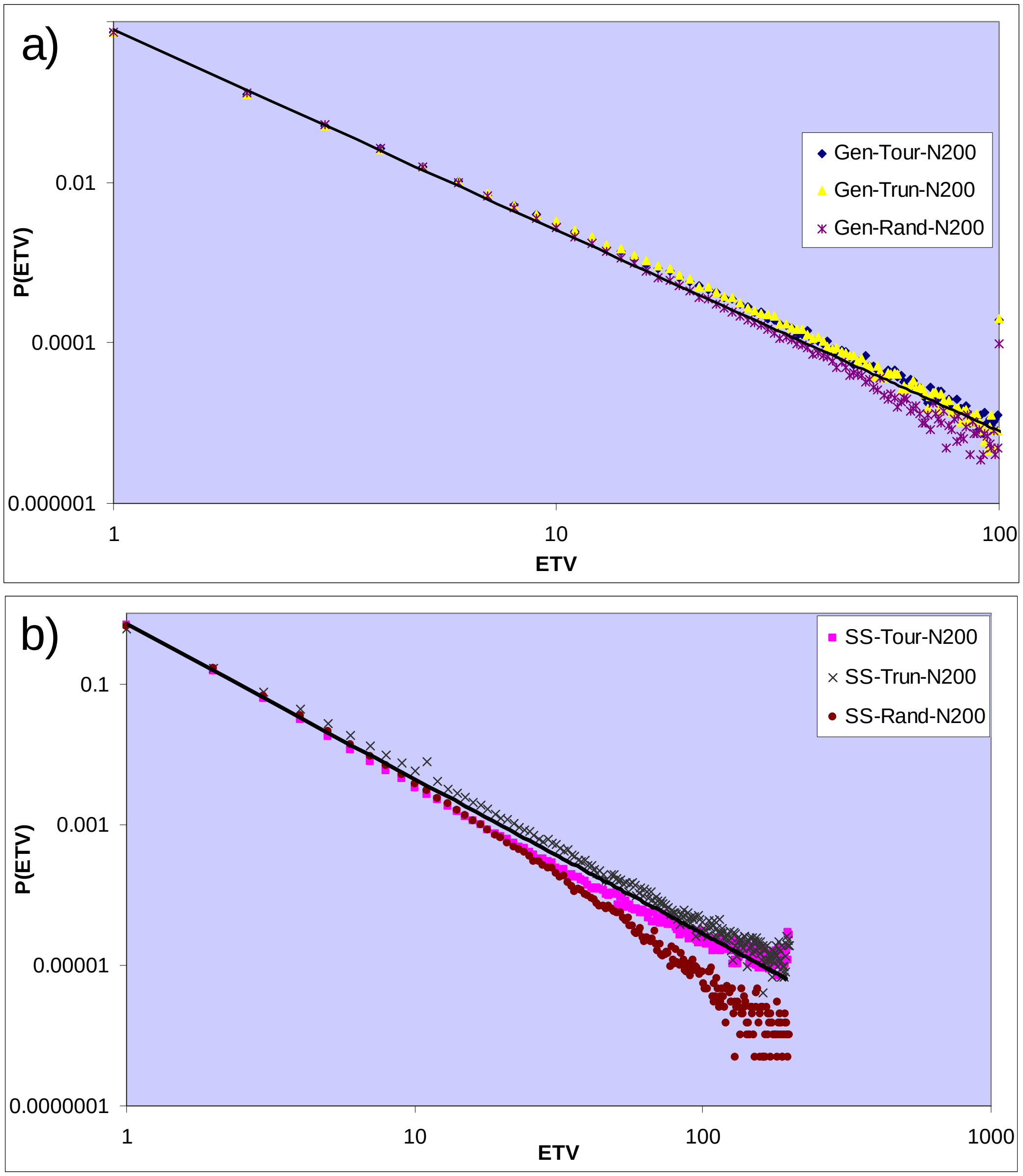

**Figure 4-1 ETV size distributions for a number of panmictic EA designs. a) EA designs with population size *N*=200, generational population updating (Gen), and selection methods Tournament (Tour), Truncation (Trun) and Random (Rand) selection. Solid line represents a power law with exponent 2.5. b) EA designs with population size *N*=200, steady state (SS) population updating, and selection methods Tournament (Tour), Truncation (Trun) and Random (Rand) selection. Solid line represents a power law with exponent 2.3. Results from each EA design are taken over 20,000 generations of evolution on the 30-D Hyper Ellipsoid test function.**

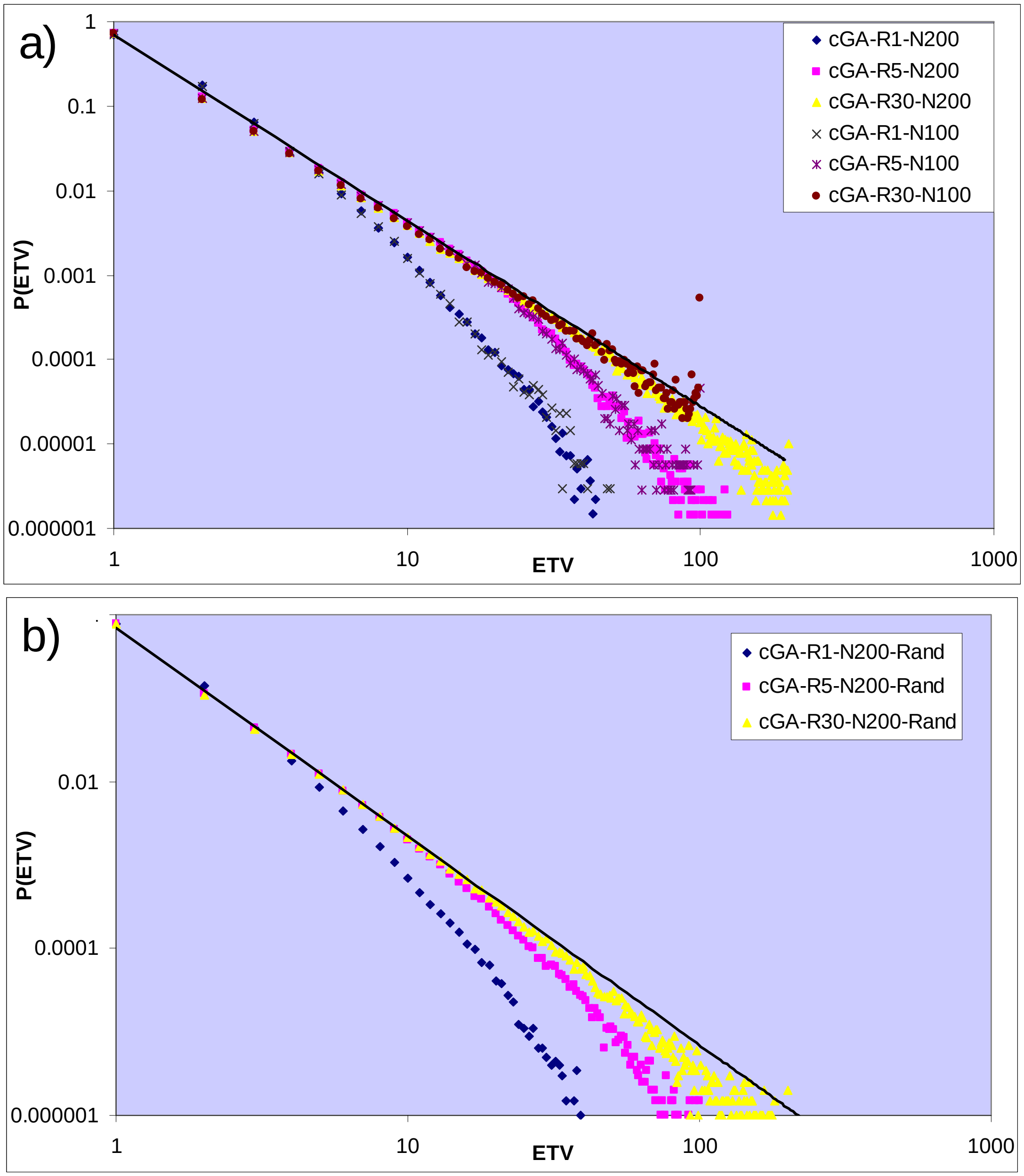

**Figure 4-2 ETV size distributions for a number of spatially distributed EA designs. a) Cellular Genetic Algorithm (cGA) designs with population sizes (*N*=100, *N*=200), and neighborhood radius (*R*=1, *R*=5, *R*=30). Solid line represents a power law with exponent 2.2. b) Cellular Genetic Algorithm (cGA) designs with random selection (Rand), population size (*N*=200), and neighborhood radius (*R*=1, *R*=5, *R*=30). Solid line represents a power law with exponent 2.5. Results from each EA design are taken over 20,000 generations of evolution on the 30-D Hyper Ellipsoid test function.**

Given that spatial restrictions are so far the only EA design factor significantly influencing the ETV distribution, it was decided to consider other mechanisms for restricting interactions within an EA population. A common approach for restricting interactions are so called crowding methods where offspring are forced to compete with similar individuals in the population. The results in Figure 4-3 show that crowding does have a significant impact on the ETV distribution, but only for spatially distributed EA designs. For all but the smallest ETV value (ETV=1), the use of crowding decreases the ETV's probability of occurrence by roughly an order of magnitude in the cGA. However, for Panmictic EA designs, the use of crowding did not appear to have a significant impact on the ETV distribution which is demonstrated using Deterministic Crowding (see inset of Figure 4-3).

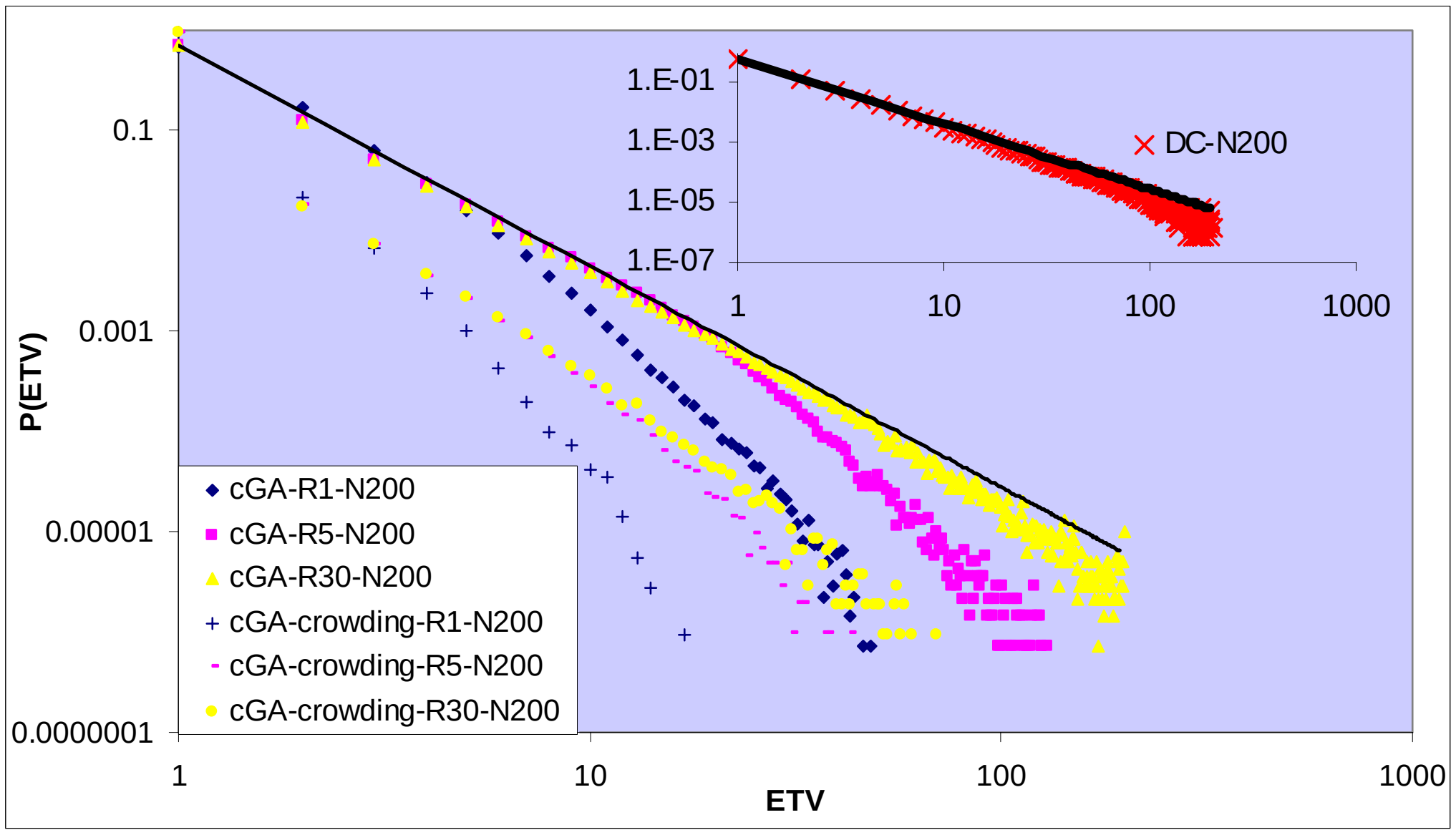

**Figure 4-3 ETV distributions shown primarily for spatially distributed EA designs. All EA designs have population size *N*=200. Cellular Genetic Algorithm (cGA) designs vary in the use of crowding and the neighborhood radius size (*R*=1, *R*=5, *R*=30). Results from using Deterministic Crowding (DC) are also presented in the inset. Solid line represents a power law with exponent 2.2. Results from each EA design are taken over 20,000 generations of evolution on the 30-D Hyper Ellipsoid test function.**

All results presented thus far have been taken with evolution occurring on the 30-D Hyper-Ellipsoid test function. This test function was selected because each of the EA designs were able to evolve for long periods of time (achieving 1 million to 10 million events) which allowed for greater clarity in the distribution results. The next section addresses the impact of the fitness landscape.

### 4.1.2.2 Fitness Landscape Dependencies

The fitness landscape that an EA population evolves on will obviously impact the trajectory that the population takes through parameter space. Hence, it came as a surprise to find how little the fitness landscape influenced the ETV distribution results. Test functions were selected from Appendix A and include unimodal and multimodal functions, linear and nonlinear functions, functions with strong and weak epistasis, as well as deceptive and non-deceptive functions. Results shown in Figure 4-4 and Figure 4-5 demonstrate little sensitivity to the fitness landscape on which evolution occurs. Other Panmictic EA designs were also tested with similar results (results not shown). For the distributed EA results, some sensitivity to the fitness landscape was observed when strong spatial restrictions were present in the EA population (e.g. see Figure 4-5a). However, the general conclusions from the previous section remain unchanged; only spatial restrictions in the EA population result in significant changes to the ETV distribution.

Another way to test the influence of the fitness landscape on ETV results is to use a random selection pressure as was done in the previous section. The use of random selection in an EA design is similar to evolving on a completely flat fitness landscape. The use of different search operators is also expected to have an impact that is similar to changing the fitness landscape. Some preliminary work has tested the use of different search operators (results not shown) and this was found to have a similar effect to varying the test function although these results displayed even less sensitivity (possibly due to the range of operators used).

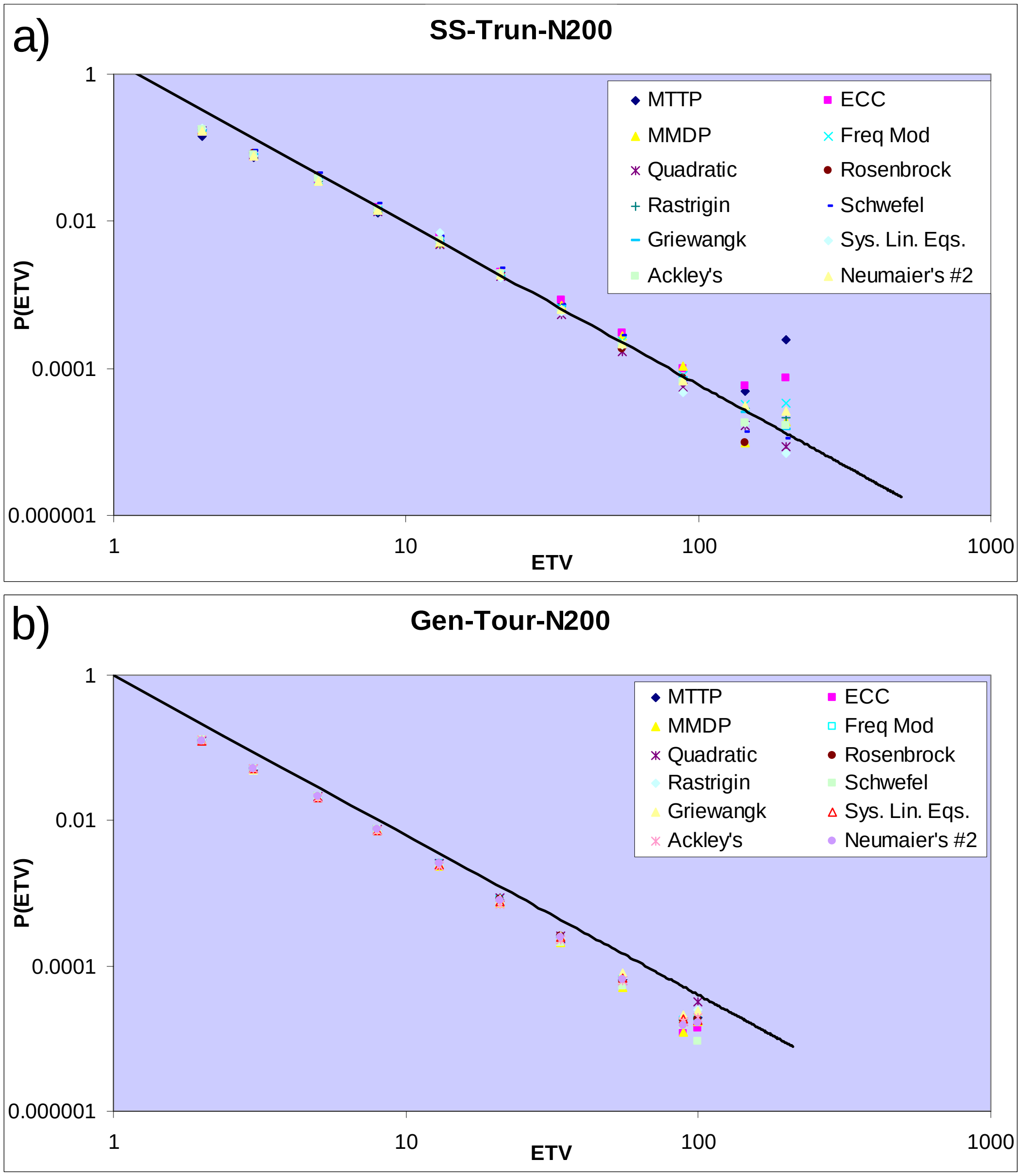

**Figure 4-4 ETV distributions shown for selected EA designs on a range of test functions taken from Appendix A. Evolution occurred over 2000 generations and results shown are averages taken over 10 runs. To help in viewing results from a large number of test functions, data is grouped into bins. a) Results for an EA design using steady state (SS) population updating, truncation selection (Trun), and population size *N*=200. Solid line represents a power law with exponent 2.2. b) Results for an EA design using generational (Gen) population updating, tournament selection (Tour), and population size *N*=200. Solid line represents a power law with exponent 2.2.**

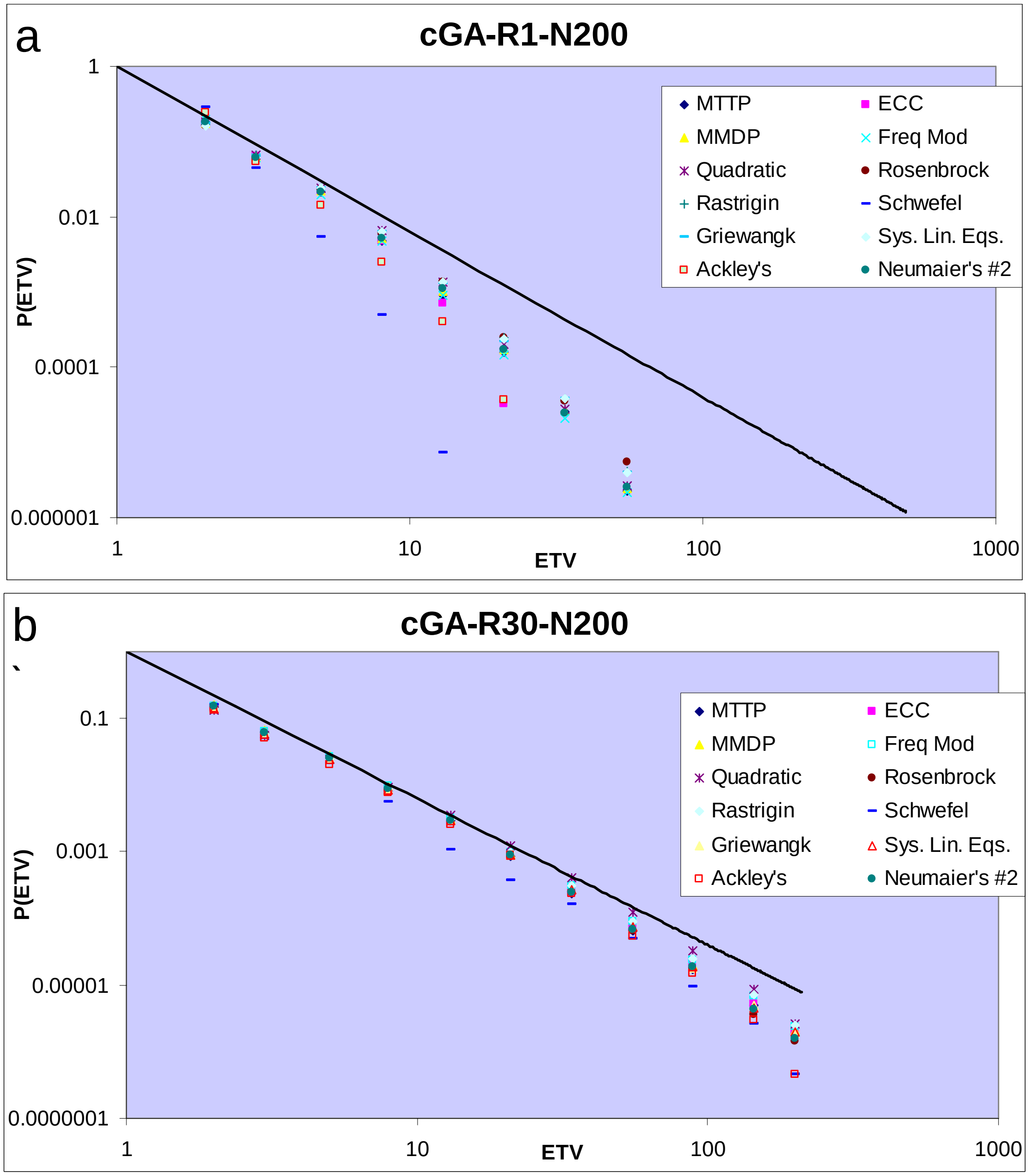

**Figure 4-5 ETV distributions shown for selected EA designs on a range of test functions taken from Appendix A. Evolution occurred over 2000 generations and results shown are averages taken over 10 runs. To help in viewing results from a large number of test functions, data is grouped into bins. a) Results for a distributed EA design (cGA) using neighborhood radius *R*=1, and population size *N*=200. Solid line represents a power law with exponent 2.2. b) Results for a distributed EA design (cGA) using neighborhood radius *R*=30, and population size *N*=200. Solid line represents a power law with exponent 2.2.**

#### 4.1.2.3 Impact of time length of evolution

In this section, tests were conducted with evolution taking place over different lengths of time. As seen in Figure 4-6, during the initial stages of evolution, the ETV distribution displays power law deviations for large ETV sizes however these deviations disappear as evolution is observed over longer periods of time.

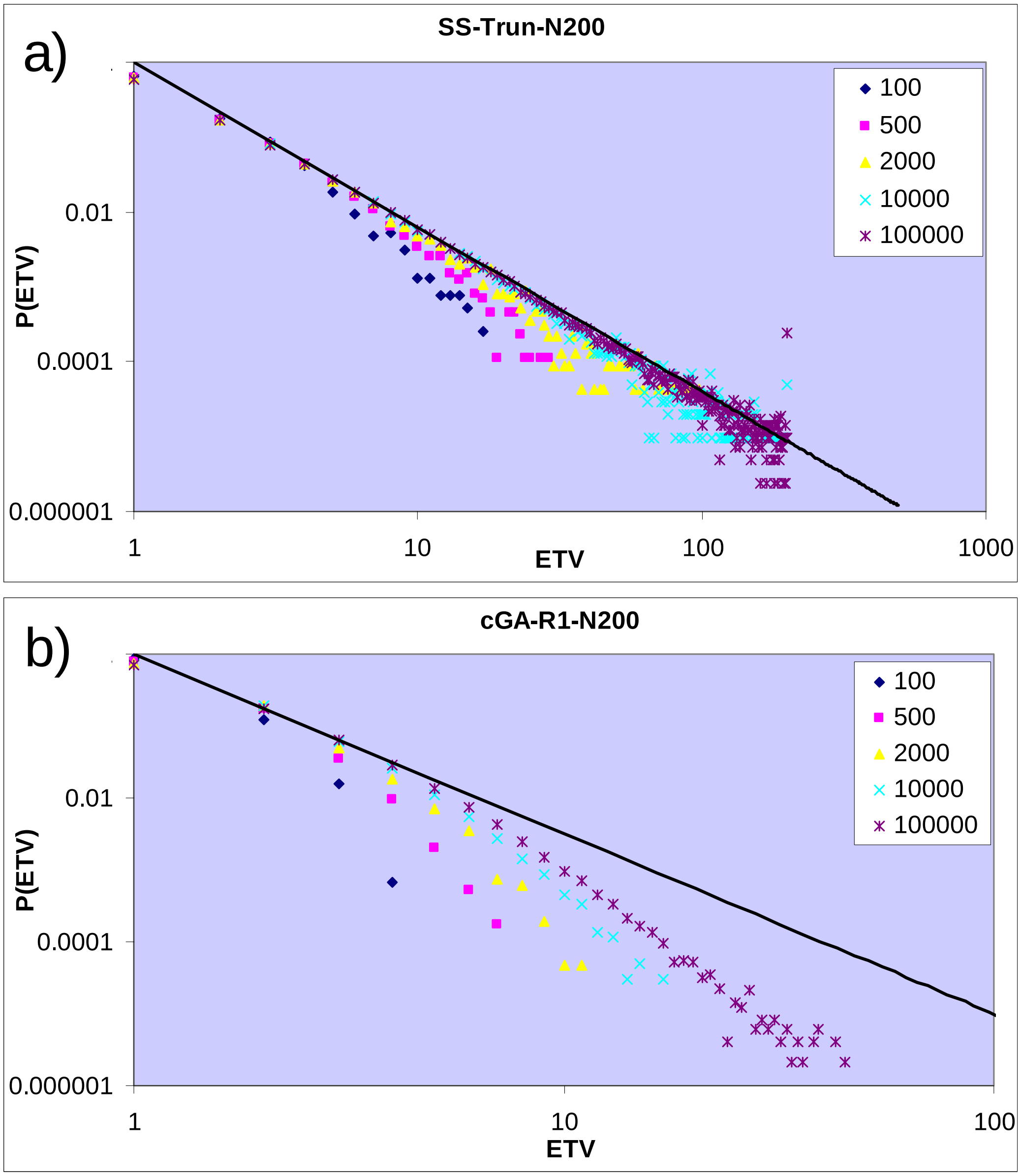

**Figure 4-6 ETV distribution results as a function of the time span of evolution. a) Results for an EA design using steady state (SS) population updating, truncation selection (Trun), and population size *N*=200. Solid line represents a power law with exponent 2.2. b) Results for a distributed EA design (cGA) using neighborhood radius *R*=1, and population size *N*=200. Solid line represents a power law with exponent 2.5. Data sets are labeled by a number which indicates the number of ETV measurements that are used to generate the distribution. For each EA run, the first 100 events are given to the first data set, the next 500 are given to the next data set and so on. Results for each EA design are averages over ten runs.**

The fact that ETV distribution results have only a brief transient where the distribution is sensitive to time, but is insensitive thereafter, indicates that the distribution approaches a stationary state. However, record statistics of ETV in Figure 4-7 provide evidence that maximum ETV sizes have an initial time dependency. This could mean that the system does not initially start with population dynamics being defined by a power law distribution but instead that the system evolves to achieve that state over time.

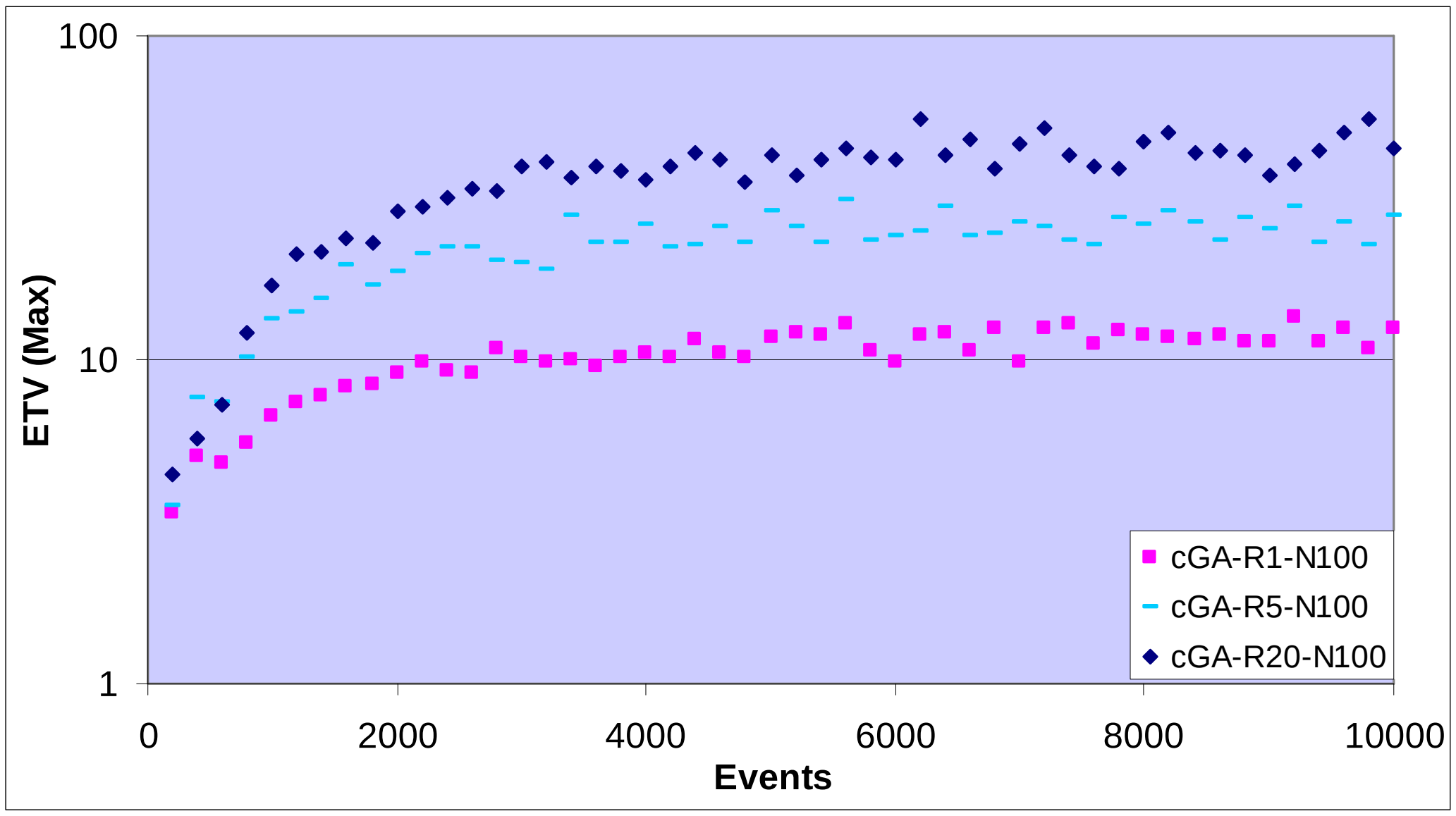

**Figure 4-7 Record ETV statistics for cellular Genetic Algorithms (cGA) with population size *N*=100 and neighborhood radius (*R*=1, *R*=5, *R*=30). ETV(Max) is the largest ETV found in every 200 events. Values are averages over 10 experimental replicates.**

It has been determined that the reason for this initial dynamical behavior is actually due to the lack of a genealogical or a historical coupling between individuals in the initial population. To confirm this, Figure 4-8 shows ETV distribution results where each new offspring has a probability $P_{new}$ of being historically uncoupled from the rest of the population. Historical uncoupling is simply done by preventing offspring from inheriting historical data from their parents. From a population dynamics perspective, this is equivalent to an EA design which includes a steady introduction of new individuals into the EA population. As seen in Figure 4-8, a small amount of historical uncoupling can result in power law deviations for the largest ETV sizes. However, as $P_{new}$ is increased, the extent that the distribution deviates from a power law is found to increase only slightly.

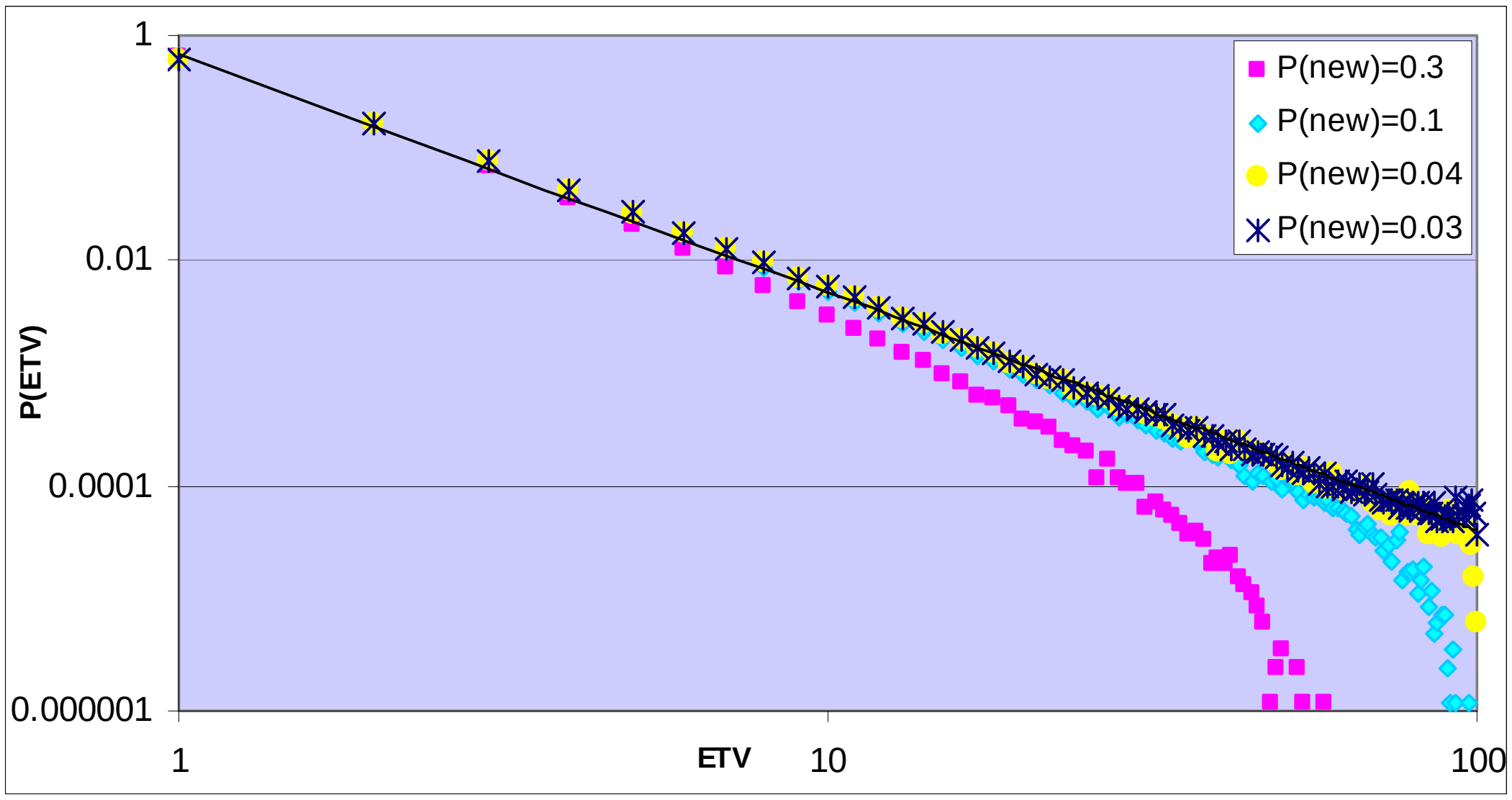

**Figure 4-8 ETV distributions with varying amounts of historical uncoupling in EA population dynamics. Experiments are conducted with a steady state EA using truncation selection and population size *N*=100. Evolution took place over 20,000 generations on the 30-D Hyper Ellipsoid test function. When conducting the standard ETV calculation, historical event information is copied from the genetically dominant parent to its offspring. In these experiments, the step of historical transfer is skipped with probability $P_{new}$. The solid line in the graph represents a power law with exponent = 2.1**

#### 4.1.2.4 Other Experimental Conditions

Additional tests were also conducted (results not shown) to help ensure that the ETV distribution results that have been presented so far in this chapter were not biased due to other experimental factors. This included experiments on selected EA designs at population sizes up to *N*=500, running evolution up to 100,000 generations, and experiments with the ETV calculation parameter $T_{obs}$ set as high as 500. These experiments resulted in no observable changes to ETV distribution results.

### 4.1.3 ETV Age Results

In addition to measuring the size of an individual's impact, one can also measure the amount of time that an individual is able to impact population dynamics. This is measured by recording the number of generations required for an individual's ETV calculation to finish, which is referred to in this thesis as the *ETV age*. This section investigates this aspect of EA dynamics more closely, again with the aim of determining what experimental conditions impact the ETV age distribution.

Looking at Figure 4-9 and Figure 4-10, these results demonstrate that the ETV age repeatedly approximates a power law however different sensitivities to EA design conditions have emerged (compared with ETV size distribution results shown previously). Although there is still no sensitivity to the population size, these results reveal that the selection method and population updating strategy do have an impact on the ETV age distribution for Panmictic populations. This is seen for instance in the results presented in Figure 4-10b where EA designs with steady state population updating and tournament selection are found to have a clear power law deviation for large ages. On the other hand, the introduction of spatial restrictions to the EA population does not have any influence on this characteristic of population dynamics as seen in Figure 4-9. This is surprising considering the importance of spatial restrictions in the previous ETV size distribution results. Also shown in Figure 4-9, the addition of crowding to the cGA has a completely unexpected impact on the age distribution and appears to result in an almost log-periodic behavior that on average still tends toward a power law distribution. On the other hand, the addition of crowding in Panmictic Populations (e.g. Deterministic Crowding) was found to have little influence on the age distribution. In summary, these results indicate that most ETV age distributions are well approximated by power laws although changes to the distribution shape do occur under certain conditions.

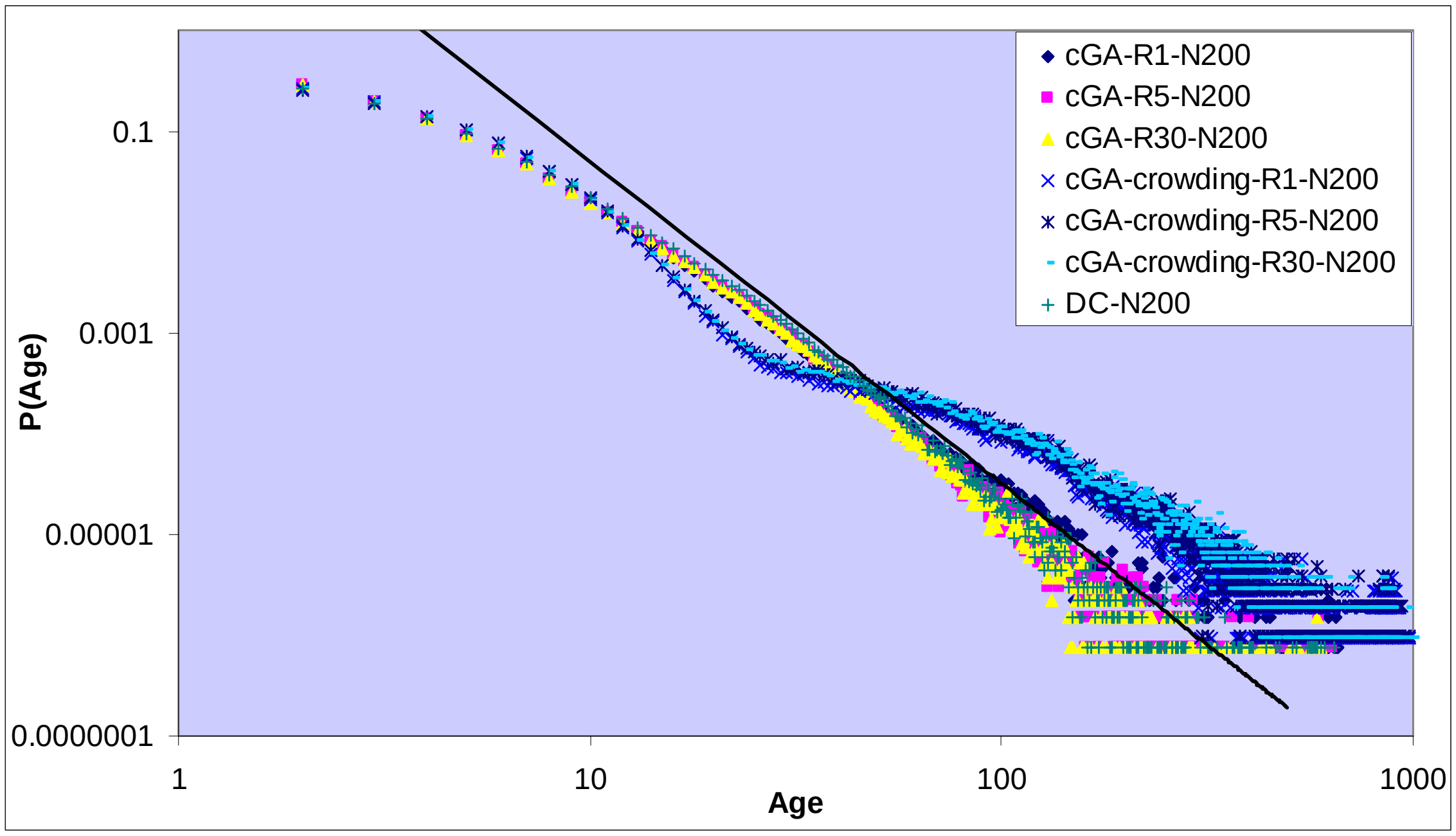

**Figure 4-9 ETV age distributions shown primarily for spatially distributed EA designs. The age of an ETV is defined by the number of generations from the initial event to the completion of the ETV calculation. All EA designs have population size *N*=200. Cellular Genetic Algorithm (cGA) designs vary in the use of crowding and the neighborhood radius size (*R*=1, *R*=5, *R*=30). Results from using Deterministic Crowding (DC) are also provided for a population size of *N*=200. Solid line represents a power law with exponent 3.2.**

As a final comment on these results, it is also worth mentioning that although the ETV has a maximum size equal to *N*, the ETV age measured here is only constrained by the amount of time that the system is observed. For these experiments, evolution was observed for up to 20,000 generations and ETV ages were found approaching 1000 without any evidence of power law deviations at large ages. Based on the observed distributions, it is concluded that the maximum age of events in EA dynamics is only limited by the amount of time that evolution is allowed to take place.[12]

[12] The maximum age can also be limited by the ETV calculation procedure, for instance by limits placed on the size of the historical records kept in the EA population. In these experiments, the maximum record size was set to $T_{obs}$=240 and under these conditions, it was found that roughly one in every 50,000 events failed to finish the ETV calculation before reaching a maximum record position.

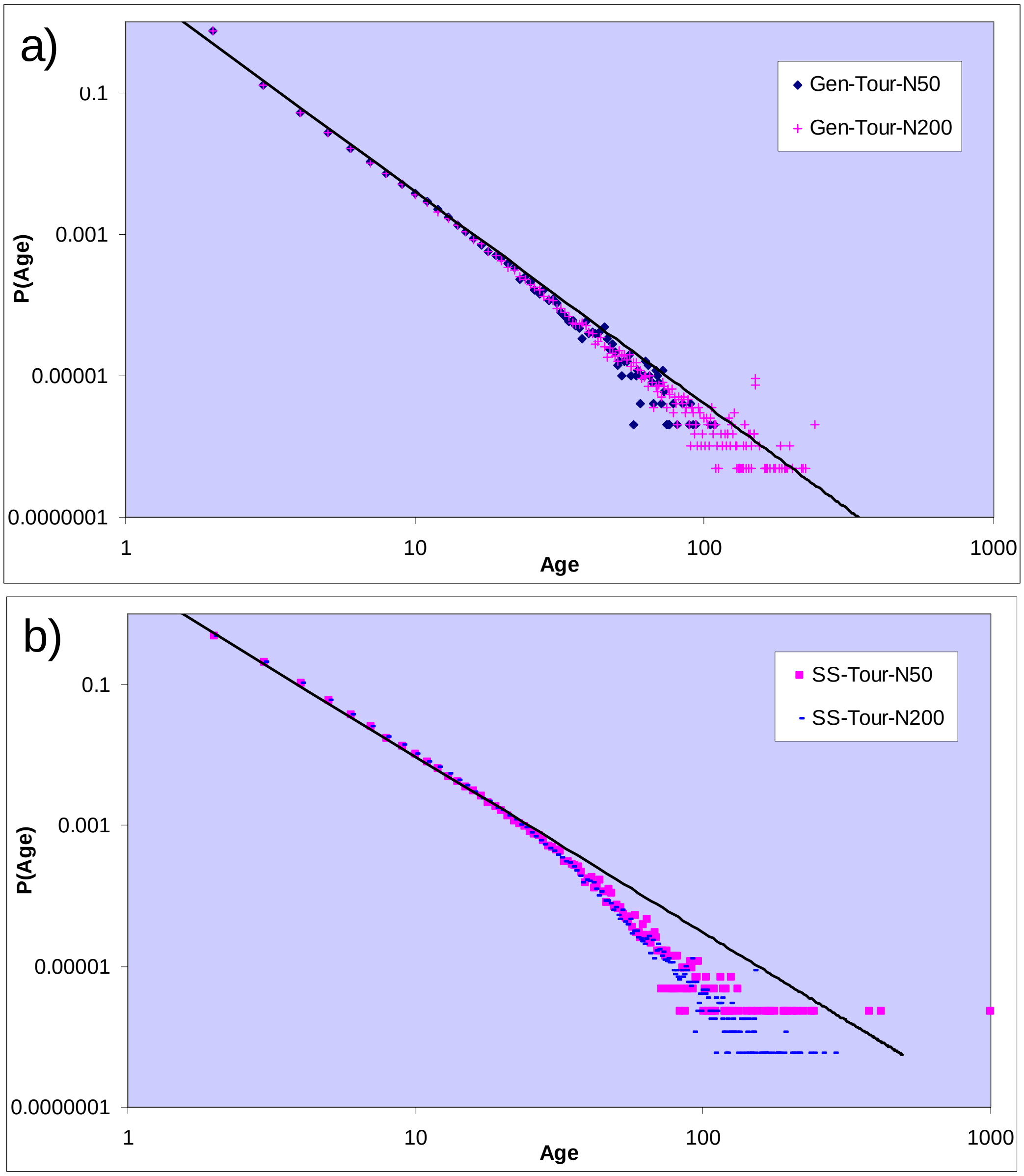

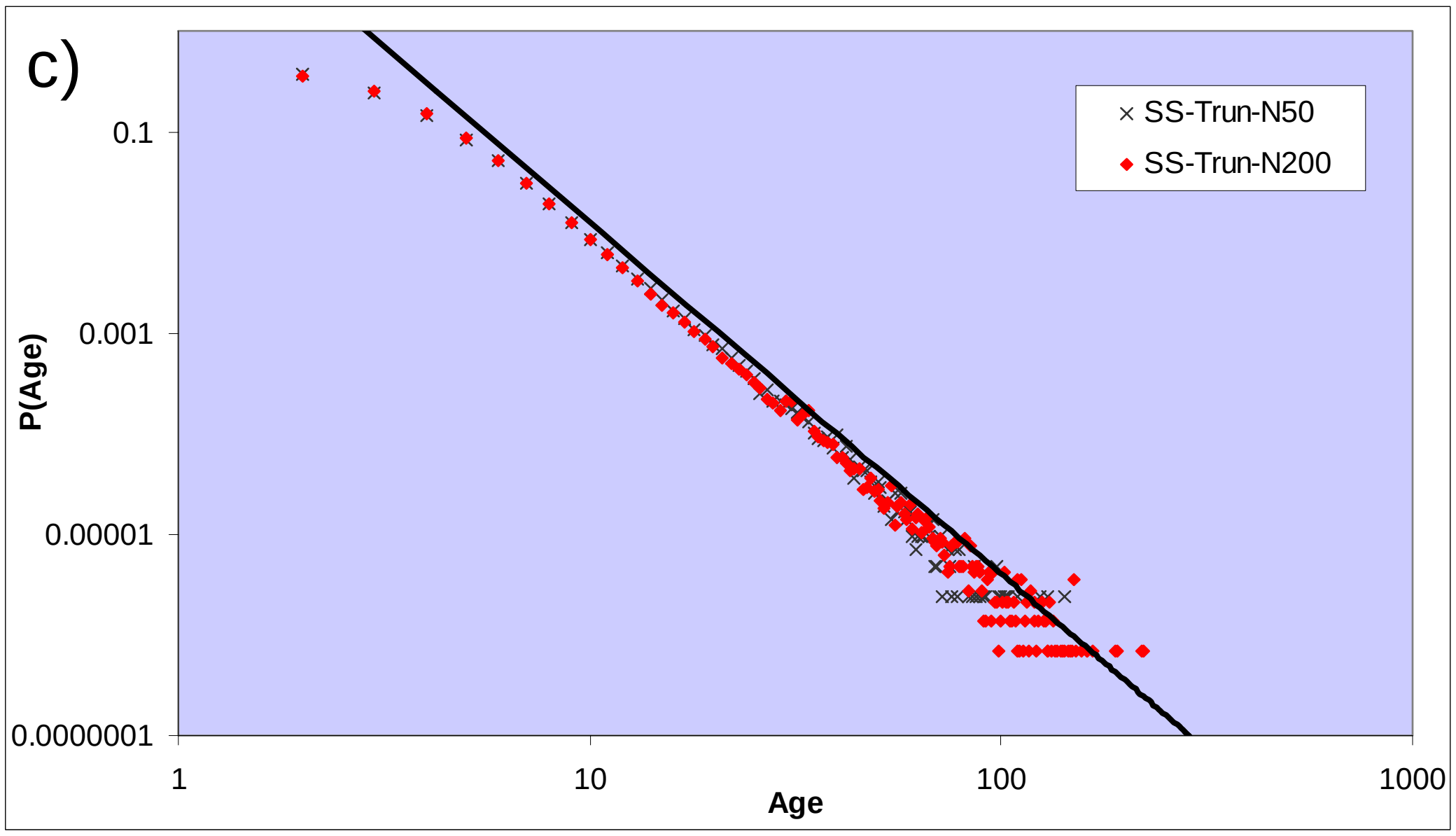

**Figure 4-10 ETV age distributions for several EA designs. The age of an ETV is defined by the number of generations from the initial event to the completion of the ETV calculation. a) EA designs with population sizes (*N*=200, *N*=50), generational population updating (Gen) and Tournament selection (Tour). Solid line represents a power law with exponent 3. b) EA designs with population sizes (*N*=200, *N*=50), steady state population updating (SS), and Tournament selection (Tour). Solid line represents a power law with exponent 2.5. c) EA designs with population sizes (*N*=200, *N*=50), steady state population updating (SS), and Truncation selection (Trun). Solid line represents a power law with exponent 3.5.**

#### 4.1.3.1 Caveats

It should be mentioned that, despite considerable efforts, the experiments were not exhaustive and so it is possible that other EA designs and certain landscape characteristics could result in ETV distributions which deviate from a power law or are otherwise different from what was presented here. As an example, EA designs which parameterize the amount of interaction between population subgroups (i.e. island model population structure) could be one unaccounted for situation where power laws would only be observed with the appropriate parameter tuning.

### 4.1.4 Conclusions

A number of conclusions can be drawn from the results presented in this chapter. First, it was found that the probability of an individual's impact on EA dynamics fits a power law (exponent between 2.2 and 2.5). This is a robust property of the system which is largely insensitive to most experimental conditions including changes to population size, search operators, fitness landscape, selection scheme, population updating strategy, and the presence of crowding mechanisms.

Two experimental conditions were however found to result in power law deviations for large ETV sizes. The first is the steady introduction of new individuals that have no relation to others in the population (i.e. historically uncoupled). The second condition is the introduction of spatial restrictions into an EA population. Using either of these conditions effectively removes the possibility of single individuals dominating the dynamics of the entire population. The associated power law deviations can be understood as an indicator of parallel computation within the system.

The amount of time than an individual influences EA dynamics (i.e. ETV age) also was found to fit a power law with most individuals influencing the system for only brief periods of time. However, as suggested by the power law relation, there is a non-negligible probability that an individual will influence EA dynamics over very large time scales. This behavior was found to be robust and was almost completely insensitive to all experimental conditions tested.

## 4.2 Discussion: Comparisons between EA and nature

From the last section it was concluded that EA population dynamics exhibit power laws in ETV spatial and temporal properties with little sensitivity to experimental conditions. Some of the measurements that have been taken of the spatial and temporal properties of natural evolution have also been found to exhibit power law relations. This section briefly reviews the results from natural evolution and compares and contrasts them the results from this chapter.

It is important to point out that no known measurements of natural evolution are exactly equivalent to ETV and so strong conclusions about similarities or differences in behavior are not possible. Instead, this section is provided to simply review and discuss current evidence that spatial and temporal patterns in EA population dynamics are similar to those observed in natural evolution.

Section 4.2.1 reviews past studies on extinction size distributions that are derived from the fossil record and compares this with ETV size distribution results. Section 4.2.2 reviews past studies on the distribution of species life times and compares this with ETV age distribution results. Finally, Section 4.2.3 looks at some topological properties of taxonomic structure in natural evolution and compares this with the genealogical structure of EA populations.

### 4.2.1 Extinction Sizes

The first large scale feature of evolutionary dynamics that is discussed deals with the characterization of extinction event sizes. Extinction event sizes are measured as the number of species (or percentage of species) which become extinct over a predefined time interval. Using fossil data that has been compiled by Sepkoski [169], several studies have analyzed extinction records [170], [171] and have found the distribution of extinction sizes to be very broad and well approximated by a power law. This is shown for instance in Figure 4-11, which is taken from [171]. In this figure, a best fit "kill curve", which was developed in [170], is used to create the extinction distribution for Paleozoic marine species. Using this model, a power law distribution is clearly observed for all but the very largest extinction events.

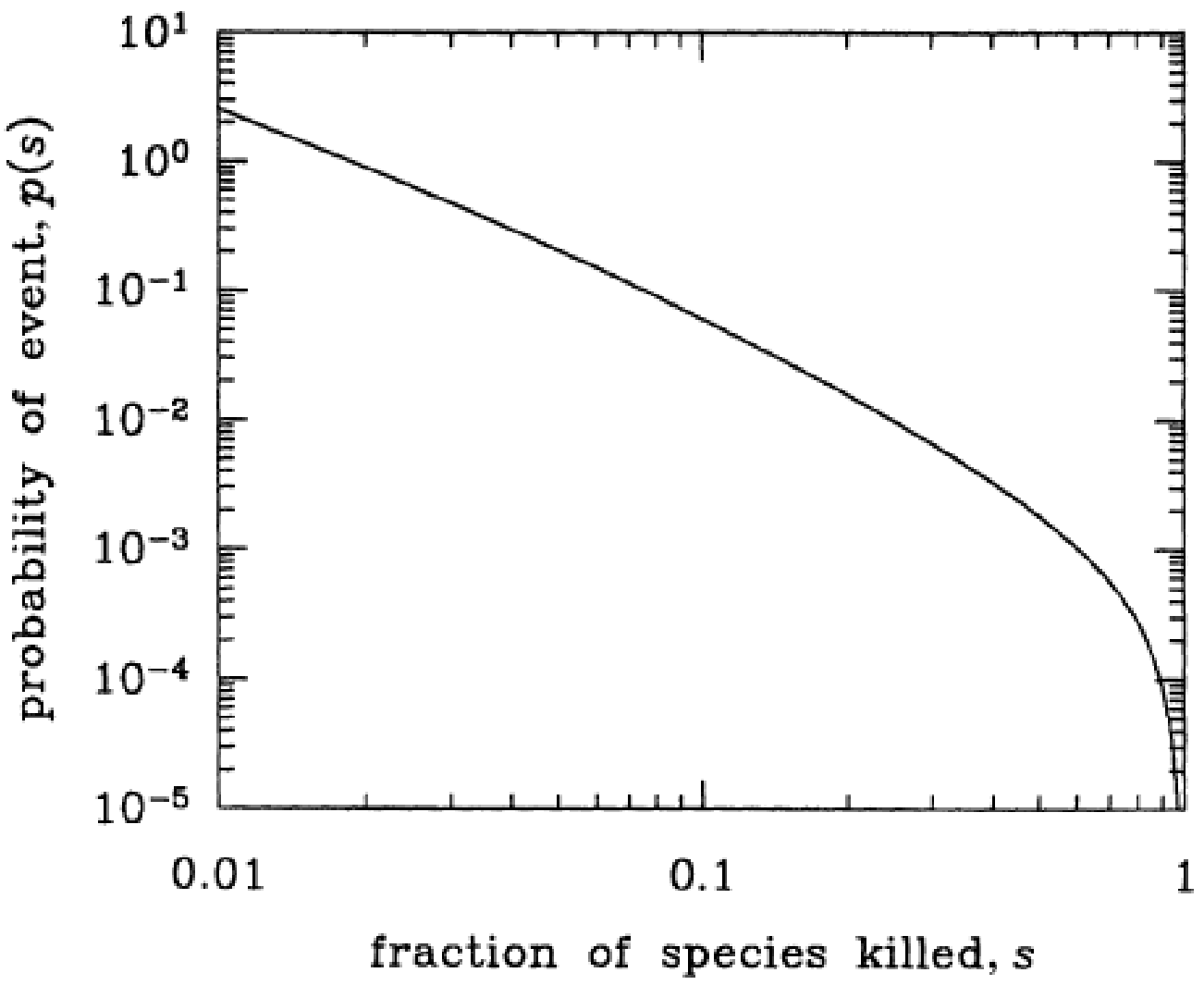

**Figure 4-11 Probability of an extinction event as a function of the fraction of all species killed. The distribution is derived based on a best of fit kill curve (see [170]) using fossil data of marine species from the Paleozoic era. Reprinted by permission from the Royal Society (Proceedings: Biological Sciences) [171], copyright (1996).**

**Comparing Extinction sizes and ETV:** The ETV measurement is similar in some ways to the measurement of extinction sizes in natural evolution however there are also some important differences. The most significant similarity is that both the ETV and the extinction size are a measure of the magnitude of changes that are taking place within each of their respective systems. On the other hand, while ETV is measuring the spread of genetic material in a population, the extinction size is measuring the removal of species (which can also be thought of as a loss of genetic material). Another important difference is that ETV measures the changes resulting from a particular event while extinction sizes from the fossil record look at changes occurring over a time window.

**Comparing Results:** Despite these difference, the broad degree distributions for ETV sizes is found to approximate a power law (exponent = 2.2 to 2.5) which is arguably similar to what has been observed in nature (exponent ~ 2) for extinction sizes. However, it does appear from Figure 4-11 that extinction sizes in natural evolutionary dynamics exhibit power law deviations for large extinctions while this is largely not the case in most of the ETV distribution results.

Some have suggested [172] that the power law deviations in the fossil record are a result of the system being in a state of disequilibrium and that more time is needed before a power law is observed. This is similar to the argument in [173] and shown in Figure 4-12, that power law deviations can occur due to finite size effects suggesting that a larger time window is needed before the fossil record can display a clear power law. Assuming for the moment that the power law behavior in EA and natural evolution has a similar origin, the ETV results from this chapter would then seem to support the disequilibrium argument. As demonstrated in Figure 4-6, running an EA for smaller periods of evolution results in ETV distributions with power law deviations for large ETV sizes.

However, it is possible that other factors contribute to power law deviations in the fossil record. In the experiments with EA, it was also found that spatial restrictions were a primary cause of power law deviations. Obviously, some amount of spatial restriction is present in natural evolution (e.g. geographical isolation) and it is speculated that this at least contributes to the observed power law deviations for extinction sizes in nature.

### 4.2.2 Species Lifetime Distributions

Another way to characterize natural evolutionary dynamics is to measure the lifetimes of species (or other taxa) as shown for example in Figure 4-12. Several studies [173], [169] have found a broad distribution of lifetimes however there is disagreement as to whether some of the reported results fit an exponential function [174], [175] or a power law, [173], [176].

The difference actually has great relevance to our understanding of evolution. As explained in Section 4.1.2 of [177], an exponential distribution would indicate that the age of a species has no impact on its likelihood of survival. In other words, older species are not better adapted to their environment compared to newer species. However, a more broad distribution such as a power law would indicate a correlation exists between age and extinction probability meaning that older species are better adapted compared to newer species. The results taken from [173] and presented in Figure 4-12 indicate that a correlation does exist and that the age distribution does fit a power law.

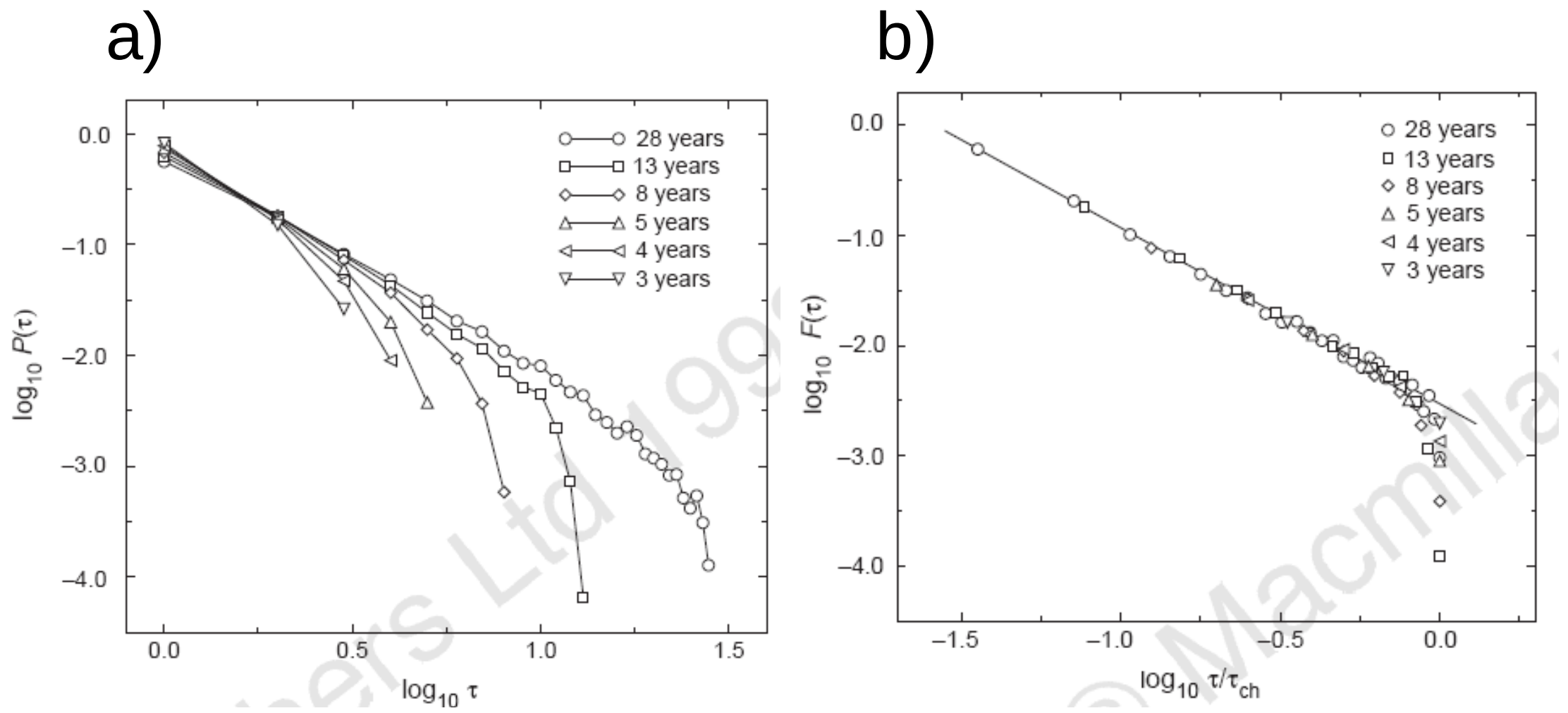

**Figure 4-12 Local lifetime distributions for species based on North American bird populations. a) Lifetime distributions for data taken over different timescales. Power law deviations are clearly present. b) Lifetime distributions with rescaling of data to account for finite size effects. Data is now well approximated by a power law. Reprinted by permission from Macmillan Publishers Ltd: (NATURE) [173], copyright (1998).**

**Comparing Species Lifetimes and ETV ages:** The ETV age measurement from Section 4.1.3 and the lifetime of a species in natural evolution are both measurements of relevant timescales of events in their respective systems. However, the two measurements are also different for a number reasons. For instance, if one thinks of events in EA population dynamics as being speciation events, then the species lifetime measurement for an EA would simply be the life time of individuals in the EA population. The ETV age, on the other hand, measures the total lifetime of all species that contain a strong genealogical link to an original speciation event. There are no known studies of the fossil record which have attempted an analogous metric of natural evolutionary dynamics.

**Comparing Results:** Despite these differences, it is still interesting to note that both ETV age distributions and species lifetime distributions are very broad and are well approximated by power laws. However the power law exponents are quite distinct with the ETV age distribution having an exponent of 2.5 to 3.5 while the results on bird species taken from [173] have an exponent of 1.6.

### 4.2.3 Fractal Taxonomic Structures

Yet another way to characterize natural evolutionary dynamics is to measure the topological properties of its taxonomic structure. As briefly reviewed in [177], several studies have looked at the frequency distribution of the number of species within a genus [178] as well as frequency distributions in higher taxa [179]. From these results, some of which are reproduced in Figure 4-13, it appears that the frequency distribution fits a power law with exponents reported to vary between 1.5 and 2.3.

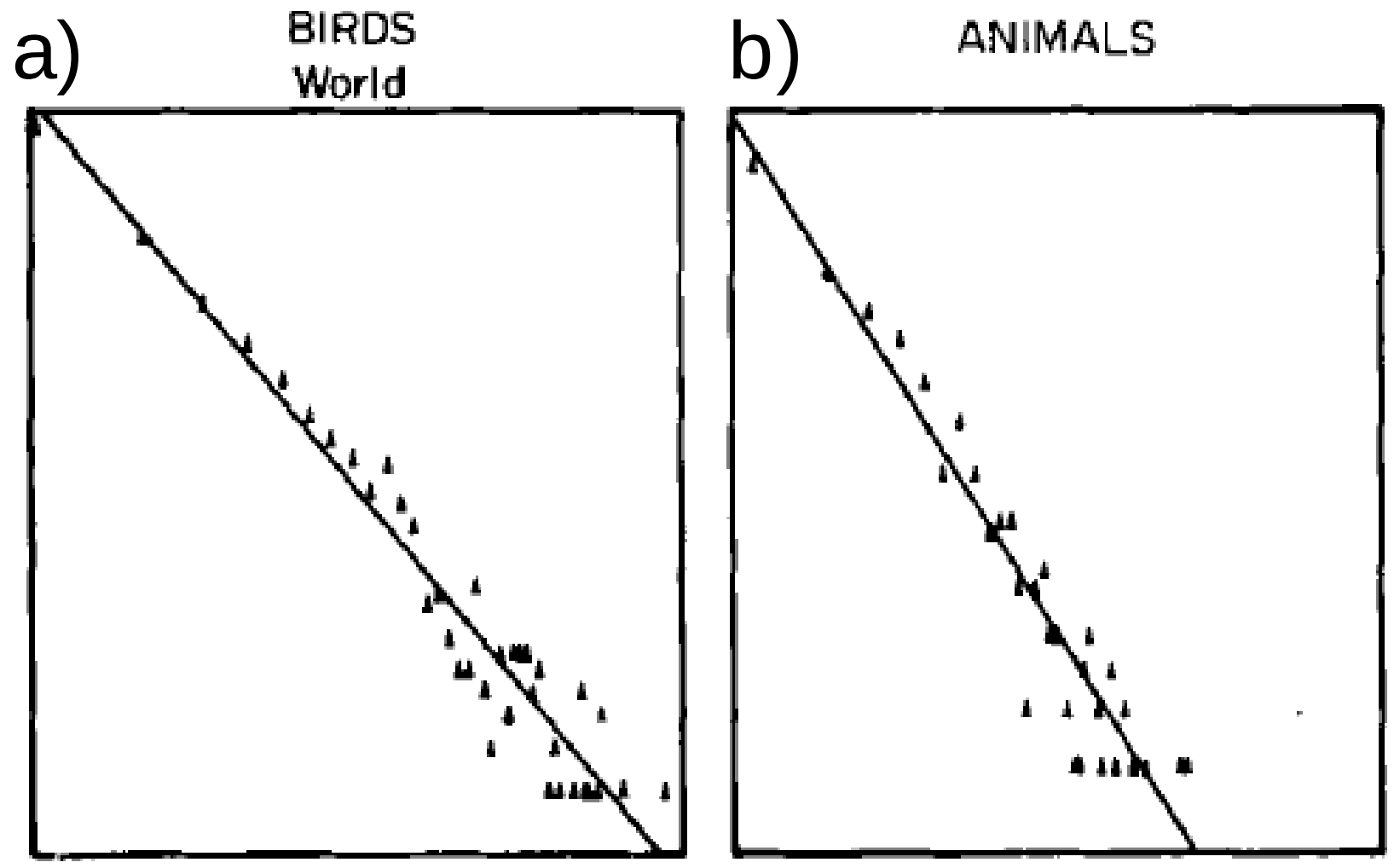

**Figure 4-13 log-log plots of the frequency of a selected taxon with different numbers of sub-taxa. a) Frequency of genera with different numbers of species for birds. The frequency is given on the vertical axis and the number of bird species within the genera is given on the horizontal axis. b) Frequency of orders with different numbers of families for animals. The frequency is given on the vertical axis and the number of animal families within the order is given on the horizontal axis. Data points with frequencies *f*=1 are omitted. Similar distributions for other data sets are presented in [179]. Reprinted by permission from Elsevier: (J. theor. Biol.) [179], copyright (1990).**

**Comparing EA Genealogy and Evolutionary Taxonomy:** To compare the topological properties of taxonomic structure (in natural evolution) with the genealogical structure of EA populations, it is necessary to clarify exactly what topological properties are being measured in Figure 4-13.

If one thinks of taxonomic structure in terms of a branching process, then the number of sub-taxa within a taxon (the horizontal axis of Figure 4-13) is equivalent to the number of branches that extend away from a node in the taxonomic tree. This is what is measured in Figure 4-13 with the addition of a few restrictions on the data used. In particular, data is restricted to particular groupings (e.g. birds, animals) and particular levels of hierarchy within the taxonomic structure (e.g. species/genera, families/order).

Figure 4-13a provides results for the number of species within a genera which are the lowest and second lowest levels of the taxonomic hierarchy (resp.). An equivalent measure in EA genealogical graphs would be that of the number of offspring created by a parent. This aspect of EA genealogy has been studied in detail in [180] where the distribution was found to fit a power law under a range of conditions. They also found that spatial restrictions in the EA population can result in power law deviations but only in cases where high levels of spatial restrictions were imposed.[13] Comparisons at higher levels of EA genealogy would be possible by reconstructing the genealogical graphs of an evolving EA population, however this was not considered here.

It is also interesting to note that similar studies have looked at human genealogy through an analysis of surname distributions [181] (also see [182] and references therein). In these studies, it has been determined that the surname distribution is also well approximated by a power law. In summary, the taxonomy of natural evolution has been found to have a fractal structure (as evidenced by the stated power laws) which is also observed in both human genealogies and the genealogical trees of Evolutionary Algorithms.

### 4.2.4 Summary of Conclusions

The ETV distribution results presented in this chapter were found to be similar to the extinction distributions in natural evolution. However the distribution of extinction sizes in natural evolution also displays power law deviations for large extinction sizes. From the limited sources of power law deviations in the ETV results with EA, it was speculated that that these deviations could be caused by either an insufficient amount of time that evolution has taken place or the presence of geographical isolation.

The ETV age distribution results were generally found to be similar to the species lifetime distributions in natural evolution with the power law exponent of the distribution being the only significant difference. Finally, a review of natural evolution taxonomy, human genealogy, and EA genealogy has found that each displays fractal characteristics as evidenced by power law distributions of structural features. In particular, fractal characteristics are found to be pervasive throughout natural evolution taxonomy.

[13] Notice that similar conclusions have also been made in this chapter with ETV distribution results.

## 4.3 Self-Organized Criticality

The review from the previous section has indicated that a number of spatial and temporal properties in natural evolution do not have a characteristic scale (as evidenced by a power law). The experiments conducted in this chapter also provide evidence that some spatial and temporal properties in Evolutionary Algorithms do not have a characteristic scale. These macro scale features have great relevance to the behavior of these systems and to our understanding of evolution. For example, a power law relation in species lifetimes provides strong evidence that older species are better adapted (on average) than younger species. This information has also helped to improve our understanding EA behavior. For instance, in the previous chapter the ETV size distribution provided convincing evidence that most interactions between an EA and its environment are effectively neutral. This new understanding was used to build a more effective approach to the adaptation of EA design parameters.

It could potentially be of great benefit to understand how this behavior emerges in natural evolution and Evolutionary Algorithms. One contender for explaining the macro features of evolutionary dynamics is the Theory of Self-Organized Criticality (SOC). This theory is briefly described next, followed by a set of conditions that any dynamical system is expected to satisfy in order to be compatible with SOC theory. Section 4.3.2 provides evidence that Evolutionary Algorithms meet a number of these conditions, with evidence based primarily on the ETV distribution results of this chapter.

### 4.3.1 SOC Definition

The Theory of Self-Organized Criticality was first put forth by Bak, Tang, and Wiesenfeld [183] and has been used to explain a range of physical phenomena such as flicker noise ($1/f$ noise) which is observed in the light emitted from quasars, the intensity of sunspots, current through resistors, sand flow in an hour glass, the flow of rivers, and stock exchange price indices (see [183] and references therein). The theory claims that some coupled dynamical systems are driven or attracted to a critical state where the system displays self-similarity in both space and time. This behavior is in contrast to other critical phenomena (e.g. phase transitions) where an environmental parameter (e.g. temperature) must be tuned in order for the system to reach a critical state. A more detailed discussion of critical phenomena is beyond the scope of this thesis however an introduction to the topic can be found in [184].[14]

**Conditions for SOC:** Although there is no generally agreed upon litmus test for SOC behavior, a number of basic conditions are expected. Given a system of loosely coupled components, an SOC system will evolve to a (critical) stationary state where interactions at a local level (i.e. localized disturbances) can propagate and reach any size (including the size of the entire system) with a non-negligible probability. Such a state is popularly indicated through the presence of power laws in spatial and temporal properties.

For an SOC system, critical dynamics should not be fragile to experimental conditions; otherwise this would indicate some sort of tuning is necessary. Hence, a broad range of experimental conditions should be tested before any claims of SOC are made. Finally, since SOC systems are attracted to a critical state, but do not start in one, it is expected that some transient exists where the system is initially not critical (i.e. power law deviations exist during the transient). Using these general conditions as a guide, the next section considers whether Evolutionary Algorithms are compatible with SOC theory using arguments based on the ETV results from this chapter.

### 4.3.2 Compatibility of EA with SOC

EA populations are already known to be loosely coupled dynamical systems and so it is assumed that some form of self-organization will take place. The question is simply whether the attractor for the system is a critical one. To provide support for this statement, one must show that i) the distribution of disturbance sizes fits a power law and ii) the power law is a robust property that occurs under many conditions. Disturbances to an EA population can be measured by ETV as described below.

**Defining ETV as a measure of disturbance:** First a brief explanation is needed of how ETV is a measure of disturbance size in EA populations. The explanation given here for describing population dynamics on a graph is equivalent to the description provided for the measurement of ETV that is given in the previous chapter.

[14] Critical phenomena also has relevance to other aspects of evolution which are not covered in this thesis. The interested reader can find studies on its relevance to evolvability and fitness landscapes in [17], and its relevance to evolutionary dynamics on neutral networks in [80].

The spatio-temporal process of EA population dynamics can be represented by a sparse directed graph where individuals are represented by nodes and directed connections between nodes indicate that one individual (with outgoing connection) has influenced the creation of another (with incoming connection). As has been done in similar studies, weak interactions (i.e. connections) in the system are ignored meaning in this case that only the dominant parent is considered to be connected to an offspring.[15] This results in a graph topology where each node has only one input connection. This graphical model of EA dynamics is identical to the EA genealogical tree shown in the last chapter.

The creation of each new node (i.e. individual) is assumed to represent a new disturbance to the system (i.e. genotypic and phenotypic change to population makeup). To observe the growth of a disturbance, we look at the total number of nodes in the current state of the system (i.e. population members) that have a path leading to this node. This represents the current size of the system that is affected by a disturbance at a given point in time (i.e. generation). The maximum impact of the disturbance would be calculated in an identical fashion to the ETV measurement. Therefore, one can see that disturbances to genotypic and phenotypic characteristics of the population can propagate from one node to another and the eventual size of the disturbance is precisely what ETV measures.

**Evidence that EA populations are self-organized to a critical state:** Tests conducted in this chapter under a broad range of experimental conditions have indicated that the ETV size and age distributions are well approximated by power laws and that the power laws are a robust property of the system. The experimental results shown in Figure 4-6 also indicate that a short transient occurs before the population dynamics organize to a stationary power law distribution. During this transient, larger ETV are less likely to be observed. This indicates that disturbances initially remain localized and that the EA population starts in an ordered state but is quickly driven to a critical one. In summary, the results from this chapter demonstrate that Evolutionary Algorithms meet the conditions necessary for compatibility with SOC theory.

[15] The use of a threshold criteria for considering only large interactions in a dynamical system is a standard approach employed when one only wants to study the existence of interactions and not the relative strengths of interactions, the latter being more complicated. For example see [185].

**Alternative Explanations:** Despite this apparent compatibility, it is also worth pointing out that both spatial and temporal properties of EA population dynamics were derived from genealogical graphs which are a type of branching process. Furthermore, it is known that branching processes exhibit criticality when the average death rate equals the average growth rate of new branches (e.g. see [186] and [187]). This condition is the same as the requirement in static EA population sizes that the number of individuals removed from an EA population is equal to the number added to the population. Hence, it is likely that the use of a static EA population size is an important contributing factor to the ETV results presented in this chapter.

Although the underlying causes are not fully understood, the ETV results from this chapter allow for tentative statements to be made regarding the sufficient conditions for a system's historical coupling to self-organize to a critical state. In particular, this behavior has been shown to take place in a closed system that contains a reproducing Panmictic population with a static population size. More experimentation is needed to further expand our understanding of this aspect of EA dynamics. It would also be a significant contribution if this form of dynamical behavior could be related back to the dynamics of an EA in parameter space.

## 4.4 Relevance to EA research

What is most remarkable from the results of this chapter is not that measurements of EA population dynamics (as measured by ETV) fit a power law. What is remarkable is how little sensitivity the measurement results displayed to the selection pressure, the fitness landscape, or the medium (artificial or natural) in which evolution took place.

Furthermore, many other complex systems that are unrelated to evolution, ranging from earthquakes to solar flares to turbulence, also spontaneously organize to display similar spatial and temporal patterns (see [185] and references therein). This is of importance to Evolutionary Computation research because it indicates that at least some properties of natural evolution are not reliant on the specific context of "nature" and instead are a consequence of very general and easily reproduced conditions and driving forces.

### 4.4.1 Impetus for SOTEA Chapter

In recent years there has been growing evidence that many significant features of biology are not a consequence of natural selection but instead are a result of physical conditions and constraints. For instance, genome complexification models have been developed which, when randomly evolved (without a particular selection pressure), are able to generate topological characteristics that are similar to gene regulatory networks [188] and protein interaction networks [189].

In another important development, characteristics such as modularity and hierarchy, which are heavily exploited in natural evolution, have been found to emerge from simple localized rules [190] and do not require the presence of natural selection. Models of genome complexification have also been proposed recently where modularity is expected to emerge without the influence of natural selection [20].

These findings suggests that the unique quality of life is not generated solely from natural selection in reproducing populations but is also heavily reliant on the physical laws and constraints that are imposed on these evolving systems.

If our goals as EA researchers are to mimic the salient features of life in order to exploit it for purposes of optimization, it behooves us to actively explore what other conditions (beyond Darwinian theory) are necessary to acquire the robust and adaptive properties of life. Furthermore, with growing evidence that Darwinian principles only provide a partial explanation for life, one could reasonably speculate that only so much progress is possible in EA research from tweaking the traditional controls of natural selection (e.g. through the development of search operators and selection pressures).

On the other hand, the results from this chapter have shown that other factors such as spatial restrictions can cause significant changes to large scale dynamical features of a system. These results have also indicated that some form of self-organization already occurs in EA population dynamics which is notably distinct from its well-known organization in parameter space. This raises the question as to what other self-organizing processes occur in nature and could be of benefit to EA performance. The final chapter of this thesis focuses squarely on these issues.

# Chapter 5 Self-Organizing Topology Evolutionary Algorithms

Within the last several years, it was discovered that the interaction networks of complex biological systems have evolved to take on several non-random topological characteristics, some of which are believed to positively impact system robustness. A number of network growth models have also been discovered that can successfully recreate many of these structural characteristics using simple rules and in the absence of a selection pressure.

In the previous chapter, it was shown that spatial constraints (i.e. population topology) have a significant impact on EA population dynamics. This chapter focuses on ways in which the population topology can self-organize to exhibit topological characteristics similar to complex biological systems. The aim of this chapter is not only to mimic the structural characteristics of biological systems but also to acquire some of the desirable qualities found in these systems.

The next section presents a critical review of previous work related to the application, characterization, and evolution of interaction networks. This is followed by Section 5.2 which presents the motivations and aims of this chapter. Section 5.3 then describes the first of two network models. The first model is designed to sustain population diversity in rugged fitness landscapes which it accomplishes, in part, by mimicking the process of genome complexification in natural evolution. The second model is described in Section 5.4 and is designed for the purpose of evolving important topological properties such as modularity. This model is also designed with a focus on EA performance with tests conducted on a number of artificial test functions and engineering design problems. The results from these experiments provide strong evidence that the new Self-Organizing Topology Evolutionary Algorithms (SOTEA) are able to exhibit robust search behavior with strong performance over both short and long time scales.

## *5.1 Critical Review of Previous Work*

This section provides a review of topics that are relevant to the work presented in this chapter. It starts by reviewing approaches to constraining interactions in population-based systems with a focus on spatially distributed systems that are defined on a network. The section then reviews metrics for characterizing network topology. The section concludes by presenting a number of network models that have been successful in mimicking the topological characteristics of complex biological systems.

### 5.1.1 Interaction Constraints

In recent years, there has been an increasing awareness of the importance of locality or interaction constraints when modeling complex systems. Restricting interactions in population-based systems has been a key factor in topics such as robustness against parasitic invasion [191], [192], enabling speciation [193], sustaining population diversity in rugged landscapes [136 2007), 2007)], the emergence of cooperative behavior [18], and robustness to random attack [75]. Furthermore, convincing arguments have been made for its role in natural evolution and in particular, its impact on system evolvability [17].

Parallel developments have also taken place in population based search heuristics such as Evolutionary Algorithms, where it has been recognized that restricting interactions between population members can result in significant changes to algorithm behavior. This has been observed in several seemingly disparate topics such as the age restrictions present in the Age Layered Population Structure (ALPS) for Genetic Algorithms [99], the genealogical and phenotypic restrictions present in Deterministic Crowding [84], coarse-grained restrictions in interactions between heterogeneous subpopulations [100], and explicit static topologies for constraining interactions in the cellular Genetic Algorithm (cGA).

#### 5.1.1.1 Population Networks for Evolutionary Algorithms

This chapter limits its focus to interaction constraints that are implemented by defining an EA population on a network. The use of explicitly defined interaction networks provides a useful framework for understanding system constraints and their impact on system dynamics. This is, in part, because network representations of systems can be probed using a number of tools developed in statistical mechanics.

Defining an EA population on a network impacts an EA through the localization of genetic operators. For instance, actions such as reproduction and selection only occur among individuals directly connected (i.e. linked) or near each other in the network. The three types of population structures typically considered for EA populations are shown on the top row of Figure 5-1.

The fully connected graph in Figure 5-1a represents the canonical EA design, which is referred to in this thesis as the Panmictic GA. Here, each individual (represented by nodes in the graph) can interact with every other individual such that no definition of locality is possible. The network in Figure 5-1b represents a typical island model population structure where individuals exist in fully-connected subgroups which are largely isolated from other population subgroups. Here the large arrows represent interactions which take place between subgroups but occur at a time scale much greater than that of interactions within subgroups. As a consequence of this setup, the locality of island model networks is defined on a scale that is significantly larger than the individual. The final EA structure shown in Figure 5-1c represents a cellular EA population structure which is referred to in this thesis as the cellular Genetic Algorithm or cGA. Similar to cellular Automata, the network of interactions takes on a lattice structure with interactions constrained by the dimensionality of the lattice space. With the cellular GA, each individual has a unique environment defined by its own unique set of interactions which is referred to as a neighborhood.

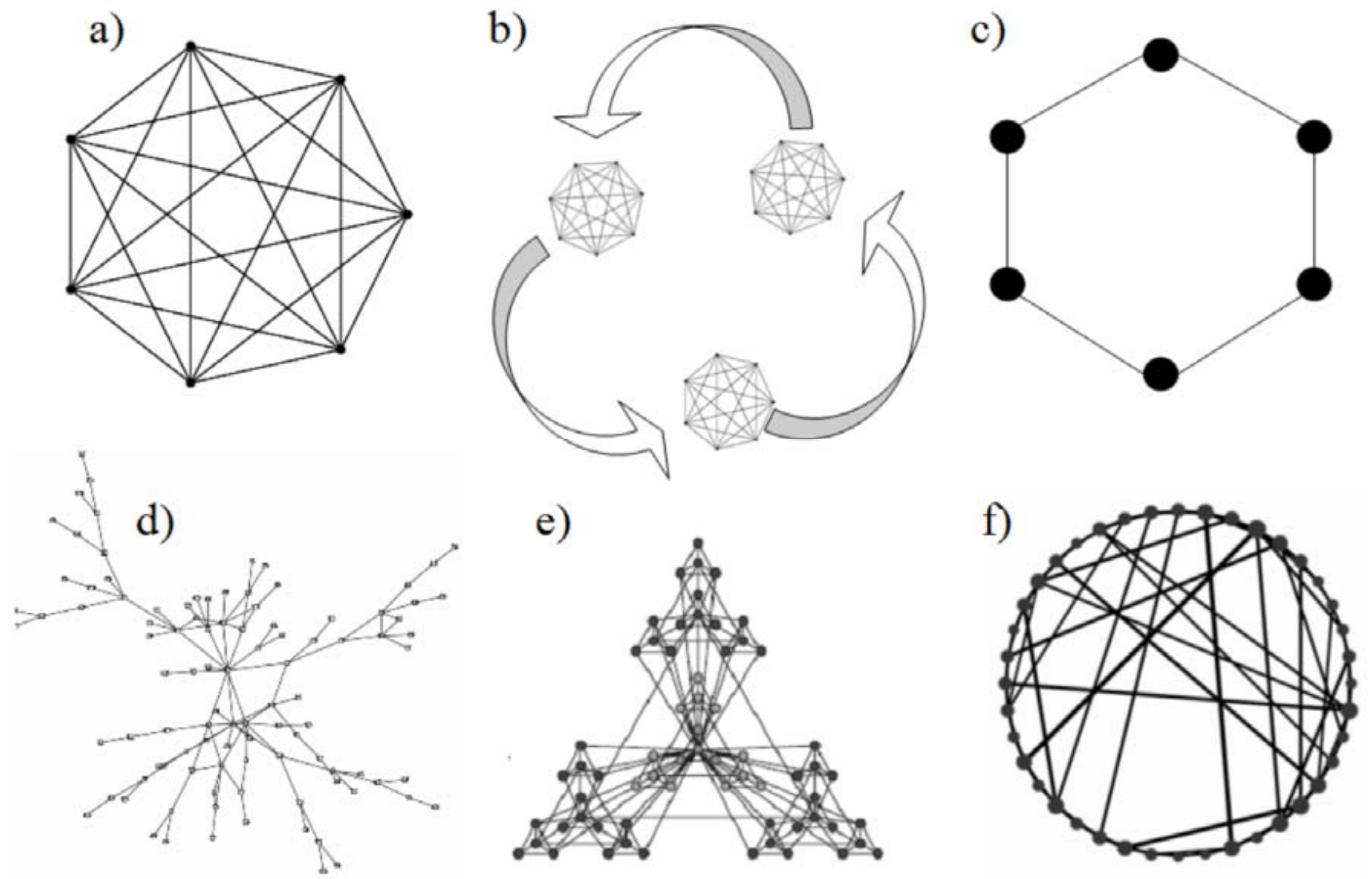

**Figure 5-1: Examples of interaction networks. The networks on the top represent commonly used EA population structures and are known as (from left to right) Panmictic, island model, and cellular population structures. Networks at the bottom have been developed with one or more characteristics of complex biological networks and are classified as (from left to right) Self-Organizing Networks (presented here), Hierarchical Networks [194], and Small World Networks [195]. Figure 1e is reprinted with permission from AAAS.**

The ratio of neighborhood size (i.e. number of connections per node) to system size (i.e. total number of nodes) can be seen as a measure of locality and it is worth noting that this ratio decreases in the EA population structures from left to right on the top row of Figure 5-1. Although the three population structures clearly have different degrees of locality, they also have some important similarities. For each population structure, the nodes within the network each have the exact same number of interactions and the same type of interactions (i.e. regular graphs). Furthermore, the networks for all three cases are static and predefined.

### 5.1.2 Structural Characteristics of Complex Networks

#### 5.1.2.1 Properties of real networks

Many natural and man made systems consist of a large number of dynamical interacting components. Examples are seen in biology (metabolic networks, protein interaction networks, gene regulatory networks, food webs, neural networks), social systems (coauthorship, relationships) and man made systems (World Wide Web, Internet, power grids).

Despite the significant simplifications necessary to create network representations of these systems and despite their inherent differences in scale, environmental context and functionality, most real networks have a great deal of similarity in their topological properties. These similarities include 1) small characteristic path lengths, 2) high clustering coefficients, 3) fat-tailed degree distributions (e.g. power law), 4) degree correlations, 5) low average connectivity, as well as other properties reviewed in [196]. Each of these features are notably distinct from random graphs and regular lattices. The next section reviews a number of topological properties that are commonly measured when studying networks. Comprehensive reviews of this topic are provided in [196], [197], and [198].

#### 5.1.2.2 Topological Property Metrics

To help understand the interaction networks of complex systems, a few simple measures are introduced which are commonly used to assess network structural characteristics. Throughout this paper, networks are represented by an adjacency matrix $J$ such that individuals $i$ and $j$ are connected (not connected) when $J_{ij}=1$ ($J_{ij}=0$).

**Characteristic Path Length:** The path length is the shortest distance between two nodes in a network. The characteristic path length $L$ is the average path length over all node pair combinations in a network. Generally, $L$ grows very slowly with increasing system size (e.g. population size) $N$ in complex systems. For instance, networks exhibiting the "Small World" property, such as the network in Figure 5-1f, have $L$ proportional to log $N$ [199].

**Degree Average:** The degree $k_i$ is the number of connections that node $i$ has with other nodes in the network which is defined in (5-1). The degree average $k_{ave}$ is simply $k$ averaged over all nodes in the network. The degree average is expected to remain small, even for large networks, as reviewed in [196].

$$k_i = \sum_{j=1}^{N} J_{i,j} \tag{5-1}$$

**Degree Distribution:** The degree distribution has been found to closely approximate a power law distribution for biological complex systems with power law and exponential distributions often fitting abiotic complex systems [198]. Networks which display a power law $k$ distribution are often referred to as scale free networks in reference to the scale invariance of $k$.

**Clustering Coefficient:** Many complex biological systems have high levels of modularity which is typically indicated by the clustering coefficient. The clustering coefficient $c_i$ is a measure of how well the neighbors of a given node are locally interconnected. More specifically, this is defined in (5-2) as the ratio between the number of connections $e_i$ among the $k_i$ neighbors of node $i$ and the maximum possible number of connections between these neighbors which is $k_i(k_i\text{-}1)/2$.

$$c_i = \frac{2e_i}{k_i(k_i - 1)} \tag{5-2}$$

Although in practice, more efficient calculation methods are used, $e_i$ can be formally defined using the adjacency matrix $J$ as shown in (5-3).

$$e_i = \sum_{j=1}^{N}\left( J_{ij} \sum_{k=1}^{N} J_{ik} J_{jk} \right), i \neq j \neq k \tag{5-3}$$

**Clustering-Degree Correlations:** A common feature of biological and social systems is the existence of an hierarchical architecture. Such an architecture implies that sparsely connected nodes form tight modular units or clusters and communication paths between these modular units are maintained via the presence of a few highly connected hubs [199]. Figure 5-1e shows a network with these hallmark signs of modularity and hierarchy which was grown using the deterministic models presented in [194].

The existence of hierarchy in a network is typically measured by looking at the correlation between the clustering coefficient and the node degree. Based on the description given above, an hierarchical network is expected to exhibit higher connectivity for nodes with low clustering (i.e. hubs) and vice versa. Furthermore, for the feature of hierarchy to be a scale invariant property of the system, $c$ should have a power law dependence on $k$.

**Degree-Degree Correlations:** For many complex networks, there exist degree correlations such that the probability that a node of degree $k$ is connected to another node of degree $k`$ depends on $k$. This correlation is typically measured by first calculating the average nearest neighbors degree $k_{NN,i}$ which is defined in (5-4).

$$k_{NN,i} = \frac{1}{k_i} \sum_{j=1}^{N} J_{i,j} k_j \quad \textbf{(5-4)}$$

Networks are classified as assortative if $k_{NN}$ increases with $k$ or disassortative if $k_{NN}$ decreases with $k$. Degree correlations are often reported as the value of the slope $\upsilon$ for $k_{NN}$ as a linear function of $k$.

**Random Networks:** Thus far, only qualitative statements have been given regarding the topological properties of complex networks. In practically all cases, when topological properties are mentioned as being large or small (as has been mentioned above), the statements are referring to property values in relation to those values observed in random graphs and particularly the models developed by Erdös and Rényi [200], [201]. As reviewed in [197], random graphs have i) a characteristic path length $L_{Rand}$ similar to that observed in complex networks and approximated by (5-5), ii) a Poisson degree distribution (as opposed to the fat tailed degree distribution in complex networks), and iii) a clustering coefficient $c_{Rand}$ given by (5-6) which is orders of magnitude smaller than what is typically seen in complex networks [195]. Random graphs also do not exhibit any degree correlations or correlations between the degree and the clustering coefficient.

$$L_{Rand} \approx \frac{\ln(N)}{\ln(k_{Ave})} \quad \textbf{(5-5)}$$

$$c_{Rand} = \frac{k_{Ave}}{N} \quad \textbf{(5-6)}$$

### 5.1.3 Network Evolution Models

In order to mimic complex systems, it is important to understand how they obtain their interesting behaviors and properties. For both man-made and biological complex systems, it is generally understood that the development of interaction networks in these systems occurs through a process of constrained growth. Examples would include growth of the World Wide Web, the developmental process in multi-cellular organisms, and the complexification of the genome.

Over the last decade, substantial progress has been made in the development of network growth models which can evolve to display characteristics similar to real systems. Exemplars of this success can be seen in the Barabasi-Albert (BA) Model [202], the Duplication and Divergence (DD) Model [203], the intrinsic fitness models of [204] and the random walk models of [190]. Common to many successful models is the emergence of relevant network characteristics, such as those previously mentioned (e.g. $L \sim \log N$, Power law $k$ distribution), through the use of simple, locally defined rules which constrain structural dynamics (including, but not limited to, network growth). Furthermore, these structural dynamics are driven by one or more state properties of the nodes. This simply means that connections in the network change and nodes are added or removed with a bias derived by property values assigned or calculated for each node. Properties that have been used in models include the degree of a node $k$ [202], measures of node modularity [205], as well as measures of node fitness [204]. The remainder of this section presents several network evolution models that are important contributions to the field and have been particularly important in development of the SOTEA models presented in this thesis.

#### 5.1.3.1 The BA Model:

The Barabási-Albert (BA) model works on the basis of network growth and preferential attachment. These principles are inspired by experiences with real systems and they are prevalent in a number of large complex systems. Examples of systems driven by growth include the World Wide Web, collaboration networks, genome complexification and many more. The concept of preferential attachment is also observed in many systems such as citation networks where a new manuscript is more likely to cite well-known and already well-cited papers compared with its less-cited peer papers. Preferential attachment is also seen in a number of other social networks as mentioned in [202].

**Model Description:** Starting with a small number ($m_0$) of nodes, at every time step a new node is added with $m$ ($< m_0$) links that connect the new node to $m$ nodes already present in the system. To incorporate preferential attachment, the probability $P_{N1,N2}$ that a new node $N1$ is connected to an existing node $N2$ depends on the degree $k_{N2}$ of node $N2$. Furthermore, it is assumed that this dependence is linear as expressed in (5-7).

$$P_{N1,N2} = \frac{k_{N2}}{\sum_{i=1}^{N} k_i} \tag{5-7}$$

**Model Characteristics:** This model creates networks with power law $k$ distributions with exponent similar to that observed in real systems. These networks also have a path length $L \sim \log \log N$ which obviously grows very slowly with increasing system size $N$. However, these models produce networks with no correlation between $k$ and $c$ [199].

#### 5.1.3.2 The Duplication and Divergence Model

**Genome Complexification:** As far as complex systems are concerned, the genome and its associated expression represent the largest and most interesting case of network evolution. Hence it is of great interest that large data sets of these systems are now available as well as the tools necessary for probing their structural organization. For instance, recent analysis of protein interaction networks and metabolic networks have found them to be characterized as having power law degree distributions, high modularity, low characteristic path lengths, and a low degree average [206], [195], [207].

Evolutionary processes associated with genome complexification are known to be initiated by a process of gene duplication and gene mutation, which is also referred to as Duplication and Divergence (DD). In the review presented in [208] , studies are cited which have found that roughly 40% of the human genome can be confirmed as being derived from past duplication events (with even higher values observed in other species).

Due to the nature of the divergence process, it is also likely that the 40% estimate is a conservative lower bound on the actual number of duplicated genes in the human genome. Knowing that gene divergence occurs by a series of random mutations, the process approximates a random walk which acts to reduce (over time) any ability to recognize two genes as having a similar historical origin. Hence, it is speculated here that DD is the predominate cause of genome complexification.

Although several duplication and divergence models are available from the literature, the one presented in [189] has been selected due to its simplicity and its proven capacity to generate power laws for $k$.

**Model Description:** Starting with an unspecified initial network size, at every time step a node is randomly chosen and duplicated. Links of the duplicated node are removed with probability $\delta$. New links are added to the duplicated node between itself and randomly selected nodes in the network with probability $\alpha$. In this model, $\delta$ is set to 0.53 and $\alpha$ is set to $0.06/N$ where $N$ is the network size at a given time step.

The duplication step in the model represents gene duplication where the original and duplicated genes retain the same structural properties meaning they initially have an identical set of interactions. The rewiring steps (involving probabilities $\delta$ and $\alpha$) represent mutations in the duplicated gene which cause its set of interactions to diverge from those of the original gene. The parameter settings listed above for $\delta$ and $\alpha$ are set in [189] based on empirical observations of protein interactions networks in yeast and the interaction networks of other complex systems.

**Simplifying Assumptions:** Although this model for genome complexification was selected in part due to its simplicity, it is still important to highlight the simplifying assumptions that have been made.

- The first simplification is that it does not allow for the presence of multiple duplications at a single instance in time. This is known to occur in natural evolution with duplication involving sizes up to and including the entire genome [209], [210]. A model considers multi-gene duplications is presented in [203].

- Another difference between this DD model and nature is that gene divergence mechanisms can potentially take place for both the duplicated and original genes. Also divergence occurs as a slow and possibly continuous process instead of happening in a single time step.

- Finally the removal of genes from the genome is also neglected thereby implying that natural selection pressures present in real genome complexification will have roughly the same impact on topology as random rewiring of the duplicated gene.

It is not clear what impact these additional features would have on the model and its structural characteristics, however it is interesting to notice (as indicated below) that this model does provide many similarities to complex biological systems without the presence of natural selection pressures or the other assumptions listed above.

**Model Characteristics:** In [189], structural characteristics of this DD model are compared with data available on the yeast proteome. Results indicate the model creates a network with characteristics similar to yeast for values of $k_{ave}$, $k$ distribution, $c_{ave}$, and $L$. Correlation measurements were not considered in this study.

#### 5.1.3.3 The Fitness Model:

The model presented in [204] proposes a "good-get-richer" mechanism for network dynamics where nodes of higher fitness are more likely to become highly connected. This is presented as an alternative to preferential attachment or the so called "rich-get-richer" schemes present in the BA and DD models where an historical bias in network connectivity drives future connectivity.

The fitness model is based on the concept of mutual attraction. One example they provide of such a system is a sexual interaction network where it is assumed interactions take place due to mutual attraction between two partners. They argue that knowledge of sexual promiscuity (i.e. knowledge of $k$ which is a precondition for preferential attachment) would actually have the reverse impact on the probability of interactions in such networks. Another example they provide is the protein interaction networks inside cells where interactions are driven by chemical affinity. This second example is less convincing however because protein specificity is an evolved trait meaning that both historical (i.e. evolutionary) bias as well as principles of mutual attraction play a role for such systems. The fitness model description given in [204] is provided below.

**Model Description:** In this version of the fitness model, they neglect mechanisms for network growth and instead start off with a fixed number of nodes *N* containing no links. For every node, a fitness value $x_i$ is assigned randomly from a probability distribution $\rho(x)$. For every possible pair of nodes *i*, *j*, a link is created with probability $f(x_i, x_j)$ which is given by (5-8) where $x_M$ is the maximum value possible for *x*.

$$f(x_i, x_j) = \frac{(x_i x_j)}{x_M^2} \quad \textbf{(5-8)}$$

**Model Characteristics:** When $\rho(x)$ takes an exponential form, this model creates networks with power law *k* distributions with the power law exponent similar to that observed in real systems. The resulting networks also have clear correlations in $k_{NN}$-*k*, and in *c*-*k* [204].

## 5.2 Motivations and Aims

Currently implemented network structures for EA populations have proved beneficial to EA performance, however the population structures do not actually resemble the interaction networks of complex biological systems as indicated in the previous review. This puts into question how "nature-inspired" these EA designs are and what additional benefits might be derived from more accurate representations of the structure and dynamics of complex systems.

Over the last several years, the interaction networks of many complex systems have been studied. It is now known that these systems display some interesting non-random characteristics that are similar among many biological and even manmade systems [197]. These characteristics are believed to be highly relevant to the behavior of these systems and particularly important to emergent qualities such as robustness.

A primary aim of this chapter is to improve upon the performance and behavior of distributed Evolutionary Algorithms by mimicking the self-organizing processes of complex systems within the population topology. Evolutionary Algorithms which have this behavior are referred to in this thesis as Self-Organizing Topology Evolutionary Algorithms (SOTEA).

To date, few have investigated the importance of dynamic population topologies for an EA. One exception is seen in [135] where the grid shape of a cellular GA adapts in response to performance data using a predefined adaptive strategy. In that work, structural changes are globally controlled using statistics on system behavior. SOTEA algorithms, on the other hand, adapt to local conditions through a coevolution of network structure and population dynamics.

## 5.3 SOTEA model I

This section describes the first of two SOTEA designs that are developed and tested in this thesis. As previously mentioned, a general aim of this chapter is to create EA population networks which are topologically similar to the interaction networks of complex biological systems. This first SOTEA model also looks at how this can benefit EA behavior on optimization problems containing rugged fitness landscapes.

Section 5.3.1 describes SOTEA and a cGA variant that is used for comparison purposes. Section 5.3.2 describes the experimental setup including a description of a tunable fitness landscape that is used to test algorithm behavior when exposed to different amounts of landscape ruggedness. The results are presented in Section 5.3.3 including algorithm performance, analysis of population topology, and the impact of SOTEA design features and fitness landscape features on population diversity. From these experiments, it is found that SOTEA exhibits strong performance and is able to sustain high levels of population diversity for evolution on rugged fitness landscapes. The population topology for SOTEA is also found to have some similarities to known features of complex biological systems. This is followed by a discussion of these results in Section 5.3.4 as well as conclusions in Section 5.3.5.

### 5.3.1 Model Description

For all EA designs, the population is defined on a network. Besides the trivial case where the network is fully connected (i.e. Panmictic GA), two other network designs are used and are referred to as the cellular GA (cGA) and the Self-Organizing Topology Evolutionary Algorithm (SOTEA). For the cGA and SOTEA, the population is initially defined in a ring structure with each node connected to exactly two others (e.g. Figure 5-1c). A change to the network structure (i.e. network dynamics) simply refers to the addition or removal of nodes or links. The rules for defining network dynamics are described next.

#### 5.3.1.1 SOTEA and cGA Network Dynamics

For both the cGA and SOTEA, a node is only added to the network when a new offspring is added to the population and a node is only removed when an individual dies. Network changes due to offspring creation are referred to as reproduction rules and changes due to death of individuals are referred to as competition rules. The reproduction and competition rules define how network dynamics occur and are described next.

**Reproduction Rule:** The reproduction rule (described in Figure 5-2) is used in SOTEA and the cGA when a new offspring is created. The first step in the reproduction rule involves making a copy of a parent and then mutating that copy to create an offspring.[16]

Structural changes from the reproduction rule involve the addition of a new node (offspring) to the network, connection of the offspring to its parent, and then (depending on the EA design) the possibility of additional connections being added to the offspring node and the possibility of connections being removed from the parent node. Complete details of the addition and removal of connections in the reproduction rule are provided in Figure 5-2.

[16] Notice this means an offspring only has a single parent

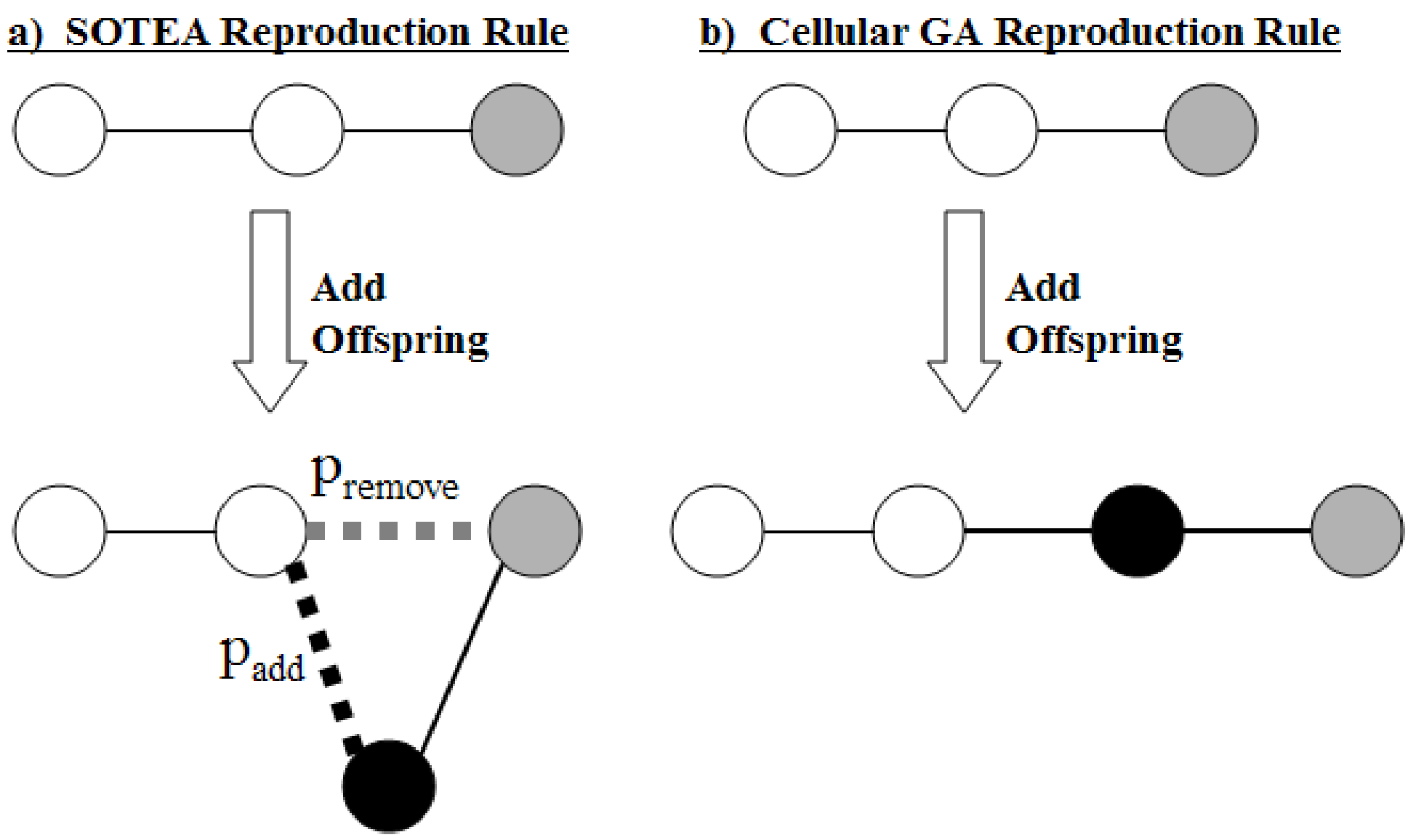

**Figure 5-2: Reproduction rules that change the population structure for SOTEA and the cellular GA. a) SOTEA Reproduction: When an offspring is created (by asexual reproduction), a new node (shown in black) is added to the network through a connection to its parent (shown in gray). Each of the parent's connections are then inherited by the offspring (black dotted line) with probability $P_{add}$ followed by each of the inherited connections being lost by the parent (gray dotted line) with probability $P_{remove}$. Unless stated otherwise, the parameters are set as $P_{add} = P_{remove} = 10\%$. This particular rule is loosely based on established models for genome complexification [203]. b) cellular GA Reproduction: When an offspring is created, a new node (shown in black) is added to the network and connected to its parent (shown in gray). One of the parent's connections is then transferred to the offspring, which allows the network to maintain a ring topology.**

It is worth noting that the reproduction rule represents the only difference between SOTEA and the cellular GA. With SOTEA, the addition of new nodes causes changes to the network topology (see Figure 5-2a). These changes to network structure turn out to be a crucial source of structural innovation needed for the self-organization of the SOTEA network.

**Competition Rule:** The competition rule (described in Figure 5-3) is the same for SOTEA and the cellular GA. With this rule, a randomly selected individual tries to kill its weakest (i.e. least fit) neighbor. If instead, the selected individual is worse than its worst neighbor, then it will die. Structural changes from the competition rule involve removal of the dead individual and the transfer of its connections to the individual that survived.

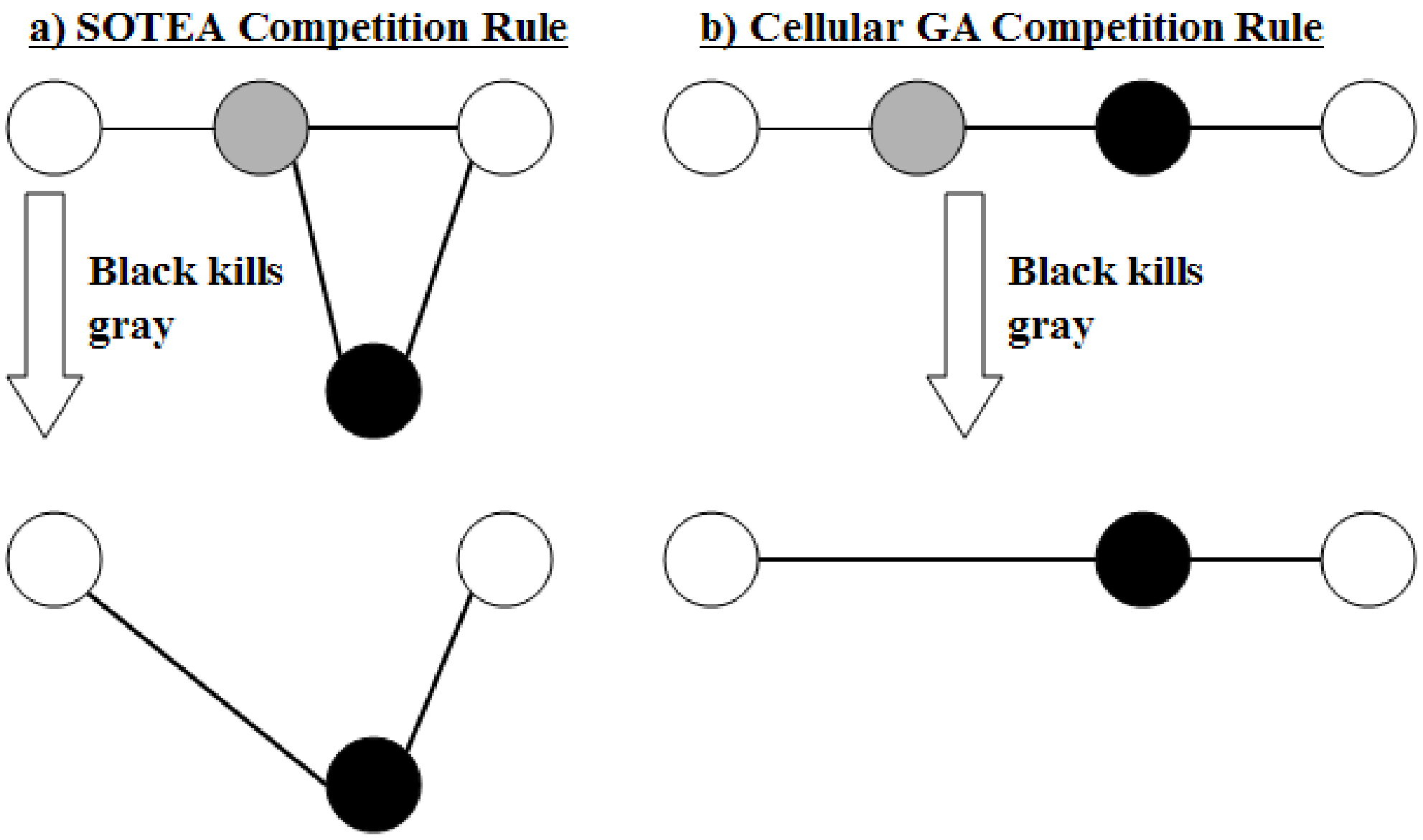

**Figure 5-3: Competition rules that change the population structure for SOTEA and the cellular GA. The details of the competition rule are the same for SOTEA and the cGA, however examples are given for both EA designs in this figure. Competition rule: The first step is to select an individual at random. This individual then decides to compete for survival with its least fit neighbor. When these two individuals compete for survival such as the nodes shown in black and gray, the less fit individual is killed. The winning individual (shown in black) inherits all connections from the losing individual (shown in gray) that weren't already in the winning individual's neighborhood. Finally, the losing individual is removed from the network.**

This rule is particularly important because it allows for structural changes to depend on node states. Figure 5-4 is provided to help clarify this point. Notice that once a node has been selected for the competition rule, this node must decide who to compete with. The decision of who to compete with depends on which of the nodes is worst in the neighborhood. As a result, structural changes are always driven towards those nodes with the lowest fitness. Notice that if an individual decided to kill one of its neighbors at random then this decision would no longer depend on the node states and the network structural dynamics would no longer depend on (i.e. be coupled to) the population dynamics.

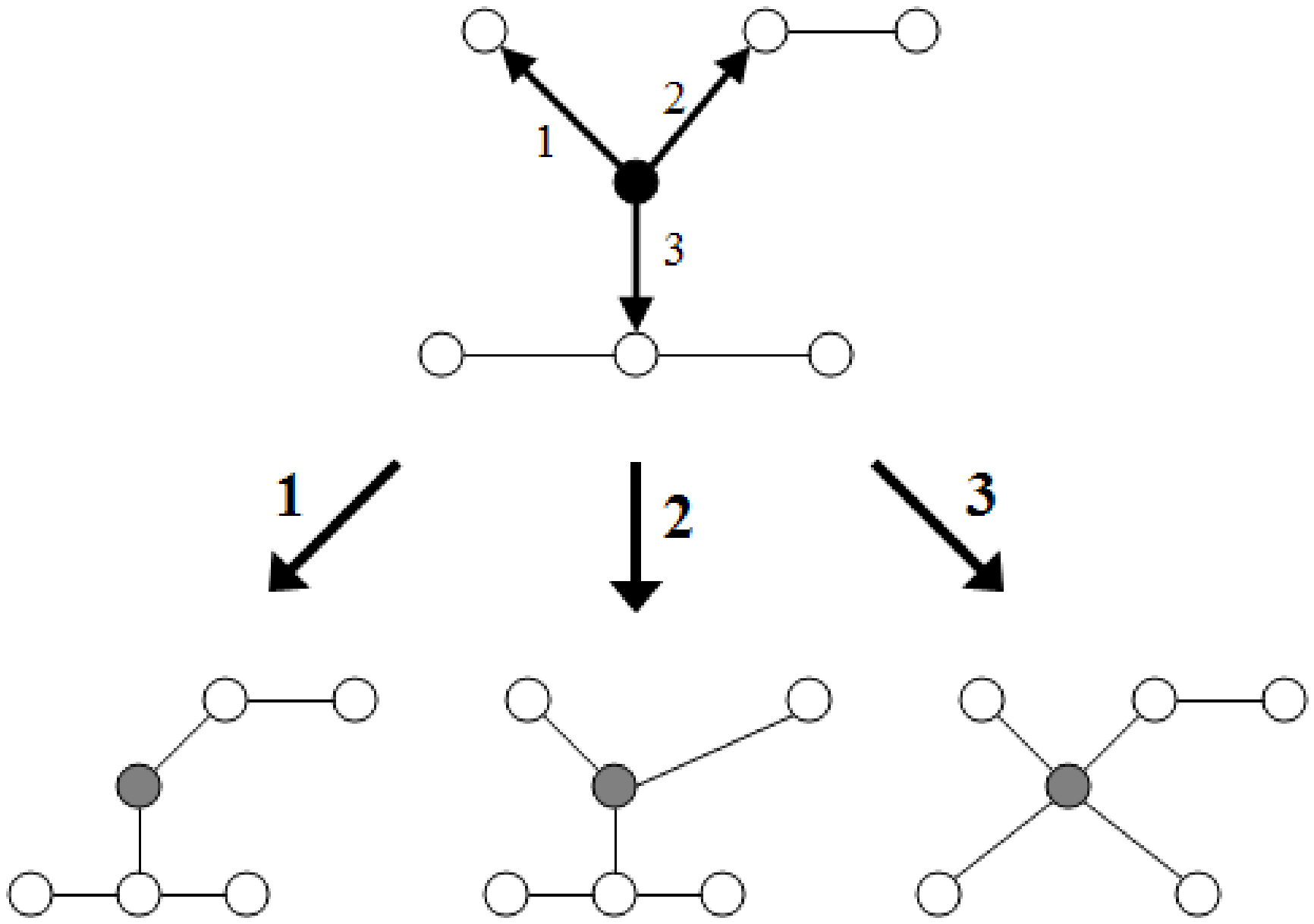

**Figure 5-4: This figure shows how structural changes from SOTEA's competition rule depend on the fitness of individuals in the network. Starting with the network at the top, the individual represented by the black node must decide which of its neighbors it will try to kill. The networks at the bottom show what would happen if neighbor 1, 2, or 3 had been the least fit in the black node's neighborhood. Each of the choices creates a new structure that is different from the other choices. Notice that for the networks on the bottom, the black node has been changed to gray. This is to indicate that either the black node or the white neighbor could have won the competition (the structure is the same in either case).**

#### 5.3.1.2 SOTEA and cGA State Dynamics

To mimic the interaction networks of complex systems, it is important to recognize that state dynamics occurring on these networks play a significant role in the system's behavior. In complex systems, the states of a node are (by definition) dependent upon the states of neighboring (i.e. connected) nodes. Significant progress has taken place recently in understanding the state dynamics of complex systems. Some current directions of research include exploring the synchronization of component states [211], [212], robustness of dynamical expression [213], [75] and the coupled dynamics of states and network structures [214], [215], [216].

In the previous section it was pointed out that the competition rule forces changes in network structure to depend on the fitness (i.e. the state) of population members. As a consequence, network structural dynamics are driven by population dynamics. This section considers how a reverse coupling of state and structural dynamics could be achieved. In other words, how can a node's state depend on the states of the other nodes it is connected to.

Along these lines, a measure of fitness called *Epistatic Fitness* is defined to be sensitive to the fitness values of individuals in a neighborhood. The definition of epistatic fitness that is used in the SOTEA algorithm is provided in (5-9).

$$\text{Epistatic Fitness} = \frac{k - Rank - 1}{k} \tag{5-9}$$

In (5-9), *Rank* refers to the rank of an individual's objective function value among all of its $k$ neighbors. Here the objective function is not a direct measure of fitness but only an intermediate value used to compute (epistatic) fitness. A rank of 1 indicates that the individual is better than all its neighbors, resulting in epistatic fitness taking on its maximum value of 1. A rank of $k$+1 indicates that the individual is worse than all its neighbors, resulting in epistatic fitness taking on its minimum value of 0. The term epistatic fitness is used in reference to the measure's similarity to genetic epistasis.[17]

Using epistatic fitness (5-9) results in the fitness of an individual being dependent on the network structure. In other words, the fitness is contextual. Figure 5-5 provides an example to help clarify how (5-9) causes an individual's fitness to be dependent upon the network structure.

[17] In the Genome, genetic epistasis refers to interactions between genes which have a noticeable impact on the phenotype. Similarly, nodes in the population network will now interact in a way such that they impact each other's fitness values.

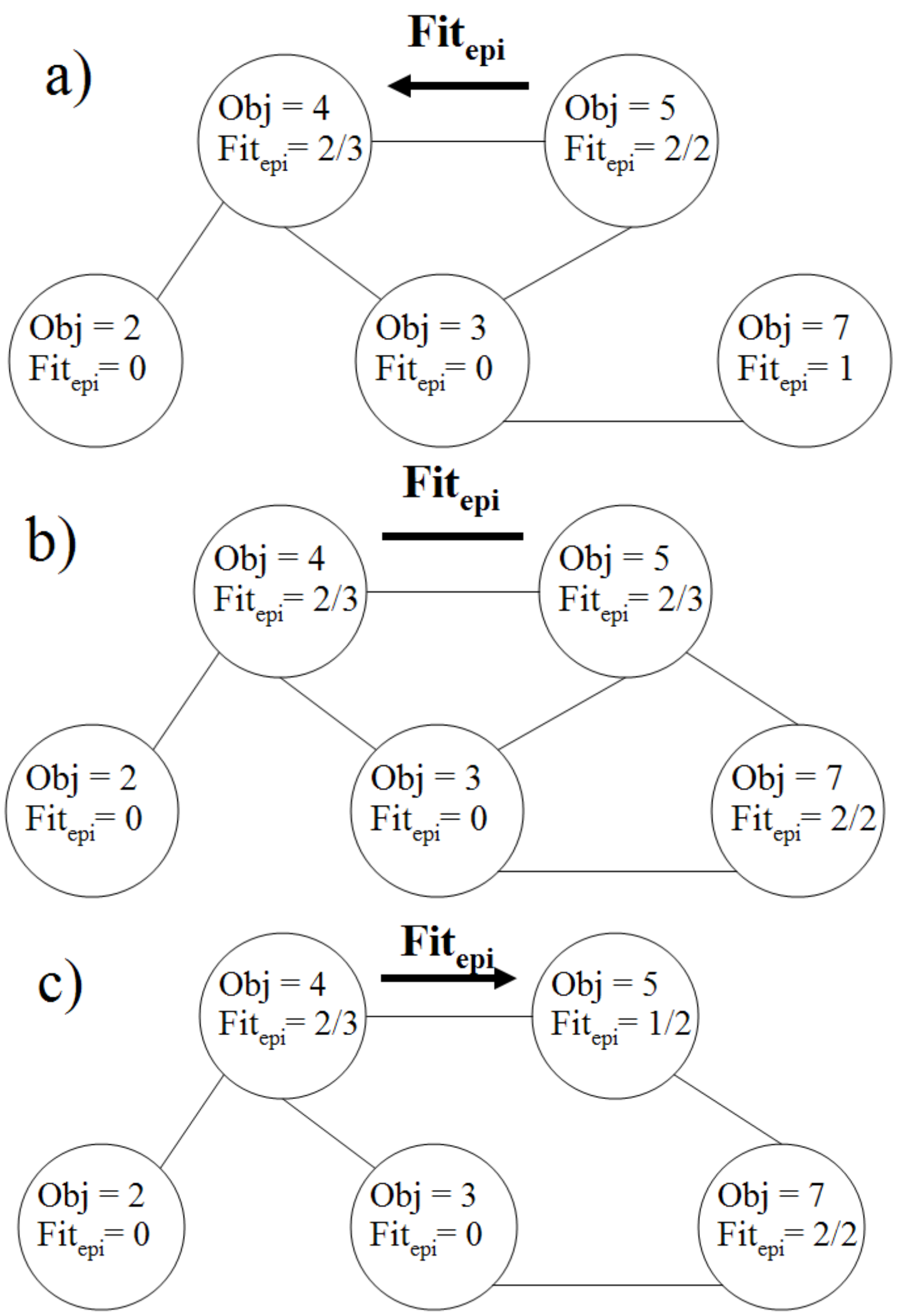

**Figure 5-5: This figure shows how the epistatic fitness ($Fit_{epi}$) defined by (5-9) causes the fitness of an individual to depend on its local neighborhood. Parts a-c of the figure show a population of five individuals defined on a network. The Objective Function Value (Obj) and epistatic fitness defined by (5-9) are provided in the top and bottom (resp.) of each node (i.e. individual). For the top two individuals in part a), an arrow is drawn towards the individual on the left to indicate it has the lower epistatic fitness. The top left individual's epistatic fitness is 2/3 because its objective function value is better than 2 of its 3 neighbors. In part b), a new connection has been added to the network causing the epistatic fitness values for the two top individuals to now be equal. Finally in part c), a connection has been removed from the network, causing the top left individual to have an epistatic fitness that is now higher than the top right node. If the top two nodes were to compete for survival based on epistatic fitness, it should now be clear that the decision of who survives (i.e. who is more fit) will depend on the neighborhoods of the individuals.**

It is important to mention that an interesting situation arises when SOTEA is used with epistatic fitness. In this case, the fitness values depend on network structure (due to epistatic fitness) and structural changes depend on fitness values (due to the competition rule). The result is a coupling of structural changes to states plus a coupling of state definitions to structure. It is believed that such a dual coupling is unique among existing network evolution models. The competition and reproduction rules that are used for SOTEA and the cellular GA are summarized in the pseudocodes in Figure 5-6.

**Pseudocode: Competition Rule (SOTEA and Cellular GA)**
- -Select Individual (randomly from Parents + Offspring)
- -Compare selected individual with its least fit* neighbor
- -Better individual inherits all links of worse individual
- -Worse individual is removed from population (and node removed from network)
- *fitness is epistatic fitness

**Pseudocode: SOTEA Reproduction Rule**
- -Add new node (offspring) to network
- -Link offspring and parent
- -Offspring inherits Parent links with probability $P_{add}$=0.1
- -If inherited, Parent loses link with probability $P_{remove}$= 0.1

**Pseudocode: Cellular GA Reproduction Rule**
- -Add new node (offspring) to network
- -Link offspring and parent
- -Offspring inherits one of Parent's links
- -Parent loses inherited link

**Figure 5-6: Pseudocode for SOTEA and cellular GA network dynamics.**

## 5.3.2 Experimental Setup

This section presents the remaining aspects of the Evolutionary Algorithm designs as well as the test function generator used in these experiments.

### 5.3.2.1 NK Landscape Test Function

The NK landscape, originally developed by Kauffman in [81], is a test function generator with a tunable amount of ruggedness and a tunable problem size. The following description of the NK landscape has been adapted from [217]. The NK landscape is a function $f: B^N \rightarrow \mathbb{R}$ where $B = [0,1]$, $N$ is the bit string length, and $K$ is the number of bits in the string that epistatically interact with each bit. To help reduce confusion with other notation in this thesis, the $N$ and $K$ parameters of the NK landscape are relabeled as $N_{NK}$ and $K_{NK}$ (resp.)

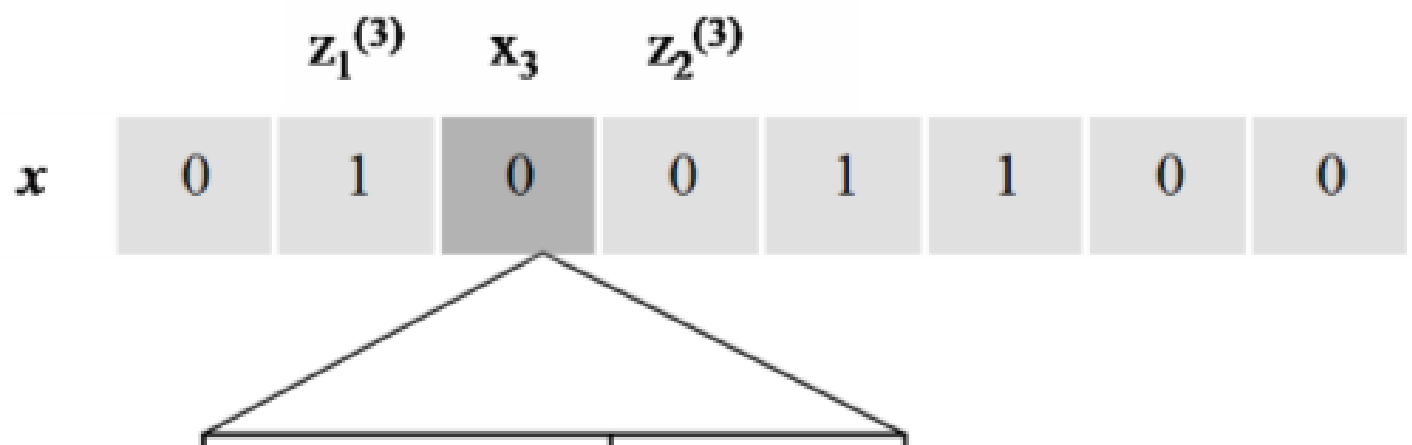

| $z_1^{(3)}, x_3, z_2^{(3)}$ | $f_3$ |
|---|---|
| 000 | 0.94 |
| 001 | 0.36 |
| 010 | 0.83 |
| 011 | 0.20 |
| **100** | **0.67** |
| 101 | 0.14 |
| 110 | 0.71 |
| 111 | 0.44 |

**Figure 5-7: An example of the fitness lookup tables for determining the fitness contribution $f_i$ from bit $x_i$. Given an NK landscape with $N_{NK}$ =8 and $K_{NK}$ =2, $f_3(x_3, z_1^{(3)}, z_2^{(3)})$ is the fitness contribution for $x_3$. $z_1^{(3)}$ and $z_2^{(3)}$ are the two bits that epistatically interact with $x_3$. As shown in the figure, they have been selected as the bits to the left and right of $x_3$ (i.e. $z_1^{(3)} = x_2$ and $z_2^{(3)} = x_4$). The lookup table consists of 2^( $K_{NK}$ +1) entries, each associated with a unique combination of bit states for $x_3$, $z_1^{(3)}$ and $z_2^{(3)}$. Each entry in the lookup table is a number between [0,1] drawn from a uniform distribution.**

Each bit $x_i$ provides a fitness contribution $f_i : B^{K_{NK}+1} \longrightarrow \mathbb{R}$ whose value depends on the state of bit $x_i$ and the states of the $K_{NK}$ bits interacting with $x_i$. The $K_{NK}$ bits interacting with $x_i$ are labeled as $z_1^{(i)}, z_2^{(i)}, ..., z_{K_{NK}}^{(i)}$. NK Landscapes are stochastically generated with the fitness contribution $f_i$ of bit $x_i$ being a number drawn from a uniform distribution in the range [0,1]. To determine the fitness contribution $f_i$, a lookup table is used such as the one shown in Figure 5-7. The final fitness value $f(x)$ is an average of each of the fitness contributions as defined below. For a given instance of the NK landscape, the maximum fitness value is not known however fitness values are bounded between [0,1].

$$f(x) = \frac{1}{N_{NK}} \sum_{i=1}^{N_{NK}} f_i\left(x_i, z_1^{(i)}, z_2^{(i)}, ..., z_{K_{NK}}^{(i)}\right), \quad x_i \in (0,1), \quad i \in \{1,..,n\} \qquad \textbf{(5-10)}$$

In the original description [81], the $K_{NK}$ bits that epistatically interact with $x_i$ are those adjacent to $x_i$ in the bit string as seen in Figure 5-7. In this work, each $z_i$ is randomly selected to be any of the bits (other than $x_i$) and not just those adjacent or nearby. Notice that without epistatic interactions ($K_{NK}$ =0), the problem is completely decomposable and trivial to solve. However, as $K_{NK}$ increases, so too does the phenotypic interdependence of genes. Genetic encoding of the NK landscape is simple with each bit $x_i$ representing a binary gene and $N_{NK}$ being the size of the genome.

For most of these experiments $N_{NK}$ = 30, $K_{NK}$ = 14. These parameters have been selected based on a tradeoff between the problem size, degree of ruggedness, and memory costs of the model which are proportional to $N \times 2^{K_{NK}}$. More detailed descriptions of the NK landscape model and its properties can be found in [217], [218], [79].

#### 5.3.2.2 Core EA Design

A binary coded EA is used with population size $N$ varying over the range [50,400]. Only asexual reproduction is considered via parent duplication plus mutation with a bit flip mutation rate of $2/N_{NK}$ for $N_{NK}$ binary genes. Evolution occurs using a pseudo steady state updating strategy where the parent population of size $N$ is randomly uniformly sampled (with replacement) $N$ times to generate $N$ offspring. The parents + offspring then compete for survival to the next generation. A high level pseudocode for each of the EA designs is provided below.

```
Pseudocode for all three EA designs:
Initialize population
If SOTEA or cellular GA: Connect individuals with ring topology
Loop
    Loop N times
        Randomly select an individual i
        Generate offspring by mutation
        If SOTEA: apply SOTEA reproduction rule (Figure 5-6)
        If cellular GA: apply cellular GA reproduction rule (Figure 5-6)
    End loop
    Loop N times
        Randomly select an individual i
        If Panmitic GA: Select random neighbour
        If SOTEA or cellular GA: Select worst neighbour
        Eliminate worse of i or its chosen neighbour
        If SOTEA or cellular GA: assign links of loser to winner
    End loop
    Gen=Gen+1
Until maximum number of generations
```

A few comments should be made about the similarities and differences between the three EA designs. First, it should be noted that the cGA and SOTEA only differ in the reproduction rule used, which is described in Figure 5-6. In particular, SOTEA uses a reproduction rule that is loosely based on genome duplication and divergence while the cGA uses a rule that ensures the ring topology is maintained.

The Panmictic GA also clearly differs from the distributed EA designs in that its population is defined on a static fully connected network. However there is another difference in the Panmictic GA that needs to be explained and justified. As demonstrated in the pseudocode above, the Panmictic GA uses a selection method that is similar to binary tournament selection (without replacement). This selection method was used in order to provide the Panmictic GA with a better chance of maintaining genetic diversity. If the Panmictic GA incorporated the same selection procedure as the distributed EA designs, this would be equivalent to truncation selection. Truncation selection was not used with the Panmictic GA because preliminary results (not shown) have indicated that this selection method causes poor performance and low genetic diversity when a Panmictic GA is run on the NK fitness landscape.

### 5.3.3 Results

The experimental results start off by assessing the topological characteristics of the EA populations to determine if any of the EA designs are able to acquire the topological characteristics of complex biological systems. Experiments are also conduced to see if any other behavioral qualities of complex biological systems are acquired. One important quality that would be of great value in an EA design is the capacity to sustain diversity within a competitive environment. This section investigates whether any of the EA designs can sustain high levels of genetic diversity and also investigates whether this provides a benefit to algorithm performance.

#### 5.3.3.1 Topological Characteristics of Interaction Networks

This section looks at the structural characteristics of the interaction networks for each of the EA designs and compares them to what is observed in complex systems. For each of the structural characteristics presented in Table 5-1, only the SOTEA network was found to have characteristics similar to that seen in complex systems.

The last column in Table 5-1 highlights the fact that every individual has the same neighborhood size $k$ in the Panmictic GA and the cellular GA, however $k$ takes on a distribution of values for SOTEA. The distribution for $k$ is fat tailed (closely fitting an exponential function), meaning that there is large heterogeneity in the neighborhood size. Keeping in mind that only neighbors can compete in a structured EA, the neighborhood size $k$ impacts the selection pressure within the population. Since there is large heterogeneity in neighborhood sizes for SOTEA, it is reasonable to suspect that there will also be significant heterogeneity in selection pressure.

**Table 5-1: Topological Characteristics for the interaction networks of the Panmictic GA, cellular GA, and SOTEA. For comparison, common topological characteristics of complex networks are also provided (taken from [198], and references therein). *L* is the characteristic path length, *k* is the node degree, $k_{ave}$ is the average node degree, *N* is the population size, and *R* is a correlation coefficient for the stated proportionalities.**

| System | L | $k_{ave}$ | k distribution |
|---|---|---|---|
| Panmictic GA | L = 1 | $k_{ave}$ = N-1 | k = N-1 |
| cellular GA | L ~ N | $k_{ave}$ = 2 | k = 2 |
| SOTEA | L ~ log N ($R^2$=0.969) | $k_{ave}$ ~ log log N ($R^2$=0.989) | Exponential ($R^2$=0.991) |
| Complex Networks | L ~ log N | $k_{ave}$ << N | Fat Tail (e.g. Power Law, Exponential) |

### 5.3.3.2 Genetic Diversity

This section looks at the genetic diversity that is maintained in each of the EA designs. Measuring genetic diversity of the population is done in a straightforward manner. Genetic Diversity is calculated as the average Hamming Distance between population members divided by the average Hamming Distance between random points in solution space. For a single binary gene, two randomly selected gene values have a 50% chance of being different making the Hamming Distance between random individuals of $N_{NK}$ genes equal to $N_{NK}$ /2. The Hamming Distance is defined in (5-11) as a summation of 1 minus the Kronecker Delta function $\delta(X_{i,h}, X_{j,h})$. The Kronecker Delta function has a value of 1 if $X_{i,h} = X_{j,h}$ and 0 otherwise. $X_{i,h}$ and $X_{j,h}$ represent the $h^{th}$ gene for individuals *i* and *j* (resp.).

$$Ham(i,j) = \sum_{h=1}^{N_{NK}} \left(1 - \delta(X_{i,h}, X_{j,h})\right) \tag{5-11}$$

$$Div = \frac{\sum_{i=1}^{N} \sum_{j=1, j \neq i}^{N} Ham(i,j)}{N(N-1)\left(N_{NK}/2\right)} \tag{5-12}$$

Diversity results are shown in Figure 5-8 with each of the EA designs using epistatic fitness. Results are given for genetic diversity of the entire population as well as diversity for the 20% best individuals in the population. It is useful to measure diversity for the top 20% because it is often very difficult to maintain diversity among the best individuals in a population. As expected, the results demonstrate that the Panmictic GA is not able to sustain genetic diversity, particularly in the top 20% of the population. The cellular GA has much higher levels of diversity although this is significantly reduced in the top 20%.

SOTEA exhibits sizeable improvements in diversity compared to the other EA designs, particularly for 20% best individuals in the population.

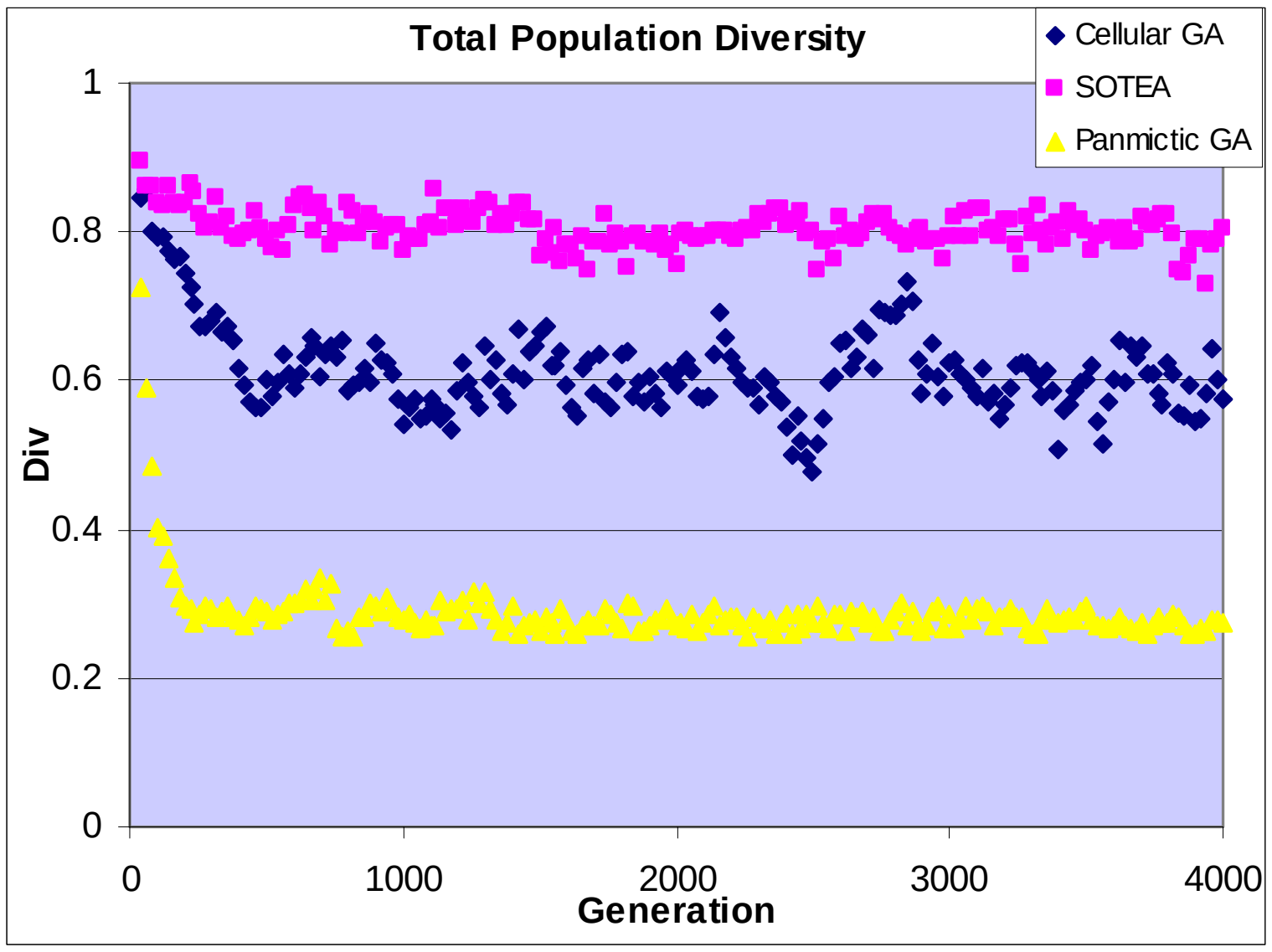

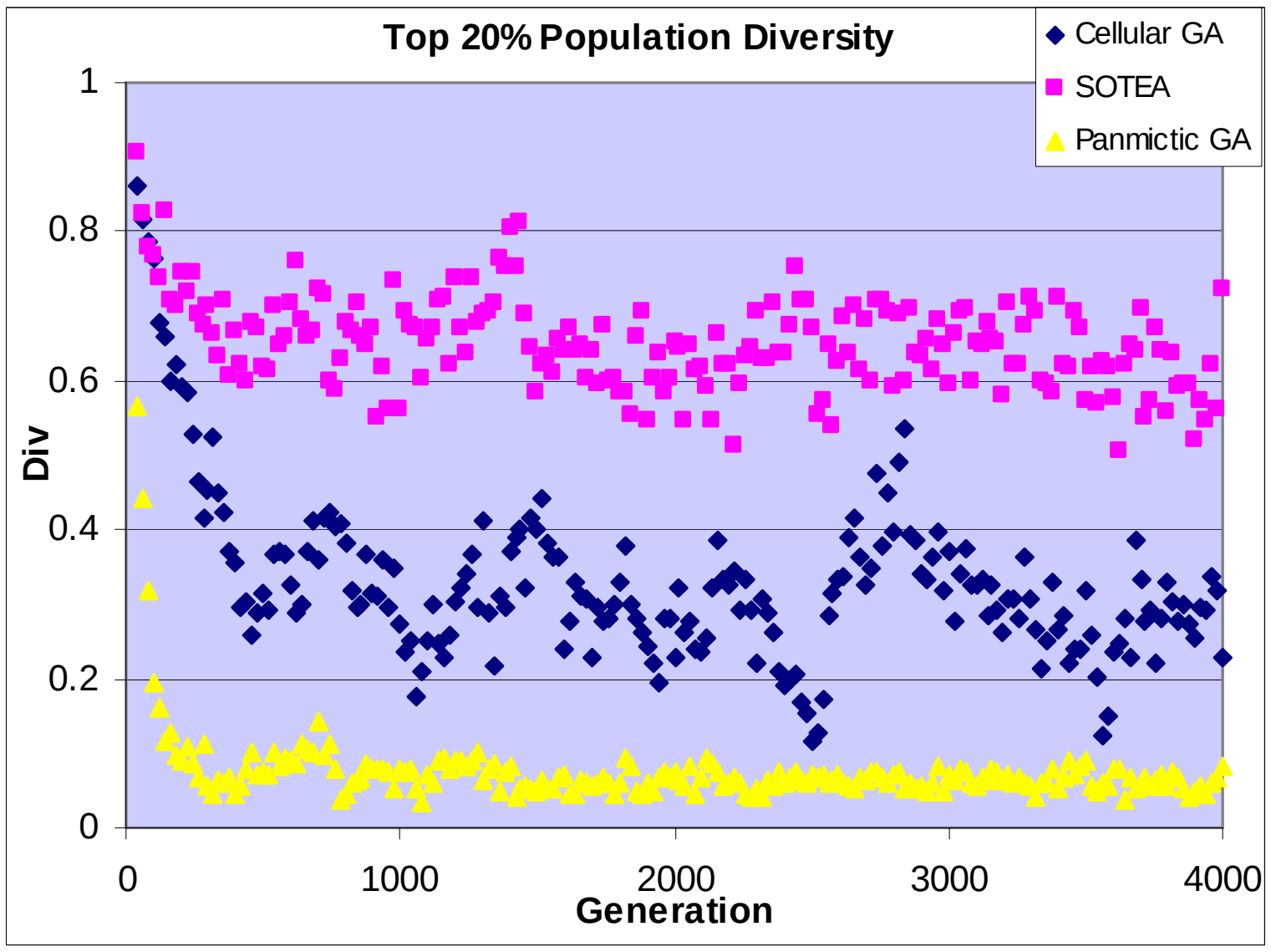

**Figure 5-8: Genetic Diversity Results are shown over 4000 generations for Panmictic GA, SOTEA, and cellular GA. Diversity for each EA is an average over 10 runs with diversity calculated from (5-12) using the entire population (top graph) or the 20% best individuals in the population (bottom graph). Experiments are conducted on NK models with $N_{NK}$ =30, $K_{NK}$ =14. For each EA design the population size is set to $N$=100 and epistatic fitness is used as defined by (5-9).**

### 5.3.3.3 Performance Results

Performance results are shown in Figure 5-9 with each of the EA designs using epistatic fitness. These results demonstrate that the Panmictic GA is not able to continually locate improved solutions while SOTEA and the cellular GA both are able to make steady progress throughout the 5000 generations considered. However, SOTEA was found to have better performance than the cellular GA during the later stages of evolution.

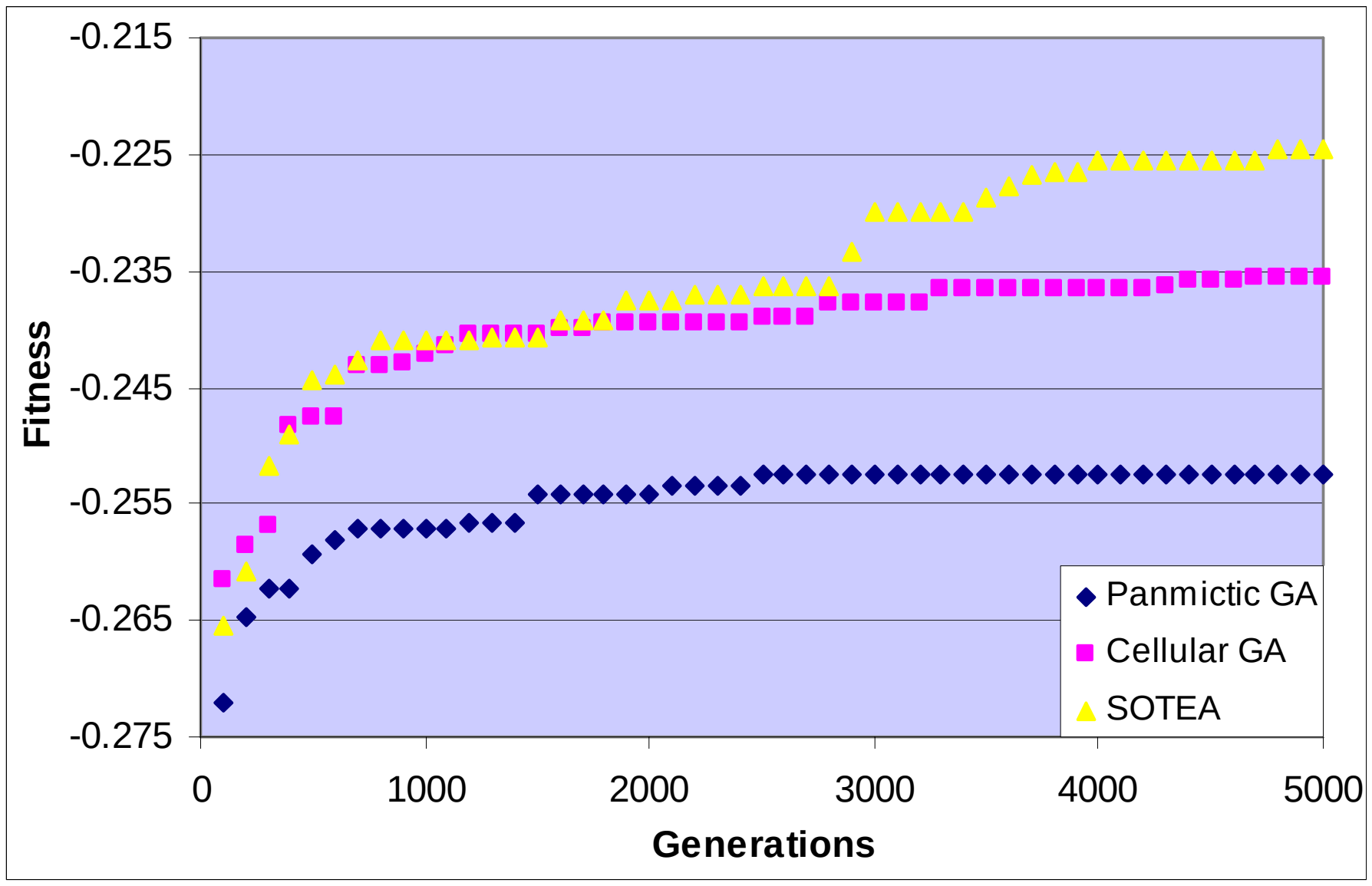

**Figure 5-9: Performance results are shown over 5000 generations for Panmictic GA, SOTEA, and cellular GA each operating with Epistatic Fitness. Performance for each EA is an average over 10 runs with performance calculated as the best objective function value in a run. Experiments are conducted on NK models with $N_{NK}$ =30, $K_{NK}$ =14. For each EA design the population size is set to $N$=100 and epistatic fitness is used as defined by (5-9).**

#### 5.3.3.4 Impact of Ruggedness

This section considers the impact that landscape ruggedness has on genetic diversity of the population for each of the EA designs. Landscape ruggedness is varied by changing the $K_{NK}$ parameter of the NK model as shown in Figure 5-10. These results clearly show that as the NK landscape becomes completely smooth (i.e. $K_{NK} \rightarrow 0$), each of the EA designs loses the capacity to sustain genetic diversity. However as ruggedness increases, each EA design approaches its own asymptotic limit indicating its maximum capacity for genetic diversity. Notice that the asymptote for SOTEA was not observed over the range of $K_{NK}$ values tested. Larger values of $K_{NK}$ were not considered due to computational costs.

Knowing that a diversity measure equal to 1 approximates a uniform distribution in genotype space, the fact that SOTEA has diversity close to 0.8 among its top 20% individuals indicates that SOTEA is able to distribute the search process across many promising regions of genotype space.

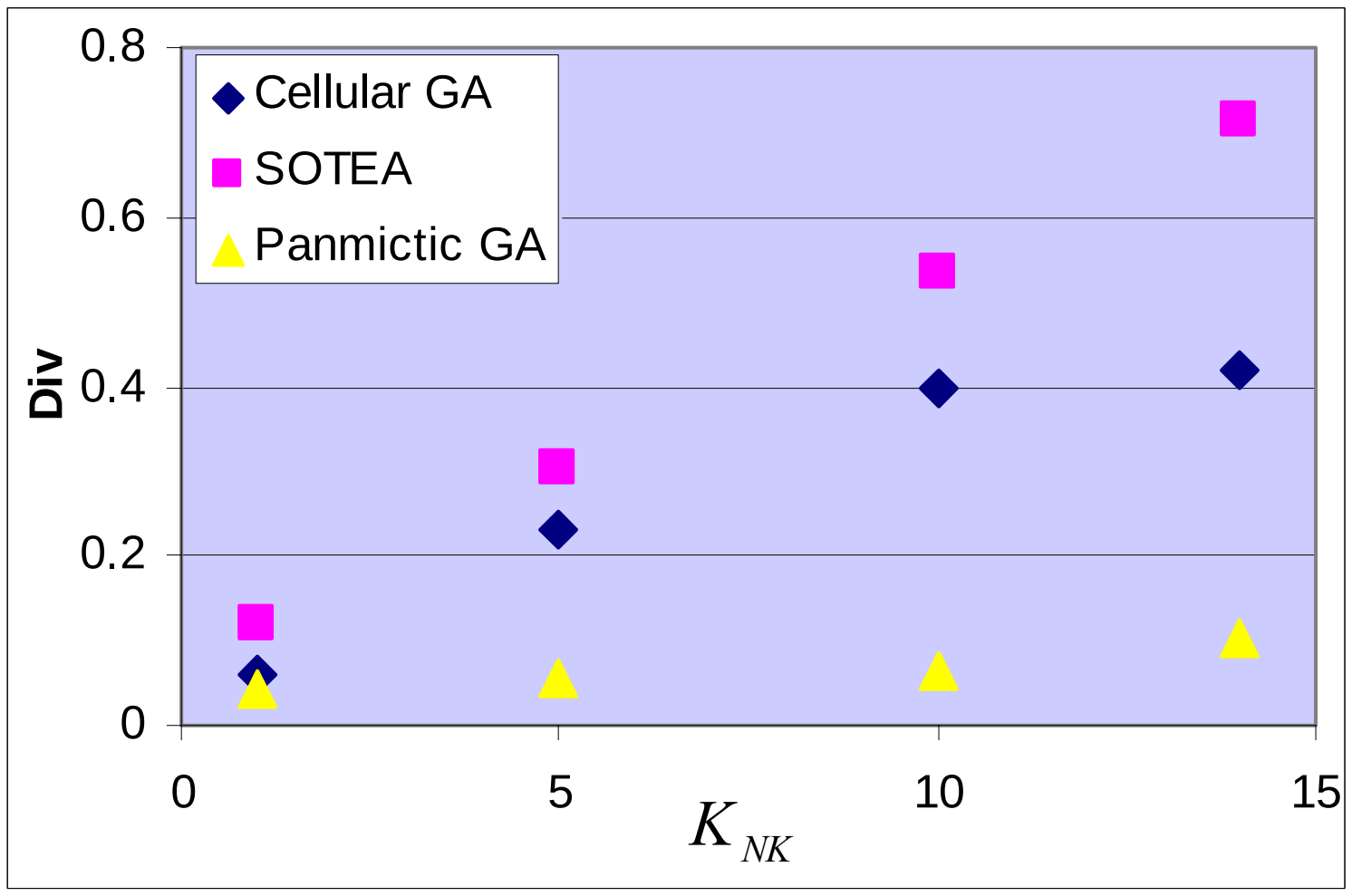

**Figure 5-10: Genetic diversity results are shown for different amounts of landscape ruggedness for the Panmictic GA, SOTEA, and the cellular GA. Diversity is an average of calculations using (5-12) that are taken at every 20 generations (up to 1000 generations) from the 20% best individuals in the population. This measure then also averaged over 5 runs. Experiments are conducted on NK models with $N_{NK}$ =30, and $K_{NK}$ varying as shown in graph. Increasing $K_{NK}$ indicates increasing levels of landscape ruggedness. For each EA design, the population size is set to $N$=100 and epistatic fitness is used as defined by (5-9).**

#### 5.3.3.5 Impact of Epistasis

All results presented thus far have considered EA designs with individual fitness defined by (5-9) (i.e. epistatic fitness). Figure 5-11 extends the analysis of population diversity for cases where the individual fitness is defined in the standard way (as the raw objective function value). Compared to the results with epistatic fitness (see Figure 5-8), both SOTEA and cellular GA have significantly less diversity and are hard to distinguish from the diversity present in the Panmictic GA. This result provides evidence that epistasis can play an important role in sustaining diversity in structured populations including in the cellular GA.

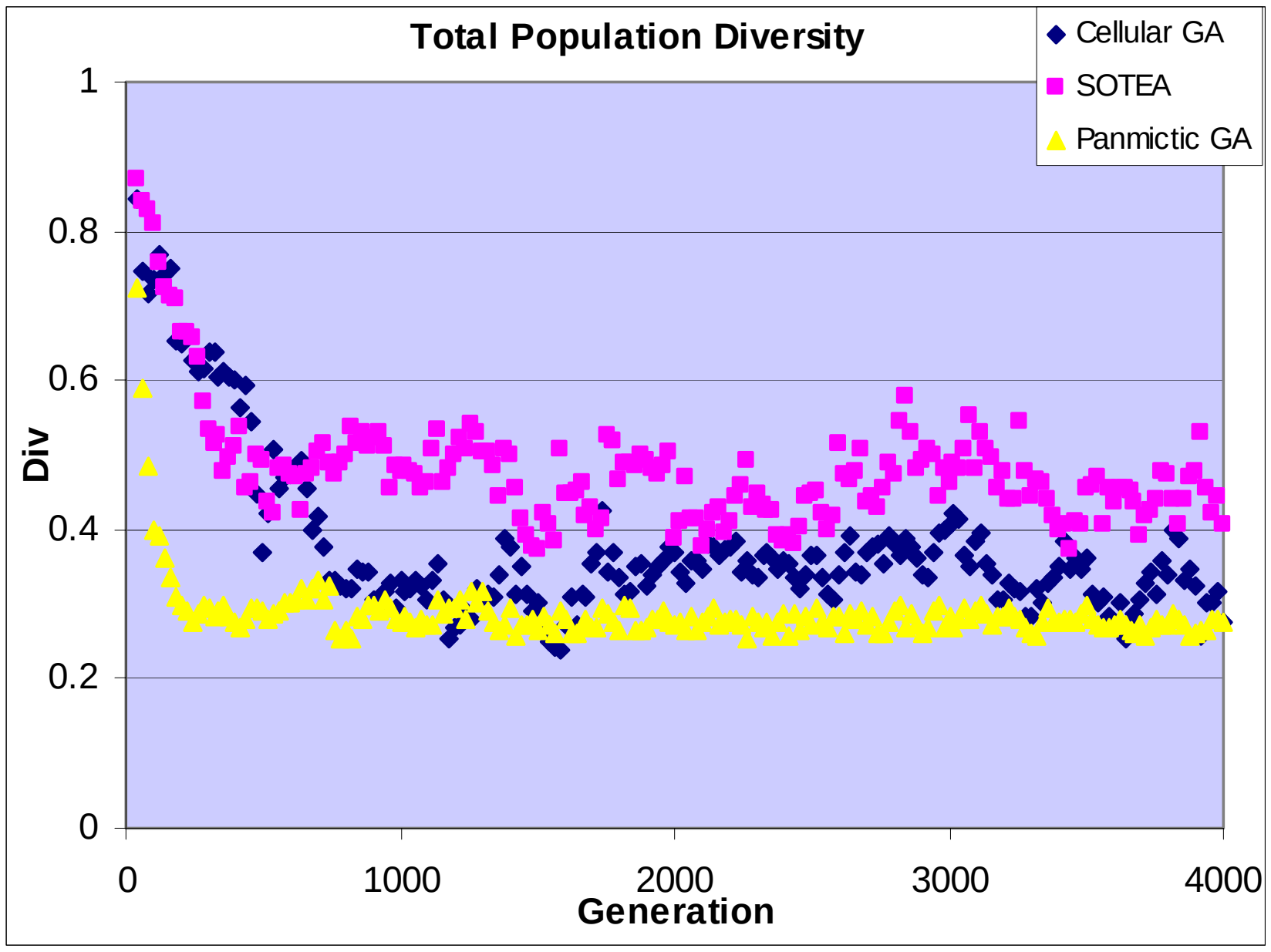

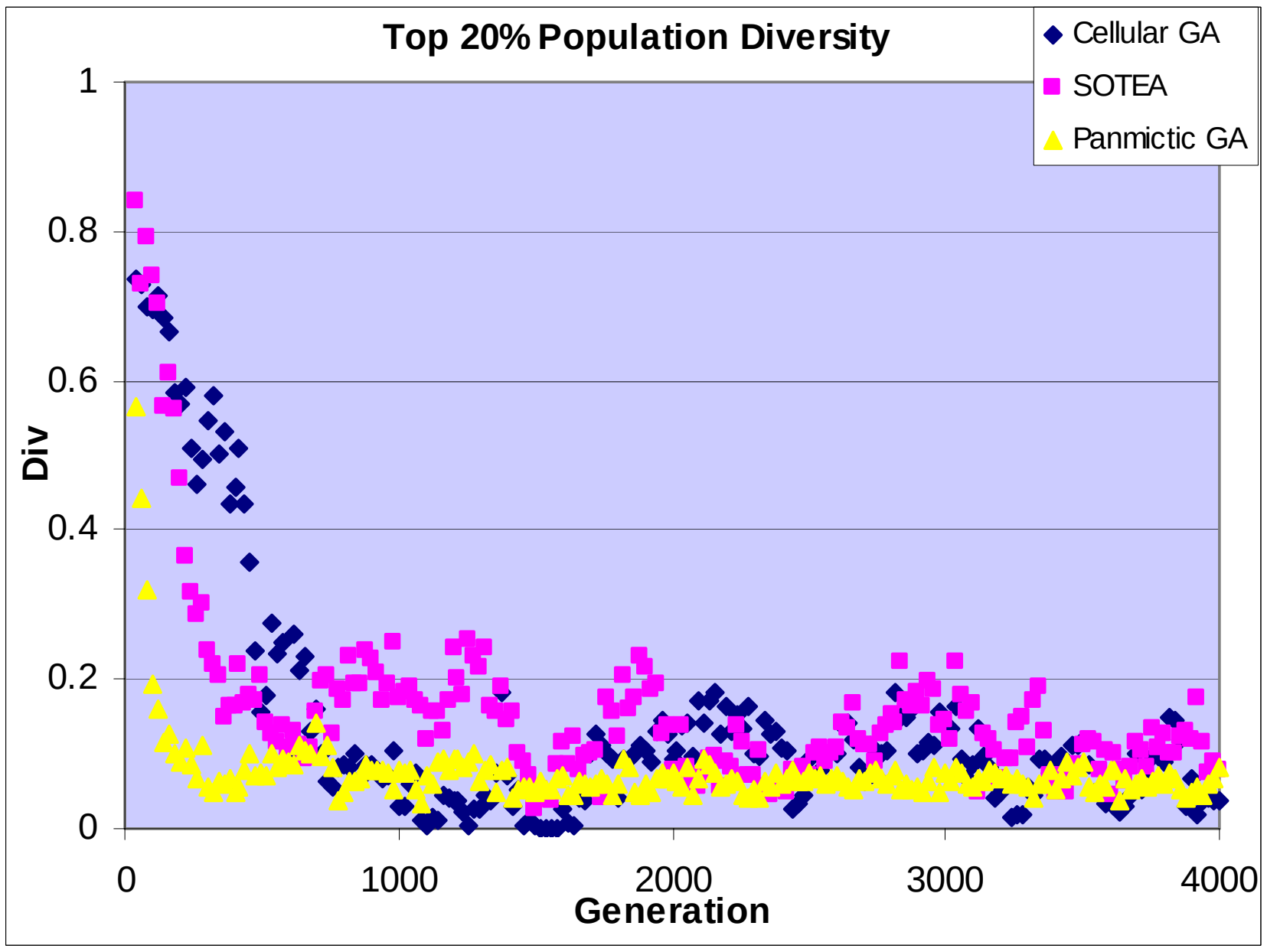

**Figure 5-11: Genetic diversity results are shown over 4000 generations for Panmictic GA, SOTEA, and cellular GA each operating without epistatic fitness. Diversity for each EA is an average over 10 runs with diversity calculated from (5-12) using the entire population (top graph) or the 20% best individuals in the population (bottom graph). Experiments are conducted on NK models with $N_{NK}$ =30, $K_{NK}$ =14. For each EA design, the population size is set to $N$=100 and fitness is defined as the Objective Function Value. The results shown here for the Panmictic GA are identical to results shown in Figure 5-8. This is because the fitness rankings of individuals in a fully connected population are the same regardless of whether epistatic fitness (5-9) is used or the Objective Function Value is used. Because the fitness rankings are the same, the outcome of competitions will also be the same (hence no change to EA behavior).**

#### *5.3.3.5.1 Selection Pressure Patterns*

A better understanding of the impact of epistatic fitness on SOTEA is possible by observing its influence on the selection pressure within the SOTEA network. The networks in Figure 5-12 are examples of SOTEA networks grown with and without epistatic fitness.

To represent selection pressure in the system, each node is selected in a mock competition trial and arrows are drawn to its worst neighbor. Arrows are drawn in this way because, in SOTEA and the cellular GA, competition occurs by first selecting an individual and then having it compete against its worst neighbor. Arrows in black represent selection pressure directed away from the network center, while arrows in green indicate selection pressure that is not directed away from the center.

For networks evolved with epistatic fitness, selection pressure points away from the network center but without epistatic fitness, selection pressure points both toward and away from the network center. It was also found that older and better fitness nodes tend to be located more towards the center of the network. Additional experiments are needed in order to better understand this behavior of SOTEA, however it is believed that the selection pressure patterns shown here ultimately play an important role in explaining why genetic diversity is maintained at such high levels in SOTEA.

It should also be mentioned that the two networks shown in Figure 5-12 are taken after 100 generations of SOTEA evolution. Typically the amount of time required for the self-organization of network structure to take place was less than 100 generations however no attempt was made at determining the exact time when this transient was complete. Beyond 100 generations, it was found that topological characteristics of the SOTEA network as well as network visualizations were very consistent.

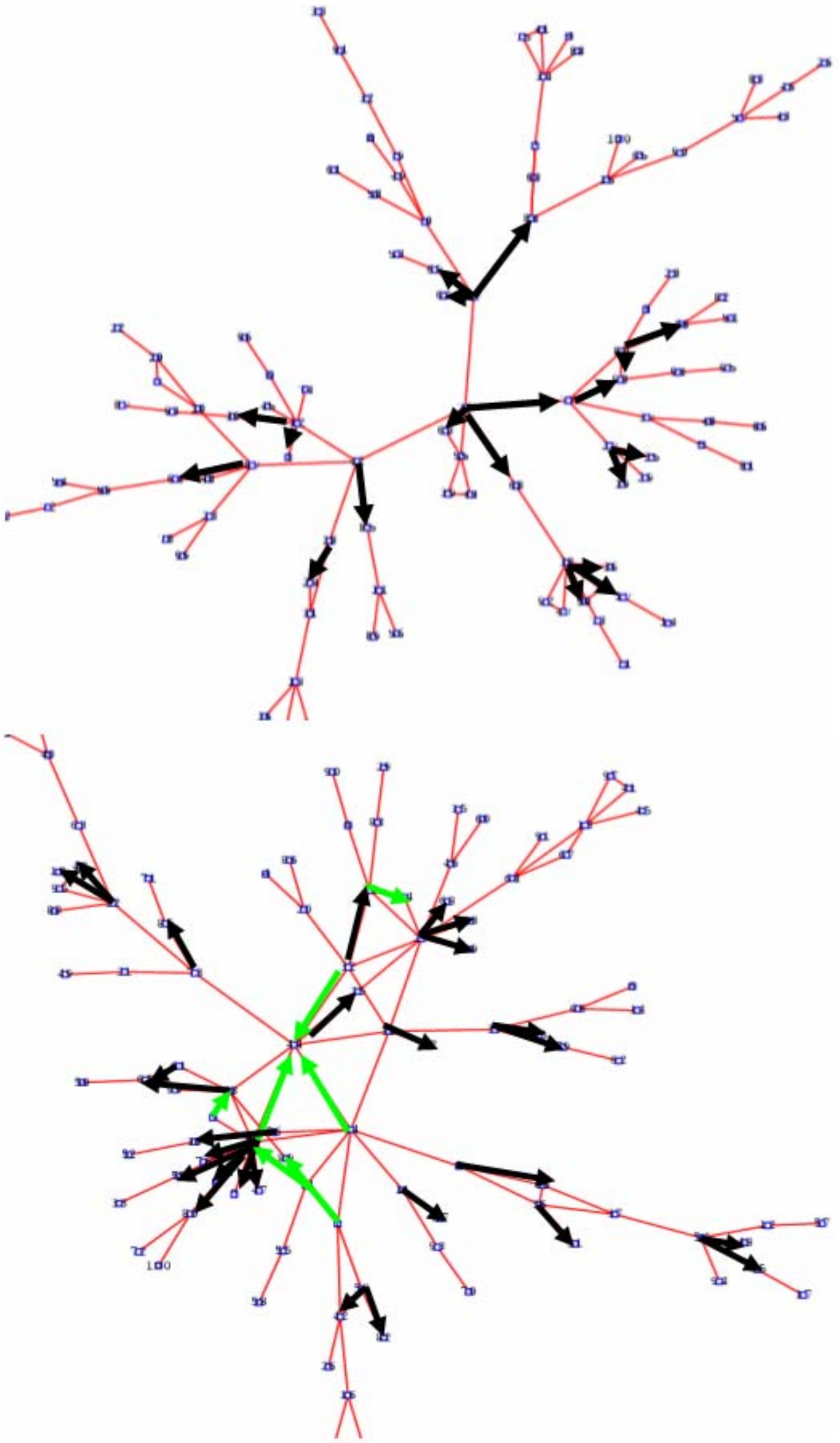

**Figure 5-12: Selective pressure patterns in the SOTEA network with (top) and without (bottom) epistasis. Selective pressure in the network is shown with arrows in black for pressure directed away from the network center and green for other directions of pressure. Selective pressure directions have only been calculated for nodes located near the network center. The arrows are drawn by selecting a node and drawing an arrow from this node to its worst neighbor. The worst fit neighbor is determined by epistatic fitness (5-9) for the top graph and by the Objective Function Value for the bottom graph.**

#### 5.3.3.6 The Impact of SOTEA model parameters

The SOTEA model includes parameters $P_{add}$ and $P_{remove}$ for controlling how much the connections of an offspring are different from the connections of its parent. These parameters are conceptually similar to a mutation rate for network topology and they will control the amount of structural innovation that is introduced to the network. This in turn is expected to impact the range of topologies that are possible during evolution. To get a sense of the impact of these parameters, Figure 5-13 presents networks that were evolved with different settings for $P_{add}$ and $P_{remove}$. As seen in Figure 5-13a, when no innovation is possible ($P_{add}$ = $P_{remove}$ = 0), the network consistently takes on a structure that resembles a simple branching process. When both parameters are increased to 0.1, as seen in Figure 5-13b, some clustering begins to emerge. However, when the parameters are increased a little more to 0.2 (see Figure 5-13b), the structure changes dramatically and is dominated by a single highly connected cluster.

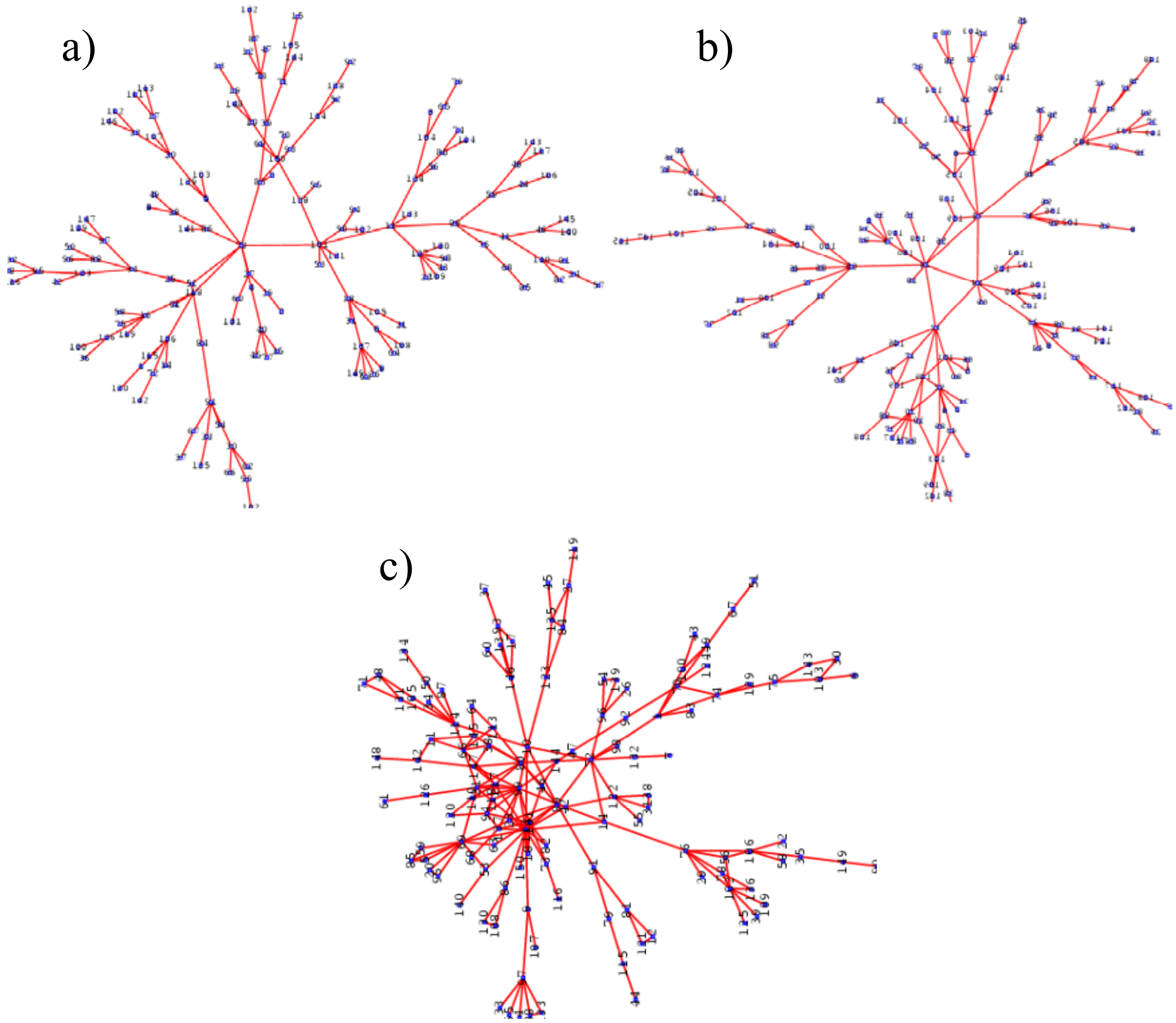

**Figure 5-13 SOTEA networks evolved using different parameter settings for the reproduction rule. In the reproduction rule, a parent's connections are inherited by its offspring with probability $P_{add}$ followed by each of the inherited connections being lost by the parent with probability $P_{remove}$. Population interaction networks were evolved for a) $P_{add}$ = $P_{remove}$ = 0.0%, b) $P_{add}$ = $P_{remove}$ = 10%, c) $P_{add}$ = $P_{remove}$ = 20%.**

### 5.3.4 Discussion

#### 5.3.4.1 SOTEA Guidelines

It would be useful to be able to generalize the results shown in this work and develop a framework under which network dynamics would be beneficial to an EA population. Along these lines, several aspects of SOTEA have been highlighted as important features of the design.

First, the most important precondition for network self-organization was to have changes in network structure be driven by the same forces that drive population dynamics; namely a fitness-based selection pressure.

The SOTEA model also used a dual coupling between network states and structure which substantially improved the behavior of the system. In particular, network dynamics depended upon node states (due to the SOTEA competition rule) and node states depended upon network structure (due to the use of epistatic fitness). It is worth mentioning that the coevolution of structure and states is a topic of great significance to the study of complex networks. To this author's knowledge, a dual coupling between states and structure is not present in any other model of complex systems available in the literature.

It is also interesting to note that selecting the worst neighbor in the competition rule is also similar to the extremal dynamics used in most models of self-organized critical systems as reviewed in [177]. By eliminating the worst individual in a neighborhood, SOTEA may actually be using an important driving force for some self-organizing processes in nature. Additional experimentation is needed to substantiate these claims.

#### 5.3.4.2 The NK Model as an optimization research tool

There are valid concerns about the extent to which the NK model, as currently defined, represents real optimization problems of interest. One concern is that the topological properties of the NK landscape's gene interaction network are similar to a random graph and do not correspond with the topological properties of many real world systems.

As reviewed in this chapter, many complex systems have a number of similarities in their topological properties and these properties are also clearly non-random. Given an appropriate problem representation, it is speculated that this (approximate) universality could extend to a large and interesting class of optimization problems based on the fact that many problems of interest are large complex systems of interacting components. To test this simple idea, it would be necessary to develop course-grained network approximations of fitness landscapes for a number of real-world optimization problems. Real world problems could then be probed using tools developed in statistical mechanics in order to determine whether any common structural properties exist.

Under the reasonable assumption that some topological properties are repeated in many real world problems, it should be possible to develop network evolution models which can evolve similar structures. Given the success of recent models in mimicking topological properties of complex systems, this task should not be too difficult. The result of these efforts would be a fitness landscape generator, similar in principle to the NK landscape, but one that has the capacity to generate problem instances with properties that are similar to real world problems. It is also worth mentioning that this suggestion has some similarities to the proposal by Kauffman for probing gene regulatory networks [219].

### 5.3.5 Conclusions

This work was intended as an initial investigation into the self-organization of interaction networks for an Evolutionary Algorithm. Motivating this research was a desire to acquire structural characteristics of complex biological systems which are believed to be relevant to their behavior. In addition, this work also aimed to create an artificial system with a capacity for sustainable coexistence of distinct components within a competitive environment (i.e. sustainable diversity).

Population diversity was not imposed upon the EA as is traditionally done but instead emerges in the system as a natural consequence of population dynamics. The environmental conditions which enable sustainable diversity are similar to what is observed in complex biological systems. These conditions involved a self-organizing interaction network and a contextual definition of individual fitness which was referred to as epistatic fitness. In addition, high levels of diversity also required evolution to take place within a rugged fitness landscape.

## 5.4 SOTEA Model II

The previous SOTEA model demonstrated that fitness was a natural property for tying network structural dynamics to the evolutionary dynamics of an EA population. Different forms of fitness measurements were considered including a contextual form (called epistatic fitness) where an individual's fitness is defined by its own local environment.

The next SOTEA model focuses on the emergence of additional structural properties that were not present in the first SOTEA model. One of these properties is modularity which is a structural feature that is heavily exploited in natural evolution. To encourage modularity, the second model uses both fitness and measures of community cohesion to drive network dynamics. The new SOTEA is also designed for use with multi-parent search operators which are standard in most EA designs and were missing in the first SOTEA model.

Section 5.4.1 describes the new SOTEA model including new driving forces for network dynamics and new rules for implementing changes to network structure. Section 5.4.2 presents the experimental setup including the remaining aspects of the core EA design. Results are presented in Section 5.4.3 and include both an analysis of performance and a comparison of topological properties between SOTEA and complex biological systems. The performance results provide evidence that the new SOTEA exhibits robust search behavior with strong performance on many problems. Discussion and conclusion sections finish the chapter in Sections 5.4.4 and 5.4.6.

### 5.4.1 Model Description

This section describes the new SOTEA network model used to couple the network topology to the population dynamics of an Evolutionary Algorithm. With the new model, network dynamics are driven by a measure of node fitness and by a measure of node modularity as described in Section 5.4.1.1. These dynamics are implemented by rewiring localized regions of the network as described in Section 5.4.1.2.

As previously mentioned, the network is described by an adjacency matrix $J$ such that individuals $i$ and $j$ are connected (not connected) when $J_{ij}$=1 ($J_{ij}$=0). This work only deals with undirected networks such that $J_{ij} = J_{ji}$. The terms *individual* and *node* are used interchangeably to refer to individual members of the EA population situated within the population network. Also, the terms *links* and *connections* are used interchangeably to refer to directly connected nodes (i.e. individuals that are neighbors in the population).

#### 5.4.1.1 Driving Forces

Two properties are used to drive network dynamics in this SOTEA model which are described in this section.

##### *5.4.1.1.1 k Adaptation*

For most real networks, the degree $k$ is not constant (unlike what is seen in lattices) but instead takes on a distribution of values often fitting exponential or power law distributions. How nodes come to obtain $k$ values that are higher than others depends on the system under study and could be historically motivated or could be motivated by some form of intrinsic node fitness. The former has been theorized to take place in the process of genome complexification which has been modeled primarily by the previously mentioned DD model. Other popular models such as the BA model and its associated mechanism of "preferential attachment" also appear to derive $k$ distributions using an historical bias. An alternative is the "good get richer" concept [204] introduced in the fitness model in Section 5.1.3.3, where individuals of high fitness are driven to obtain higher $k$.

In this model, an adaptive set point $K_{Set}$ is used to define a node's desired number of links such that high fitness individuals in the population are encouraged to acquire a larger number of connections as defined in (5-13). Although conceptually similar to the work in [204], the node's fitness will have an ability to evolve due to the dynamics of the EA population which is something previously unexplored in network evolution models. The $K_{Set}$ parameter is defined in (5-14) as a quadratic function of rank which has a lower bound of $K_{Min}$ = 3 and an upper bound $K_{Max}$ which must be set by the user. The Rank term in (5-14) refers to an individual's fitness based ranking with Rank = 1 being the best individual and Rank = $N$ the worst individual in the population.

$$Min \quad \left| k_i - K_{Set,i} \right| \tag{5-13}$$

$$K_{Set,i} = K_{Min} + \left( (K_{Max} - K_{Min}) \left( \frac{N - Rank}{N} \right)^2 \right) \tag{5-14}$$

#### 5.4.1.1.2 Weighted Clustering Coefficient

As mentioned in the review of network properties in Section 5.1.2, many complex biological systems have a high level of modularity (as measured by the clustering coefficient) and a clear hierarchical structure. Networks with hierarchical structure are expected to have a clustering coefficient that is inversely related to a node's degree (i.e. *c*-*k* correlations). The second property used to drive network dynamics attempts to explicitly address both observations.

Based on the previously stated driving force for *k* adaptation, it is expected that nodes in SOTEA networks with higher fitness will also have higher *k*. To encourage high levels of modularity as well as the inverse relationship between *c* and *k* (needed for hierarchy), network rewiring is driven to maximize a weighted version of the clustering coefficient as defined in (5-15). In this new version of the clustering coefficient, a connection's contribution to *c* is weighted to give less importance to connections involving nodes of higher fitness. The weight *W* for each connection is defined in (5-17) which alters the $e_i$ term of the clustering coefficient equation, as seen in (5-16). Notice the similarity between *W* and the intrinsic fitness measure which is defined in (5-8) and was first presented in [204].

$$Max \;\; c_i^* = \frac{2e_i^*}{k_i(k_i - 1)} \tag{5-15}$$

$$e_i^* = \sum_{j=1}^{N} J_{ij} \sum_{k=1}^{N} J_{ik} J_{jk} W_{jk}, \; i \neq j \neq k \tag{5-16}$$

$$W_{jk} = \frac{Rank_j \times Rank_k}{N^2} \tag{5-17}$$

#### 5.4.1.2 Mechanics of Network Rewiring

The previous section described two properties which will act as driving forces for network dynamics. In particular, each node will be driven to obtain a specific value of $k$ based on its fitness and defined in (5-14) and each node will be driven to maximize the value of a weighted clustering coefficient given in (5-15). To obtain these goals, changes to network structure must take place involving the addition, removal, and local transfer of links in a network. Although conceptually simple, these rules must satisfy a number of constraints in addition to and sometimes superseding the driving forces previously stated. These rules are described next and are also demonstrated in Figure 5-14.

The "Add Link Rule" and the "Remove Link Rule" are two rewiring rules that have been created in order to allow the $k$ value for each node to reach $K_{Set}$.

**Add Link Rule:** Starting with a selected node *N1*, a two step random walk is taken, moving from node *N1* to node *N2* to node *N3*. If *N1* wants to increase its number of links ($k_{N1} < K_{Set}$) and *N3* wants to increase its number of links ($k_{N3} < K_{Set}$) then add a link between *N1* and *N3*.

**Remove Link Rule:** For a selected node *N1* with $k_{N1} > K_{Set}$, a two step random walk is taken, moving from node *N1* to node *N2* to node *N3*. If *N3* is already connected to *N1* ($J_{N1,N3} = 1$) and $k_{N3} > K_{Set}$ then remove the connection between *N1* and *N3*. Notice the presence of *N2* with $J_{N2,N1} = J_{N2,N3} = 1$ ensures that connections removed using this rule do not result in network fragmentation.

The "Transfer Link Rule" allows for the improvement of clustering locally within the network. However, this rule is not allowed to result in net violations to $k$ adaptation.

**Transfer Link Rule:** For a selected node *N1* a two step random walk is taken, moving from node *N1* to node *N2* to node *N3*. If $k_{N3} < K_{Set}$, then the connection between *N1* and *N2* is transferred to now be between *N1* and *N3* (i.e. $J_{N1,N2} = 1$, $J_{N1,N3} = 0$ changes to $J_{N1,N2} = 0$, $J_{N1,N3} = 1$). To determine if the transfer will be kept, the local modularity is calculated using (5-15) for *N1*, *N2* and *N3* both BEFORE and AFTER the connection transfer. If $\left(c^*_{N1} + c^*_{N2} + c^*_{N3}\right)$ increases after the connection transfer then the transfer is kept, otherwise it is reversed. In this way connections are only added which strengthen the weighted clustering metric and don't cause a net increase in $K_{Set}$ violations.

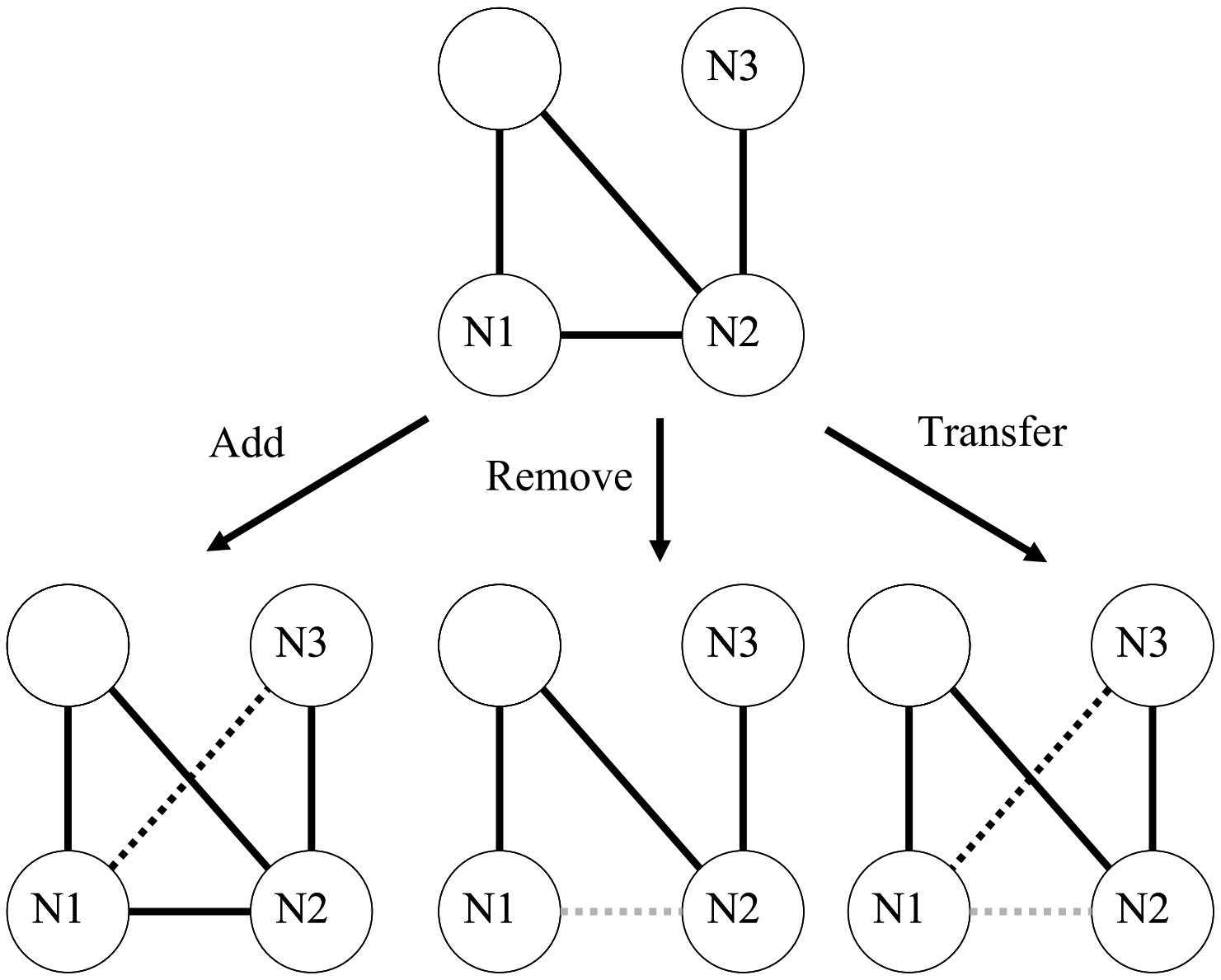

**Figure 5-14 Adaptive Network Rules: A selected node *N1* will attempt to add, remove or transfer its connections based on the satisfaction of constraints and the improvement of properties. Add Rule: The dotted line represents a feasible new connection in the network assuming nodes *N1* and *N3* both would like to increase their number of connections. Remove Rule: The gray dotted line represents a feasible connection to remove in the network assuming nodes *N1* and *N2* both have an excess of connections. Transfer Rule: The connection between *N1* and *N2* (gray dotted line) being transferred to now connect *N1* and *N3* (black dotted line) represents a feasible transfer assuming this action results in an overall improvement to local clustering.**

Since there are several constraints that the random walks (in the rewiring rules) must satisfy, up to a maximum of 10 random walks are conducted starting from *N1* for each instance of rule execution in an attempt to satisfy the conditions. An upper bound on the number of walks is needed because there is no guarantee that a random walk exists which satisfies all conditions.

## 5.4.2 Experimental Setup

### 5.4.2.1 Algorithm Designs

**SOTEA:** A high level pseudocode for SOTEA is provided below. The algorithm starts by defining the initial population *P* on a ring topology with each node connected to exactly two others (e.g. Figure 5-1c). For a given generation *t*, each node is subjected to both network rewiring rules and standard genetic operators. For a given node *N1*, each of the three network rewiring rules are executed (defined in Section 5.4.1.2). Afterwards, *N1* is selected as a parent and a second parent *N2* is selected by conducting a two step random walk across the network. An offspring is then created using the parents and a single search operator selected at random from the list in Table 5-2.[18] The better fit between the offspring and *N1* is stored in a temporary list Temp(*N1*) while the network rewiring rules and genetic operators are repeated on each of the remaining nodes in the population. To begin the next generation, the population is updated with the temporary list. This process repeats until some stopping criteria is met. In this case, the stopping criteria is a maximum of 3000 generations.

**Pseudocode for SOTEA**

```
t=0
Initialize P(t) (at random)
Initialize population topology (ring structure)
Evaluate P(t)
Do
    For each N1 in P(t)
        Add Link Rule(N1)
        Remove Link Rule(N1)
        Transfer Link Rule(N1)
        Select N1 as a first parent
        Select parent N2 by conducting a two step random walk from N1
        Select Search Operator (at random)
        Create and Evaluate offspring
        Temp(N1) = Best_of(offspring, N1)
    Next N1
    t=t+1
    P(t) = Temp()
Loop until stopping criteria
```

**cellular GA:** A cellular GA is also tested in these experiments which is identical to SOTEA except for two design changes. First, the cGA does not use any of the network rewiring rules (Add, Remove, Transfer) that are used in SOTEA. This means the cGA has a static ring topology. Also, when creating an offspring, the second parent *N2* is selected among all neighbors within a radius *R* of *N1* using linear ranking selection. A high level pseudocode is for the cGA is provided below.

[18] No parameter tuning was attempted and all search operators are used with equal probability.

**Pseudocode for cGA**

```
t=0
Initialize P(t) (at random)
Initialize population topology (ring structure)
Evaluate P(t)
Do
    For each N1 in P(t)
        Select N1 as first parent
        Select N2 from Neighborhood(N1,R)
        Select Search Operator (at random)
        Create and evaluate offspring
        Temp(N1) = Best_of(offspring, N1)
    Next N1
    t=t+1
    P(t) = Temp()
Loop until stopping criteria
```

**Panmictic EA:** SOTEA is also compared against a number of Panmictic EA designs. The core of the Panmictic EA is given by the pseudocode below. For this pseudocode, the parent population of size $\mu$ at generation $t$ is defined by $P(t)$. For each new generation, an offspring population $P`(t)$ of size $\lambda$ is created through variation (search) operators and is evaluated to determine fitness values for each offspring. The parent population for the next generation is then selected from $P`(t)$ and $Q$, where $Q$ is subset of $P(t)$. $Q$ is derived from $P(t)$ by selecting those in the parent population with an age less than $\kappa$.

**Pseudocode for Panmictic EA**

```
t=0
Initialize P(t)
Evaluate P(t)
Do
    P`(t) = Variation(P(t))
    Evaluate (P`(t))
    P(t+1) = Select(P`(t) ∪ Q)
    t=t+1
Loop until stopping criteria
```

Eight EA designs are tested which vary by the use of generational (with elitism) vs. pseudo steady state population updating, the use of binary tournament selection vs. truncation selection, and by the number of search operators. Details are given below for each of the design conditions.

**Population updating:** The generational EA design (with elitism for retaining the best parent) has the parameter settings $N=\lambda=2\mu$, $\kappa=1$ ($\kappa=\infty$ for best individual). The pseudo steady state EA design has the parameter settings $N=\lambda=\mu$, $\kappa=\infty$. Unless otherwise stated, $N=50$. Each experiment ran for 3000 generations meaning that 150,000 objective function evaluations are required to obtain a final solution in each experimental run.

**Selection:** Selection occurs by either binary tournament selection (without replacement) or by truncation selection. Both selection methods are described in Chapter 2.

**Search Operators:** For each EA design, an offspring is created by using a single search operator. Two designs were considered: i) a seven search operator design and ii) a two search operator design. For the seven operator case, an offspring is created by an operator that is selected at random from the list in Table 5-2 (no parameter tuning was attempted). For the two operator case, uniform crossover is used with probability = 0.98 and single point random mutation is used with probability = 0.02. Search operator descriptions are provided in Appendix B.

**Table 5-2: Names of the seven search operators used in the cellular GA, SOTEA, and selected Panmictic EA designs are listed below. More information on each of the search operators can be found in Appendix B.**

| **Search Operator Names** |
|---|
| Wright's Heuristic Crossover |
| Simple Crossover |
| Extended Line Crossover |
| Uniform Crossover |
| BLX- α |
| Differential Operator |
| Single Point Random Mutation |

**Constraint Handling:** Each of the engineering design case studies involve nonlinear inequality constraints meaning that solution feasibility must be addressed. Feasibility is dealt with by defining fitness using the Stochastic Ranking method presented in [64] as opposed to defining fitness by the objective function. The parameter settings for Stochastic Ranking were taken from the suggestions in [64].

#### 5.4.2.2 Engineering Design Case Studies

Experiments for assessing SOTEA performance are conducted on six engineering design problems and six artificial test problems taken from the literature (described in Appendix A) and compared against cellular Genetic Algorithms and Panmictic Evolutionary Algorithm designs.

The first four engineering design problems were chosen due to a prior difficulty in solving these problems using Panmictic Evolutionary Algorithms and the difficulty that others have had in solving these problems in general. Some of these test problems are small enough that mathematical programming techniques have been used to solve for the global optimal solution (problems 1, 2, and 3) which has helped in the assessment of algorithm performance. In preliminary experiments conducted on the last two engineering design problems, some Panmictic EA designs were able to find solutions that were better than those reported in the literature. These problems were included in this work to see if additional improvements could be made using SOTEA.

### 5.4.3 Results

**General Performance Statistics:** This section attempts to draw general conclusions about the three EA design classes (Panmictic EA, cellular GA, and SOTEA) tested in these experiments. The first statistic shown in column two of Table 5-3 measures the percentage of runs that an EA design class was able to find the optimal solution (optimal defined as the best solution value found in all experiments). This percentage is an average over all test problems. The second statistic shown in column three measures the percentage of runs where an EA design class finds a solution that ranks in the top 5% of all solutions found in these experiments. The third statistic shown in column four is a $p$ value for the Mann-Whitney U-test where the statistical hypothesis is that the given EA design class is superior to the other two EA design classes. The fourth statistic shown in column five measures the percentage of problems where the best EA design belonged to a particular design class[19].

[19] Notice that eight Panmictic EA designs were used in these experiments while only four cellular GA and four SOTEA designs were used. This should bias columns 3 and 6 of Table 5-3 to favor the Panmictic EA.

The sixth column looks at the percentage of problems where an EA design class was able to find the best solution at least one time.

For each of the statistics in Table 5-3, SOTEA is found to be significantly better than the other EA design classes based on the 12 problems tested in these experiments. Particularly impressive are the results in column five which indicate that the SOTEA design is the best EA design in about 80% of the problems tested.

**Table 5-3 Overall performance statistics for the Panmictic EA, the cellular GA, and SOTEA. Column two measures the percentage of runs where the optimal solution was found. The optimal solution is defined as the best solution found in these experiments. Column three measures the percentage of runs where the solution ranks in the top 5% of solutions from all EA designs. In column four, "*p*" indicates the *p* value for the Mann-Whitney U-test where the hypothesis is that the given EA design class is superior to the other two EA design classes. Column five measures the percentage of problems where the best EA design belonged to a particular design class. Column six measures the percentage of problems where an EA design class was able to find the best solution at least one time. Statistics in columns 1-3 are an average value over all test problems.**

| EA Design | % of runs where EA | | U-Test | % of problems where EA | |
|---|---|---|---|---|---|
| | found best | was top 5% | p | was best design | found best |
| Panmictic EA | 4.2% | 5.1% | 0.87 | 8.3% | 16.7% |
| cellular GA | 9.4% | 11.6% | 0.50 | 12.5% | 66.7% |
| SOTEA | 16.9% | 25.2% | 0.13 | 79.2% | 83.3% |

#### 5.4.3.1 Engineering Design Performance Results

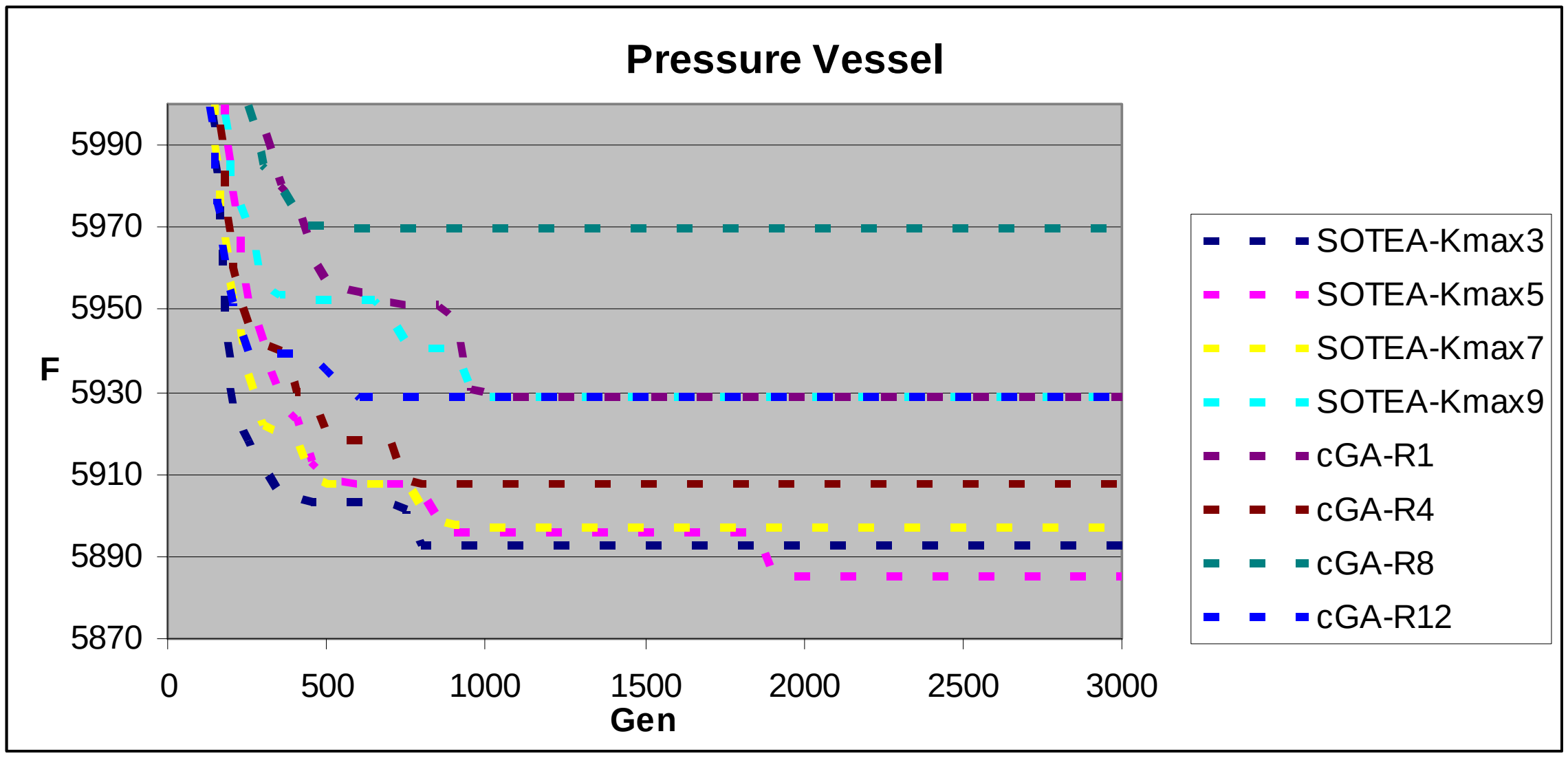

**Figure 5-15 Performance results for the Pressure Vessel design problem are shown over 3000 generations for SOTEA with different settings of $K_{max}$, and for cellular GA with different values of the neighborhood radius $R$. Performance for each EA is an average over 20 runs of the best fitness (objective function) value in the population. Infeasible solutions are neglected from the calculations, however all runs obtained feasibility within the first 100 generations. The global optimal solution has a fitness of 5850.38.**

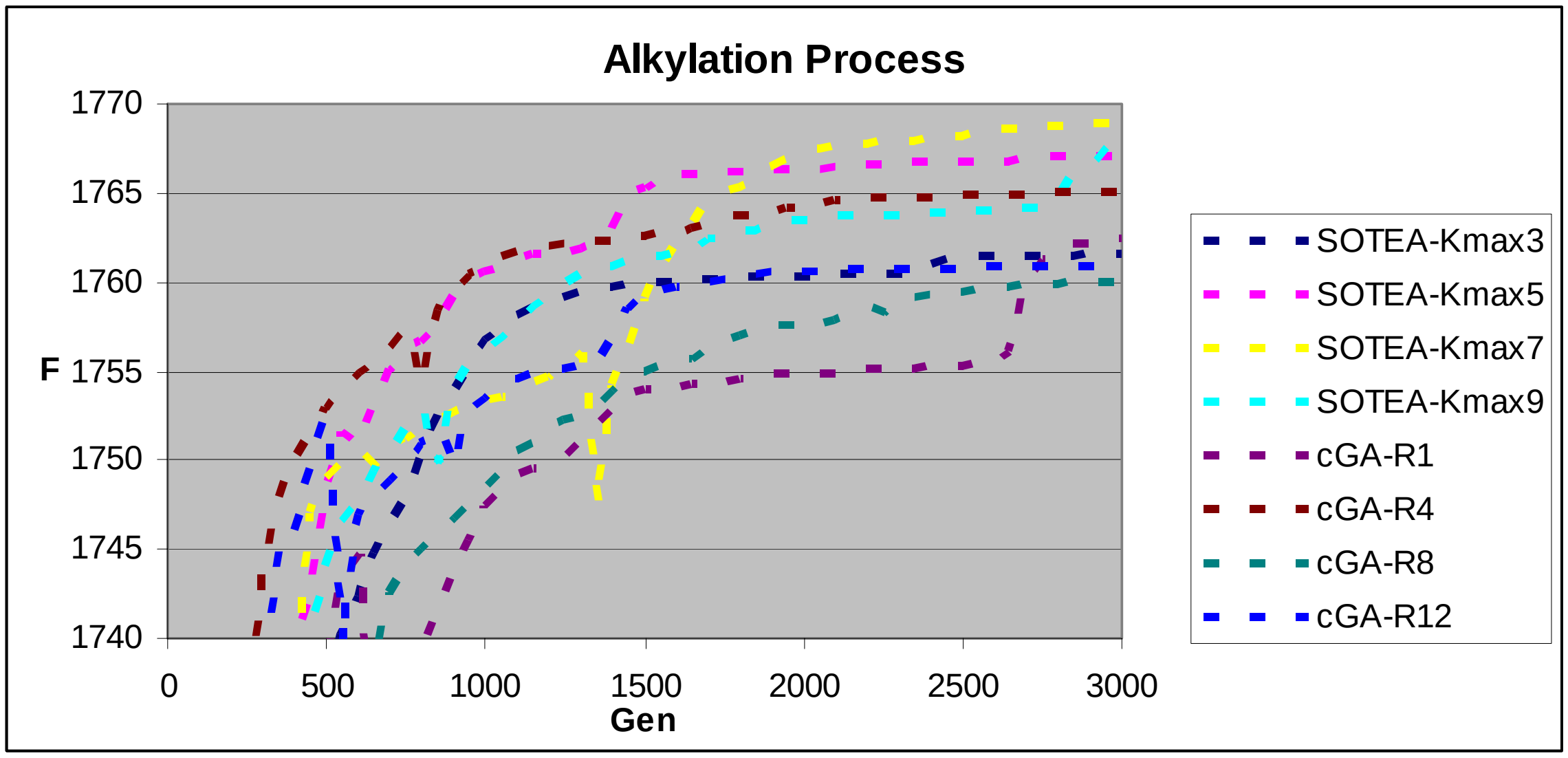

**Figure 5-16 Performance results for the Alkylation Process design problem are shown over 3000 generations for SOTEA with different settings of $K_{max}$, and for cellular GA with different values of the neighborhood radius $R$. Performance for each EA is an average over 20 runs of the best fitness (objective function) value in the population. Infeasible solutions are neglected from the calculations, however all runs obtained feasibility within the first 1400 generations. Several instances can be observed where fitness values momentarily decrease. This is the result of EA runs turning from infeasible to feasible where the new feasible solution is lower than the average performance for that EA design and generation. The global optimal solution has a fitness of 1772.77.**

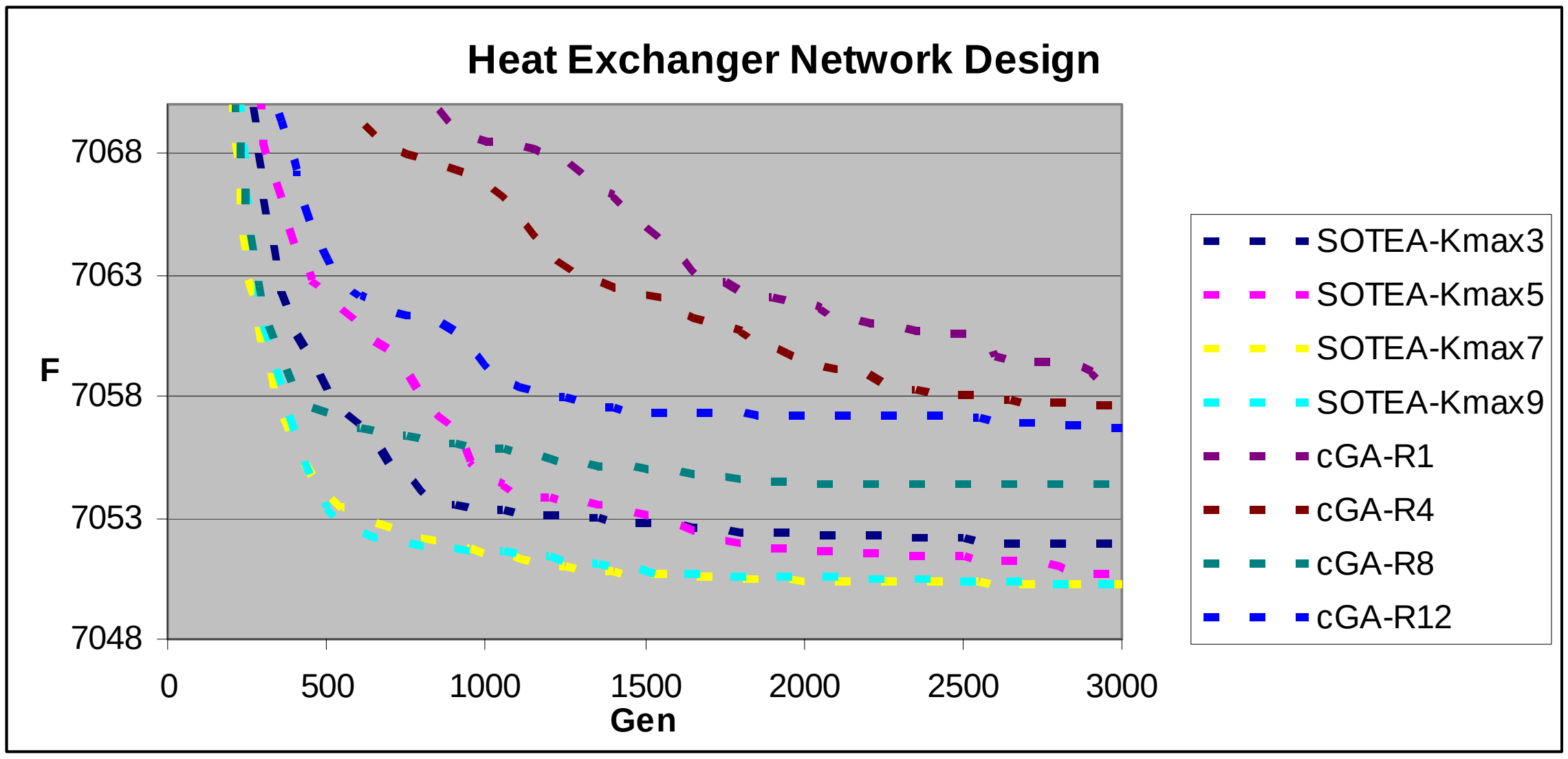

**Figure 5-17 Performance results for the Heat Exchanger Network design problem are shown over 3000 generations for SOTEA with different settings of $K_{max}$, and for cellular GA with different values of the neighborhood radius $R$. Performance for each EA is an average over 20 runs of the best fitness (objective function) value in the population. Infeasible solutions are neglected from the calculations, however all runs obtained feasibility within the first 100 generations. The global optimal solution has a fitness of 7049.25.**

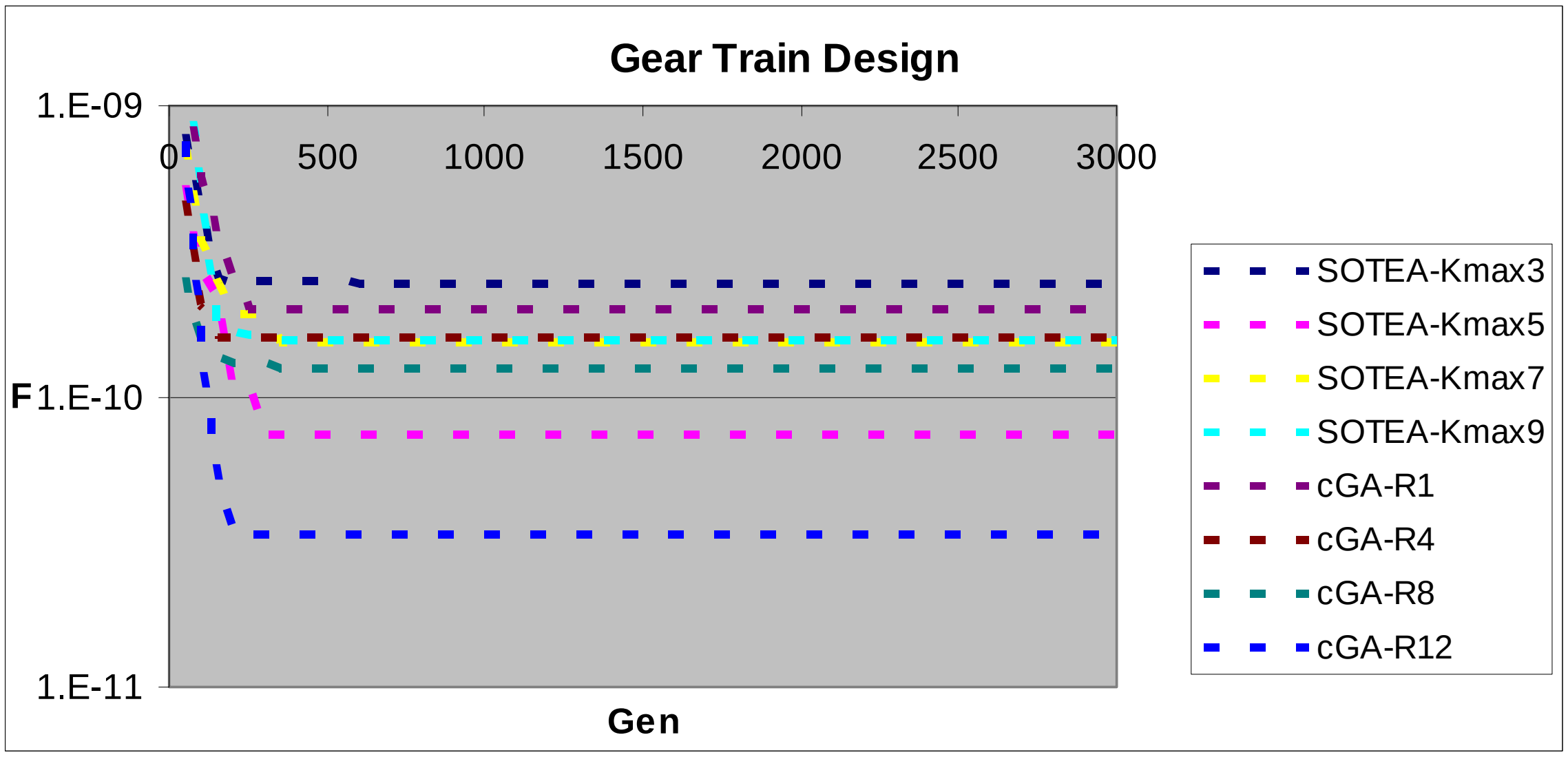

**Figure 5-18 Performance results for the Gear Train Design design problem are shown over 3000 generations for SOTEA with different settings of $K_{max}$, and for cellular GA with different values of the neighborhood radius $R$. Performance for each EA is an average over 20 runs of the best fitness (objective function) value in the population. Infeasible solutions are neglected from the calculations, however all runs obtained feasibility within the first 50 generations. The global optimal solution is unknown, however the best result previous to this work, is reported in [220] as 2.70E-12.**

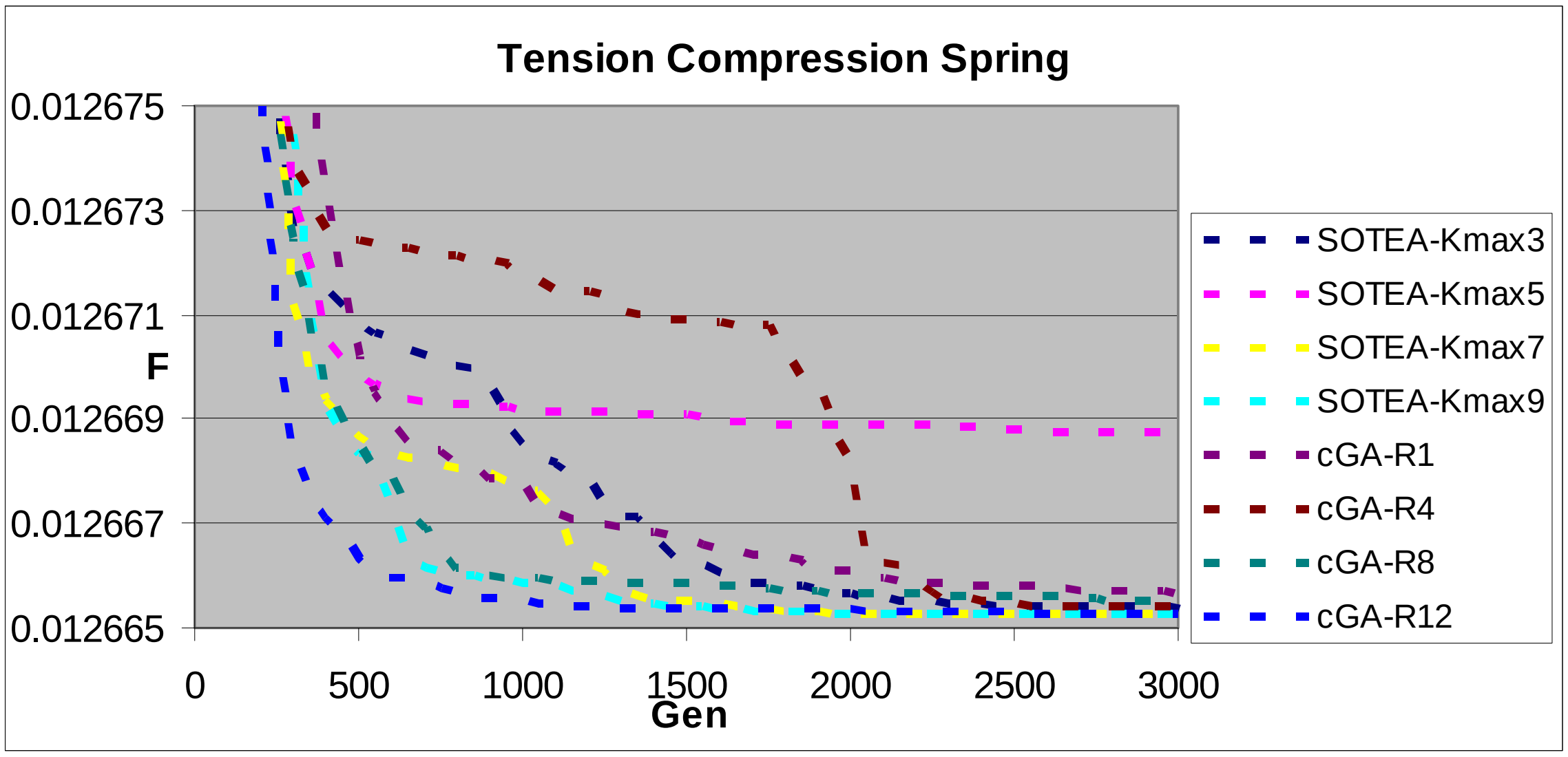

**Figure 5-19 Performance results for the Tension Compression Spring Design design problem are shown over 3000 generations for SOTEA with different settings of $K_{max}$, and for cellular GA with different values of the neighborhood radius *R*. Performance for each EA is an average over 20 runs of the best fitness (objective function) value in the population. Infeasible solutions are neglected from the calculations, however all runs obtained feasibility within the first 50 generations. The global optimal solution is unknown, however the best result previous to this work, is reported in [221] as 0.01270.**

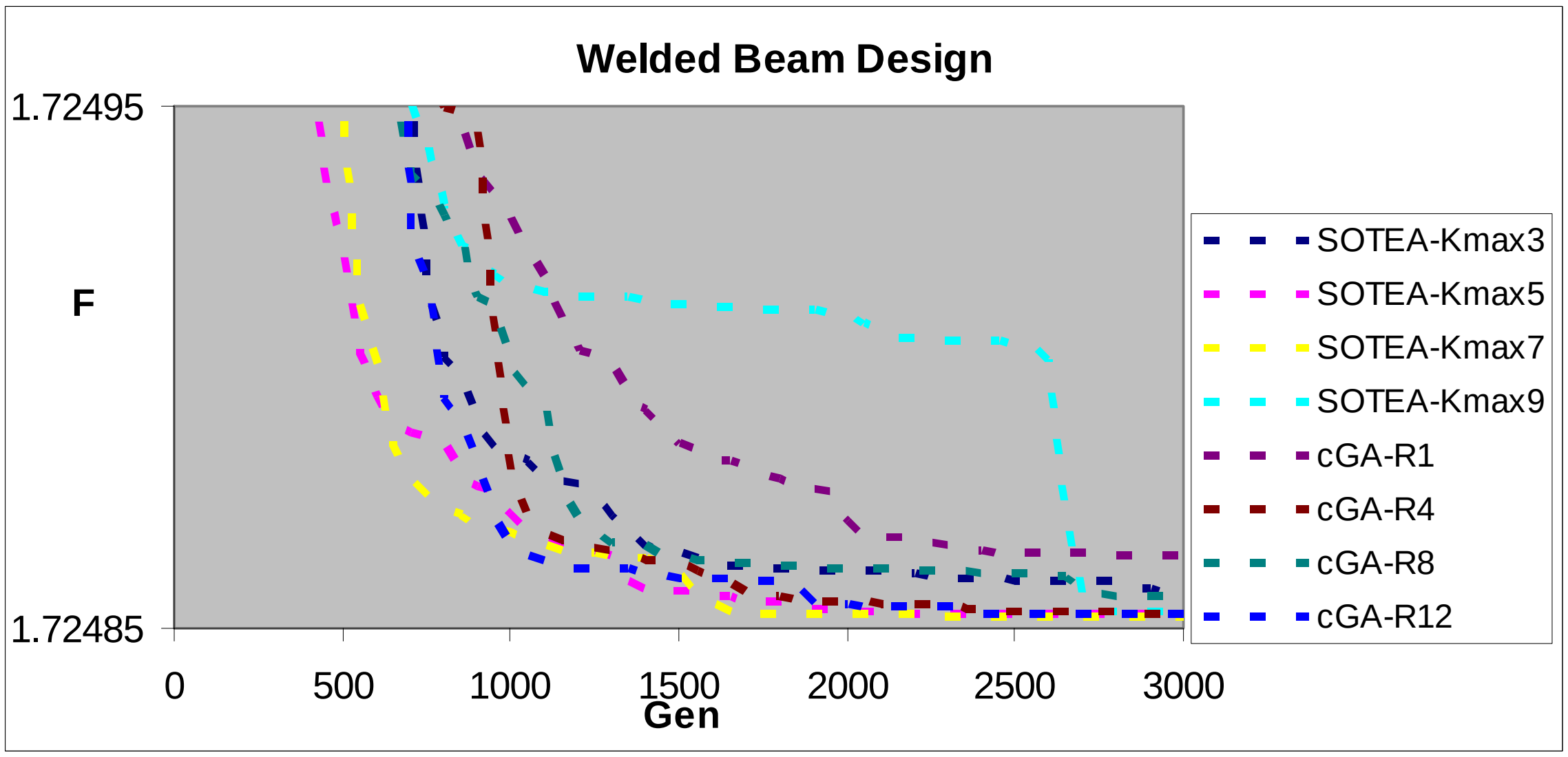

**Figure 5-20 Performance results for the Welded Beam Design design problem are shown over 3000 generations for SOTEA with different settings of $K_{max}$, and for cellular GA with different values of the neighborhood radius *R*. Performance for each EA is an average over 20 runs of the best fitness (objective function) value in the population. Infeasible solutions are neglected from the calculations, however all runs obtained feasibility within the first 50 generations. The global optimal solution is unknown, however the best result previous to this work, is reported in [222] as 1.7255.**

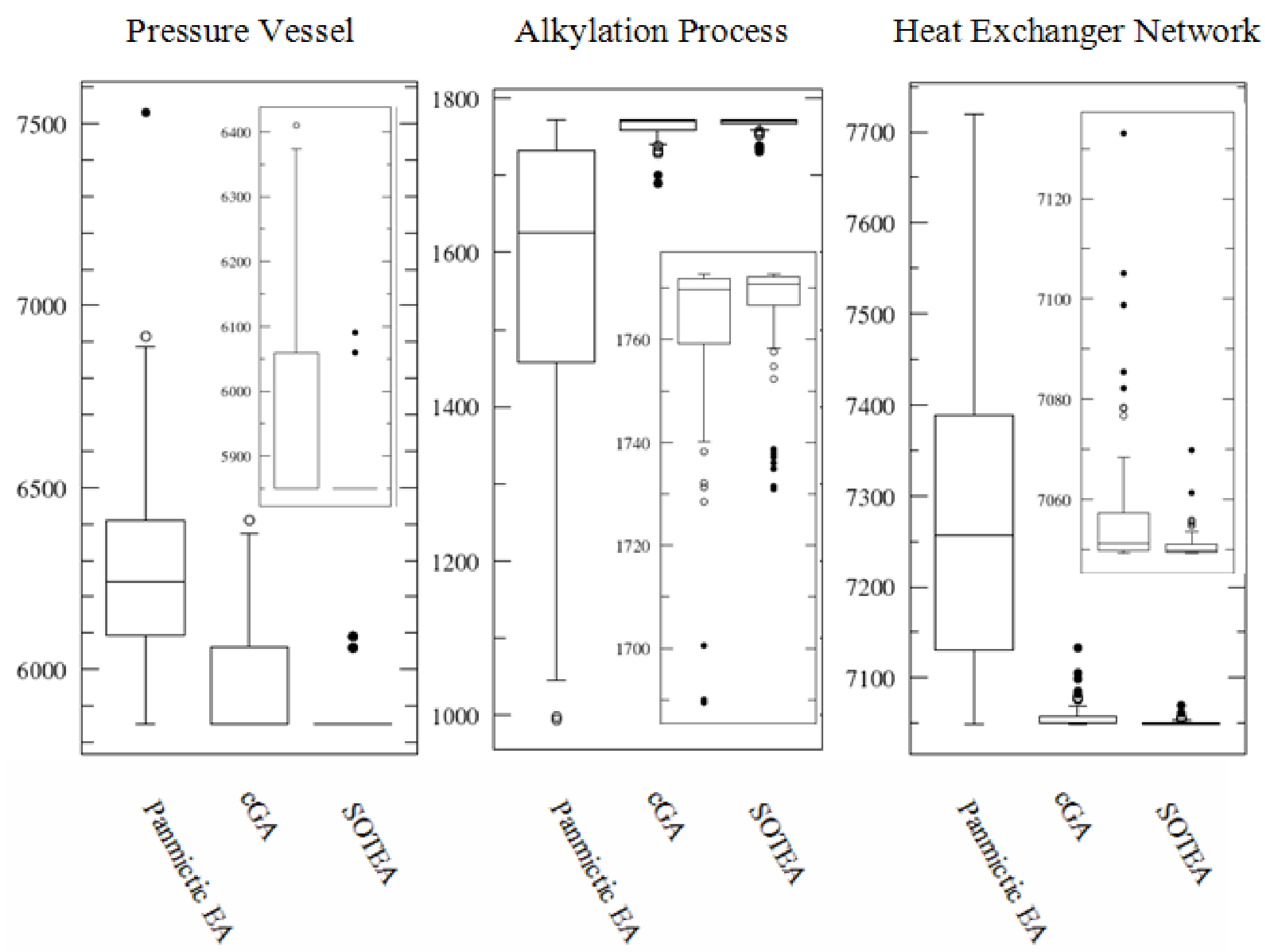

**Figure 5-21 Final performance results for the Pressure Vessel (Left), Alkylation Process (Middle) and Heat Exchanger Network (Right) design problems are shown with box plots of performance data grouped by Panmictic EA, cellular GA, and SOTEA. The box plots represent final algorithm performance (after 3000 generations) over 20 runs for all cGA, SOTEA, and Panmictic EA designs. This includes data from the four cGA designs (with different parameter settings for neighborhood radius *R*), the four SOTEA designs (with different parameter settings for $K_{Max}$), and the eight Panmictic EA designs described in Section 5.4.2.1. Insets are provided for the cGA and SOTEA box plots to highlight the difference in results between these two algorithms. Also notice that the Pressure Vessel and Heat Exchanger Network problems are Minimization problems while the Alkylation Problem is a Maximization problem.**

**Results for Pressure Vessel Design Problem:** For the Pressure Vessel design problem, all but one of the SOTEA algorithms outperformed all of the cellular GA designs as seen in Figure 5-15. Performance also tended to improve as network connectivity was reduced for both SOTEA and the cGA. In light of this trend, it is not surprising to see the performance of the suite of Panmictic EA designs performed very poorly on this problem as seen in Figure 5-21. Comparing results between Figure 5-15 and Table C-7 (in Appendix C), the best final solution for a Panmictic EA design is beaten by all SOTEA designs after only 300 generations.

Comparisons to work from previous authors highlights the strong performance of both of the distributed Evolutionary Algorithms. Of the 8 papers referenced in and including [223], only one other algorithm has been able to reach the objective function values obtained by the distributed EA designs employed here. Performance comparison tables are provided in Appendix C.

**Results for Alkylation:** For the Alkylation Process design problem, all but one of the SOTEA algorithms outperformed the cellular GA designs as seen in Figure 5-16. In this problem there was no clear trend between performance and network connectivity. It is also clear that many of the algorithms were able to find improvements throughout the run suggesting that convergence did not occur within the 3000 generations considered. Hence, it is possible that the conclusions drawn here would change if evolution was considered over a larger time scale. The Panmictic EA designs again performed relatively poorly on this problem as seen in Figure 5-21.

Comparisons to work from previous authors highlights the strong performance of the distributed Evolutionary Algorithms. Of the stochastic search methods described in the 5 papers referenced in [224] including their own Differential Evolution Algorithm, none reached the fitness values obtained by the distributed EA designs employed here. However, two αBB (Branch and Bound Non-Linear Programming) algorithms were cited which did find the global optimum. Performance comparison tables are provided in Appendix C.

**Results for HEN:** For the Heat Exchanger Network design problem, all of the SOTEA algorithms outperformed the cellular GA designs as seen in Figure 5-17. Performance also tended to improve as network connectivity was increased for both SOTEA and the cGA. Such a trend seems to suggest that interaction constraints are not needed for this problem which makes the poor performance of the Panmictic EA designs (see Figure 5-21) a little surprising. Comparing results between Figure 5-17 and Table C-7 (in Appendix C), the best final result for a Panmictic EA design is beaten by all SOTEA designs after only 400 generations.

Comparisons to other work are less favorable in this case. For instance, in [224], they introduce a Differential Evolution Algorithm that can find the optimal solution 100% of the time in under 40,000 evaluations. None of the algorithms employed here were able to obtain that level of performance for this problem. In fact, the best algorithm (SOTEA with $K_{max}$ = 7) was only able to find the optimal solution 65% of the time in 150,000 evaluations.

To make a fair comparison to the results in [224], the results from this thesis were also analyzed at 40,000 evaluations and under these conditions only two of the SOTEA algorithms (and none of the cellular GAs) were able to find an optimal solution in that amount of time (with the optimal being found only 10% of the time). It is worth mentioning that this was one of the simplest design problems tested with only a marginal level of epistasis between parameters (e.g. see problem definition in Appendix A).

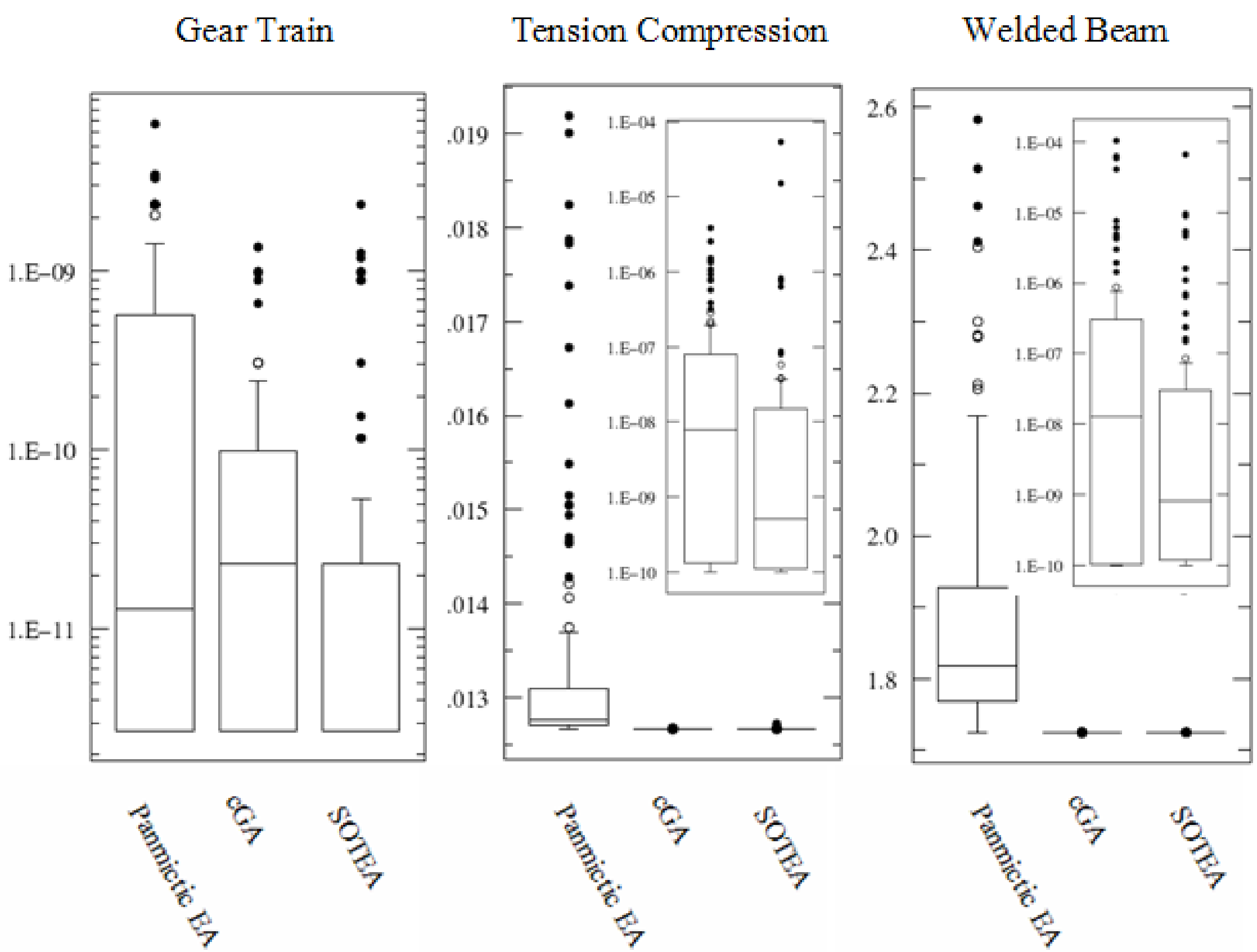

**Figure 5-22 Final performance results for the Gear Train (Left), Tension Compression Spring (Middle) and Welded Beam (Right) design problems are shown with box plots of performance data grouped by Panmictic EA, cellular GA, and SOTEA. The box plots represent final algorithm performance (after 3000 generations) over 20 runs for all cGA, SOTEA, and Panmictic EA designs. This includes data from the four cGA designs (with different parameter settings for neighborhood radius *R*), the four SOTEA designs (with different parameter settings for $K_{Max}$), and the eight Panmictic EA designs described in Section 5.4.2.1. When necessary, insets are provided for the cGA and SOTEA box plots (with data shifted and plotted on a log scale) to highlight the difference in results between these two algorithms. All three design problems are Minimization problems.**

**Results for Gear Train Design Problem:** For the gear train design problem, there was no clear distinction in performance between the cellular GA and SOTEA. One of the cellular GA designs (*R*=12) was found to have better average performance than any of the SOTEA designs as seen in Figure 5-18 however comparison of end performance between the cellular GA, SOTEA, and the Panmictic EA shows very little difference as seen in Figure 5-22. Of the two papers referenced in and including [220], one previous method has been able to find the solutions achieved in this work.

**Results for Tension Compression Spring Design Problem:** For the tension compression spring design problem, all but one of the distributed EA designs were found to converge to nearly identical values as seen in Figure 5-19. The SOTEA design with $K_{Max}$=5 was found to have worse performance than the other designs. However, comparison of end performance as shown in Figure 5-22 shows SOTEA did have a better median performance compared to the cellular GA.

Comparisons to work from previous authors highlights the strong performance of both of the distributed Evolutionary Algorithms. Of the three papers referenced in and including [221], no previous method has been able to find the solutions achieved in this work.

**Results for Welded Beam Design Problem:** For the welded beam design problem, each of the distributed EA designs were found to converge to nearly identical values as seen in Figure 5-20. Both distributed EA designs strongly outperformed the Panmictic EA as seen in Figure 5-22.

Comparisons to work from previous authors highlights the strong performance of both of the distributed Evolutionary Algorithms. Of the 3 papers referenced in and including [222], no previous method has been able to find the solutions achieved in this work.

### 5.4.3.2 Artificial Test Function Results

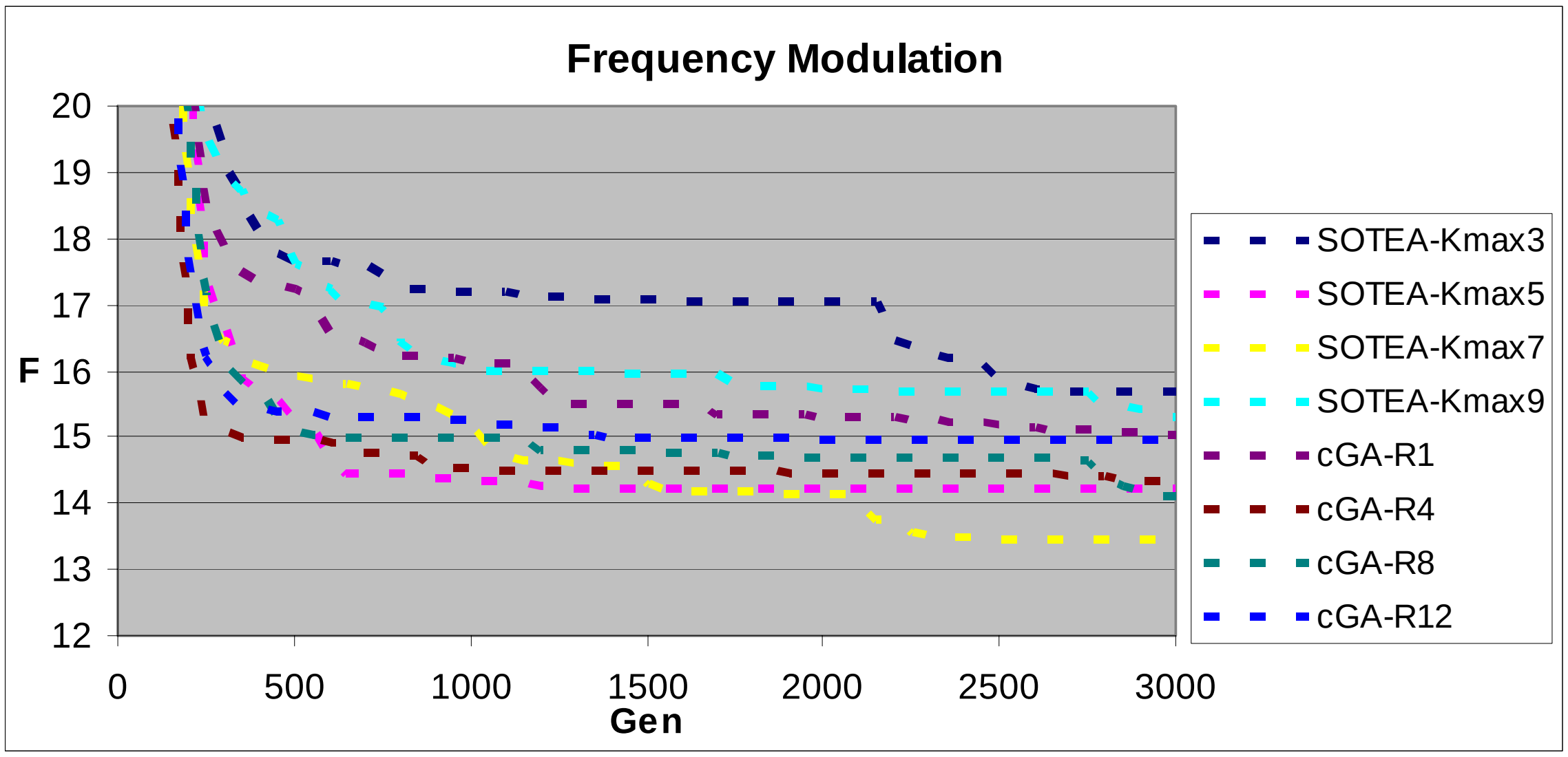

**Figure 5-23 Performance results for the Frequency Modulation problem are shown over 3000 generations for SOTEA with different settings of $K_{max}$, and for cellular GA with different values of the neighborhood radius *R*. Performance for each EA is an average over 20 runs of the best fitness (objective function) value in the population. The global optimal solution is 0.**

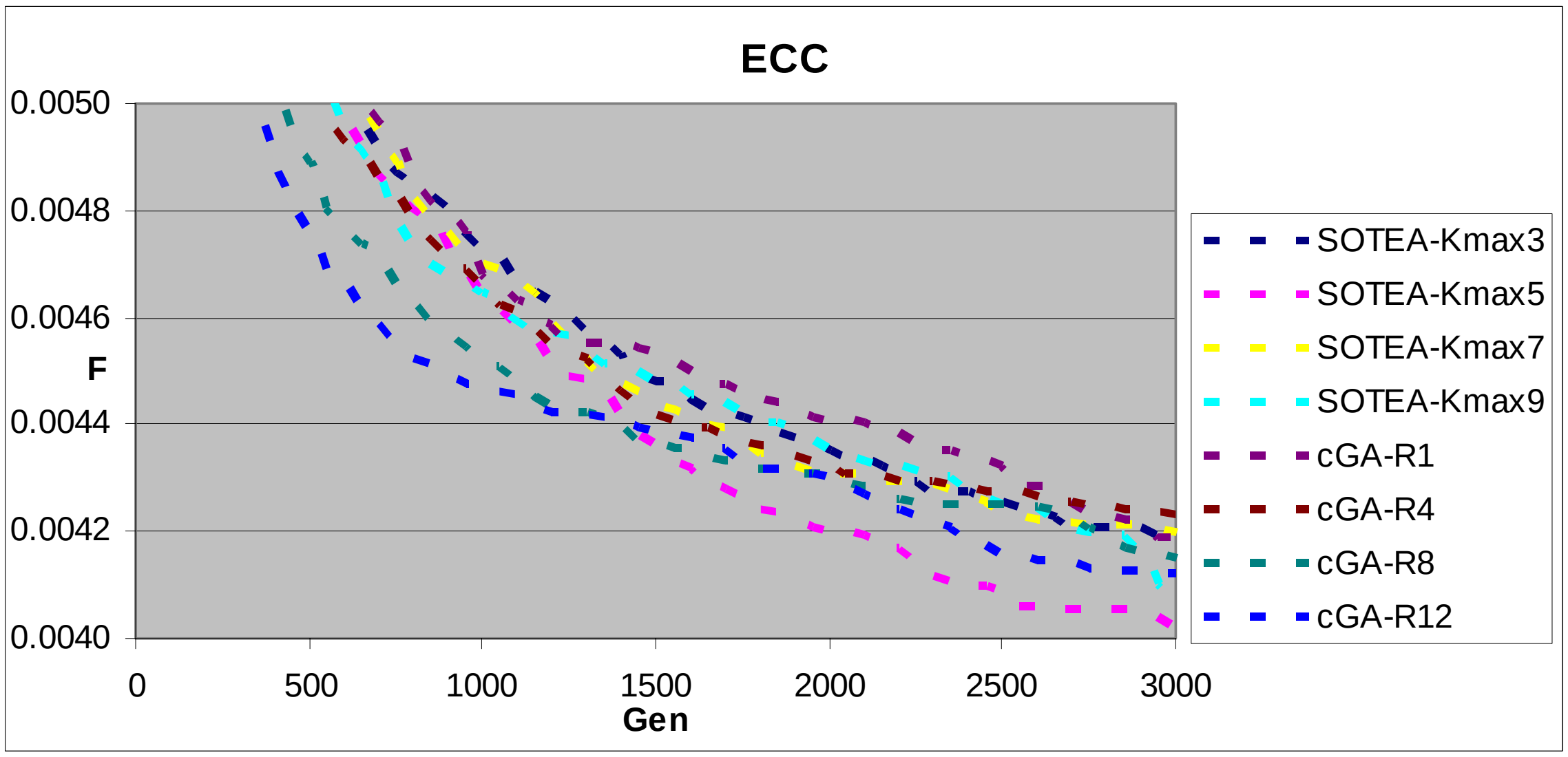

**Figure 5-24 Performance results for the error correcting code (ECC) problem are shown over 3000 generations for SOTEA with different settings of $K_{max}$, and for cellular GA with different values of the neighborhood radius *R*. Performance for each EA is an average over 20 runs of the best fitness (objective function) value in the population. The global optimal solution is 0.067416. Results are shifted so that global optima is 0.**

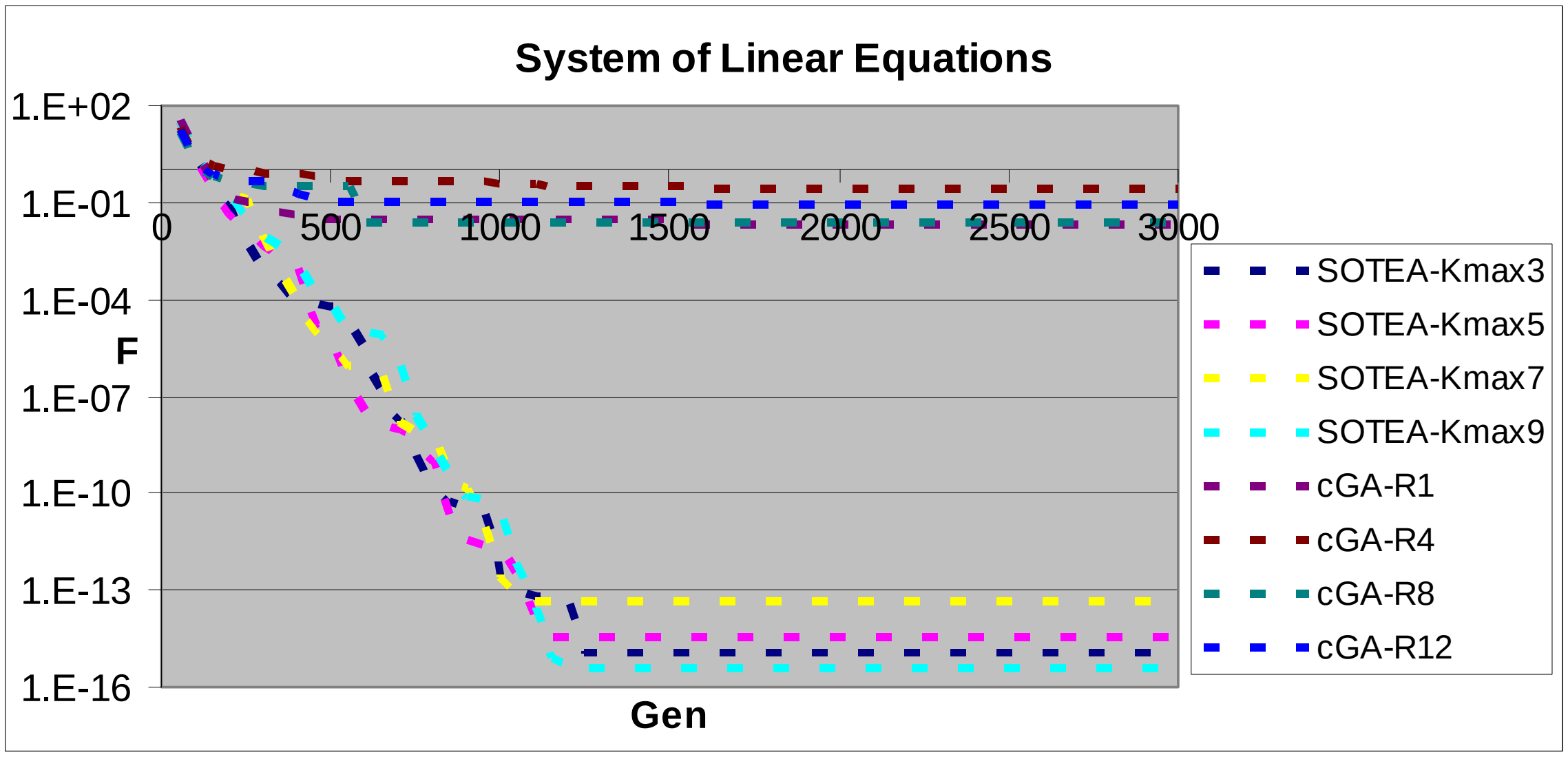

**Figure 5-25 Performance results for the system of linear equations problem are shown over 3000 generations for SOTEA with different settings of $K_{max}$, and for cellular GA with different values of the neighborhood radius $R$. Performance for each EA is an average over 20 runs of the best fitness (objective function) value in the population. The global optimal solution is 0.**

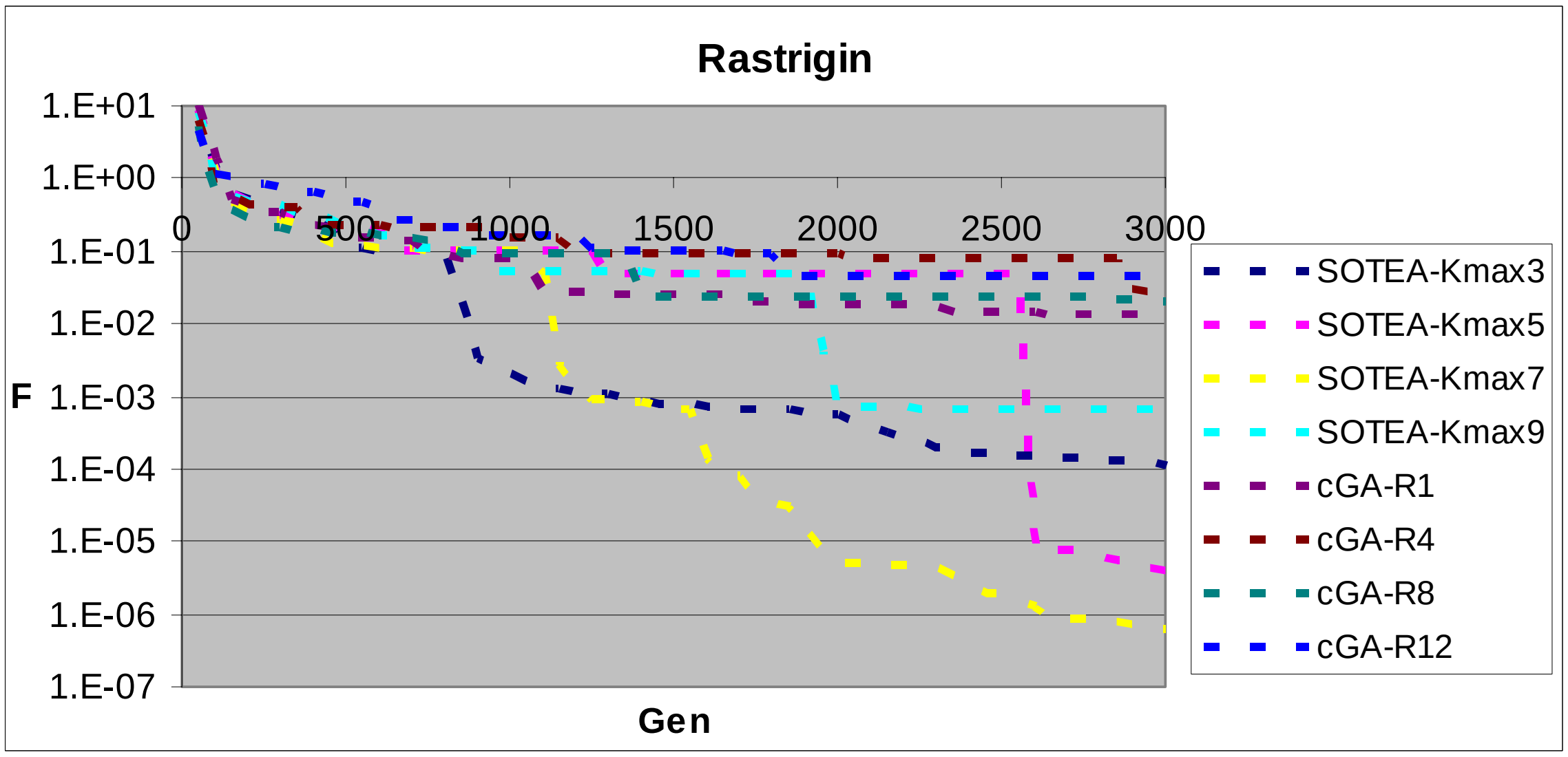

**Figure 5-26 Performance results for the Rastrigin function are shown over 3000 generations for SOTEA with different settings of $K_{max}$, and for cellular GA with different values of the neighborhood radius $R$. Performance for each EA is an average over 20 runs of the best fitness (objective function) value in the population. The global optimal solution is 0.**

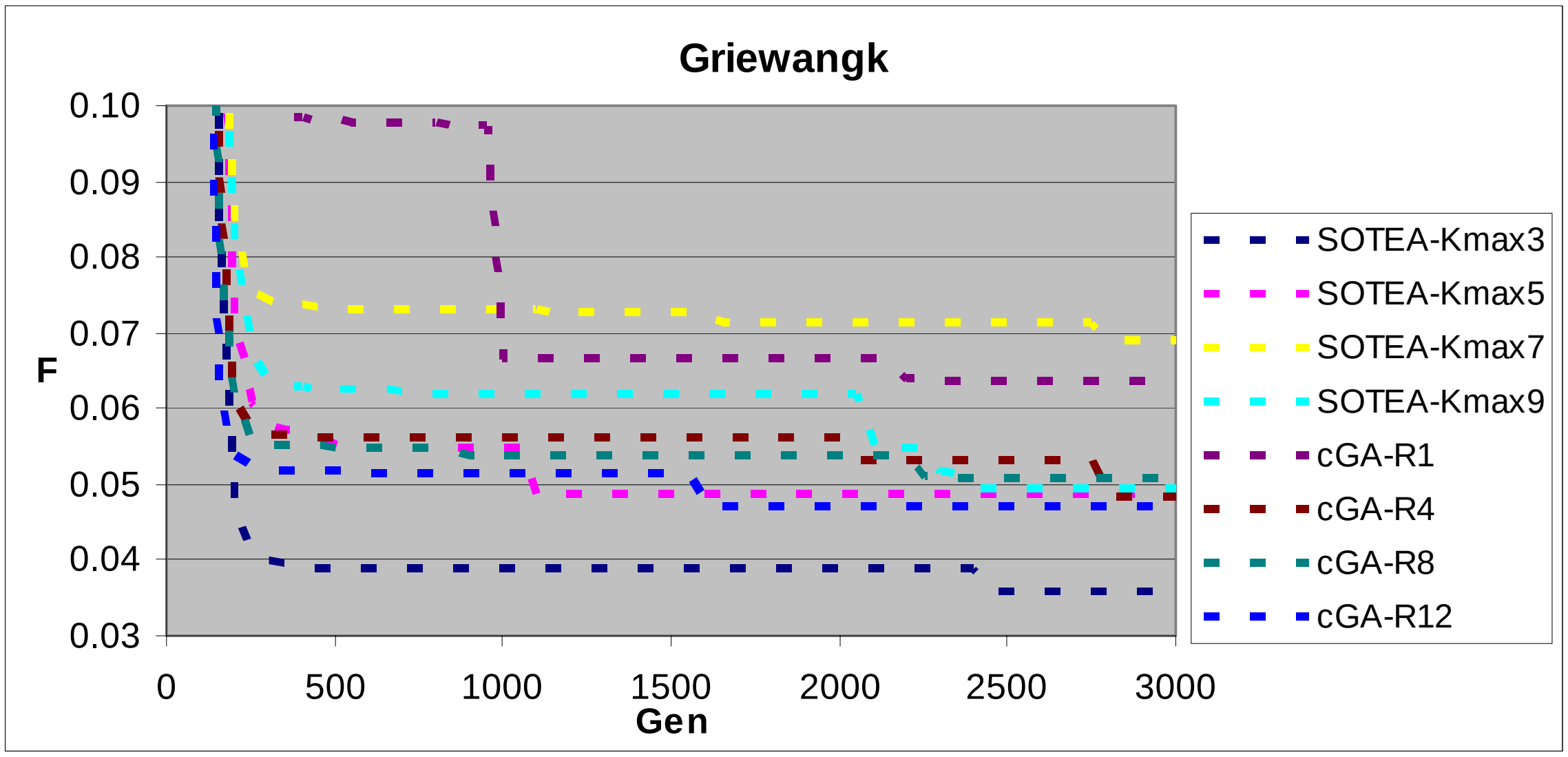

**Figure 5-27 Performance results for the Griewangk function are shown over 3000 generations for SOTEA with different settings of $K_{max}$, and for cellular GA with different values of the neighborhood radius *R*. Performance for each EA is an average over 20 runs of the best fitness (objective function) value in the population. The global optimal solution is 0.**

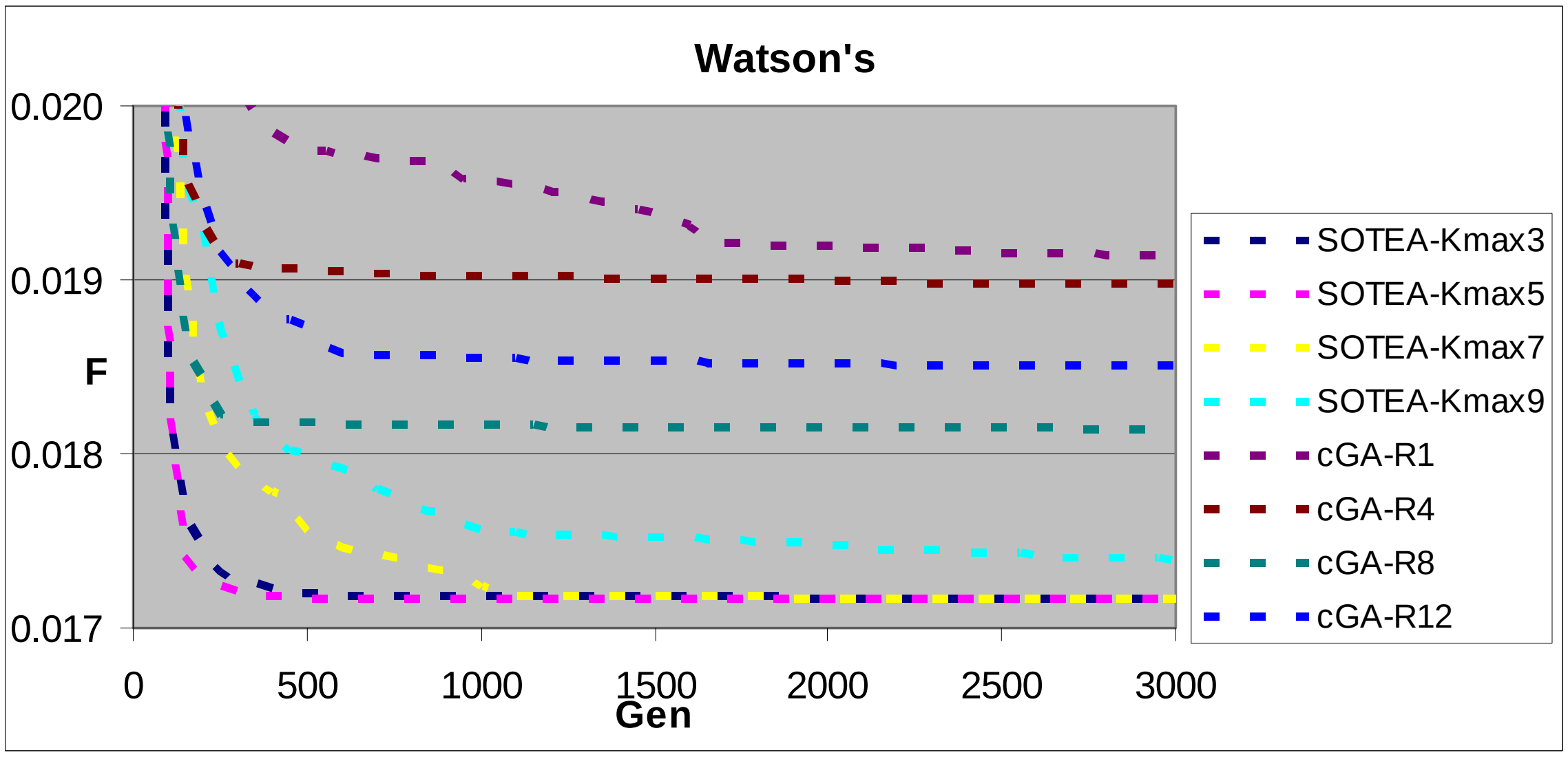

**Figure 5-28 Performance results for Watson's function are shown over 3000 generations for SOTEA with different settings of $K_{max}$, and for cellular GA with different values of the neighborhood radius *R*. Performance for each EA is an average over 20 runs of the best fitness (objective function) value in the population. The global optimal solution is 0.01714.**

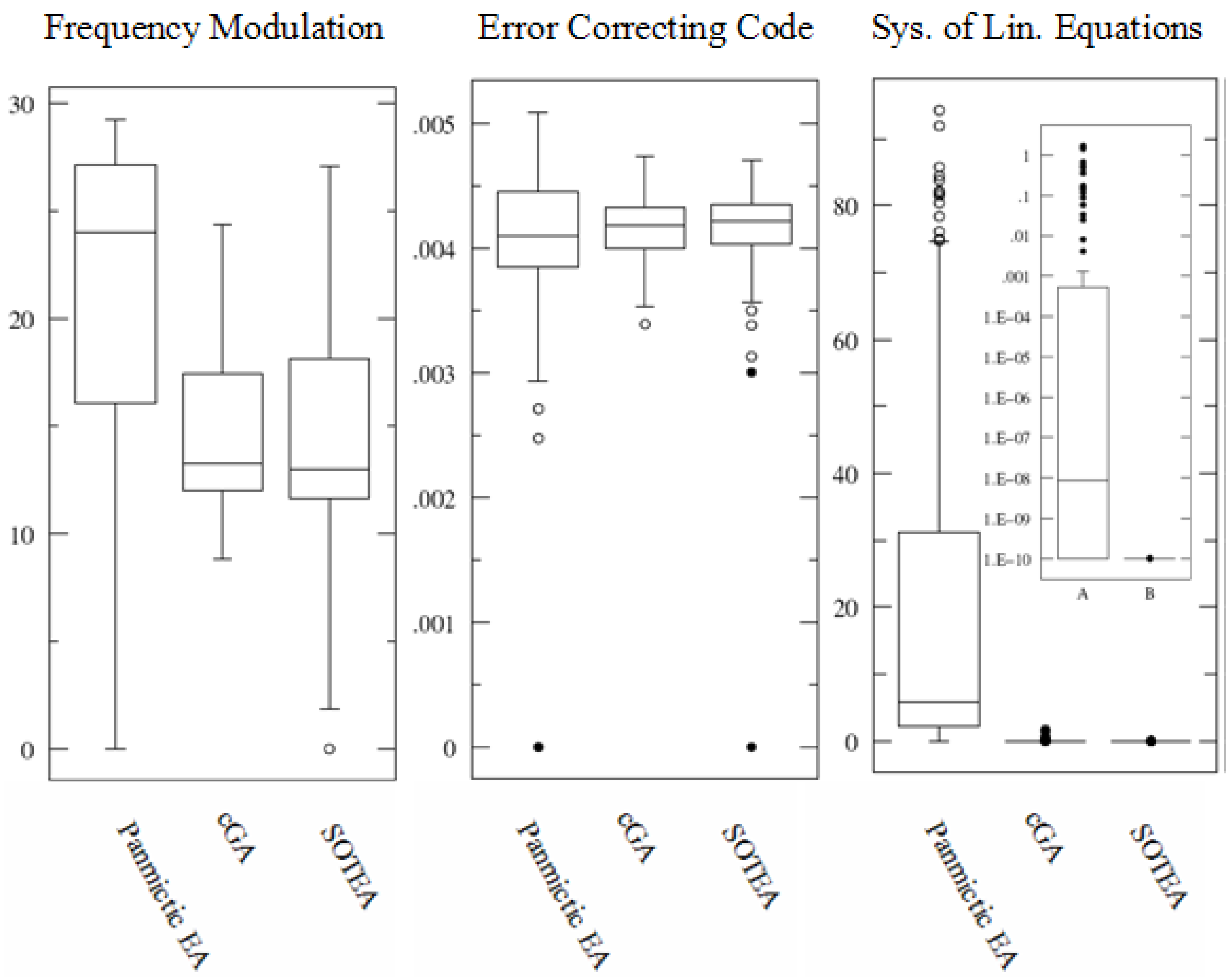

**Figure 5-29 Final performance results for the Frequency Modulation (Left), Error Correcting Code (Middle) and System of Linear Equations (Right) test functions are shown with box plots of performance data grouped by Panmictic EA, cellular GA, and SOTEA. The box plots represent final algorithm performance (after 3000 generations) over 20 runs for all cGA, SOTEA, and Panmictic EA designs. This includes data from the four cGA designs (with different parameter settings for neighborhood radius *R*), the four SOTEA designs (with different parameter settings for $K_{Max}$), and the eight Panmictic EA designs described in Section 5.4.2.1. When necessary, insets are provided for the cGA and SOTEA box plots (with data plotted on a log scale) to highlight the difference in results between these two algorithms. All three design problems are Minimization problems.**

**Results for Frequency Modulation:** For the frequency modulation problem, SOTEA designs are found to be both the best and worst performers (compared to the cellular GA) throughout the optimization runs as seen in Figure 5-23. The larger distribution of SOTEA performance is also evident in the final performance results shown in Figure 5-29. Here it can also see that while both SOTEA and the cellular GA have much better median performance than the Panmictic EA, only SOTEA and one of the Panmictic EAs were able to find the global optimal solution.

**Results for ECC:** For the error correcting code problem, both SOTEA and the cellular GA designs are able to make steady progress toward the optimal solution with little difference between the two designs as seen in Figure 5-24. However, in the final distribution of results shown in Figure 5-29, only SOTEA and one of the Panmictic EAs were able to find the global optimal solution.

**Results for System of Linear Equations:** For the system of linear equations test function, SOTEA designs overwhelmingly outperform the cellular GA as seen in Figure 5-25 and Figure 5-29. Also seen in Figure 5-29, both distributed EA designs were able to strongly outperform the Panmictic EA designs.

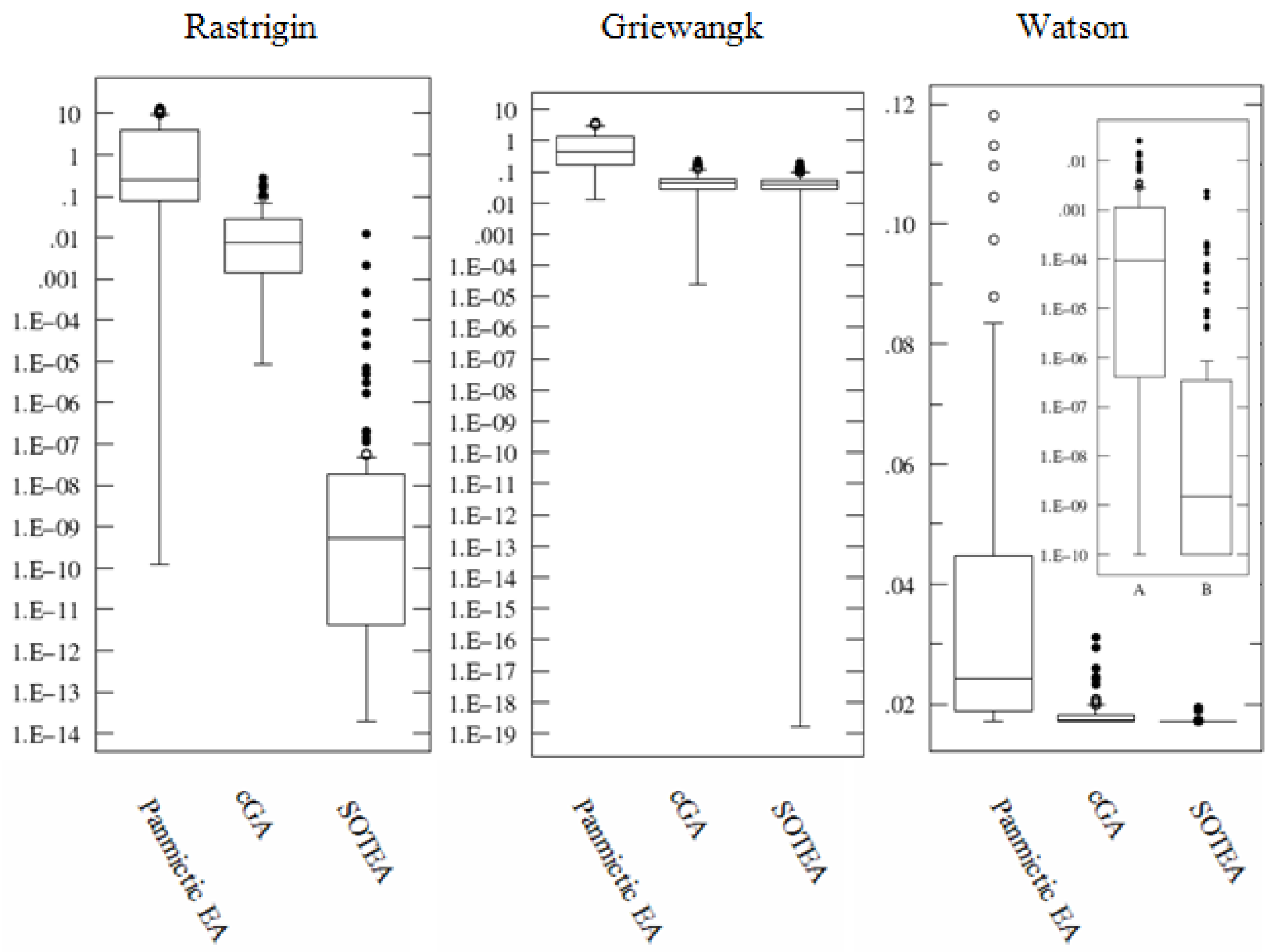

**Figure 5-30 Final performance results for the Rastrigin (Left), Griewangk (Middle) and Watson (Right) test functions are shown with box plots of performance data grouped by Panmictic EA, cellular GA, and SOTEA. The box plots represent final algorithm performance (after 3000 generations) over 20 runs for all cGA, SOTEA, and Panmictic EA designs. This includes data from the four cGA designs (with different parameter settings for neighborhood radius *R*), the four SOTEA designs (with different parameter settings for $K_{Max}$), and the eight Panmictic EA designs described in Section 5.4.2.1. When necessary, insets are provided for the cGA and SOTEA box plots (with data shifted and plotted on a log scale) to highlight the difference in results between these two algorithms. All three design problems are Minimization problems.**

**Results for Rastrigin:** For the Rastrigin test function, SOTEA designs overwhelmingly outperform the cellular GA and the Panmictic EA as seen in Figure 5-26 and Figure 5-30. Although both distributed EA designs have significantly better median performance than the Panmictic EA designs, there is some indication that the Panmictic EA can occasionally find better quality solutions than the cellular GA as seen in Figure 5-30.

**Results for Griewangk:** For the Griewangk test function, SOTEA designs are very similar in performance to the cellular GA as seen in Figure 5-27. Both distributed EA designs perform better than the Panmictic EA designs as seen in Figure 5-30. However, from Figure 5-30 it also appears that SOTEA can occasionally find better quality solutions than the cellular GA.

**Results for Watson:** For Watson's test function, SOTEA designs overwhelmingly outperform the cellular GA as seen in Figure 5-28. Both distributed EA designs perform better than the Panmictic EA designs as seen in Figure 5-30.

#### 5.4.3.3 Structural Analysis

This section presents the structural characteristics of SOTEA and compares this with the cellular GA, the Panmictic EA, and values observed in complex biological systems. These results indicate that, unlike standard EA population topologies, SOTEA obtains several characteristics observed in complex biological systems.

**Methods for SOTEA Topological Analysis:** Because network dynamics in SOTEA take place due to changes in node fitness and because node fitness is constantly evolving (due to population dynamics), the SOTEA network never fully converges to a stable structure. In order to determine topological characteristics, measurements are taken every 50 generations for SOTEA run 10 times over 1000 generations. To consider the impact of system size, topological properties for population sizes of $N$ = 50, 100 and 200 have been measured with results shown in Figure 5-31. Here it is seen that most properties show little dependency on the population size except for $L$ which is generally smaller for smaller systems. Figure 5-31 also indicates that the topological properties of SOTEA are sensitive to the setting of $K_{Max}$ which is the only extra parameter of the SOTEA design. The topological property values for SOTEA with $N$=50 are also reported in Table 5-4, which are taken as an average over all $K_{Max}$ settings considered in this work ($K_{Max}$ = 3, 5, 7, 9).

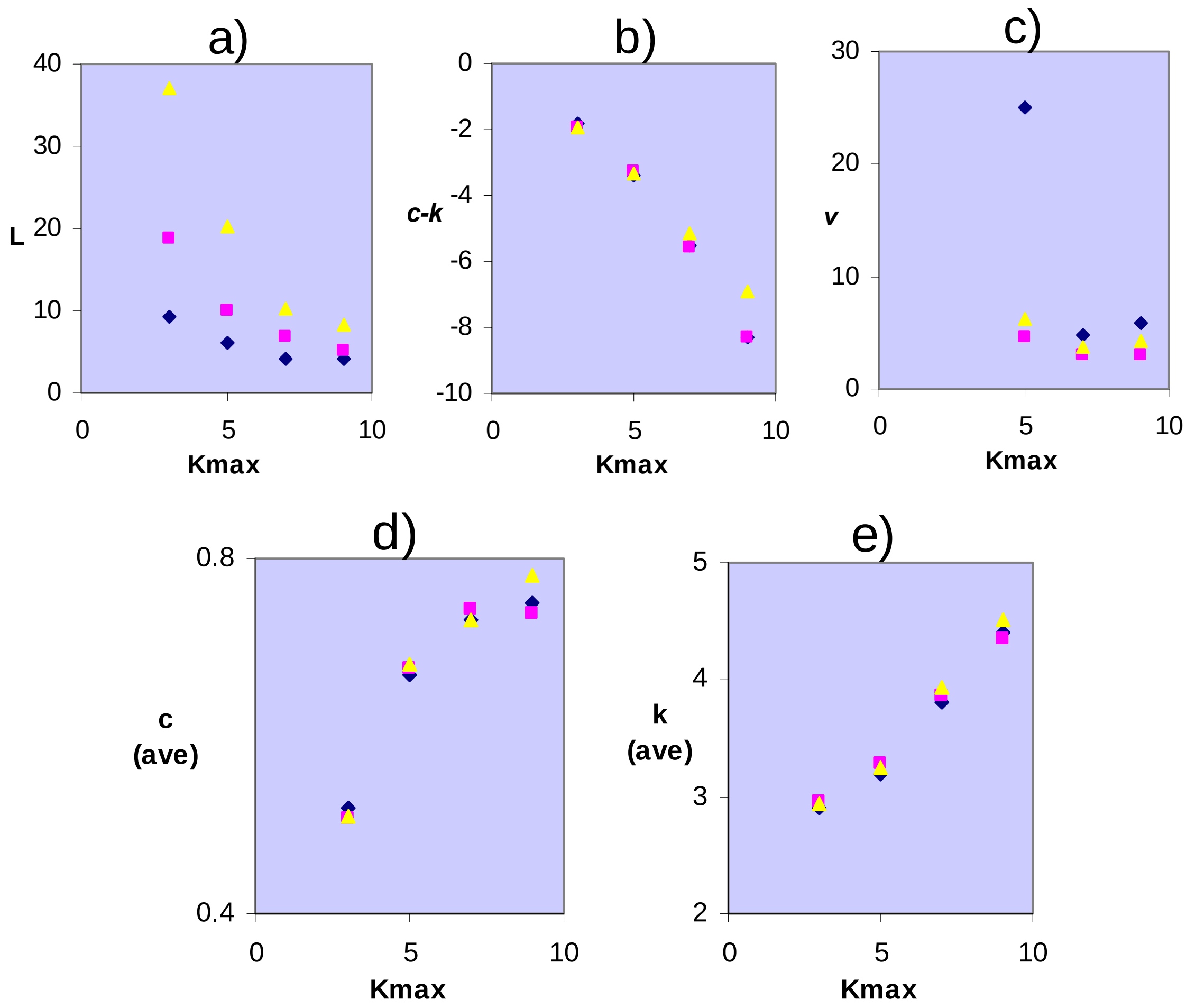

**Figure 5-31 Topological properties for SOTEA with different values of $K_{Max}$ and population sizes of $N$ = 50 (◆), 100(■), and 200(▲). Characteristics include a) the characteristic path length ($L$), b) the correlation between $c$ and $k$ ($c$-$k$), c) the slope of the degree correlation ($\upsilon$), d) the average clustering coefficient $c_{ave}$ and e) the degree average $k_{ave}$.**

**Table 5-4: Topological characteristics for the interaction networks of the Panmictic EA, cellular GA, and SOTEA. SOTEA networks are averages taken over all settings for $K_{Max}$ as described elsewhere. For comparison, common topological characteristics of several biological systems are also provided (taken from [196] and references therein). Characteristics include the characteristic path length *L*, the degree average $k_{ave}$, the linkage distribution (k dist.), the average clustering coefficient $c_{ave}$, correlation between *c* and *k* (*c-k*), and degree correlations ($k$-$k_{NN}$). For the *k* distribution, $\gamma$ refers to the exponent for *k* distributions that fit a power law. Two values for $\gamma$ are given for the metabolic network and refer to the in/out-degree exponents (due to this being a directed network). Results for degree correlations are given as the slope $\upsilon$ of $k_{NN}$ vs *k*. *N* is the population size, and R is a correlation coefficient for the stated proportionalities.**

| System | N | L | $k_{ave}$ | k dist. | $c_{ave}$ ($c_{rand}$) | c-k | k-$k_{NN}$ |
|---|---|---|---|---|---|---|---|
| Panmictic EA | 50 | L = 1 | $k_{ave}$ = N-1 | k = N-1 | 1 (1) | no | no |
| cellular GA | 50 | L ~ N | $k_{ave}$ = 2 | k = 2 | 0 (0.04) | no | no |
| SOTEA | 50 | 5.97 | 3.6 | Poisson | 0.687 (0.07) | c = -4.75k | $\upsilon$ = 11.8 |
| Complex Networks | Large | L ~ log N | $k_{ave}$ << N | Power Law, $2<\gamma<3$ (Scale Free Network) | $c_{ave}$ >>$c_{rand}$ | Power Law (Hierarchical) | either $\upsilon > 0$ or $\upsilon < 0$ |
| Protein | 2,115 | 2.12 | 6.80 | Power Law, $\gamma = 2.4$ | 0.07 (0.003) | Power Law | $\upsilon < 0$ |
| Metabolic | 778 | 7.40 | 3.2 | Power Law, $\gamma$ = 2.2/2.1 | 0.7 (0.004) | Power Law | $\upsilon < 0$ |

#### *5.4.3.3.1 Topological Properties of SOTEA*

This section briefly comments on some of the topological properties of SOTEA and the relevance of these properties to algorithm behavior.

**Characteristic Path Length *L*:** The total distance genetic material must travel across the network is always small as indicated by small *L* suggesting there is always a potential for any two nodes to influence each other over a relatively small time scale. However, it should be mentioned that a small path length does not necessarily mean strong interactions occur between different regions of the network (as suggested below). Additional studies on the population dynamics of these systems are needed to verify the impact of small *L*.

**Clustering Coefficient:** The high value of the average clustering coefficient is potentially very important to population dynamics and algorithm behavior in general. For example, consider the impact of clustering on the random walks used for reproduction in SOTEA. Random walks starting from within highly clustered regions of the network are unlikely to travel outside the cluster (due to high levels of interconnectivity among neighbors). Such a topological feature may act to reduce the amount of communication between clusters, a behavior reminiscent of the island model GA.

**Degree Average:** The low value for $k_{ave}$ suggests the SOTEA network maintains a sparsely connected architecture with high levels of locality similar to that of the cellular GA.

**Degree distribution:** $k$ approximates a Poisson distribution which is not similar to the fat tailed distributions observed in complex systems or the distributions observed in the first SOTEA algorithm developed in Section 5.2. The distributions results suggest relatively little heterogeneity in $k$ is present such that the level of locality is roughly uniform within the system.

Previous studies, as reviewed in [197], have indicated that placing upper bounds on $k$ can result in strong deviations from a power law. This SOTEA model introduces very tight constraints on the values of $k$ (e.g. upper and lower bounds, quadratic set point) so $k$ distribution results should not be surprising. Future work will try to allow for higher levels of connection heterogeneity in the system, which is expected to become increasingly relevant to system behavior as larger population sizes are considered.

**Degree-Degree correlations:** The assortative character of the SOTEA networks ($\upsilon > 0$) suggests high fitness nodes are driven to preferentially interact with other high fitness nodes. Such a population topology might provide a natural robustness to the search process allowing for the coexistence of explorative and exploitive behaviors within a single system.

**Clustering-Degree correlations:** The linear relation between $c$ and $k$ suggests that some marginal levels of hierarchy exist within the network however its presence is unlikely to persist with larger population sizes. In the current SOTEA design, it is not clear what role (if any) that hierarchy would play in algorithm behavior however this could change if nodes were able to take on a diverse range of behaviors and actions.

#### *5.4.3.3.2 SOTEA Scaling*

It is also helpful to analyze networks visually to understand network structure. Figure 5-32 shows SOTEA networks after 400 generations of evolution with varying population sizes ($N$=50, 100, 200) and $K_{Max}$ = 7. Figure 5-33 shows the same conditions but with $K_{Max}$ = 5.

One very noticeable consequence of the SOTEA model is that many nodes are found in four neighborhood clusters and in particular, there appears to be a “kite” motif present in the network.[20] It is expected that this is in part due to the degree lower bound of $K_{Min} = 3$ in the SOTEA model.

In the network visualizations, node sizes are adjusted to reflect individual fitness with larger nodes representing individuals with better fitness. It was disappointing to see that higher fitness nodes did not clearly take network hub positions even though the network rewiring rules encourage high fitness nodes to acquire more connections and be less attracted to clusters. Also, one can notice that as population size increases, residual ring-like structures can still be observed in the network, even after 400 generations. This indicates that initial topological bias continues to impact the network structure over long periods of time for larger systems. It is suspected that this structural bias can significantly impact algorithm behavior which may be investigated in future work.

[20] A motif refers to an over-represented sub-graph within a network. In other words, there exists a structural pattern within the network that is repeated at a frequency that is unlikely to occur by chance alone.

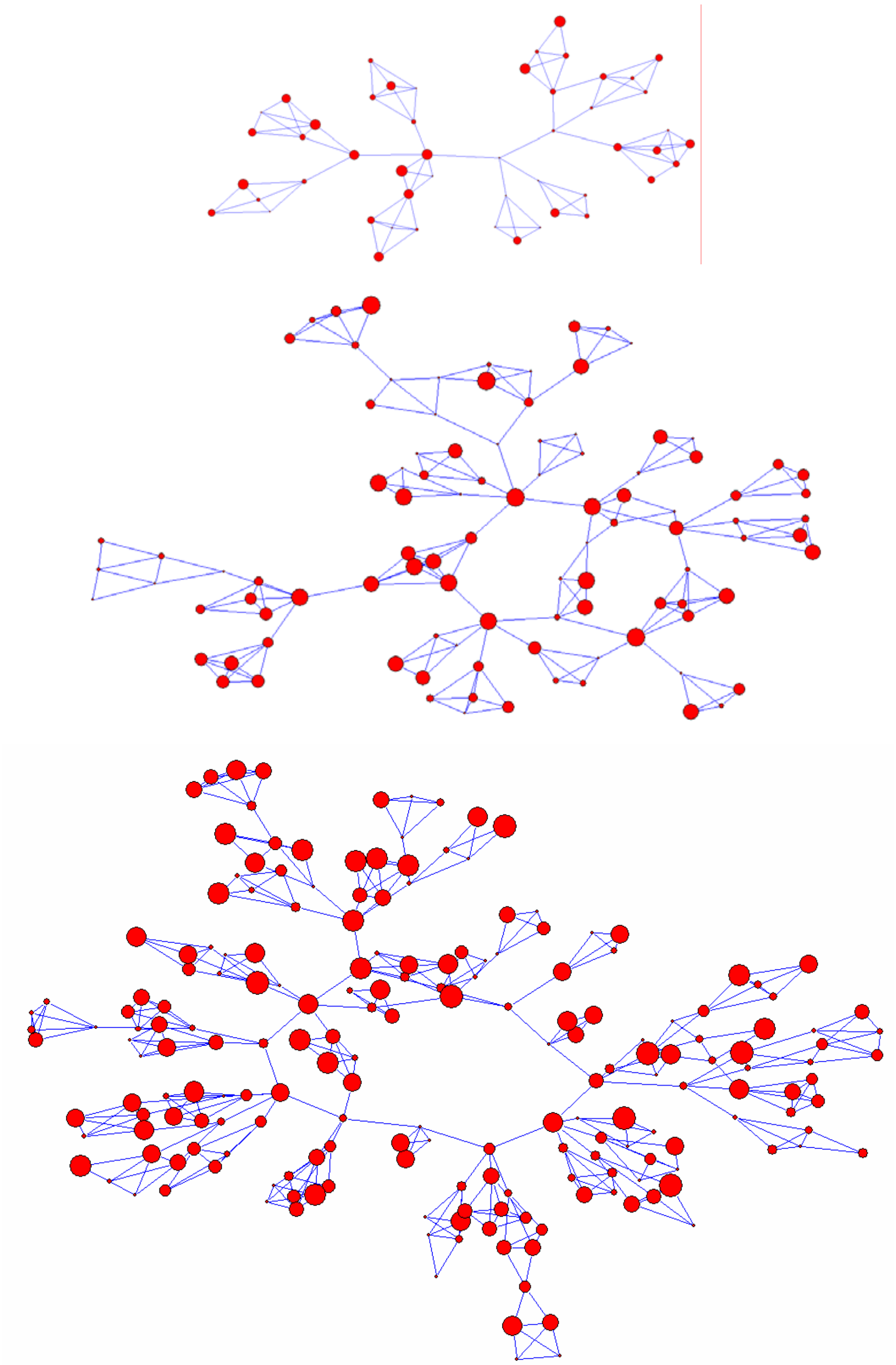

**Figure 5-32 SOTEA Network Visualizations with $K_{Max}$ = 7 for population sizes $N$ = 50, $N$ = 100, and $N$ = 200. Network visuals were created using Pajek Software.**

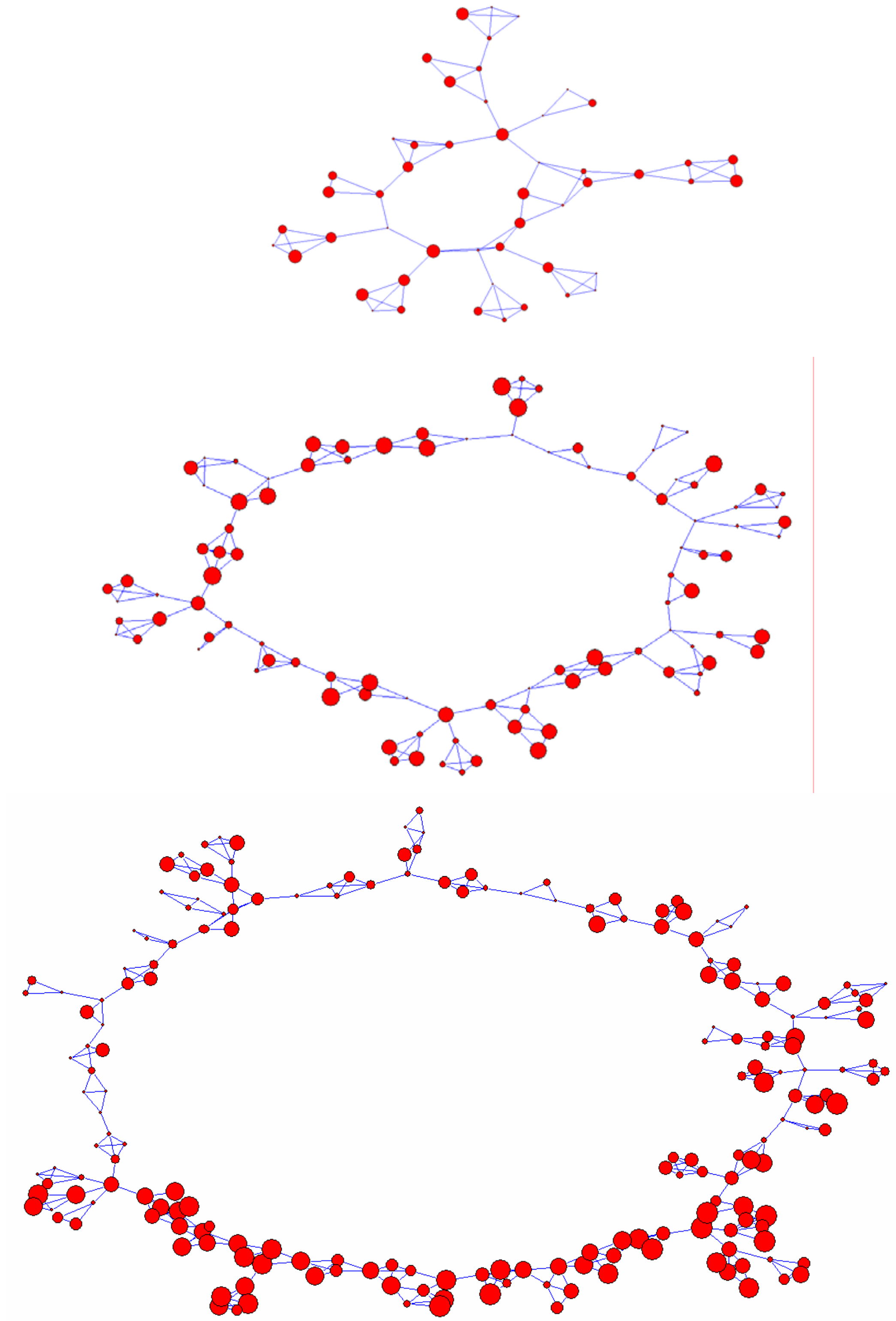

**Figure 5-33 SOTEA Network Visualizations with $K_{Max}$ = 5 for population sizes $N$ = 50, $N$ = 100, and $N$ = 200. Network visuals were created using Pajek Software.**

Another interesting but potentially worrisome attribute of this SOTEA model can be observed in the network with $N$=200 in Figure 5-33. Here one can see that, for the lower half of the network, the nodes are generally much larger indicating they are of relatively higher fitness than the upper section. Under such conditions, the use of a global fitness ranking to control $K_{Set}$ in the SOTEA algorithm could cause entire regions of the network to become highly connected while leaving the rest of the network with very little connectivity. The possibility of this situation occurring could be mitigated by replacing the global fitness measure with one that is locally defined. This would not impact selection pressure since selection is based on local pair-wise comparisons and not based on the magnitude of fitness values. Furthermore, a localized fitness measure would remove the only global information currently used in SOTEA which would make SOTEA more efficient in physically distributed implementations of the algorithm.

### 5.4.4 Discussion

**SOTEA Network Model:** The model for network dynamics used in this SOTEA algorithm was developed using several guiding principles. First, it was desired to have topological changes be driven by, and enacted on, local regions of the network. This not only occurs for many real-world complex systems, it also is a prerequisite for physically parallel implementations of the algorithm. This led to the use of network rewiring rules based on short random walks as well as node property values which are almost completely derived from local information (except for fitness ranking).

Second, it was recognized (in both SOTEA models) that for many complex systems, self-organization is at least partly driven by component fitness or attractive forces between system components. Clearly the concept of fitness also plays an important role in optimization. This made an individual's fitness a natural choice for coupling the structural dynamics of the network to the dynamics of the EA population.

**Distributed EA research:** A large amount of research efforts have been devoted to the study of distributed Evolutionary Algorithms. These efforts include the study of fine-grained (e.g. cellular grids), coarse-grained (e.g. island models), and hybrid structures (e.g. hierarchical). The highly modular topology of the second SOTEA model combined with short random walk interactions within the system could create fuzzy or partial islands within the system where interactions within a cluster are much more likely to occur compared to interactions between clusters. Quantifying the prevalence of this behavior could be accomplished by calculating the characteristic residence time of random walkers on local regions of the network using methods outlined in Section 2.3 in [225]. Assuming that clusters can become fairly isolated from other clusters, this would allow for a more nature-inspired approach to the integration of fine-grain and coarse-grain structures within an EA population (as opposed to explicitly defined hierarchical topologies).

### 5.4.5 Future Work

There are other issues which have not been addressed here and will be left to future work. One issue is that the network models are not directed so that information can flow in any direction across the network. This is often not the case for many biological systems due to thermodynamic law and other irreversible processes.

It would also be interesting to investigate whether the structural bias in the initial population changes the algorithm's performance sensitivity to initial conditions. It is speculated that the combination of genetic bias and structural bias (in the initial population) could offer an extended range of flexibility to the algorithm. For example, combining these two features could force interactions between certain initial genotypes to take place with a frequency that is much greater than would occur under other circumstances. In other words, this provides some control over which regions in solution space are able to initially interact. Combining this with a highly modular adaptive network like SOTEA could also allow for some control over the timing of future interactions between genetic material. In short, optimizing structural bias in SOTEA could provide some limited capacity to control not only which points in solution space are able to interact but also the timing in which these interactions occur. This might also provide a viable path for mitigating the effects of deleterious (e.g. deceptive) attractor basins within a particular fitness landscape.

Finally, it should be mentioned that network models currently exist (e.g. see [190]) which can acquire the structural characteristics of complex systems without the presence of driving forcing that encourage these characteristics to emerge (e.g. the weighted clustering coefficient in SOTEA). SOTEA would be significantly improved if it could allow for the emergence of important topological properties using a simplified model while still exhibiting robust performance on optimization problems.

### 5.4.6 Conclusions

The Self-Organizing Topology Evolutionary Algorithm or SOTEA is a distributed EA containing a population structure that coevolves with EA population dynamics. With the population defined on a network, rules are used to modify the network topology based on the current states of the population.

**SOTEA Network Model:** The second SOTEA model presented in this chapter was designed with an emphasis on locality of network information and locality in network dynamics. Network dynamics were driven by i) an adaptive connectivity where higher fitness individuals were encouraged to obtain higher levels of connectivity and ii) an adaptive definition of community which attempted to encourage high levels of clustering in nodes of low fitness.

To this author's knowledge, this model was unique among network models in that it considered two driving forces instead of just one. Also, this SOTEA model was the first network model which evolves due to a dynamic state value of the nodes (i.e. fitness). The dynamics of node fitness were a natural consequence of the dynamics of the EA population.

**Topological Analysis:** The second SOTEA model allows for a self-organization of population network topology resulting in a large degree of clustering, small characteristic path length, and correlations between the clustering coefficient and a node's degree. Each of these characteristics are similar to what is observed in complex biological systems.

However, a number of topological properties observed in complex systems were not attained in the current model including properties achieved in the first SOTEA algorithm such as a fat-tailed degree distribution. Future work will attempt to address these shortcomings as well as attempt to create a clearer framework for the integration of multiple driving forces in an adaptive network. The network rewiring rules were also developed in a somewhat ad hoc fashion and future work will look to develop a more intuitive framework for structural dynamics.

**Performance:** A number of engineering design problems and artificial test functions were selected to test the effectiveness of the new SOTEA algorithm against another distributed design, the cellular GA. Results indicate the SOTEA algorithm was able to provide improved performance and more consistent results compared with the cGA. Both of the distributed Evolutionary Algorithms strongly outperformed a suite of eight other Evolutionary Algorithms tested.

# Chapter 6 Summary of Findings

The primary goal of this thesis was to improve the performance and general robustness of Evolutionary Algorithms using principles inspired by nature. Contributions from this thesis include: i) reducing the practical difficulties associated with designing an Evolutionary Algorithm by developing a more effective procedure for adapting EA design parameters, ii) determining the aspects of EA design which impact population dynamics and enable parallel search behavior, and iii) mimicking the structural self-organization in complex biological systems as a means to obtain advanced behaviors and improved performance in distributed Evolutionary Algorithms.

**Designing an Effective Adaptive Process:** Chapter 3 proposed mechanisms for making a more effective adaptive process for supervisory control of EA design parameters. To make the adaptive process more effective, two modifications were proposed. The first modification was to use an empirical measure of an individual's importance on future population dynamics instead of estimating its importance through the use of fitness measurements. The second modification was to reduce the influence of non-informative interactions between the adaptive system and its environment. This was deemed to be particularly relevant due to evidence presented in this chapter that non-informative measurements dominated the data received by the adaptive system. This second modification to the adaptive system was accomplished by using statistical arguments that quantified the importance of measurements.

Not only did the new adaptive method outperform all other methods on the majority of problems tested, it was also found to be much more robust compared to the other adaptive methods. In particular, it's performance was not strongly sensitive to the class of problems (artificial test functions vs. engineering design problems) that it was tested on.

**The impact of EA design on population dynamics:** Chapter 4 started with the goal of understanding how EA design factors can influence EA population dynamics. It was concluded that the probability distribution of an individual's impact on population dynamics fits a power law regardless of almost all experimental conditions. This result indicates that a small number of individuals are capable of driving EA population dynamics while most other individuals have only a small impact. The existence of power law deviations in the probability of large impact sizes (i.e. large ETV) was seen as an indicator that such systems were not capable of being driven by single individuals and instead were able to exhibit higher levels of parallel search behavior.

The most significant factor that enables parallel search behavior was the topology of the EA population. As the population topology came closer to approximating a Panmictic population, the system became increasingly driven by only a select few individuals. On the other hand, as spatial restrictions were increased in the population, single individuals were no longer capable of dominating the dynamics of the entire population. It is speculated that this could account for the strong and robust performance gains that have been repeatedly observed in distributed Evolutionary Algorithms over the years. Another factor which was found to create a smaller degree of parallel search behavior was the introduction of completely randomized new individuals into an EA population.

**The Self-Organization of Interaction Networks for EA population topology:** The aim of Chapter 5 was to create EA populations with topological features that are similar to those observed in complex biological systems. It was speculated that mimicking this aspect of nature could provide additional improvements to algorithm behavior compared to those already observed in distributed EA designs. This chapter has demonstrated that the self-organization of population topology can induce a number of interesting new behaviors in an EA and has the potential to significantly improve its performance on challenging optimization problems. It is hoped that this work will inspire others to investigate the use of network models for the self-organization of population structure and that these research efforts will help to narrow the gap between EA and natural evolutionary processes.

# APPENDIX A TEST FUNCTION DEFINITIONS

## A.1 Artificial Test Functions

**Table A-1 Artificial Test Function Characteristics Table. Epi: Epistasis or Tight Linkage (i.e. Non-Separable), Con = Continuous, *n* = problem dimensionality (* indicates *n* is a parameter of the problem), Ref. = reference to problem description used, MM = multimodal fitness landscape, Params = parameters of the problem.**

| Name | Epi | Con | n | MM | Ref. | Params. |
|---|---|---|---|---|---|---|
| MTTP | Yes | No | * | Yes | [135] | $n = 200$ |
| ECC | Yes | No | * | Yes | [135] | $n = M*N$<br>$M = 24$<br>$N = 12$ |
| MMDP | Yes | No | * | Yes | [226] | $n = 6k$<br>$k$=20 |
| Frequency Modulation | Yes | Yes | 6 | Yes | [56] | |
| NK Landscape | varies | No | * | Yes | [81] | $n = N$<br>$N$, $K$ varies |
| Rosenbrock | Yes | Yes | * | No | [56] | $n = 2$ |
| Rastrigin | No | Yes | * | Yes | [56] | $n = 20$ |
| Schwefel | No | Yes | * | Yes | [227] | $n = 20$ |
| Griewangk | Yes | Yes | * | Yes | [56] | $n = 10$ |
| Bohachevsky's | Yes | Yes | 2 | Yes | [56] | |
| Watson's | No | Yes | 5 | No | [56] | |
| Colville's | Yes | Yes | 4 | Yes | [56] | |
| System of linear equations | No | Yes | 10 | No | [56] | |
| Ackley's Function | No | Yes | 25 | Yes | [56] | |
| Neumaier's Function #2 | No | Yes | 4 | Yes | [228] | |
| Hyper Ellipsoid | No | Yes | * | No | [83] | $n = 30$ |

### A.1.1 Minimum Tardy Task Problem (MTTP)

The Minimum Tardy Task Problem (MTTP) [229] is a task scheduling problem where the objective is to execute as many tasks as possible within the time constraints and precedence relations. Each task $T_j$, $j \in \{1,2,\ldots,n\}$ has a time length $L_j$ (time needed to execute task), a deadline $D_j$ (before which the task must be completed), and a weight $W_j$ (indicating the penalty cost from not completing the task). $L$, $D$, and $W$ all take on positive integer values and the scheduling tasks are executed in sequential order. The scheduling problem is then to find a subset $S$ of $T$ which executes within the allocated time and minimizes the sum of all penalties for tasks that were not completed. A penalty term $P$ is added for infeasible solutions involving tasks that are started but not finished within the allocated time. $P$ is given as the sum of all task weights thereby ensuring that infeasible solutions are assigned a worse fitness than feasible solutions.

$$Min\ F(x) = P + \sum_{j\varepsilon T-S} W_j$$

$$P = \sum_{j\varepsilon T} W_j$$

$$x_i \in (0,1)$$

Candidate schedules $S$ are represented as a binary vector $x$ indicating which tasks are to be executed. A candidate schedule is therefore the set of all tasks where $x_j$=1. As enforced by the problem generation method, tasks are ordered in the candidate schedule by their deadlines and are also executed in that order. (This is easy to see since $D_j$ as defined below is a monotonically increasing function of $j$). The size of the problem $n$ can be controlled using the problem generation method specified below (and taken from [230]) where $n$ is a multiple of 5.

**Problem Generation Method**

$$L_j = \begin{cases} L_i & if \quad j > 5 \\ 3j & else \end{cases}$$

$$D_j = \begin{cases} D_i + 24m & if \quad j > 5 \\ 5j & else \end{cases}$$

$$W_j = \begin{cases} W_i(m+1) & if \quad j > 5 \\ 60 & if \quad j = 1 \\ 40 & if \quad j = 2 \\ 7 & if \quad j = 3 \\ 3 & if \quad j = 4 \\ 50 & if \quad j = 5 \end{cases}$$

$i$ = ($j$ Mod 5) + 1,  $m$ = $j$ / 5,  $j$ =1,…,$n$

Optimal Solution = 2$n$

**MTTP as implemented in Code**

```
Function F(x)                       'x_j j∈{1,2,…,n)  ' parameters to optimize

        T^0 = 0                     'start time for a task
        For j = 1 To n
          If x_j = 1 Then           'Is task being executed?
            If (T^0 + L_j)< D_j  Then  'can we complete task in time?
              T^0 = T^0 + L_j
            Else
              Infeasible = True
              Cost = Cost + W_j
            End If
          Else
            Cost = Cost + W_j
          End If
        Next i
      P = sum(W_j)

      If Infeasible Then  Cost = Cost + P
      F(x) = 2n – Cost              'Fopt = 0
```

### A.1.2 Error Correcting Code Problem (ECC)

The Error Correcting Code Problem (ECC) [231] is a problem where we try to minimize the error in reading coded messages (error due to input noise) by maximizing the distance between code words in the code parameter space. Given a set of $M$ binary code words, each of length $N$, the objective is to maximize the Hamming distance $d$ between any pair of code words.

$$Max\ F(X) = \frac{1}{\sum_{i=1}^{M} \sum_{j=1, j\neq i}^{M} \frac{1}{d_{ij}^2}}$$

$$x_i \in (0,1), \quad i \in \{1,...,n\}$$

$$n = M*N$$

Optimal (for $M$=24, $N$=12, $F$ = 0.067416)

### A.1.3 Massively Multimodal Deceptive Problem (MMDP)

$$Min\ F(X) = \sum_{i=1}^{k} y_{j+1} - k$$

$$j = \sum_{m=1}^{6} x_t, \quad t = 6(i-1) + m$$

$Y = (y_1, y_2, y_3, y_4, y_5, y_6) = (1, 0, 0.360384, 0.640576, 0.360384, 0, 1)$

$x_i \in (0,1), \quad i \in \{1,...,n\}$

$n = 6k$

### A.1.4 Frequency Modulation

The problem is to specify six parameters of the frequency modulation sound model represented by $y(t)$.

Original Parameters: $X = (x_1, x_2, x_3, x_4, x_5, x_6) = (a_1, w_1, a_2, w_2, a_3, w_3)$

$$Min\ F(X) = \sum_{t=0}^{100} \left(y(X,t) - y(X_0,t)\right)^2$$

$$y(X,t) = x_1 \sin\left(x_2 t\theta + x_3 \sin\left(x_4 t\theta + x_5 \sin\left(x_6 t\theta\right)\right)\right)$$

$\theta = \left(\frac{2\pi}{100}\right)$ $X_0 = (1.0, 5.0, 1.5, 4.8, 2.0, 4.9)$

$-6.4 \le x_i \le 6.35, \quad i \in \{1,...,n\}$

$x_i \in \mathbb{R}, \quad n = 6$

Optimal $(F, x_1, x_2, x_3, x_4, x_5, x_6) = (0, 1.0, 5.0, 1.5, 4.8, 2.0, 4.9)$

### A.1.5 Quadratic Function

$$Min\ F(X) = x_1^2 + x_2^2 + x_3^2$$

$-5.12 \le x_i \le 5.12, \quad i \in \{1,...,n\}$

$x_i \in \mathbb{R}, \quad n = 3$

Optimal $(F, x_1, x_2, x_3) = (0, 0, 0, 0)$

### A.1.6 Generalized Rosenbrock's Function

$$Min\ F(X) = \sum_{i=1}^{n-1} \left(100\left(x_{i+1} - x_i^2\right)^2 + \left(x_i - 1\right)^2\right)$$

$$-2 \le x_i \le 2, \quad i \in \{1,...,n\}$$

$x_i \in \mathbb{R}, \quad n = 2$

Optimal $(F, x_1, x_2) = (0, 1.0, 1.0)$

### A.1.7 Rastrigin's Function

$$Min\ F(X) = 10n + \sum_{i=1}^{n} \left(x_i^2 - \cos(2\pi x_i)\right)$$

$$-5.12 \le x_i \le 5.12, \quad i \in \{1,...,n\}$$

$x_i \in \mathbb{R}, \quad n = 20$

Optimal $(F, x_i, ...x_n) = (0, 0,\ldots,0)$

### A.1.8 Schwefel's Function

$$Min\ F(X) = \sum_{i=1}^{n} -x(i)Sin\left(Abs\left(x(i)\right)^2\right)$$

$$-500 \le x_i \le 500, \quad i \in \{1,...,n\}$$

$x_i \in \mathbb{R}, \quad n = 20$

Optimal $(F, x_i, ...x_n) = (0, 0,\ldots,0)$

### A.1.9 Griewangk's Function

$$Min\ F(X) = \frac{1}{4000}\sum_{i=1}^{n} x_i^2 - \prod_{i=1}^{n} \cos\left(\frac{x_i}{\sqrt{i}}\right) + 1$$

$$-600 \le x_i \le 600, \quad i \in \{1,...,n\}$$

$x_i \in \mathbb{R}, \quad n = 10$

Optimal $(F, x_i, ...x_n) = (0, 0,\ldots,0)$

### A.1.10 Bohachevsky's Function

$$Min\ F(X) = x_1^2 + 2x_2^2 - 0.3\cos(3\pi x_1)\cos(4\pi x_2) + 0.3$$

$$-50 \le x_i \le 50, \quad i \in \{1,...,n\}$$

$x_i \in \mathbb{R}, \quad n = 2$

Optimal ($F, x_i, x_2$) = (0, 0, 0)

### A.1.11 Watson's Function

$$Min\ F(X) = \sum_{i=1}^{30}\left(\sum_{j=1}^{5}\left(ja_i^{j-1}x_{j+1}\right) - \left[\sum_{j=1}^{6} a_i^{j-1}x_j\right]^2 - 1\right)^2 + x_1^2$$

$$a_i = \frac{i-1}{29}$$

$$-2 \le x_i \le 2, \quad i \in \{1,...,n\}$$

$x_i \in \mathbb{R}, \quad n = 6$

Optimal ($F, x_i, ..., x_n$) = (2.288E-3, -0.0158, 1.012, -0.02329, 1.260, -1.513, 0.09928)

### A.1.12 Colville's Function

$$Min\ F(X) = 100\left(x_2 - x_1^2\right)^2 + \left(1 - x_1\right)^2 + 90\left(x_4 - x_3^2\right)^2 + \left(1 - x_3\right)^2 +$$
$$10.1\left(\left(x_2 - 1\right)^2 + \left(x_4 - 1\right)^2\right) + 19.8\left(x_2 - 1\right)\left(x_4 - 1\right)$$

$$-10 \le x_i \le 10, \quad i \in \{1,...,n\}$$

$x_i \in \mathbb{R}, \quad n = 4$

Optimal ($F, x_i, ..., x_n$) = (0, 1,…, 1)

### A.1.13 System of linear equations

$$Min\ F(X) = \sum_{i=1}^{n} \sum_{j=1}^{n} \left(a_{ij} x_j\right) - b_j$$

$Ax = b$ is given by:

$$\begin{vmatrix} 5 & 4 & 5 & 2 & 9 & 5 & 4 & 2 & 3 & 1 \\ 9 & 7 & 1 & 1 & 7 & 2 & 2 & 6 & 6 & 9 \\ 3 & 1 & 8 & 6 & 9 & 7 & 4 & 2 & 1 & 6 \\ 8 & 3 & 7 & 3 & 7 & 5 & 3 & 9 & 9 & 5 \\ 9 & 5 & 1 & 6 & 3 & 4 & 2 & 3 & 3 & 9 \\ 1 & 2 & 3 & 1 & 7 & 6 & 6 & 3 & 3 & 3 \\ 1 & 5 & 7 & 8 & 1 & 4 & 7 & 8 & 4 & 8 \\ 9 & 3 & 8 & 6 & 3 & 4 & 7 & 1 & 8 & 1 \\ 8 & 2 & 8 & 5 & 3 & 8 & 7 & 2 & 7 & 5 \\ 2 & 1 & 2 & 2 & 9 & 8 & 7 & 4 & 4 & 1 \end{vmatrix} \begin{vmatrix} 1 \\ 1 \\ 1 \\ 1 \\ 1 \\ 1 \\ 1 \\ 1 \\ 1 \\ 1 \end{vmatrix} = \begin{vmatrix} 40 \\ 50 \\ 47 \\ 59 \\ 45 \\ 35 \\ 53 \\ 50 \\ 55 \\ 40 \end{vmatrix}$$

$-9 \leq x_i \leq 9, \quad i \in \{1,...,n\}$

$x_i \in \mathbb{R}, \quad n = 10$

Optimal $(F, x_i, ..., x_n) = (0, 1, \ldots, 1)$

### A.1.14 Ackley's Function

$$Min\ F(x) = -20 \exp\left(-0.2\sqrt{n^{-1}\sum_{i=1}^{n} x_i^2}\right) - \exp\left(n^{-1}\sum_{i=1}^{n}\cos(2\pi x_i)\right) + 20$$

$-32.768 \leq x_i \leq 32.768, \quad i \in \{1,...,n\}$

$x_i \in \mathbb{R}, \quad n = 25$

Optimal $(F, x_i, ..., x_n) = (0, 0, \ldots, 0)$

### A.1.15 Neumaier's Function #2

**Original definition**

$$Min\ \ F(x) = \sum_{k=1}^{n} \left(b_k - \sum_{i=1}^{n} x_i^k\right)^2$$

b = (8, 18, 44, 114)

$0 \leq x_i \leq n, \quad i \in \{1,...,n\}$

$x_i \in \mathbb{R}, \quad n = 4$

Optimal $(F, x_1, x_2, x_3, x_4) = (0, 1, 2, 2, 3)$

**Modified definition (used in all experiments)**

$$Min\ \ F(x) = \sum_{k=1}^{n} (b_k - \alpha_k)^2$$

$$\alpha_k = \sum_{j=1}^{k} \left( \sum_{i=1}^{n} x_i^j \right)$$

b = (8, 18, 44, 114)

$0 \le x_i \le n, \quad i \in \{1,...,n\}$

$x_i \in \mathbb{R}, \quad n = 4$

Optimal unknown

### A.1.16 Hyper Ellipsoid

$$Min\ F(X) = \sum_{i=1}^{n} x_i^2 (i^2)$$

$-1 \le x_i \le 1, \quad i \in \{1,...,n\}$

$x_i \in \mathbb{R}, \quad n = 30$

Optimal ($F, x_1, ..., x_n$) = (0, 0,…, 0)

## A.2 Engineering Design Test Problems

### A.2.1 Turbine Power Plant

$$Min\ F(X) = x_3 f_1 + x_4 g_1$$

$$g_1(X) = 0.8008 + 0.2031x_2 + 0.000916x_2^2$$

$$g_2(X) = 0.7266 + 0.2256x_2 + 0.000778x_2^2$$

$$f_1(X) = 1.4609 + 0.15186x_1 + 0.00145x_1^2$$

$$f_2(X) = 1.5742 + 0.1631x_1 + 0.001358x_1^2$$

**Subject to:**

$BFG = (1 - x_3) f_2 + (1 - x_4) g_2 \le 10.0$

(18, 14, 0, 0) $\le$ ($x_1, x_2, x_3, x_4$) $\le$ (30, 25, 1, 1)

$x_i \in \mathbb{R}, \quad n = 4$

Optimal Solution ($F, x_1, x_2, x_3, x_4$) = (3.05, 30, 20, 0, 0.58)

### A.2.2 Alkylation Process

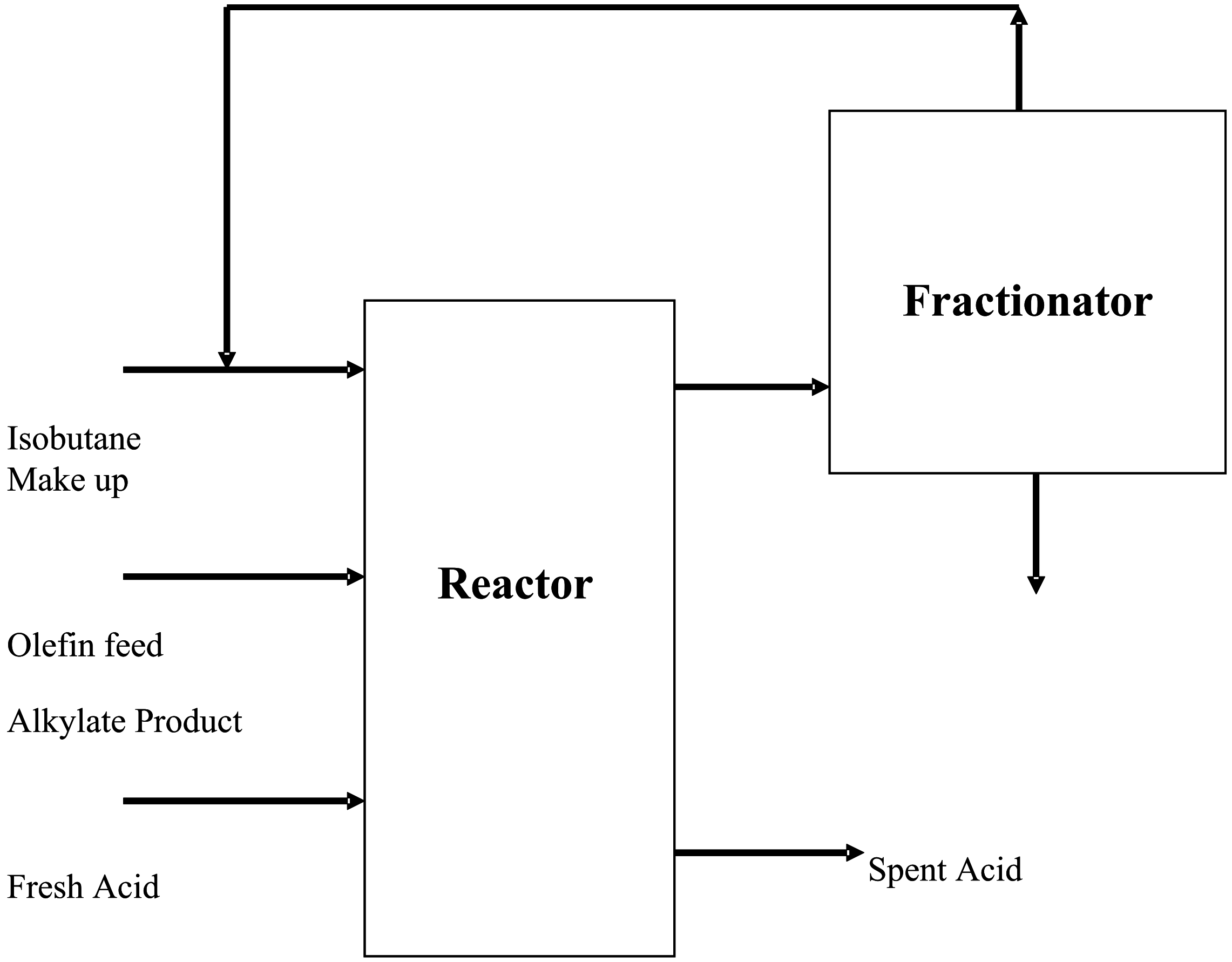

**Figure A-1 Simplified diagram of an alkylation process (recreated from [224])**

The alkylation process design problem, originally defined in [232], has the goal of improving the octane number of an olefin feed stream through a reaction involving isobutene and acid. The reaction product stream is distilled with the lighter hydrocarbon fraction recycled back to the reactor. The objective function considers maximizing alkylate production minus the material (ie feed stream) and operating (ie recycle) costs. Design parameters all take on continuous values and include the olefin feed rate $x_1$ (barrels/day), acid addition rate $x_2$ (thousands of pounds/day), alkylate yield $x_3$ (barrels/day), acid strength $x_4$ (wt. %), motor octane number $x_5$, external isobutene to olefin ratio $x_6$, and F-4 performance number $x_7$.

$$Max\ F(X) = 1.715x_1 + 0.035x_1x_6 + 4.0565x_3 + 10.0x_2 - 0.063x_3x_5$$

**Subject to:**

$$g_1(X) = 0.0059553571x_6^2x_1 + 0.88392857x_3 - 0.1175625x_6x_1 - x_1 \le 0$$
$$g_2(X) = 1.1088x_1 + 0.1303533x_1x_6 - 0.0066033x_1x_6^2 - x_3 \le 0$$
$$g_3(X) = 6.66173269x_6^2 + 172.39878x_5 - 56.596669x_4 - 191.20592x_6 - 10000 \le 0$$
$$g_4(X) = 1.08702x_6 + 0.32175x_4 - 0.03762x_6^2 - x_5 + 56.85075 \le 0$$
$$g_5(X) = 0.006198x_7x_4x_3 + 2462.3121x_2 - 25.125634x_2x_4 - x_3x_4 \le 0$$
$$g_6(X) = 161.18996x_4x_3 + 5000x_2x_4 - 489510x_2 - x_3x_4x_7 \le 0$$
$$g_7(X) = 0.33x_7 - x_5 + 44.333333 \le 0$$
$$g_8(X) = 0.022556x_5 - 0.007595x_7 - 1.0 \le 0$$
$$g_9(X) = 0.00061x_3 - 0.0005x_1 - 1.0 \le 0$$
$$g_{10}(X) = 0.819672x_1 - x_3 + 0.819672 \le 0$$
$$g_{11}(X) = 24500.0x_2 - 250.0x_2x_4 - x_3x_4 \le 0$$
$$g_{12}(X) = 1020.4082x_4x_2 + 1.2244898x_3x_4 - 100000x_2 \le 0$$
$$g_{13}(X) = 6.25x_1x_6 + 6.25x_1 - 7.625x_3 - 100000 \le 0$$
$$g_{14}(X) = 1.22x_3 - x_6x_1 - x_1 + 1.0 \le 0$$

(1500, 1, 3000, 85, 90, 3, 145) $\le$ ($x_1, x_2, x_3, x_4, x_5, x_6, x_7$) $\le$ (2000, 120, 3500, 93, 95, 12, 162)

$x_i \in \mathbb{R}$, $n = 7$

Optimal Solution ($F, x_1, x_2, x_3, x_4, x_5, x_6, x_7$) = (1772.77, 1698.18, 53.66, 3031.3, 90.11, 95, 10.5, 153.53)

### A.2.3 Heat Exchanger Network Design

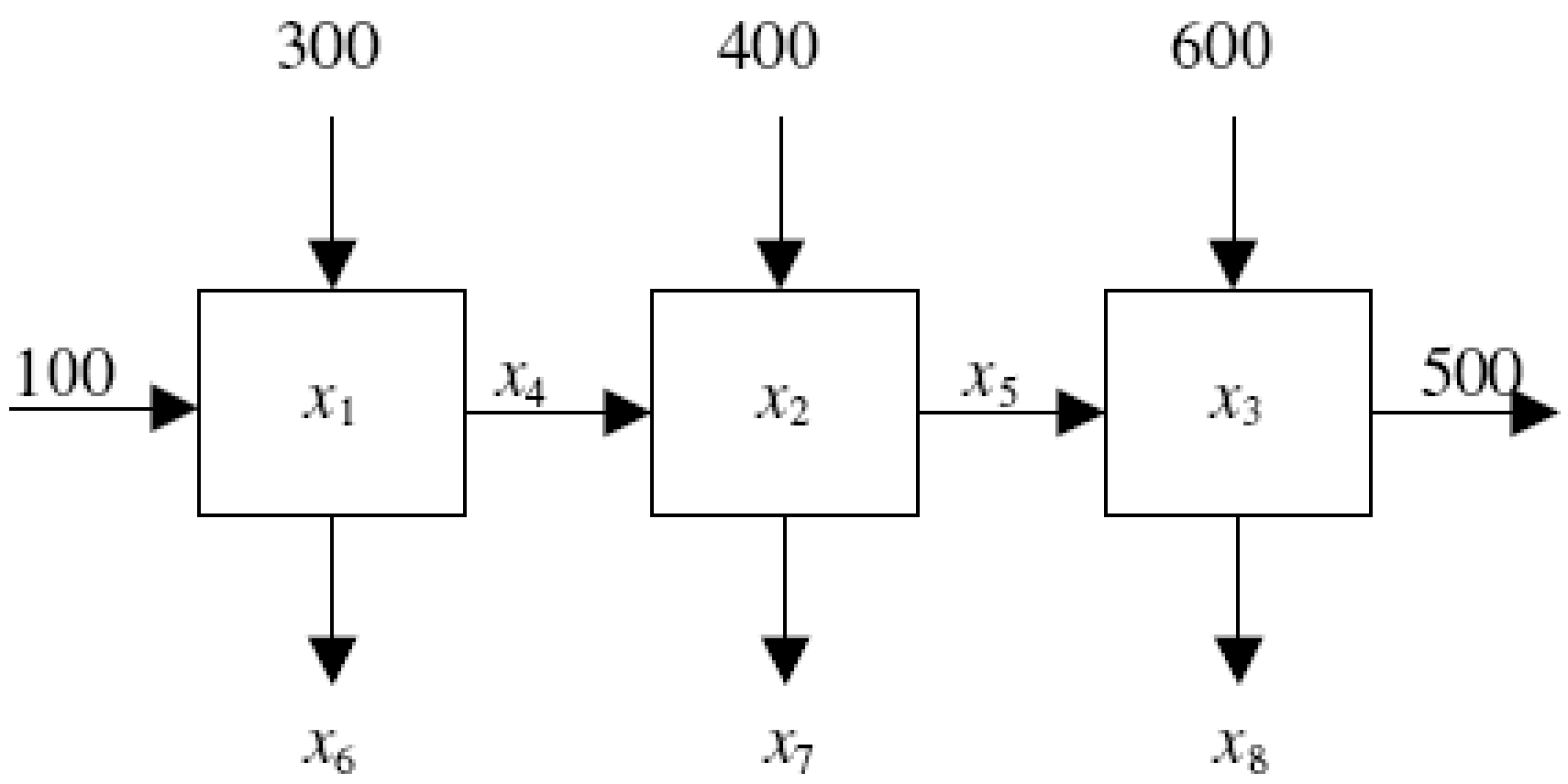

**Figure A-2 Diagram of the Heat Exchanger Network Design Problem involving 1 cold stream that exchanges heat with three hot streams. Parameters to optimize include heat exchange areas ($x_1$, $x_2$, $x_3$) and stream temperatures ($x_4$, $x_5$, $x_6$, $x_7$, $x_8$).**

The Heat Exchanger Network design problem, originally defined by [233], has the goal of minimizing the total heat exchange surface area for a network consisting of one cold stream and three hot streams. As shown in Figure A-2, there are eight design parameters consisting of the heat exchanger areas ($x_1$, $x_2$, $x_3$), intermediate cold stream temperatures ($x_4$, $x_5$) and hot stream outlet temperatures ($x_6$, $x_7$, $x_8$). The problem is presented below in a reformulated form taken from [234] where a variable reduction method has been used to eliminate equality constraints.

$$Min\ F(X) = x_1 + x_2 + x_3$$

**Subject to:**

$$g_1(X) = 100x_1 - x_1(400 - x_4) + 833.33252x_4 - 83333.333 \leq 0$$
$$g_2(X) = x_2x_4 - x_2(400 - x_5 + x_4) - 1250x_4 + 1250x_5 \leq 0$$
$$g_3(X) = x_3x_5 - x_3(100 + x_5) - 2500x_5 + 1250000 \leq 0$$

$(100, 1000, 1000, 10, 10) \leq (x_1, x_2, x_3, x_4, x_5) \leq (10000, 10000, 10000, 1000, 1000)$
$x_i \in \mathbb{R}$, $n = 5$

Optimal Solution ($F$, $x_1$, $x_2$, $x_3$, $x_4$, $x_5$) = (7049.25, 579.19, 1360.13, 5109.92, 182.01, 295.60,)

*remaining parameters are calculated from equality constraints. Their optimal values are: ($x_6$, $x_7$, $x_8$) = (217.9, 286.40, 395.60)*

### A.2.4 Pressure Vessel

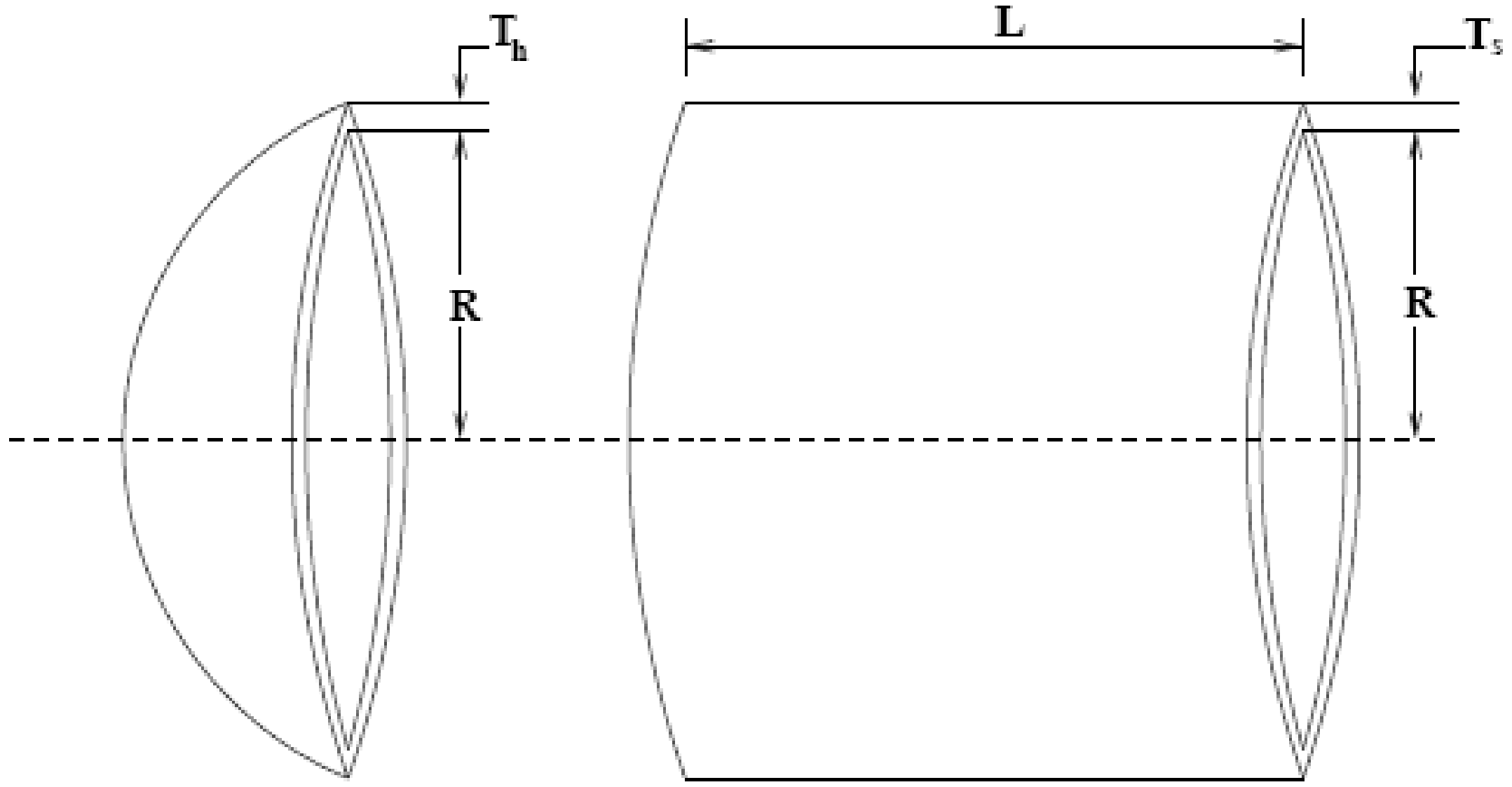

**Figure A-3 Pressure Vessel Drawing. Parameters of the problem include the thickness of the shell $T_s$, the thickness of the head $T_h$, the inner radius of the vessel $R$ and the length of the cylindrical section of the vessel $L$. This figure is taken out of [221] and is reprinted with permission from IEEE (© 1999 IEEE).**

The pressure vessel design problem, originally defined by [235], has the goal of minimizing the cost of a pressure vessel as calculated based on material, forming and welding costs. The design is subject to dimensional constraints which are set to meet ASME standards for pressure vessels. As shown in Figure A-3, there are four design parameters to optimize consisting of the thickness of the shell $T_s$, the thickness of the head $T_h$, the inner radius $R$ and the length of the cylindrical section of the vessel $L$. $T_s$ and $T_h$ take on integer values indicating the number of rolled steel plates (where each steel plate is 0.0625 inches thick) and $R$ and $L$ are continuous variables.

Original Parameters: $X = (R, L, T_h, T_s) = (x_1, x_2, x_3, x_4)$

$$Min\ F(X) = 0.6224x_1x_2(0.0625x_3) + 1.7781x_1^2(0.0625x_4)$$
$$+ 3.1661x_2(0.0625x_3)^2 + 19.84x_1(0.0625x_3)^2$$

**Subject to:**

$$g_1(X) = -0.0625x_3 + 0.0193x_1 \le 0$$
$$g_2(X) = -0.0625x_4 + 0.00954x_1 \le 0$$
$$g_3(X) = -\pi x_1^2 x_2 - \tfrac{4}{3}\pi x_1^3 + 1{,}296{,}000 \le 0$$
$$g_4(X) = x_2 - 240 \le 0$$

$(1, 1, 1, 1) \le (x_1, x_2, x_3, x_4) \le (100, 400, 20, 20)$

$x_1, x_2 \in \mathbb{R}, \quad x_3, x_4 \in \mathbb{Z}, \quad n = 4$

Optimal Solution unknown

### A.2.5 Coello's Welded Beam Design

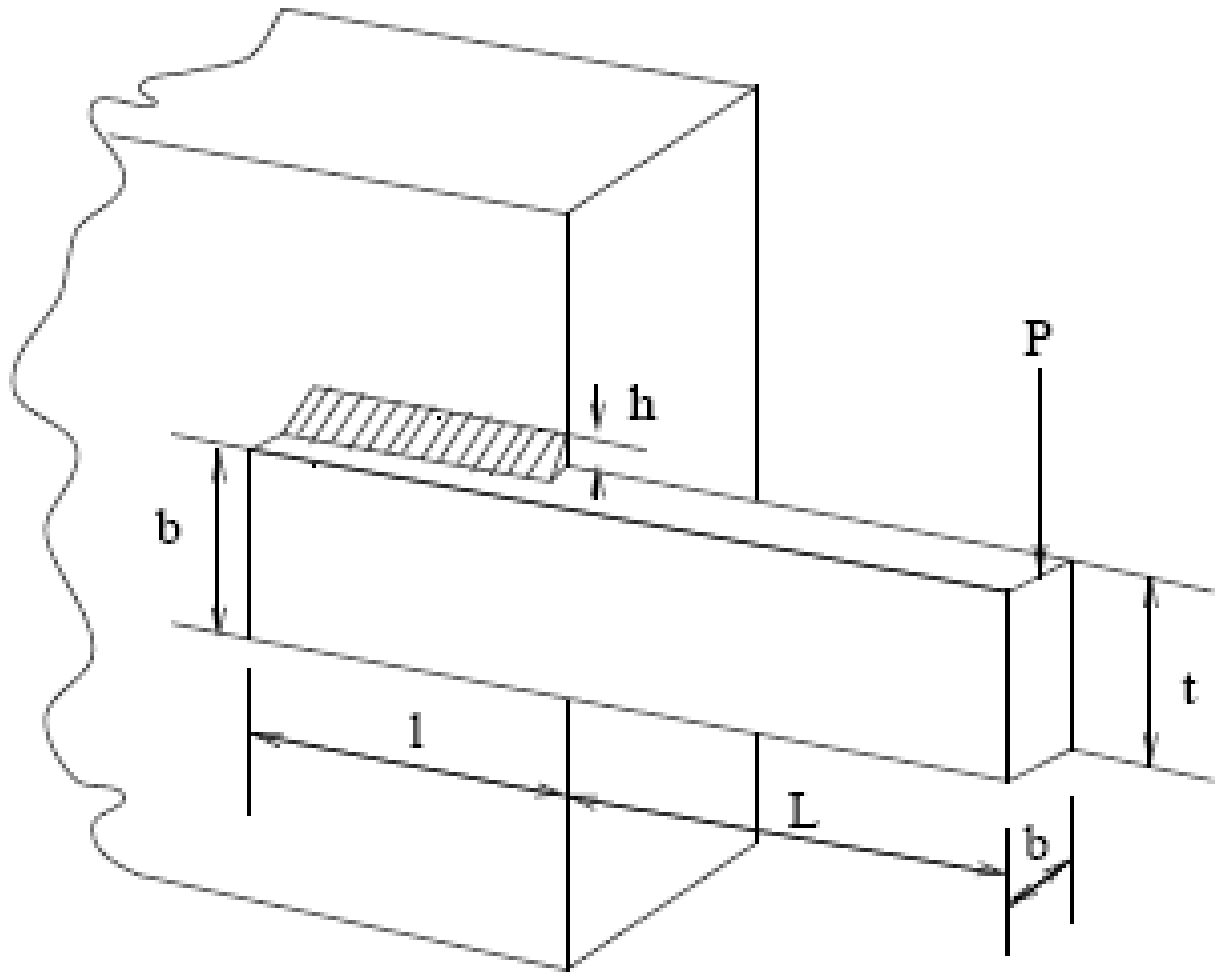

**Figure A-4: Diagram of a welded beam. The beam load is defined as *P* with all other parameters shown in the diagram defining dimensional measurements relevant to the problem. This figure is taken out of [221] and is reprinted with permission from IEEE (© 1999 IEEE).**

The Welded beam design problem has the goal of minimizing the cost of a weight bearing beam subject to constraints on shear stress $\tau$, bending stress $\sigma$, buckling load on the bar $P_c$, and dimensional constraints [221]. There are four design parameters to optimize consisting of the dimensional variables *h*, *l*, *t*, and *b* shown in Figure A-4.

The original formulation of the problem can be found in {Rekalitis, #336}. A change to the $P_c$ term in the problem formulation (stated below) appears to have occurred in [221] and some publications have implemented this new problem definition including the work presented in this thesis and others cited in Appendix C.

Original Parameters: $X = (h, l, t, b) = (x_1, x_2, x_3, x_4)$
$Min\ F(X) = 1.10471 x_1^2 x_2 + 0.04811 x_3 x_4 (14 + x_2)$

**Subject to:**
$g_1(X) = \tau(X) - \tau_{\max} \le 0$
$g_2(X) = \sigma(X) - \sigma_{\max} \le 0$
$g_3(X) = x_1 - x_4 \le 0$
$g_4(X) = 0.10471 x_1^2 + 0.04811 x_3 x_4 (14 + x_2) - 5 \le 0$
$g_5(X) = 0.125 - x_1 \le 0$
$g_6(X) = \delta(X) - \delta_{\max} \le 0$
$g_7(X) = P - P_c(X) \le 0$

**Where:**

$$\tau(X) = \sqrt{(\tau')^2 + 2\tau'\tau''\frac{x_2}{2R} + (\tau'')^2}$$

$$\tau' = \frac{P}{\sqrt{2}x_1 x_2}, \quad \tau'' = \frac{MR}{J}, \quad M = P\left(L + \frac{x_2}{2}\right)$$

$$R = \sqrt{\frac{x_2^2}{4} + \left(\frac{x_1 + x_3}{2}\right)^2} \quad J = 2\left\{\sqrt{2}x_1 x_2\left[\frac{x_2^2}{12} + \left(\frac{x_1 + x_3}{2}\right)^2\right]\right\}$$

$$\sigma(X) = \frac{6PL}{x_4 x_3^2}, \quad \delta(X) = \frac{4PL^3}{E x_3^3 x_4}$$

$$P_c(X) = \frac{4.013E\sqrt{\frac{x_3^2 x_4^6}{36}}}{L^2}\left(1 - \frac{x_3}{2L}\sqrt{\frac{E}{4G}}\right)$$

$P$ = 6000 lb, $L$ = 14 in, $\delta_{max}$ = 0.25 in, $E$ = 30 x $10^6$ psi, $G$ = 12 x $10^6$ psi,
$\tau_{max}$ = 13,600 psi, $\sigma_{max}$ = 30,000 psi

(0.1, 0.1, 0.1, 0.1) ≤ ($x_1, x_2, x_3, x_4$) ≤ (2, 10, 10 ,2)
$x_i \in \mathbb{R}$, $n = 4$

Optimal Solution unknown

### A.2.6 Tension Compression Spring

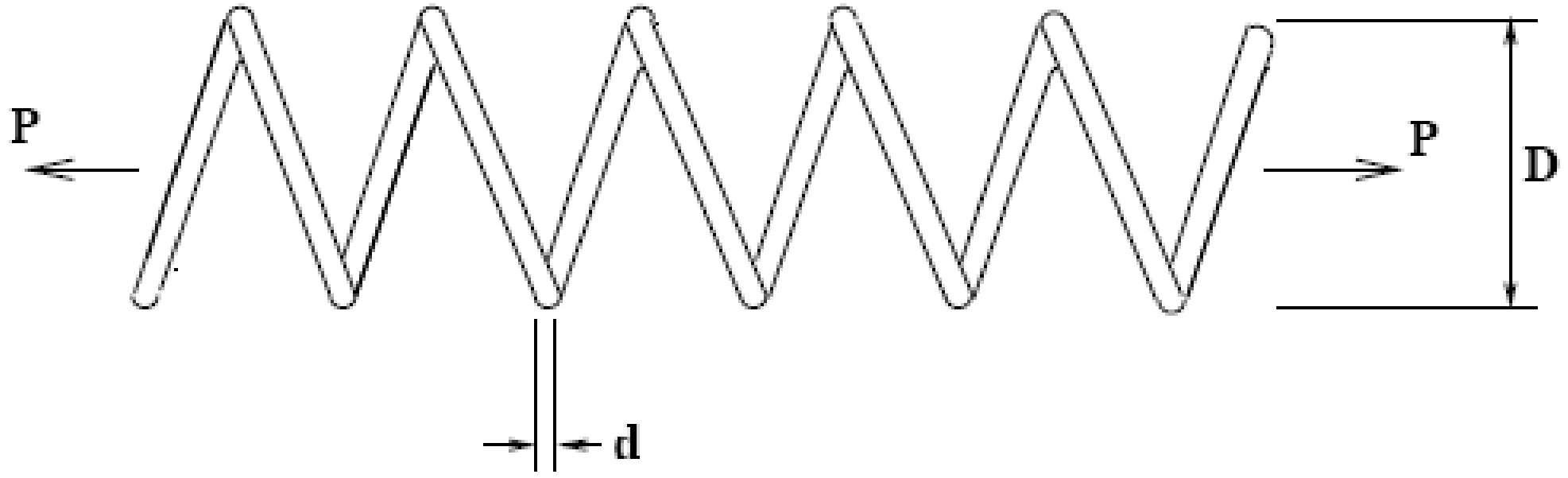

**Figure A-5 Diagram of Tension Compression Spring. Parameters of the problem include the mean coil diameter *D*, the wire diameter *d* and the number of active coils *N* which is represented by the number of loops of wire in the diagram. Forces acting on the spring are shown as *P*. This figure is taken out of [221] and is reprinted with permission from IEEE (© 1999 IEEE).**

The Tension Compression Spring problem, shown in Figure A-5, has the goal of minimizing the weight of a tension/compression spring subject to constraints on minimum deflection, shear stress, surge frequency, and dimensional constraints [221]. There are three design parameters to optimize consisting of the mean coil diameter *D*, the wire diameter *d* and the number of active coils *N*.

Original Parameters: $X = (d, D, N) = (x_1, x_2, x_3)$
$Min\ F(X) = (N+2)Dd^2$

**Subject to:**

$$g_1(X) = 1 - \frac{D^3 N}{71785 d^4} \le 0$$

$$g_2(X) = \frac{4D^2 - dD}{12566(Dd^3 - d^4)} + \frac{1}{5108 d^2} - 1 \le 0$$

$$g_3(X) = 1 - \frac{140.45 d}{D^2 N} \le 0$$

$$g_4(X) = \frac{D + d}{1.5} - 1 \le 0$$

$(0.05, 0.25, 2) \le (x_1, x_2, x_3) \le (2, 1.3, 15)$
$x_1, x_2 \in \mathbb{R},\ \ x_3 \in \mathbb{Z},\ \ \ n = 3$

Optimal Solution unknown

### A.2.7 Gear Train Design

The gear train design problem was originally defined by [235] consists of optimizing a gear train such that the gear ratio approach as close as possible to 1/6.931. There are four design parameters consisting of integer values for the number of teeth for each gear.

Original Parameters: $X = (T_d, T_b, T_a, T_f) = (x_1, x_2, x_3, x_4)$

$$Min\ F(X) = \left( \frac{1}{6.931} - \frac{x_1 x_2}{x_3 x_4} \right)^2$$

$12 \le x_i \le 60, \quad i \in \{1, \ldots, n\}$

$x_i \in \mathbb{Z}, \quad n = 4$

Optimal Solution ($F, x_1, x_2, x_3, x_4$) = (2.70 x$10^{-12}$, 19, 16, 43, 49)

# APPENDIX B SEARCH OPERATORS

For the search operator descriptions, the offspring is defined as $H = (h_1,\ldots h_n)$ and the $j^{th}$ parent is defined as $G^j = (g_i^j,\ldots g_n^j)$. For all search operators, the first parent is always the best of the selected parents.

Also in the description, variables randomly assigned over a Uniform distribution with upper and lower bounds $U$ and $L$ are stated as Uniform($U$, $L$), variables randomly assigned over a Normal distribution with mean $\mu$ and variance $\sigma^2$ are stated as N($\mu$, $\sigma^2$).

**Table B-1: List of search operators used in experiments. Details provided in this table include the search operator name, other common name, reference for description, and parameter settings if different from reference.**

| Search Operator Name | Other Name | Parameter Settings | Reference |
|---|---|---|---|
| Wright's Heuristic Crossover | Interpolation | $r = 0.5$ | [53] |
| Simple Crossover | Single point Crossover | — | [66] |
| Extended Line Crossover | Extrapolation | $\alpha$=0.3 | [46] |
| Uniform Crossover | Discrete Crossover | — | [46] |
| BLX- α | | $\alpha$=0.2 | [236] |
| Differential Evolution | | — | [50] |
| Swap | | — | [104] |
| Raise | | $A = 0.01$ | [104] |
| Creep | | $A = 0.001$ | [54] |
| Single Point Random Mutation | | — | [66] |

## B.1 Single Point Random Mutation

This operator is described in [66] and requires a single parent $G^1$. The $i^{th}$ gene is defined by:

$$h_i = \begin{cases} r & if \quad i = k \\ g_i^1 & else \end{cases}$$

$$k = Uniform[0, n], \quad k \in Z$$

$$r = Uniform\left(h_i^{MAX}, h_i^{MIN}\right)$$

## B.2 Creep

The creep operator is a variant of Gaussian mutation and was originally described in [54]. Creep requires a single parent $G^1$. The $i^{th}$ gene is defined by:

$$h_i = \begin{cases} g_i^1 + r & if \quad i = k \\ g_i^1 & else \end{cases}$$

$$k = Uniform[0, n], \quad k \in Z$$

$$r = N(0, \sigma^2)$$

$$\sigma = \frac{h_i^{MAX} - h_i^{MIN}}{1000}$$

## B.3 Raise

This operator is described in [104] and requires a single parent $G^1$. The $i^{th}$ gene is defined by:

$$h_i = g_i + r_i$$

$$r_i = N(0, \sigma_i^2)$$

$$\sigma_i = \frac{h_i^{MAX} - h_i^{MIN}}{100}$$

## B.4 Swap

This operator is described in [104] and requires two parents, $G^1$ and $G^2$. Defining $G^1$ as the more fit parent, the $i^{th}$ gene is defined by:

$$h_i = \begin{cases} g_i^1 & if \quad R_i > \alpha \\ g_i^2 & else \end{cases}$$

$$R_i = Rank(\|g_i^1 - g_i^2\|)$$

The function Rank() gives the ranking of the absolute difference between gene values of two parents (Rank = 1 being the greatest absolute difference).

Parameter Specifications: In this work, $\alpha = 1$. This means only the single most similar gene between the parents will be swapped.

## B.5 Uniform Crossover

This operator was originally described by [46] and requires two parents, $G^1$ and $G^2$. The $i^{th}$ gene is defined by:

$$h_i = \begin{cases} g_i^1 & if \quad r_i > 0.5 \\ g_i^2 & else \end{cases}$$
$$r_i = Uniform[0,1]$$

## B.6 Single Point Crossover

This operator requires two parents, $G^1$ and $G^2$. The offspring $H$ is then defined by:

$$H = (g_1^{\,1}, g_2^{\,1}, \ldots, g_i^{\,1}, g_{i+1}^{\,2}, \ldots, g_n^{\,2})$$
$$i = Uniform[0, n], \quad i \in \mathrm{Z}$$

## B.7 BLX- α Crossover

This operator was originally described by [236] and requires two parents, $G^1$ and $G^2$. The $i^{th}$ gene is defined by:

$$h_i = Uniform[g_{\min} - I * \alpha,\ g_{\max} + I * \alpha]$$
$$g_{\min} = Max(g_i^1, g_i^2)$$
$$g_{\max} = Min(g_i^1, g_i^2)$$
$$I = g_{\max} - g_{\min}$$

Specifications: The parameter α must be set by the user. In this work, $\alpha = 0.2$

## B.8 Wright's Heuristic Crossover

This operator was originally described in [53] and requires two parents, $G^1$ and $G^2$. Defining $G^1$ as the more fit parent, the $i^{th}$ gene is defined by:

$$h_i = r(g_i^1 - g_i^2) + g_i^1$$
$$r = Uniform[0,1]$$

Modifications: In this work, $r$ is a static value set at 0.5. This operator has been modified to create a single offspring instead of two offspring by defining $G^1$ as the more fit parent.

## B.9 Extended Line Crossover

This operator was originally described in [46] and requires two parents, $G^1$ and $G^2$. Defining $G^1$ as the more fit parent, the $i^{th}$ gene is defined by:

$$h_i = r\left(g_i^2 - g_i^1\right) + g_i^1$$
$$r = Uniform\left[-0.25, 1.25\right]$$

Modifications: In this work, $r$ is a static value set at 0.3. This operator has been modified to create a single offspring instead of two offspring by defining $G^1$ as the more fit parent.

## B.10 Differential Evolution Operator

This operator was originally described in [50] and requires four parents $G^1$, $G^2$, $G^3$, and $G^4$. The $i^{th}$ gene is defined by:

$$h_i = \begin{cases} g_i^2 + \alpha\left(g_i^3 - g_i^4\right) \rightarrow \beta > r \\ g_i^1 \rightarrow else \end{cases}$$
$$r = Uniform\left[0,1\right]$$

Specifications: $\alpha$ and $\beta$ are parameters that must be set. In this work, $\alpha = 1$ and $\beta = 0.5$

# APPENDIX C ADDITIONAL RESULTS FROM CHAPTER 5

## C.1 Engineering Design Problems: Performance Comparisons with the Literature

Tables are provided below which compare solution results from experiments in Chapter 5 with other results stated in the literature. The first column lists the authors (with reference), the second column states the reported fitness values, and the third column provides the number of objective function evaluations used to obtain the fitness values reported in column two. It is important to keep in mind that the results taken from the literature represent the best solution among all algorithms tested in that reference. Also, some studies implement different requirements for constraint feasibility making some of the results difficult to compare. For the constraint function values reported in this section, negative values are used to indicate the satisfaction of inequality constraints.

**Table C-1 Comparison of results for the alkylation process design problem (maximization problem). Results from other authors were reported in [224]. The best solution found in these experiments was ($F$, $x_1$, $x_2$, $x_3$, $x_4$, $x_5$, $x_6$, $x_7$) = (1772.77, 1698.18, 54.73, 3029.65, 90.25, 95, 1035, 153.53) with constraints ($g_1$, $g_2$, $g_3$, $g_4$, $g_5$, $g_6$, $g_7$, $g_8$, $g_9$, $g_{10}$, $g_{11}$, $g_{12}$, $g_{13}$, $g_{14}$) = (0, 0, 4.70E-11, 0, 0, 3.72E-11, 9.98E-8, -0, 0, 0, 0, 0, 0, 0).**

| Reference | Fitness | Objective Function Evaluations |
|---|---|---|
| Bracken and McCormick, 1968 [237] | 1769 | not reported |
| Maranas and Floudas, 1997 [238] | 1772.77 | not reported |
| Adjiman et al., 1998 [239] | 1772.77 | not reported |
| Edgar and Himmelblau, 2001 [240] | 1768.75 | not reported |
| Babu and Angira, 2006 [224] | 1766.36 | 92287 (average value) |
| SOTEA (This Thesis) | 1772.77 | 150,000 |
| cGA (This Thesis) | 1772.77 | 150,000 |
| Panmictic EA (This Thesis) | 1771.35 | 150,000 |

**Table C-2 Comparison of results for the heat exchanger network design problem (minimization problem). Results from other authors were reported in [224]. The best solution found in these experiments was ($F$, $x_1$, $x_2$, $x_3$, $x_4$, $x_5$) = (7049.25, 579.19, 1360.13, 5109.92, 182.01, 295.60) with constraints ($g_1$, $g_2$, $g_3$) = (-2.06E-3, -6.22E-3, -4.60E-3).**

| Reference | Fitness | Objective Function Evaluations |
|---|---|---|
| Angira and Babu, 2003 [234] | 7049.25 | 36620 |
| Babu and Angira, 2006 [224] | 7049.25 | 31877 |
| SOTEA (This Thesis) | 7049.25 | 150,000 |
| cGA (This Thesis) | 7049.25 | 150,000 |
| Panmictic EA (This Thesis) | 7053.47 | 150,000 |

**Table C-3 Comparison of results for the pressure vessel design problem (minimization problem). Results from other authors were reported in [223]. Results are also reported for [222] however their solution violates integer constraints for the 3rd and 4th parameters making their final solution infeasible. It should also be mentioned that equations for defining the problem have errors in [221] and [223]. Previous studies have used different bounds for the solution parameters in this problem which are stated in Column 4. These bounds can change the location of the optimal solution making it hard to compare experimental results from different authors. The best solution found in these experiments was ($F$, $x_1$, $x_2$, $x_3$, $x_4$) = (5850.37, 38.8601, 221.365, 12, 6) with constraints ($g_1$, $g_2$, $g_3$, $g_4$) = (-7.00E-8, -4.27E-3, -0.53, -18.66).**

| Reference | Fitness | Objective Function Evaluations | Parameter Bounds |
|---|---|---|---|
| Sandgren, 1990 [235] | 8129.80 | not reported | not reported |
| Fu et. al., 1991 [241] | 8084.62 | not reported | not reported |
| Kannan and Kramer, 1994 [242] | 7198.04 | not reported | not reported |
| Cao and Wu, 1997 [243] | 7108.62 | not reported | not reported |
| Lin et. al., 1999 [220] | 6370.70 | 50,000 | not reported |
| Coello, 1999 [221] | 6288.74 | 900,000 | $1 \leq x_1 \leq 99$,<br>$1 \leq x_2 \leq 99$,<br>$10.0000 \leq x_3 \leq 200.0000$,<br>$10.0000 \leq x_4 \leq 200.0000$ |
| Zeng et al., 2002 [222] | ~~5804.39~~ | not reported | $0 \leq x_1 \leq 10$,<br>$0 \leq x_2 \leq 10$,<br>$0 \leq x_3 \leq 100$,<br>$0 \leq x_4 \leq 240$ |
| Li et al., 2002 [223] | 5850.38 | not reported | not reported |
| SOTEA (This Thesis) | 5850.37 | 150,000 | $1 \leq x_1 \leq 20$,<br>$1 \leq x_2 \leq 20$,<br>$1 \leq x_3 \leq 100$,<br>$1 \leq x_4 \leq 400$ |
| cGA (This Thesis) | 5850.37 | 150,000 | |
| Panmictic EA (This Thesis) | 5857.39 | 150,000 | |

**Table C-4 Comparison of results for the welded beam design problem (minimization problem). Results from other authors were reported in [222]. The best solution found in these experiments was ($F$, $x_1$, $x_2$, $x_3$, $x_4$) = (1.72485, 0.20572973978, 3.47048651338, 9.0366239103, 0.2057296397) with constraints ($g_1$, $g_2$, $g_3$, $g_4$, $g_5$, $g_6$, $g_7$) = (0, 0, -9.99E-8, 0, 0, 0, 0).**

| Reference | Fitness | Objective Function Evaluations |
|---|---|---|
| Coello, 1999 [221] | 1.74830941 | 900,000 |
| Zeng et al. 2002 [222] | 1.72553637 | not reported |
| SOTEA (This Thesis) | 1.72485217 | 150,000 |
| cGA (This Thesis) | 1.72485217 | 150,000 |
| Panmictic EA (This Thesis) | 1.72485218 | 150,000 |

**Table C-5 Comparison of results for the tension compression spring problem (minimization problem). Results from other authors were reported in [221]. The best solution found in these experiments was ($F$, $x_1$, $x_2$, $x_3$) = (0.0126652303, 0.051838, 0.360318, 11.081416) with constraints ($g_1$, $g_2$, $g_3$, $g_4$) = (-3.16E-5, 1.47E-5, -4.06, -0.725).**

| Reference | Fitness | Objective Function Evaluations |
|---|---|---|
| Belegundu,1982 [244] | 0.0128334375 | not reported |
| Arora, 1989 [245] | 0.0127302737 | not reported |
| Coello, 1999 [221] | 0.0127047834 | 900,000 |
| SOTEA (This Thesis) | 0.0126652303 | 150,000 |
| cGA (This Thesis) | 0.0126652303 | 150,000 |
| Panmictic EA (This Thesis) | 0.0126652593 | 150,000 |

**Table C-6 Comparison of results for the gear train design problem (minimization problem). Results from other authors were reported in [220]. The best solution found in these experiments was ($F$, $x_1$, $x_2$, $x_3$, $x_4$) = (2.70 x$10^{-12}$, 19, 16, 43, 49).**

| Reference | Fitness | Objective Function Evaluations |
|---|---|---|
| Fu et. al., [241] | 4.5 x$10^{-6}$ | not reported |
| Cao and Wu, 1997 [243] | 2.36 x$10^{-9}$ | not reported |
| Deb and Goyal, 1997 [246] | 2.70 x$10^{-12}$ | not reported |
| Lin et al. 1999 [220] | 2.70 x$10^{-12}$ | 50,000 |
| SOTEA (This Thesis) | 2.70 x$10^{-12}$ | 150,000 |
| cGA (This Thesis) | 2.70 x$10^{-12}$ | 150,000 |
| Panmictic EA (This Thesis) | 2.70 x$10^{-12}$ | 150,000 |

## C.2 Panmictic EA Performance Tables

**Table C-7: Final performance results for eight Panmictic Evolutionary Algorithms run for 3000 generations with algorithm designs varying by the use of generational (Gen) or pseudo steady state (SS) population updating, the use of binary tournament selection (Tour) or truncation selection (Trun), and the number of search operators ($N_{ops}$). Performance is presented as the single best objective function value found in 20 runs $F_{Best}$ as well as the average objective function value over 20 runs $F_{Ave}$. None of the Evolutionary Algorithms listed below failed to obtain a feasible solution within 3000 generations. The single best fitness values found for each problem are in bold.**

| Gen | Sel | $N_{ops}$ | **Pressure Vessel** | | **Heat Exchanger Network** | | **Alkylation Process** | |
|---|---|---|---|---|---|---|---|---|
| | | | $F_{Best}$ | $F_{Ave}$ | $F_{Best}$ | $F_{Ave}$ | $F_{Best}$ | $F_{Ave}$ |
| SS | Tour | 7 | 6059.70 | 6190.31 | **7053.47** | 7109.20 | **1771.35** | 1750.38 |
| SS | Trun | 7 | 6059.73 | 6214.31 | 7056.09 | 7179.02 | 1760.77 | 1630.90 |
| Gen | Tour | 7 | 5953.06 | 6123.22 | 7116.72 | 7213.38 | 1711.00 | 1667.34 |
| Gen | Trun | 7 | 5964.23 | 6174.55 | 7186.97 | 7250.82 | 1641.47 | 1495.13 |
| SS | Tour | 2 | 5867.87 | 6382.61 | 7070.57 | 7233.18 | 1756.00 | 1708.38 |
| SS | Trun | 2 | **5857.39** | 6449.57 | 7093.12 | 7269.02 | 1748.95 | 1661.17 |
| Gen | Tour | 2 | 6144.69 | 6340.23 | 7235.69 | 7412.11 | 1621.77 | 1510.93 |
| Gen | Trun | 2 | 6188.86 | 6391.15 | 7184.51 | 7398.23 | 1501.24 | 1343.48 |
| | | | **Gear Train** | | **Tension Compression Spring** | | **Welded Beam** | |
| | | | $F_{Best}$ | $F_{Ave}$ | $F_{Best}$ | $F_{Ave}$ | $F_{Best}$ | $F_{Ave}$ |
| SS | Tour | 7 | **2.70E-12** | 2.62E-10 | 0.012665 | 0.012758 | **1.72485** | 1.74602 |
| SS | Trun | 7 | **2.70E-12** | 7.70E-10 | **0.012665** | 0.012778 | 1.72494 | 1.80945 |
| Gen | Tour | 7 | **2.70E-12** | 2.70E-12 | 0.012679 | 0.012710 | 1.75465 | 1.77920 |
| Gen | Trun | 7 | **2.70E-12** | 1.09E-11 | 0.012687 | 0.012725 | 1.76485 | 1.79732 |
| SS | Tour | 2 | **2.70E-12** | 1.12E-09 | 0.012701 | 0.013861 | 1.73570 | 1.96193 |
| SS | Trun | 2 | 2.31E-11 | 1.81E-09 | 0.012804 | 0.015078 | 1.73060 | 2.06087 |
| Gen | Tour | 2 | **2.70E-12** | 4.74E-12 | 0.012739 | 0.013035 | 1.83742 | 1.93124 |
| Gen | Trun | 2 | **2.70E-12** | 2.70E-12 | 0.012694 | 0.012864 | 1.75302 | 1.88472 |
| | | | **Frequency Modulation** | | **Error Correcting Code** | | **System of Linear Equations** | |
| | | | $F_{Best}$ | $F_{Ave}$ | $F_{Best}$ | $F_{Ave}$ | $F_{Best}$ | $F_{Ave}$ |
| SS | Tour | 7 | **0.00** | 15.36 | 3.53E-03 | 4.32E-03 | **8.53E-14** | 2.12E-05 |
| SS | Trun | 7 | 6.69 | 18.28 | 3.68E-03 | 4.29E-03 | 3.16E-05 | 1.32 |
| Gen | Tour | 7 | 23.07 | 26.95 | 2.47E-03 | 3.75E-03 | 10.90 | 14.58 |
| Gen | Trun | 7 | 22.87 | 25.97 | 3.44E-03 | 4.13E-03 | 2.45 | 5.27 |
| SS | Tour | 2 | 8.98 | 15.87 | **2.70E-07** | 3.84E-03 | 1.67 | 3.54 |
| SS | Trun | 2 | 0.55 | 16.49 | 3.43E-03 | 3.96E-03 | 4.26 | 5.90 |
| Gen | Tour | 2 | 23.35 | 26.33 | 4.18E-03 | 4.77E-03 | 50.21 | 74.11 |
| Gen | Trun | 2 | 21.95 | 26.77 | **2.70E-07** | 3.17E-03 | 35.69 | 51.75 |
| | | | **Rastigrin** | | **Griewangk** | | **Watson** | |
| | | | $F_{Best}$ | $F_{Ave}$ | $F_{Best}$ | $F_{Ave}$ | $F_{Best}$ | $F_{Ave}$ |
| SS | Tour | 7 | **1.25E-10** | 1.65E-06 | **0.012** | 0.052 | **1.716E-02** | 2.025E-02 |
| SS | Trun | 7 | 4.24E-02 | 1.26E-01 | 0.049 | 0.158 | 1.728E-02 | 2.922E-02 |
| Gen | Tour | 7 | 6.33E-01 | 9.17E-01 | 0.615 | 0.751 | 1.778E-02 | 1.941E-02 |
| Gen | Trun | 7 | 8.82E-02 | 1.96E-01 | 0.348 | 0.508 | 1.730E-02 | 1.828E-02 |
| SS | Tour | 2 | 3.10E-02 | 6.92E-02 | 0.131 | 0.216 | 1.804E-02 | 4.887E-02 |
| SS | Trun | 2 | 1.64E-01 | 2.83E-01 | 0.154 | 0.366 | 1.829E-02 | 4.369E-02 |
| Gen | Tour | 2 | 7.82 | 10.51 | 1.476 | 2.729 | 2.444E-02 | 5.673E-02 |
| Gen | Trun | 2 | 4.89 | 7.53 | 1.474 | 2.199 | 2.205E-02 | 4.111E-02 |

# PUBLICATIONS

## *Journals*

Whitacre, J. M., Sarker, R. A., and Pham, T. Q., "The Self-Organization of Interaction Networks for Nature-Inspired Optimization." *IEEE Transactions on Evolutionary Computation*, (Accepted March, 2007) http://www.ceic.unsw.edu.au/staff/Tuan_Pham/Whitacre_SOTEA_2007.pdf

Whitacre, J. M., Sarker, R. A., and Pham, T. Q., "The Self-Organized Criticality of Population Dynamics and its Relevance to Adaptive Evolutionary Algorithms." *IEEE Transactions on Evolutionary Computation*, (**Submitted November, 2006**)

Whitacre, J. M., Sarker, R. A., and Pham, T. Q., "The influence of population topology and historical coupling on Evolutionary Algorithm population dynamics." *Applied Soft Computing*, (**Submitted September, 2007**)

Whitacre, J. M., Sarker, R. A., and Pham, T. Q., "A Self-Organizing Topology for distributed Evolutionary Algorithms based on fitness-driven community structures." *IEEE Transactions on Evolutionary Computation*, (**Submitted September, 2007**)

## *Conference Proceedings*

Whitacre, J. M., Pham, T. Q., and Sarker, R. A., "Use of statistical outlier detection method in adaptive evolutionary algorithms." In *Proceedings of the 8th Annual Conference on Genetic and Evolutionary Computation* (Seattle, Washington, USA, July 08 - 12, 2006). GECCO '06. ACM Press, New York, NY, 1345-1352, 2006. www.ceic.unsw.edu.au/staff/Tuan_Pham/fp122-whitacre.pdf

Whitacre, J. M., Pham, T. Q., and Sarker, R. A., "Credit assignment in adaptive evolutionary algorithms." In *Proceedings of the 8th Annual Conference on Genetic and Evolutionary Computation* (Seattle, Washington, USA, July 08 - 12, 2006). GECCO '06. ACM Press, New York, NY, 1353-1360, 2006. www.ceic.unsw.edu.au/staff/Tuan_Pham/fp123-whitacre.pdf